\documentclass[final,twocolumn]{svjour3}

\usepackage{graphicx}
\usepackage{color}
\usepackage{cite}

\usepackage[cmex10]{amsmath}
\usepackage{url,xspace,amssymb,amscd}
\usepackage[linesnumbered,ruled,vlined]{algorithm2e}
\usepackage{subfigure}
\usepackage{multirow}

\usepackage{marvosym}

\newcommand{\fnorm}[2][2]{\ensuremath{ \left\| #2 \right\|_{ \mathrm{#1} } } }
\newcommand\norm[1]{\left\lVert#1\right\rVert}

\newcommand{\st}{{{\rm s.t.}\!:}\xspace}

\newcommand{\argmin}{\operatornamewithlimits{argmin}}

\newcommand\comment[1]{{ }}

\def\bx{{\bf x}}

\def\bd{{\bf d}}

\def\bs{{\bf s}}
\def\bz{{\bf z}}

\def\bq{{\bf q}}
\def\bt{{\bf t}}
\def\bff{{\bf f}}

\def\bS{{\bf S}}
\def\bD{{\bf D}}

\def\calN{{\cal N}}
\def\st{ {\rm s.t.} }

\def\paragraph{\subsubsection}

\graphicspath{{./}{./datasets/result/image_fig/}
{./datasets/result/image_fig/csh/}{./datasets/result/image_fig/sf/}{./datasets/result/image_fig/our/}{./datasets/result/image_fig/dsp/}{./datasets/result/image_fig/im_anno/}
{./datasets/result/LMO_fig/}
{./datasets/result/LMO_fig/csh/}{./datasets/result/LMO_fig/sf/}{./datasets/result/LMO_fig/our/}{./datasets/result/LMO_fig/dsp/}
{./datasets/result/feature/}
{./datasets/result/pixel/}
{./datasets/result/pascal_fig/}
}

\usepackage{etoolbox}

\newcommand{\vspacedistfront}{-0.00010\textwidth}
\newcommand{\vspacedistmid}{-0.00012\textwidth}
\newcommand{\vspacedist}{-0.00015\textwidth}

\journalname{%
}
\date{v1 July  2014; v2 December 2014; v3 April 2015}

\begin{document}
\sloppy

\title{Unsupervised Feature Learning for Dense Correspondences across Scenes}

\author{Chao Zhang, Chunhua Shen, Tingzhi Shen}

\institute{
  C. Zhang
   \at
     Beijing Institute of Technology, Beijing 100081, China
   \at
     The University of Adelaide, SA 5005, Australia
  \and
  C. Shen
  (\Letter)
    \at
     The University of Adelaide, SA 5005, Australia
    \at
     Australian Centre for Robotic Vision
    \\
    \email{\url{chunhua.shen@adelaide.edu.au}}
\and
  T. Shen
   \at
     Beijing Institute of Technology, Beijing 100081, China
}

\maketitle

\begin{abstract}

    We propose a fast, accurate matching method for estimating dense pixel correspondences
    across scenes.
    It is  a challenging problem to estimate
    dense pixel correspondences between images depicting different scenes
    or instances of the same object category.
    While most such matching methods rely on hand-crafted features such as SIFT, we learn
    features from a large amount of unlabeled image patches using unsupervised
    learning. Pixel-layer features are obtained by encoding over the dictionary,
    followed by spatial pooling to obtain patch-layer features.
    The learned features are then seamlessly embedded into a multi-layer matching framework.
    We experimentally demonstrate that the learned features, together with our
    matching model, outperform state-of-the-art methods such as
    the SIFT flow \cite{liu2011sift},
    coherency sensitive hashing \cite{korman2011coherency}
    and the recent deformable spatial pyramid matching \cite{kim2013deformable}
    methods both in terms of accuracy and computation efficiency.

        Furthermore, we evaluate the performance of a few different dictionary learning and
        feature encoding methods in the proposed pixel correspondence estimation
        framework, and analyze the impact of dictionary learning and feature encoding with respect to
        the final matching performance.

\end{abstract}

\section{Introduction}

  Estimating the dense correspondence between two images across scenes is an important task,
  which has many applications in computer vision and computational photography.
  Yet, it is a  challenging problem due to large
variations exhibited in the matching images.
  Conventional dense matching methods developed for optical flow and stereo usually
  only work well for the cases in which the two input images
  contain different views of the {\em same} object.
Here we are interested in dense matching of images with different objects or scenes.
This requires the matching algorithms to be highly robust to
different object appearances and backgrounds, illumination changes,
large displacements and viewpoint changes.
For the task of matching objects in a specific category,
the intra-class variability can be larger
than the inter-class differences. %

\begin{figure*}
\centering
\includegraphics[width=0.93\textwidth]{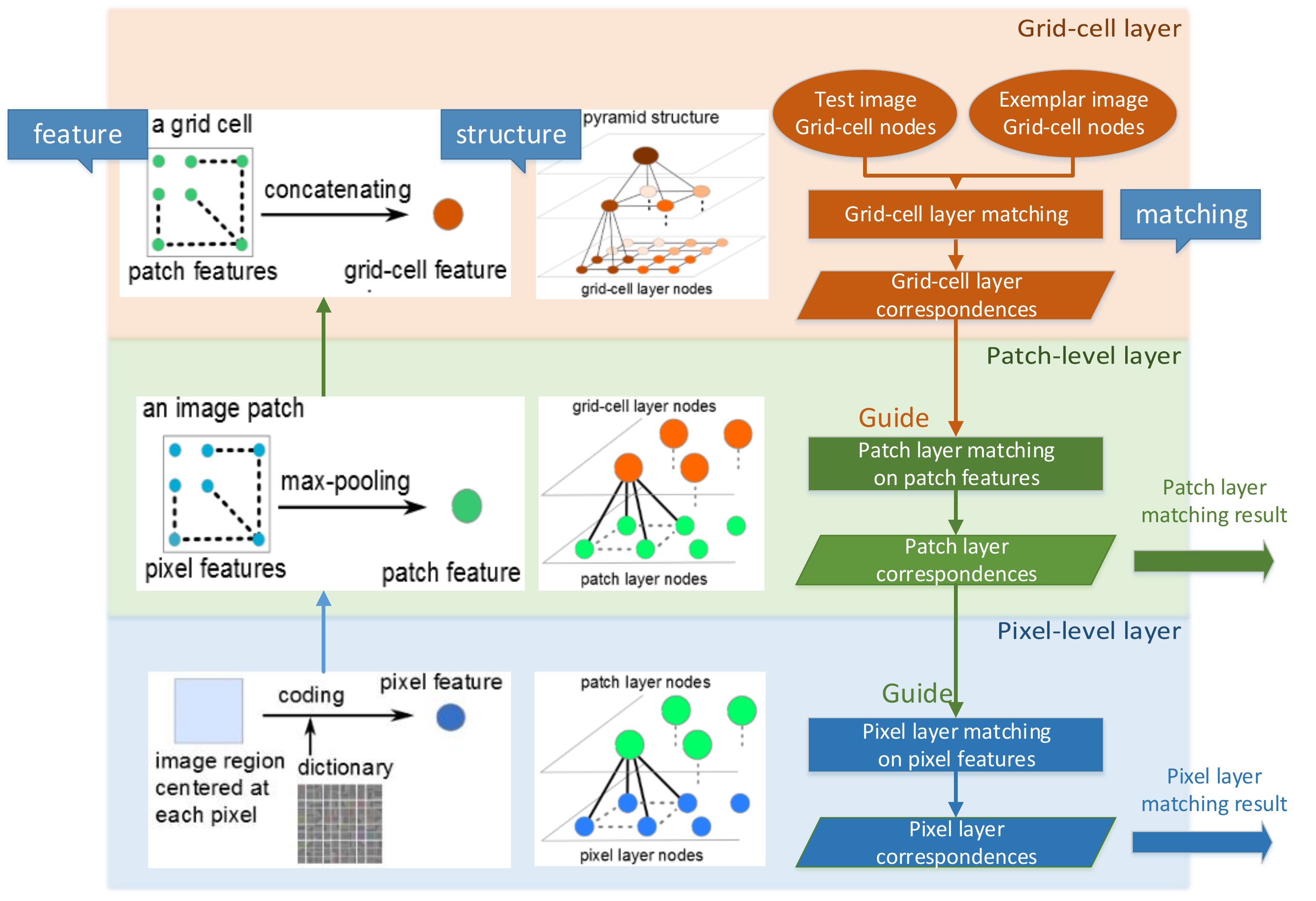}
\caption{Illustration of our multi-layer matching and unsupervised feature learning model.
First column shows the feature extraction process of each layer.
Second column shows the node structure of each layer.
Third column outlines the matching pipeline.
The learned features at the pixel-level layer within a patch are
spatially pooled to form a patch-level feature.
Here the grid-cell feature is the concatenation of patch-level features within a cell.
The matching result from the grid-cell layer guides the matching
at the patch-level layer;
and the result at the patch-level layer guides the matching at the pixel-level layer.
In our experiments, the matching accuracy obtained by the patch-level layer is already very high.
Pixel-layer matching can further improve the matching accuracy.
}
\label{fig:our_framework}
\end{figure*}

Recently a few methods were proposed to address these challenges, including
hierarchical matching \cite{liu2011sift},
fast patch matching \cite{Barnes2010TGP,korman2011coherency},
sparse-to-dense matching \cite{leordeanu2013sparsetodense} and
most recently spatial pyramid matching \cite{kim2013deformable}.
Current matching approaches typically rely on either raw image patches or hand-designed
image features (e.g., SIFT features \cite{SIFT}).
Raw pixels or patches often lack the robustness to cope with those challenging
appearance variations.
Given  a particular task, in order to  model complex real-world data, robust and distinctive
feature descriptors that can
capture relevant information
are needed.
Hand-crafted features like SIFT have achieved great success in many vision
tasks such as image classification \cite{bo2013multipath},
retrieval, and image matching.
As SIFT features have passed the test of time for  good performance,
SIFT is considered as one of the milestone results in computer vision,
which was first introduced more than a decade ago \cite{SIFT}.
Despite the remarkable success in a number of applications, SIFT is criticized for drawbacks such as
its large computational burden, and being incapable to well accommodate affine viewpoint transformation.
Researchers have been seeking improved feature descriptors.
However,
manually designing features for each data set and task can be very
expensive, time-consuming, and typically requires domain knowledge of the data.
In recent years, researchers observed that instead of manually designing features
using heuristics, learning features from a large amount of unlabeled data
with some {\em unsupervised machine learning} approaches achieves tremendous success
in various applications.
For example, in visual recognition the unsupervised feature learning pipeline
has now become the common approach
\cite{coates2011importance,coates2011analysis}.
Feature learning is attractive as it exploits the availability of data and avoids
the need of feature engineering \cite{bo2013multipath}.
For unsupervised feature learning, its main advantage  is that
unlabeled domain-specific data are usually abundant and
very cheap to obtain.
Inspired by the success
of \cite{coates2011analysis,huang2012learning,bo2013multipath,kim2013deformable},
         we propose unsupervised feature learning for
         dense pixel correspondence estimation within a
         multi-layer matching framework.
The outline of our multi-layer model is illustrated in Figure \ref{fig:our_framework}.

In our framework, features at the bottom layer (namely, the pixel layer) are extracted from raw image patches
using unsupervised feature learning methods.
We then obtain more compact representations of larger-size nodes at higher-level layers,
which achieve better robustness to noise and clutter, thus better
deal with severe variations in object or scene appearances.
Larger spatial nodes with more compact features provide better
geometric regularization when the matching objects undergo
large appearance variations, while smaller spatial nodes
with more detailed features obtain finer correspondence.
Our matching  starts from the top layer (i.e., the grid-cell layer).
The matching solution of a higher layer provides
reliable initial correspondences to the lower layer.

We apply several well-known unsupervised feature learning algorithms
to extract pixel layer features.
Then we present a detailed analysis on the impact of various parameters and configurations of our framework---the
matching model as well as the unsupervised feature learning techniques.
Despite the simplicity of our system, our framework outperforms
all previously published matching accuracy on the Caltech-101 dataset, the LMO dataset \cite{liu2009nonparametric},
and a subset of the Pascal dataset \cite{everingham2014pascal}.
Our results demonstrate that it is possible to achieve
state-of-the-art performance by using a tailored matching
framework, even with simple unsupervised feature learning techniques.

Our {\it main contributions} are thus as follows.
\begin{itemize}
  \item
 We apply unsupervised feature learning to the
problem of dense pixel correspondence estimation, rather than using hand-designed features.
Experiment results show that our method outperforms recent state-of-the-art methods
\cite{liu2011sift,korman2011coherency,kim2013deformable}
in terms of both accuracy and running time.
Our experiments demonstrate that the learned features can well handle variations of different factors.

\item

  Inspired by the recent development in multi-layer networks
and deep learning methods,
we perform matching at several levels of the image representations (grid-cell layer, patch layer, pixel layer).
Our multi-layer matching model, designed for fast and
accurate matching, is suitable for the multi-layer unsupervised feature learning pipeline.

\item

  We use the patch-layer feature as the basic unit to estimate
  correspondences in the patch-layer matching such that the computation time is considerably faster
  (due to less time spent on feature extraction and fewer variables to optimize) while
still keeping the desirable power of learned features.
Matching results at the patch layer have already
outperformed those state-of-the-art methods in the literature  in terms of both
matching accuracy and efficiency.

\item

We evaluate the performance of a few dictionary learning and feature
encoding methods in the proposed pixel correspondence estimation framework.
Moreover, we study the effect of parameter choices on the features
learned by several feature learning methods.
Several important conclusions are drawn, which are different from the
case of unsupervised feature learning for
image classification  \cite{coates2011importance}.
\end{itemize}

\subsection{Related work}

 We briefly review some relevant work in dense matching and unsupervised feature learning.
 Estimation of dense correspondences between images is essential for many computer vision
 tasks such as image registration,
 segmentation\cite{rubinstein2013unsupervised},
 stereo matching and object recognition\cite{duchenne2011graph}.
 It is challenging to estimate dense correspondences between images that contain different
 scenes.

Graph matching algorithms \cite{berg2005shapematch,leordeanu2005spectral,duchenne2011graph}
were introduced to find the dense correspondence.
Typically these methods use sparse features and
rely on geometric relationships between nodes.
 Optical flow methods have been used to estimate
 the motion field and dense correspondence in the literature.
 Recently, SIFT Flow \cite{liu2011sift} adopts the computational framework of optical flow
 and produces dense pixel-level correspondences by matching SIFT descriptors
 instead of raw pixel intensities.
 A coarse-to-fine matching scheme is used in their method to speed up the matching procedure.
Kim et al.\ \cite{kim2013deformable} proposed a deformable spatial pyramid (DSP)
model for fast dense matching.
Their model regularizes matching consistency through a pyramid graph.
The matching cost in DSP is defined by using multiple SIFT descriptors.
PatchMatch \cite{Barnes2010TGP} and more recent work
of coherency sensitive hashing (CSH) \cite{korman2011coherency}
are much faster in finding the matching patches between two images, but abandon explicit
geometric smoothness regularization for the speed,
which may lead to noisy matching results due to negligence of pixels' geometric relations.
Leordeanu et al.\ \cite{leordeanu2013sparsetodense} proposed to extend sparse matching
 to dense matching by introducing local constraints.

Image matching in general consists of two components:
local image feature extraction and feature matching.
  First, one must  define the image features based upon which the correspondence between a pair
  of images can be established. An ideal image descriptor should be robust so that it does not change from one image
  to another.
Many methods use SIFT features as local descriptors because of their
robustness to scale and illumination changes, etc. Recent work showed that, to some extent,
carefully designed descriptors may improve the matching results \cite{tola2008fast,zelnik2012sifts}.
In \cite{zelnik2012sifts}, it is shown that
SIFT features extracted at multiple scales lead to much better matches than single-scale features.
All these features were manually designed. Instead, our work here is  inspired by those feature learning approaches
that first appeared
in image classification.

In recent years, a large body of work on generic image
classification/categorization has focused
on learning features in an unsupervised  fashion \cite{coates2011importance,coates2011analysis}.
Unsupervised feature learning (or deep learning by stacking unsupervised feature learning) has emerged as a promising technique
for designing task-specific features by exploiting a large amount of unlabeled data \cite{quoc2013buildinghigh}.
The main purpose of unsupervised feature learning is
to design low-dimensional features that capture some structure
underlying the high-dimensional input data. Typical unsupervised feature learning methods include
independent component analysis \cite{NIPS2011ICA}, auto-encoders \cite{AISTATS2012ZhouSL12},
sparse coding \cite{Sparse2010,Mairal2010JMLR}, (nonnegative) matrix factorization \cite{Lee1999,Mairal2010JMLR},
and a few clustering methods \cite{NIPS2011ADAM}.
In terms of large-scale sparse coding and matrix factorization based feature learning,
an online optimization algorithm based on stochastic approximations was introduced in \cite{Mairal2010JMLR}.

Low-level image alignment such as dense stereo matching, which shares similarity with the matching
task that we concern here,
often use hand-crafted local image descriptors \cite{Tuytelaars00widebaseline}.
Traditional local feature descriptors like SIFT was shown their values for
dense wide-baseline matching, but with limited success.
This is mainly because their high computational
cost and sensitivity to occlusions.
The SURF feature \cite{Bay:2008:SRF} tries to speed up the computation
of local features.
In \cite{Tola10}, Tola et al.\ designed the DAISY feature for fast and accurate wide-baseline stereo matching.
The DAISY feature attempts to solve both the computation and occlusion problems in stereo matching.
Another computationally cheap local feature descriptor is
a modified version of the  local binary pattern (LBP) feature \cite{Heikkila:2009}.
In the sparse matching experiment, Heikkil\"a et al.\ have showed that the LBP descriptor
performs favorably compared to the SIFT
\cite{Heikkila:2009}.
     Estimating the dense correspondence between images depicting
     different scenes, which we concern here,
     is a  much more challenging problem compared to dense stereo matching.
     To our knowledge, for dense correspondence estimation across scenes,
     to date the SIFT feature is
     still the standard due to its very good performance.

Our method is closely related to the approach
of \cite{kim2013deformable} that applies only
two layers of matching, namely, grid cell layer and pixel layer.  There both layers are represented
by the same type of features.
They utilize sparse sampling to reduce
the complexity and expense of large-node representations, which may cause loss
of discriminative information.
Instead of using SIFT features as the descriptor as in \cite{kim2013deformable},
		we learn features from a large amount of small patches, which are
		randomly extracted from natural images.
    In our method, dense matching is performed at several levels of the image representations (grid-cell layer,
    patch layer, pixel layer).
		For each layer, we obtain suitable features to represent image nodes.
		Compared to the bottom layer, features in higher layer are extracted to
		achieve more robustness to the
		noise and clutter and more compactness of representation.
		By using the max-pooling operation, we obtain more compact representations
		of larger image nodes while removing irrelevant details.
		We demonstrate the efficiency and effectiveness of the learned features over
		hand-crafted features for dense matching task.

    The second approach that has inspired our work is \cite{huang2012learning}.
    The main idea is  to combine unsupervised
    joint alignment with unsupervised feature learning.
    Huang et al.\ used unsupervised feature learning, in particular,
    deep belief networks (DBNs) to obtain features that can represent
    an image at different resolutions based on network depth \cite{huang2012learning}.
    There are major differences between our work and \cite{huang2012learning} although both have used
    unsupervised feature learning. Huang et al.\ considered the problem of
    {\em congealing} alignment of images, which is to estimate the {\em parametric} image transform.
    One only needs to optimize for  a small number of continuous variables (typically the  rotation matrix and translation).
    Thus the number of variables is independent of the image size \cite{huang2012learning}.
    In Huang et al.\ \cite{huang2012learning}, gradient descent is employed for this purpose.
    In contrast, we estimate the {\em nonparametric} correspondences at the pixel level.
    The optimization problem involved in our task is a much more challenging
    discrete combinatorial problem.
    It involves thousands of discrete variables.
    Thus we use belief propagation to achieve a locally optimal solution.
    It is unclear how the method of \cite{huang2012learning} may be applied to dense correspondences.

\section{Our approach}\label{our_approach}
In this section, we first describe how to extract features for each level
in Section \ref{pixel_patch_feat}.
Then, we present our framework in detail in Section \ref{match_framework}.

\subsection{Multi-layer  image representations}\label{pixel_patch_feat}
In this work, we follow the standard unsupervised feature learning framework in
\cite{coates2011analysis,coates2011importance},
which has been successfully applied to generic image classification.
A common feature learning framework performs the following steps
to obtain feature representations:
\begin{enumerate}
  \item
The {dictionary learning step} learns a dictionary using unsupervised learning algorithms,
which are used to map input patch vectors to the new feature vectors.
\item
  The {feature encoding step} extracts features from image patches centered at each pixel using the
  dictionary obtained at the first step.
\item
  The {pooling step} spatially  pools  features together over local regions of the images to obtain
 more compact feature representations.
For classification tasks, the learned features are then used to train a classifier
for predicting  labels.
In our case, we estimate dense correspondences in a multi-layer matching framework
  using the learned multi-level feature representations.
\end{enumerate}

Next, we briefly review the pipeline of feature learning framework.
As mentioned above, there are three key components in the feature learning framework: dictionary learning,
feature encoding and pooling operation.

\paragraph{Dictionary learning}
To learn a dictionary, we first extract $N$ random patches $\bx_i$
 from a collection of natural images as training data,
 $\bx_i\in {\bf{R}}^n, (i= 1,2,...,N) $, and
then pre-process these patches as described in \cite{coates2011analysis}.
Every patch is normalized by subtracting the mean and normalized
by the standard deviation of its elements.
This is equivalent to local brightness and contrast normalization.
We also apply the whitening process as done in \cite{coates2011analysis}.
Then we use an unsupervised
learning algorithm to construct the dictionary
$\bD=[\bd_1, \bd_2, \ldots, \bd_M] \in {\bf{R}}^{n \times M}$. Here
$M$ is the dictionary size, and each column $\bd_j$ is a codeword.

We consider the following dictionary learning methods for learning
the dictionary ${\bf{D}}$:

(a) K-means clustering: We learn $M$ centroids $ \{\bd_j\} ,j= 1,2,...,M$ from the sampled
patches.
K-means has been widely adopted in computer vision for building
codebooks but less widely used in `deep feature learning' \cite{coates2011analysis}.
This may be due to the fact that K-means is less effective when the input
vectors are of high dimension.

(b) K-SVD \cite{aharon2006ksvd}: The dictionary
is trained by solving the following optimization problem using alternating minimization:
\begin{equation}
\begin{aligned}
\label{EQ:ksvd}
\min_{\bD, \bS} \quad &  \sum_{i=1}^N {
	\fnorm{\bx_i - \bD \bs_i}^2   },
\;\; \st
	\; &  \norm{\bs_i}_0 \leq k, \forall i,
\end{aligned}
\end{equation}
where  $\bS=[\bs_1, \bs_2, \ldots, \bs_N] \in {\bf{R}}^{M \times N}$ are the sparse codes.
$\norm{\bs_i}_0$  is the number of non-zero elements in $\bs_i$,
which enforces the code's sparsity. Here
$\fnorm{\cdot}$ and $\norm{\cdot}_0$ are the $L_2$
and $L_0$ norm respectively. Note that to solve for
$ \bS $, usually one seeks an approximate solution because
the optimization problem is NP-hard.

(c) Random sampling (RA):
 $M$ patches are randomly picked from the $N$ patches to form a dictionary.
 Therefore no learning/optimization is performed in this case.

Later we  test these three dictionary
learning methods on the problem of dense matching.

\paragraph{Feature encoding}
After obtaining the dictionary $\bD$, we extract patches centered
at each pixel of the pair of matching images
after applying pre-processing.
The patch vector $\bx_i$ is encoded to generate the
feature vector $\bs_i$ at the pixel layer.
We consider the following coding methods in this work:

(a) K-means triangle (KT) encoding: This can be viewed as a `soft' encoding method
while keeping sparsity of codes.
 With the $M$ basis vectors $\{\bd_j\}$ learned by the first stage, KT encodes the patch $\bx_i$ as:
 \begin{equation*}
 s_{ij} = \max\big\{0, \mu(\bz) - z_j\big\},
 \end{equation*}
where $ s_{ij}$ is the $j$-th component of feature vector $\bs_{i}$;
$z_j = \|\bx_i - \bd_j\|$ and $\mu(\bz)$ is the mean of $\bz$.
By using this encoder, roughly half of the features are set to 0.

(b) Soft-assignment (SA) encoding:
\begin{equation}
s_{ij}  = \frac{\exp(-\beta\fnorm{\bx_i-\bd_j}^2)}{\sum_{l=1}^M\exp{(-\beta\fnorm{\bx_i-\bd_l}^2)}}.
\end{equation}
Here $\beta$ is the smoothing factor controlling the softness of assignment.

(c) Orthogonal matching pursuit (OMP-k) encoding: Given the patch $\bx_i$ and
dictionary $\bD$, we use OMP-k \cite{coates2011importance}
to obtain the feature $\bs_{i}$,  which has at most $k$ non-zero elements:
\begin{equation}
\begin{aligned}
\label{EQ:omp}
\min_{\bS } \quad &  \sum_{i=1}^N {
	\fnorm{\bx_i - \bD \bs_i}^2   },
\st \quad &  \norm{\bs_i}_0 \leq k, \forall i,
\end{aligned}
\end{equation}
where $\norm{\bs_i}_0$  is the number of non-zero elements in $\bs_i$.
This explicitly enforces the code's sparsity.

We mainly use K-means for dictionary learning and K-means triangle (KT) for encoding
in our experiments. %
 In the last part of Section \ref{feature_learning},
 we evaluate the performance of different learning and encoding methods
mentioned above in dense correspondence estimation.

\paragraph{Pooling operation}
The general objective of pooling is to transform the joint feature interpretation into a new,
 more usable one that preserves important information while discarding
 irrelevant details \cite{boureau2010theoretical}.
 Spatially pooling features over a local neighbourhood to create invariance to small transformation
 of the input is employed in a large number of models of visual recognition.
 The pooling operation is typically a sum, an average, a max or more rarely
 some other commutative combination rules.
 In this paper, we apply the max-pooling operation to
 obtain the patch-layer features.

The pixel feature $\bs_i \in {\bf{R}}^M$ is the code of the patch centered at pixel $i$,
which is obtained at the feature encoding step.
At the patch layer,the image is partitioned using a uniform grid into non-overlapping square patches.
Each patch feature $\bff = [ f_1, \cdots,$ $ f_j, \cdots,$ $f_M  ] \in {\bf{R}}^M$
 is obtained by
max-pooling  all pixel features within that patch,
which is simply the component-wise
maxima over all pixel features within a patch $P$:
\begin{equation}
\begin{aligned}
\label{EQ:F}
f_j  = \max_{i\in P }\; s_{ij}
\end{aligned}
\end{equation}
where $ i$ ranges over all entries in image patch $P$.
Thus each patch feature has the same dimension as pixel features.
Note that the max-pooling operation is  non-linear.
It captures the main character of pixels in the patch
while maintaining the feature length and reducing  the feature number.
Detailed discussions on the impact of feature learning
methods are presented  in Section
\ref{feature_learning}.

\paragraph{Grid-cell layer representations}

The grid-cell layer is built on the patch-level layer.
The structure of the grid-cell layer is a  spatial pyramid, as shown in Figure \ref{fig:grid_cell_pyramid}.
Thus it can contain multiple levels. Each level contains a number of cells at different resolutions.
The cell size starts from the whole image (the top level at Figure \ref{fig:grid_cell_pyramid})
to a certain spatial size according to
the number of pyramid levels. A cell node is much larger than an image patch and may offer
greater regularization when appearance matches are ambiguous.
Each cell is represented by a grid-cell feature, called a cell node.
Grid-cell features are formed by concatenating the patch layer features within a cell.
For all the experiments in the Section \ref{Experiments},
we use the 3-level pyramid as shown in Figure \ref{fig:grid_cell_pyramid}.
At the top level of the pyramid, there is one single cell node which is the whole image.
The middle level contains 4 equal-size non-overlapping cell nodes; and the bottom level has 16 cell nodes.

To demonstrate the great
potential of unsupervised feature learning techniques in the dense matching task,
which generally requires features to preserve visual details,
we decide not to use those complex learning algorithms and models.
Instead, we employ simple learning algorithms and design a tailored matching
model for the learned features to estimate pixel-level correspondences.
The experiments  in Section \ref{Experiments} demonstrate that
it is possible to achieve state-of-the-art performance
even with simple algorithms of unsupervised feature learning.

\subsection{The proposed matching framework}
\label{match_framework}

\paragraph{The multi-layer model}
Our matching model consists of three layers: the grid-cell layer,
patch-level layer and pixel-level layer.
\begin{figure*}
\vspace{\vspacedistfront}
\centering
\includegraphics[width=0.560\textwidth]{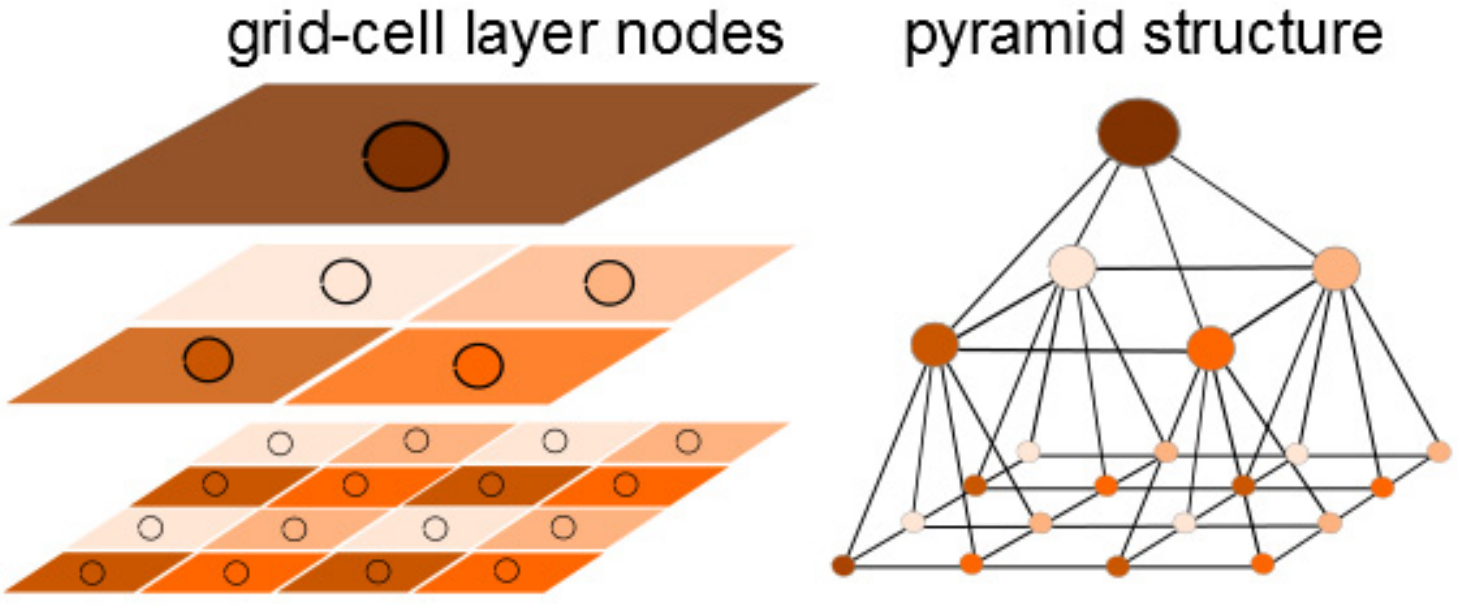} %
\vspace{.2cm}
\caption{
Illustration of the pyramid structure and node
connection of the grid-cell layer. Note that the grid-cell layer can contain multiple pyramid levels (here we have 3 levels).
}
\label{fig:grid_cell_pyramid}
\vspace{\vspacedist}
\end{figure*}

(a) \emph{model structure:}
The grid-cell layer is the top layer,
which is a conventional spatial pyramid
(we use a 3-level pyramid for all the experiments in this work).
The cell size starts from the whole image to the predefined cell size.
Grid-cell node features are formed by concatenating the patch level
features within a cell.
The patch-level layer lies underneath the grid-cell layer.
The bottom layer is pixel-level layer.

(b) \emph{node definition:}
To be clear, in our model, at the grid-cell layer, each cell can be seen as a node.
At the patch-level layer, each patch represents a node.
At the pixel-level layer, a single pixel is a node.

(c) \emph{node linkage:}
In the pyramid of the grid-cell layer,
each node links to the neighboring nodes within the same
pyramid level as well as parent and child nodes
in the adjacent pyramid levels,  as shown in Figure \ref{fig:grid_cell_pyramid}.
We define the node at the higher level layer as the parent
of the nodes within its spatial extent at the lower layer.
For the bottom two layers, namely the patch-level layer and the pixel-level layer,
each node is only linked to the parent node.

\begin{figure*}
\vspace{\vspacedistfront}
\centering
	\subfigure[]{
    \label{fig:example_patch_pixel:1}
	\includegraphics[width=0.24\textwidth]{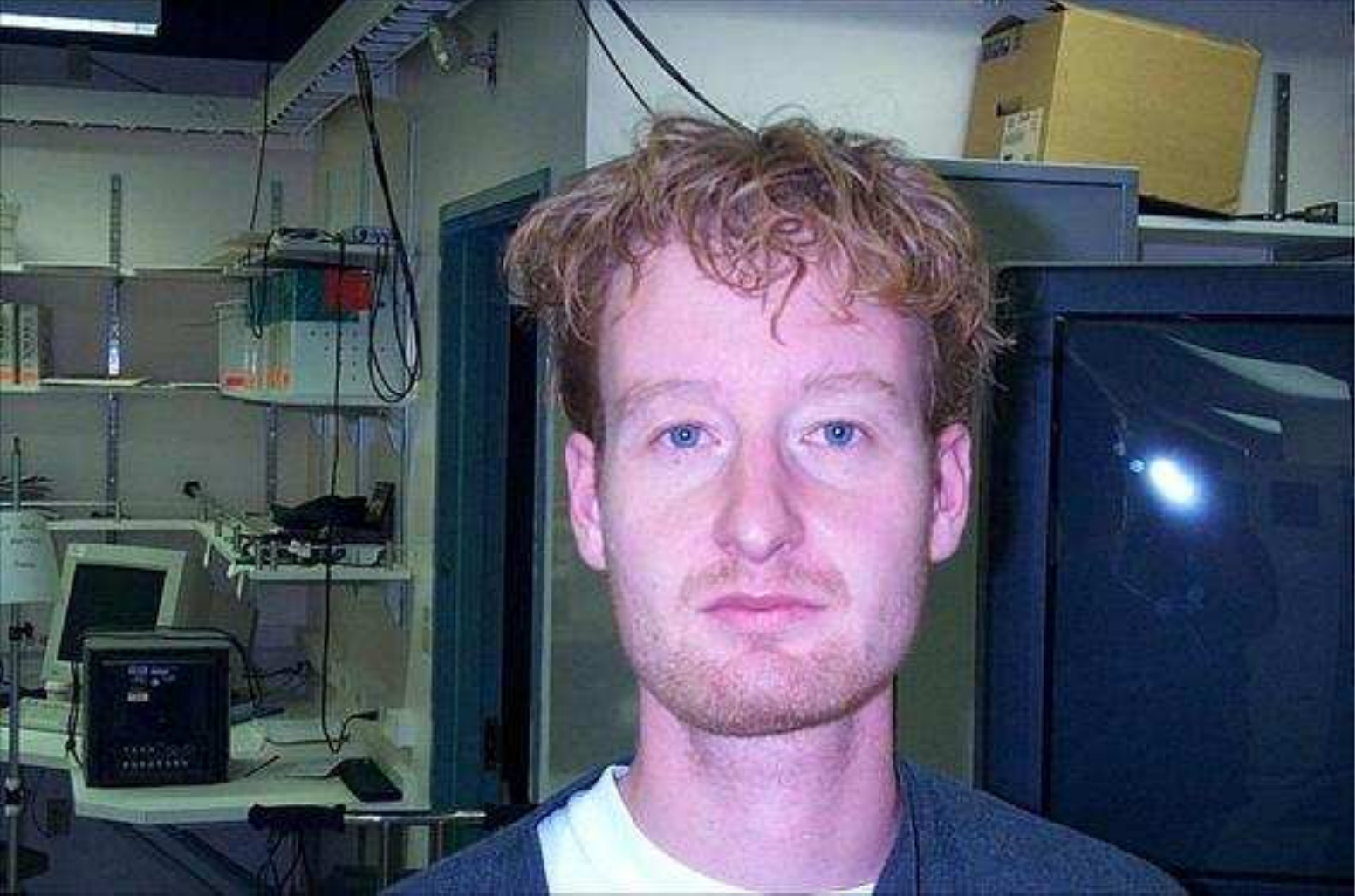} }%
	\subfigure[]{
    \label{fig:example_patch_pixel:2}
	\includegraphics[width=0.24\textwidth]{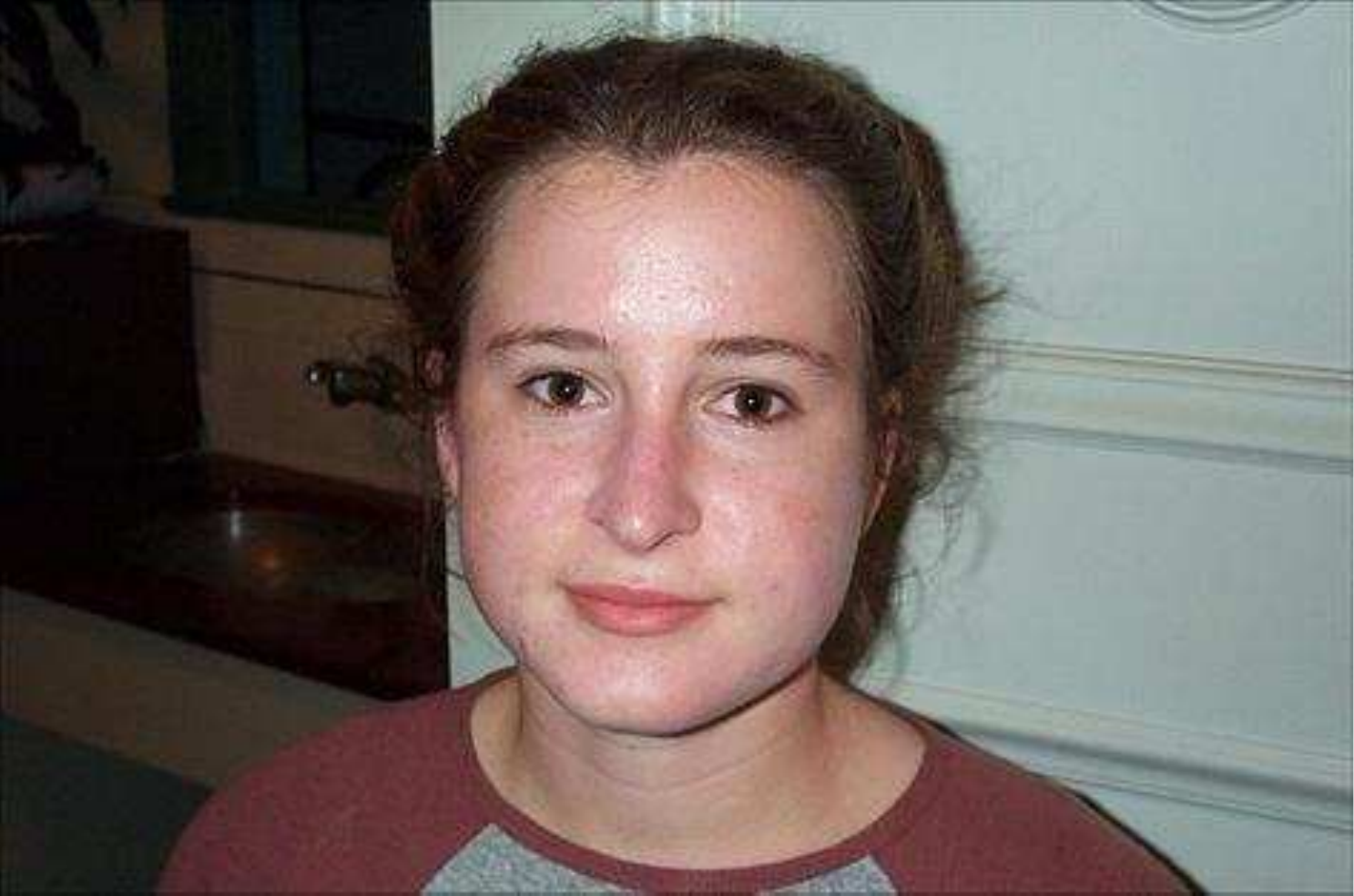} }%
	\subfigure[]{
    \label{fig:example_patch_pixel:3}
	\includegraphics[width=0.24\textwidth]{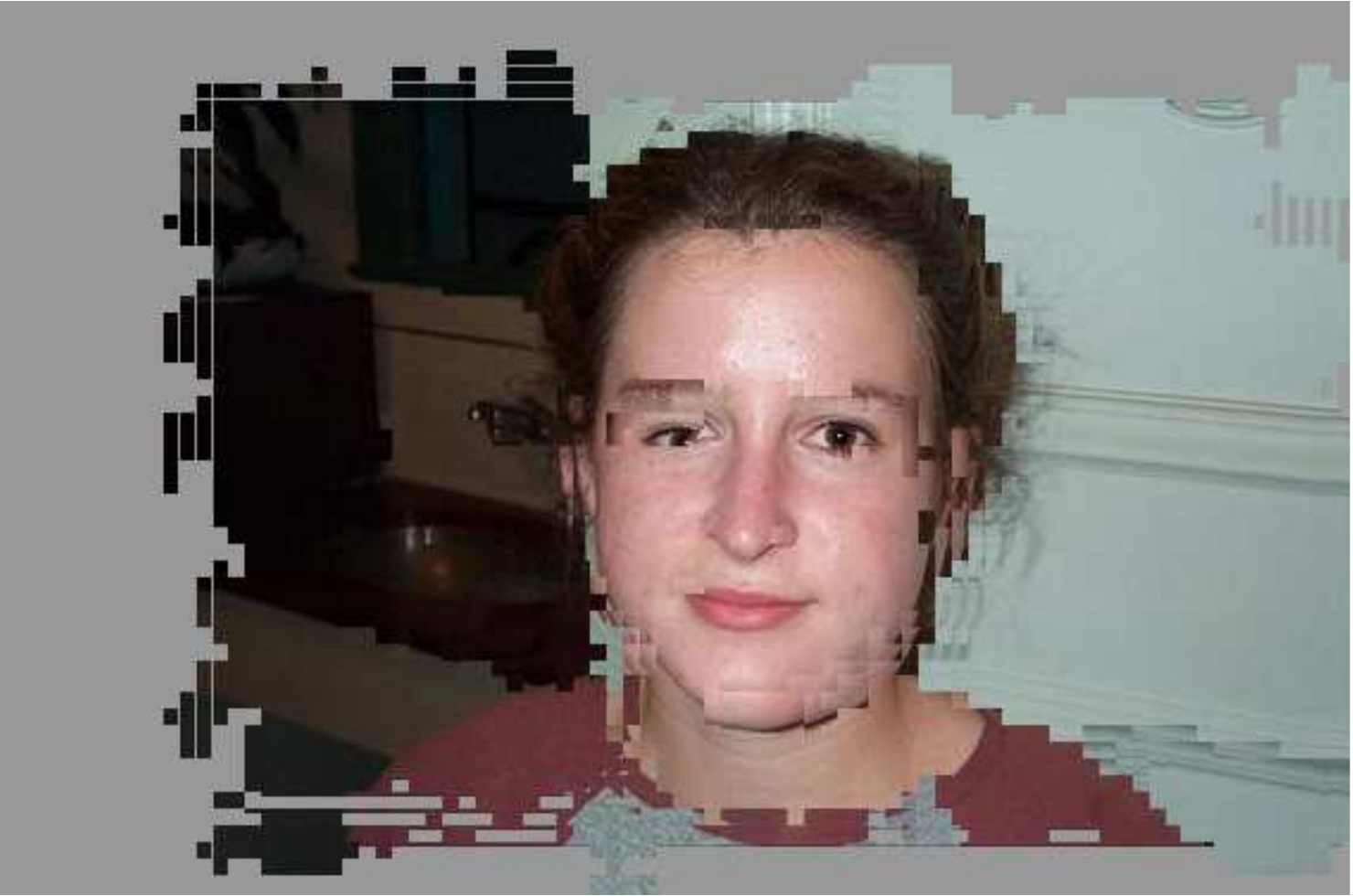} }%
	\subfigure[]{
    \label{fig:example_patch_pixel:4}
	\includegraphics[width=0.24\textwidth]{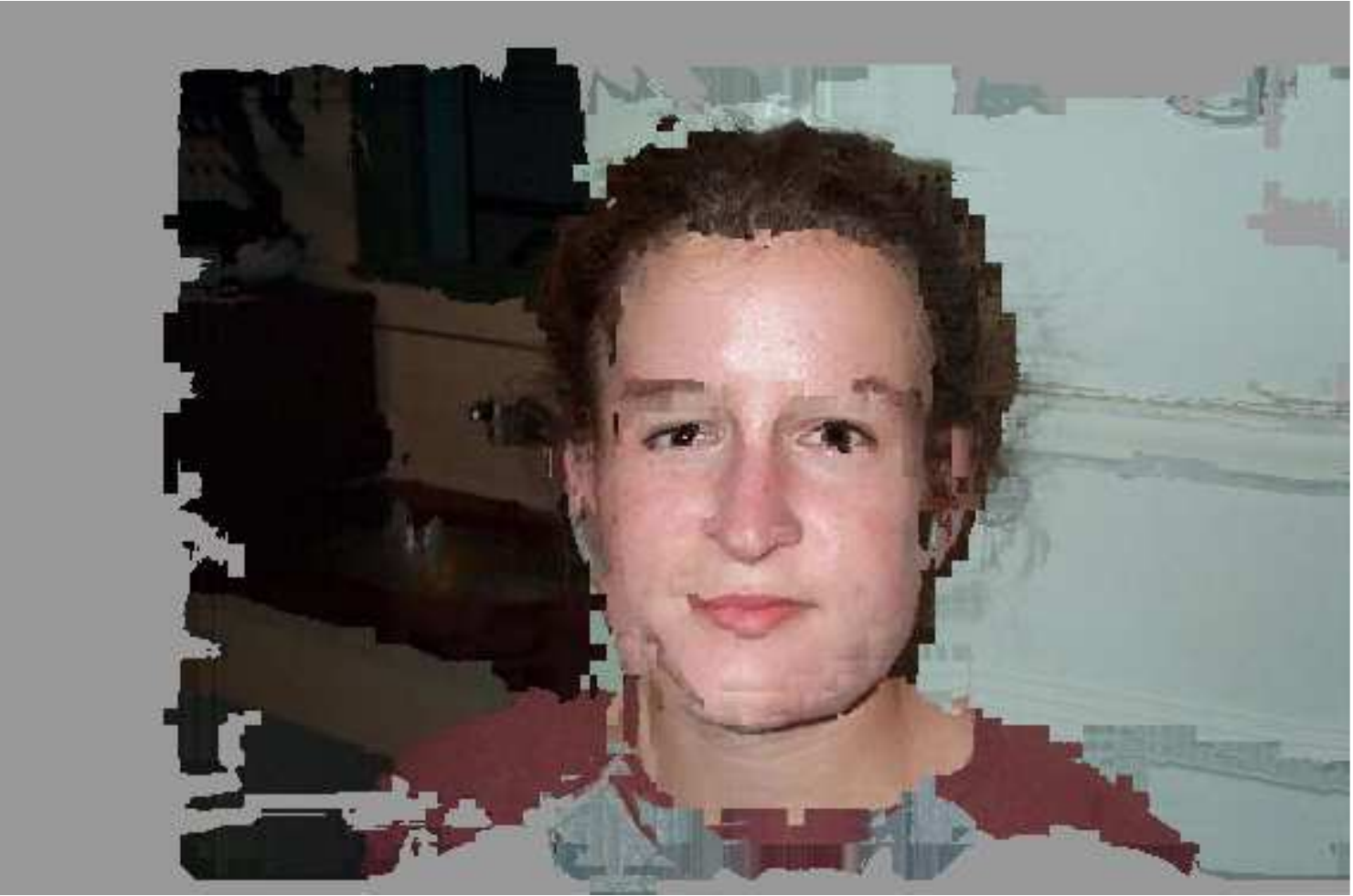} }%
\vspace{\vspacedistmid}
     \caption{Patch-level matching results vs.\ pixel-level matching results of our method.
     \subref{fig:example_patch_pixel:1} and \subref{fig:example_patch_pixel:2}
     are the test image and exemplar image respectively.
          Images \subref{fig:example_patch_pixel:3} and
          \subref{fig:example_patch_pixel:4} are the patch-level result
          and the pixel-level matching result of our method, respectively.
          We can see that the pixel-level matching provides refined visual results.
     }
\label{fig:example_patch_pixel}
\vspace{\vspacedist}
\end{figure*}

Figure \ref{fig:our_framework} shows our matching pipeline.
Our matching process starts from the grid-cell layer matching.
At this layer, the matching cost and geometric regularization
are considered for the pyramid node of different spatial extents.
Matching results of the grid-cell layer guide the patch-level matching.
In other words, results of the grid-cell layer offer
reliable initial correspondences for the patch-level matching
as generally larger spatial supports provide better robustness to image variations.
At the patch level layer, we estimate the correspondences between the patch nodes of image pair.
Guided  by the grid-cell matching results, in our framework, the patch-level matching can already
achieve high matching accuracy and efficiency.
In \cite{kim2013deformable,liu2011sift} the authors sub-sample pixels to reduce
the computation cost, which may lead to suboptimal solutions.
In contrast, we do not need to sub-sample pixels because the patch layer matching in our framework
is is extremely fast, which is one of the major advantages of our method.
At the pixel-level layer, the pixel matching refines the results of the patch-layer matching.
Figure \ref{fig:example_patch_pixel} shows the comparison of matching results from
the pixel layer and the patch layer.
We can see that the pixel layer matching provides finer visual results with heavier computation.

\paragraph{Matching objective}
For the grid-cell layer,
$\bq_i$ denotes the center coordinate of patches within the cell node $i$.
Let $\bt_i = (u_i,v_i)$ be the translation of node $i$ from the test image to the exemplar image.
By minimizing the energy function \eqref{EQ:objective},
we obtain the optimal translation of each node in the grid-cell layer.
The objective function is defined as:
\begin{equation}
\begin{aligned}
\label{EQ:objective}
 E(\bt) = \sum_{i} D_i(\bt_i) + \alpha \sum_{i,j\in \calN} V_{ij}(\bt_i,\bt_j),
\end{aligned}
\end{equation}
where $D_i$ is the data term; $V_{i,j}$ is the smoothness term; and $\alpha$ is the constant weight.
$\calN$ represents node pairs linked by graph edges
between the neighboring nodes and the parent-child nodes in the spatial pyramid.
In the above equation, the data term $ D_i ( \bt_i ) $ is defined as:
\begin{equation}
\begin{aligned}
\label{EQ:dataterm}
D_i(\bt_i) = \frac{1}{z}  \min(\norm{\bff_t(\bq_i) -
 \bff_e(\bq_i + \bt_i)}_1,\lambda ),
\end{aligned}
\end{equation}
where $z$ is the total number of patches included in the grid-cell node $i$.
$\bff_t(\bq_i)$ is the cell node feature of the test image centered at coordinate $\bq_i$ and
$\bff_e(\bq_i+\bt_i)$ is the cell node feature of the exemplar image under certain translation $\bt_i$.
$D_i( \bt_i )$  measures the similarity  between node $i$ and
the corresponding node in the exemplar image  according to translation $\bt_i$.
Here
$\lambda$ is a truncated threshold of feature distance.
We set it to the mean distance of pairwise pixels between image pairs.
$ \norm{ \cdot}_1$ is the $ L_1 $ norm.

Second,
the smoothness term is defined as:
\begin{equation}
\begin{aligned}
\label{EQ:smooth}
V_{ij}(\bt_i,\bt_j) = \min(\norm{\bt_i - \bt_j}_1,\gamma),
\end{aligned}
\end{equation}
which penalizes the matching location discrepancies among the neighboring nodes.
We use loopy belief propagation (BP) to find the optimal correspondence of each node.
Although BP is not guaranteed to converge, it has been applied with much experimental success \cite{Ihler05loopybelief}.
In our objective function, truncated $L_1$ norms are used for both the data term and the smoothness term.
The smoothness term accounts for matching outliers.
As in \cite{Felzenszwalb2006Effi,kim2013deformable},
for BP, we use a generalized distance transform technique such that
the computation cost of message passing between nodes is reduced.

 For the patch-level layer, each patch links to the parent node in the grid-cell layer.
 $\bq_i $ denotes the center coordinate of each patch in the patch-level layer.
 The patch's optimal translation can be obtained by:
 \begin{equation}
\begin{aligned}
\label{EQ:patchlayer}
D_i(\bt) &= \min(\norm{\bff_t(\bq_i) - \bff_e(\bq_i + \bt)}_1,\lambda ),\\
\bt_i &= \argmin_{\bt}(D_i(\bt) +
\alpha V_{ip}(\bt,\bt_p) )
.
\end{aligned}
\end{equation}
Here
$\bt_i$ and $\bt_p$ are the patch $i$'s optimal translation
and  its parent cell node's optimal translation respectively.
$\bff_t$ and $\bff_e$ denote the patch features in the test image and exemplar image, respectively.
Note that the results from the grid-cell layer provide reliable initial correspondences
for the patches at the patch layer.

For the pixel-level layer, each pixel links to the parent patch node.
Guided by the patch layer solution, pixel layer correspondences can be
estimated efficiently and accurately.
$\bq_i $ denotes the pixel $ i$'s coordinate in the pixel-level layer.
The $D_i(\bt)$ in pixel $i$'s optimal translation is defined as
 \begin{equation}
\begin{aligned}
\label{EQ:pixellayer}
D_i(\bt) &= \min(\norm{\bs_t(\bq_i) - \bs_e(\bq_i + \bt)}_1,\lambda ),\\
\bt_i &= \argmin_{\bt}(D_i(\bt) +
\alpha V_{ip}(\bt,\bt_p) )
.
\end{aligned}
\end{equation}
Here
$\bt_i$ and $\bt_p$ are the pixel $i$'s optimal translation
and  its parent patch node's optimal translation respectively.
$\bs_t$ and $\bs_e$ denote the pixel features in the test
image and exemplar image respectively.

\paragraph{Discussion}
Our method  improves the matching accuracy and
substantially reduces the computation time compared to the recent state-of-the-art method
\cite{kim2013deformable}.
In \cite{kim2013deformable}, they have used sparse descriptors sampling
to reduce the computational  time,
which may cause  loss of key characteristics in the nodes of the matching graph.

The grid-cell layer of our matching model is built upon the
patch layer features which cover the whole
image area without using a sparse sampling process.
The pixel-layer features obtained by an unsupervised learning algorithm
appear to be discriminative for resolving matching ambiguity between classes.
By using a pooling operation on the pixel features, we significantly reduce the
number of patch-layer features and the possible translations,
while enhancing the robustness to severe image visual variations.
{\em Matching at the patch-level layer is the core of our algorithm.}
At the patch layer, image pixels within the same patch share
the same optimal translations.
Our experiment results show that the patch-layer matching results
outperform the state-of-the-art methods with a much faster computation speed.
The pixel-level matching further refines the  matching accuracy.
Thus the pixel-level matching procedure  can be considered optional.
If the test speed is on a budget, one does not have to perform the pixel-level matching.

As can be seen from Figure \ref{fig:example_patch_pixel}, the pixel-level matching
output (Figure \ref{fig:example_patch_pixel:4})  retains finer details in object edges, compared to the
patch-level matching result (Figure \ref{fig:example_patch_pixel:3}).
Our pixel and patch feature encoding schemes
allow us to reduce the computation while improving the
matching results.
Experiment results in the next section demonstrate the advantages of our method.

\section{Experiments}\label{Experiments}
We conduct   experiments to evaluate the matching quality
(Section \ref{Comp-with-state-of-the-art}) and
to analyse the performance impact of several different
elements in the feature learning framework (Section \ref{feature_learning}).

In Section \ref{Comp-with-state-of-the-art},
we test our method on  three benchmark vision
datasets: the Caltech-101 dataset,  the
LabelMe Outdoor (LMO) dataset \cite{liu2009nonparametric},
and a subset of the Pascal dataset.

We also apply our method to semantic segmentation.
We compare our method with state-of-the-art
dense pixel matching methods, namely,
the deformable spatial pyramid (DSP) approach (single-scale) \cite{kim2013deformable},
SIFT Flow (SF) \cite{liu2011sift},
and  coherency sensitive hashing (CSH) \cite{korman2011coherency}.
Note that DSP has achieved the previously best results on dense pixel matching.
For the DSP, SF and CSH methods, we use the code provided
by their authors.

In Section \ref{feature_learning},
we present a detailed analysis of the impact of parameter settings
in feature learning, including:
(a) the choice of the unsupervised feature learning algorithm,
(b) the impact of the dictionary size,
(c) the impact of the training data size,
(d) different configurations in the patch feature extraction process.

We set the parameters of the compared methods to the values that were suggested in
the original papers.
In all of our experiments, we use 3-level pyramid in the grid-cell layer.
The parameters of our method for all experiments are fixed to
$\alpha = 0.02$, $\gamma = 0.5$.

A {\em universal} dictionary is learned from image patches extracted from
200 Background\_Google class images in the Caltech-101 dataset.
Note that `Background\_Google' contains mainly
natural images which are  irrelevant to the test images in our experiments.
The dictionary is learned before the matching process.
Once the dictionary is learned,
it is used to encode all the test images.
We use K-means dictionary learning and K-means triangle (KT) encoding for our method.
The dictionary size is set to 100. Clearly, the length of feature vectors at the pixel-level layer and patch-level layer
are equal to the dictionary size.

Then pixel-layer features are computed at each pixel of test images.
More specifically, we perform encoding on an image region
around each pixel centroid using the learned dictionary
to form the feature vector of that pixel.
So each pixel feature is extracted from an $11 \times 11$-pixels image region centered at that pixel.
Patch features are then calculated by max-pooling pixel features within each non-overlapping patch of $7 \times 7$ pixels.
The grid-cell layer is constructed by a 3-level pyramid as shown in Figure \ref{fig:grid_cell_pyramid}.

\comment{
\subsection{Evaluation metrics}

 Following \cite{kim2013deformable}, we use the label transfer accuracy (LT-ACC)
 \cite{liu2009nonparametric}, intersection over union (IOU) metric \cite{everingham2014pascal}
 and the localization error (LOC-ERR) \cite{kim2013deformable}
 to measure the quality of dense image matching.

 For each image pair (a test image and an exemplar image),
 we find the pixel correspondences between them
 using a matching algorithm and transfer
 the annotated class labels of the exemplar image pixels to the test image pixels.

The LT-ACC criterion measures how many pixels in the test image
have been correctly labelled by the matching algorithm.
On the Caltech-101 dataset, each image is divided into the foreground and background.
The IOU metric reflects the matching quality for separating the foreground pixels
from the background.
As for Caltech-101, since the ground-truth pixel correspondence information is not available,
we use the LOC-ERR metric to evaluate the distortion of correspondence pixel
locations with respect to the object bounding box.

Mathematically,
the LT-ACC metric is computed as
\begin{equation}
r = \frac{1}{\sum_{i}m_i}\sum_{i}\sum_{p\in \Lambda_i}1(o(p)=a(p), \, a(p)>0),
\end{equation}
where for pixel $p$ in image $ i$, the ground-truth annotation is $a(p)$ and
matching output is $o(p)$; for unlabeled pixels $a(p)=0$. Notation $\Lambda_i$
is the image lattice for image $ i$,
and $ m_i = \sum_{p \in \Lambda_{i} } 1(a(p)>0) $
is the number of labeled pixels for image $ i $ \cite{liu2009nonparametric}.
Here $ 1( \cdot )  $ outputs $ 1 $ if the condition is true; otherwise 0.

To define the LOC-ERR of corresponding
pixel positions, we first designate each image's pixel coordinates
using its ground-truth object bounding box.
Then, the localization error of two matched pixels is defined as:
$ e = 0.5(|x_1-x_2| +|y_1-y_2|)$, where $(x_1,y_1)$ is the pixel
coordinate of the first image and $(x_2,y_2)$ is its
corresponding location in the second image \cite{kim2013deformable}.

The intersection-over-union (IOU) segmentation measure is now the standard metric for segmentation \cite{everingham2014pascal}.

}

\subsection{Evaluation metrics}

 Following \cite{kim2013deformable}, we use the label transfer accuracy (LT-ACC)
 \cite{liu2009nonparametric}, intersection over union (IOU) metric \cite{everingham2014pascal}
 and the localization error (LOC-ERR) \cite{kim2013deformable}
 to measure the quality of dense image matching.

 For each image pair (a test image and an exemplar image),
 we find the pixel correspondences between them
 using a matching algorithm and transfer
 the annotated class labels of the exemplar image pixels to the test image pixels.

The LT-ACC criterion measures how many pixels in the test image
have been correctly labelled by the matching algorithm.
On the Caltech-101 dataset, each image is divided into the foreground and background pixels.
The IOU metric reflects the matching quality for separating the foreground pixels
from the background.
As for Caltech-101, since the ground-truth pixel correspondence information is not available,
we use the LOC-ERR metric to evaluate the distortion of correspondence pixel
locations with respect to the object bounding box.

Mathematically,
the LT-ACC metric is computed as
\begin{equation}
r = \frac{1}{\sum_{i}m_i}\sum_{i}\sum_{p\in \Lambda_i}1(o(p)=a(p), \, a(p)>0),
\end{equation}
where for pixel $p$ in image $ i$, the ground-truth annotation is $a(p)$ and
matching output is $o(p)$; for unlabeled pixels $a(p)=0$. Notation $\Lambda_i$
is the image lattice for image $ i$,
and $ m_i = \sum_{p \in \Lambda_{i} } 1(a(p)>0) $
is the number of labeled pixels for image $ i $ \cite{liu2009nonparametric}.
Here $ 1( \cdot )  $ outputs $ 1 $ if the condition is true; otherwise 0.

To define the LOC-ERR of corresponding
pixel positions, we first designate each image's pixel coordinates
using its ground-truth object bounding box.
Pixel coordinates are normalized with respect to the box's position and size.
Then, the localization error of two matched pixels is defined as:
$ e = 0.5(|x_1-x_2| +|y_1-y_2|)$, where $(x_1,y_1)$ is the pixel
coordinate of the first image and $(x_2,y_2)$ is its
corresponding location in the second image \cite{kim2013deformable}.

The intersection over union (IOU) segmentation measure is used to assess per-class accuracy
on the intersection of the predicted segmentation and the ground truth, normalized by the union.
Formally, it
is defined as
\begin{equation}
  u  = \frac{true\ pos.}{true\ pos. + false\ pos. + false\ neg.}
\end{equation}
IOU is now the standard metric for segmentation \cite{everingham2014pascal}.

\subsection{Comparison with state-of-the-art dense matching methods}\label{Comp-with-state-of-the-art}

In this section, we compare our method against state-of-the-art dense matching methods
to examine the matching quality on object matching and scene segmentation.
Detail about our method is described in Section \ref{pixel_patch_feat}.

\paragraph{Comparison with different features and matching methods} %

For this experiment, we evaluate the matching performance by
using different features in different matching frameworks.
100 test image pairs are randomly picked from the Caltech-101 dataset.
Each pair of images are chosen from the same class.
The result is shown in Table \ref{tab:caltech_matching}.
It shows that SIFT features in our
multi-layer matching model are not able to achieve the same accuracy level as the learned features.
{This result shows the advantage of learned features over hand-crafted features such as SIFT.}

Meanwhile, we use the learned features to replace the SIFT features in the DSP \cite{kim2013deformable}'s
framework, and use the same test images.
The results show that the SIFT features can obtain better matching accuracy
in DSP\cite{kim2013deformable}'s framework.
{ This result shows that the DSP method \cite{kim2013deformable} is not able to take advantage of
learned features, while our matching framework is tailored to the unsupervised features learning technique}.

\begin{table*} \center
\vspace{\vspacedistfront}
\resizebox{1\linewidth}{!}
{
\begin{tabular}{ r | c |  c |  c   | c | c  | c | c  | c }
\hline
{{{framework}}} &\multicolumn{4}{c|}{{Ours}} &\multicolumn{2}{c|}{{DSP\cite{kim2013deformable}}}
	&\multicolumn{1}{c|}{{SIFT Flow\cite{liu2011sift}}}
	&\multicolumn{1}{c}{{CSH\cite{korman2011coherency}}  }  \\
 {feature}&\multicolumn{2}{c|}{ learned feature }  &\multicolumn{2}{c|}{SIFT} & { learned feature }& {SIFT} & &   \\
  {matching level}&{patch level} &{pixel level}&{patch level} &{pixel level} & &  &  &   \\
\hline
     {LT-ACC}   & {0.801} 		&\textbf{0.803} 	& 0.757 &0.759 &0.765 & 0.792 &0.763 & 0669 \\
     {IOU}  	& {0.501}		&\textbf{0.505} 	& 0.449 &0.451 &0.436 & 0.496 &0.479 &  0.365 \\
     {LOC-ERR}  &\textbf{0.323} &0.324 				& 0.409 &0.410 &0.359 & 0.357 &0.351 & 1.002 \\
     \hline
     {Time (sec)}    &\textbf{0.07}  &0.45 			&0.09 &0.76   &0.40 & 0.65 &4.32 & 0.16\\
	\hline
\end{tabular}
}
\caption{Comparison of object matching performance of different methods
		on 100 pairs of images from the Caltech-101 dataset
		in terms of the matching accuracy and speed.
		The best results are shown in bold.}
\label{tab:caltech_matching}
\vspace{\vspacedist}
\end{table*}

The third observation is the computation speeds of compared methods.
{
The CPU time of our method (at the patch level) is about 8 times faster
than that of  DSP\cite{kim2013deformable}
and 50 times faster than  SIFT Flow of Liu et al.\ \cite{liu2011sift}.
}

For the patch-level matching, our method outperforms CSH by
about 13 points in LT-ACC, yet is
twice faster than CSH. Note that CSH is  a noticeably fast  method,  which
exploits hashing to quickly find matching patches between two images.
Our pixel-level matching  further improves the patch layer matching accuracy
and provides better visual matching quality, which is hard to be measured by LT-ACC.
The examples are shown in Figure \ref{fig:example_patch_pixel}.
By using multi-level representations, our proposed matching method  enables the learned features
(obtained by unsupervised learning methods)
to outperform those hand-crafted features (e.g., SIFT features) in dense matching tasks.
The general matching framework may not help features to achieve their best performance.
A suitable matching framework improves the feature performance in the matching task.
We conclude that our matching framework and the unsupervised feature learning pipeline are tightly coupled
to achieve the best performance.

\paragraph{Intra-class image matching}

Now we conduct extensive experiments on the Caltech-101 dataset.
We randomly pick 20 image pairs for each class (2020 image pairs in total).
Each pair of images are chosen from the same class.
The ground-truth annotation for each image is available,
indicating the foreground object and the background.

\begin{table} \center
{
\begin{tabular}{ r | c     c    c  c  }
\hline
\multirow{1}{*}{{{method}}} &\multicolumn{1}{c}{{Ours (patch layer)}} &\multicolumn{1}{c}{{DSP}}
&\multicolumn{1}{c}{{SIFT Flow}} &\multicolumn{1}{c}{{CSH}} \\
\hline
	{LT-ACC}     &\textbf{0.772} & 0.761 &0.741 &0.590\\
	{Time (s)}     &\textbf{0.033} & 0.545 &2.82  &0.163\\
	\hline

\end{tabular}
}
\caption{Intra-class image matching performance on the Caltech-101 dataset.
		The best results are in bold.}
		\label{tab:caltech_matching_2}
\vspace{\vspacedist}
\end{table}

\begin{figure}
\vspace{\vspacedistfront}
\centering
\subfigure[]{
    \label{fig:extensive_caltech:1}
	\includegraphics[width=0.32\textwidth]{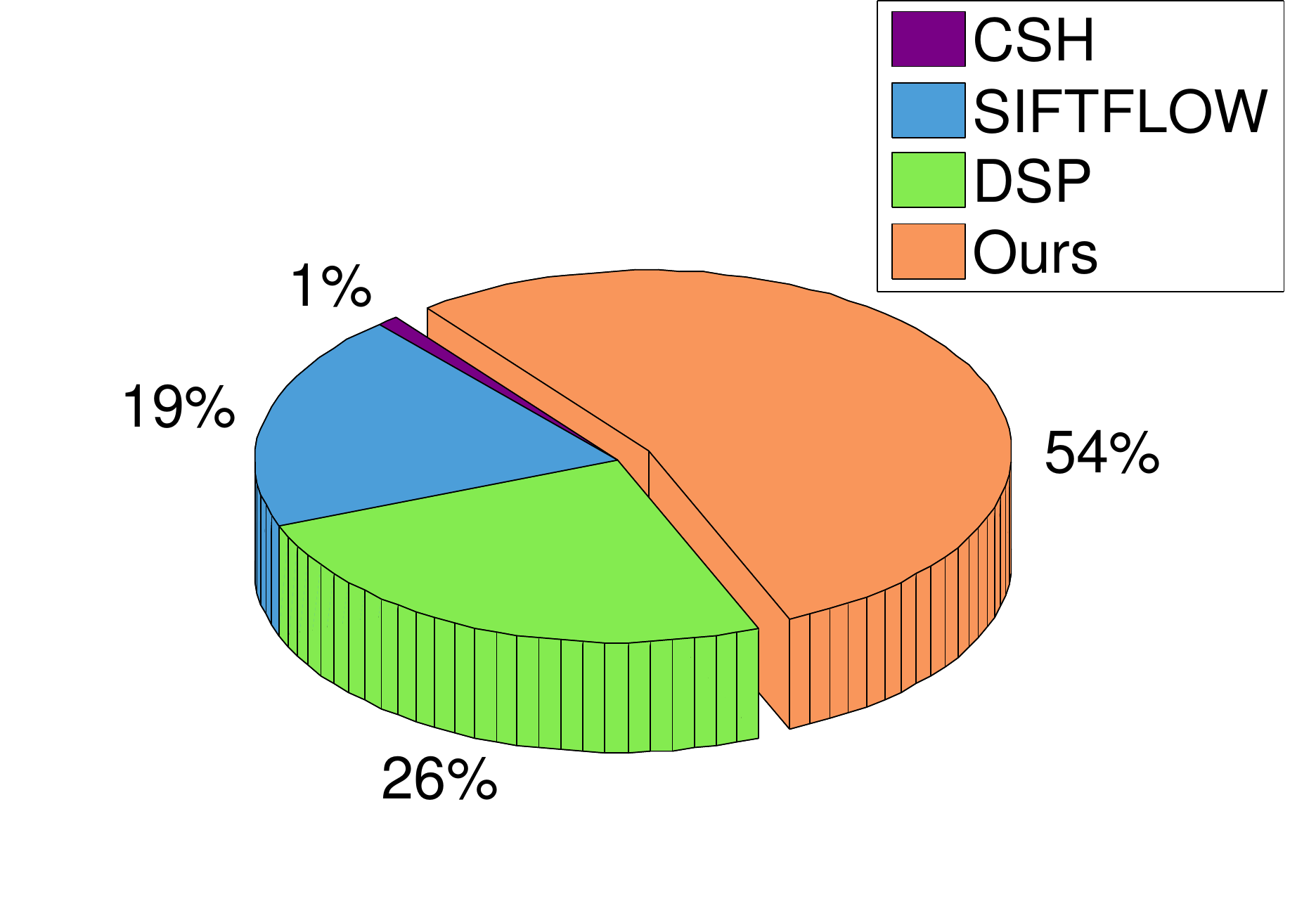} } \\%
	\vspace{\vspacedist}
\subfigure[]{
    \label{fig:extensive_caltech:2}
	\includegraphics[width=0.32\textwidth]{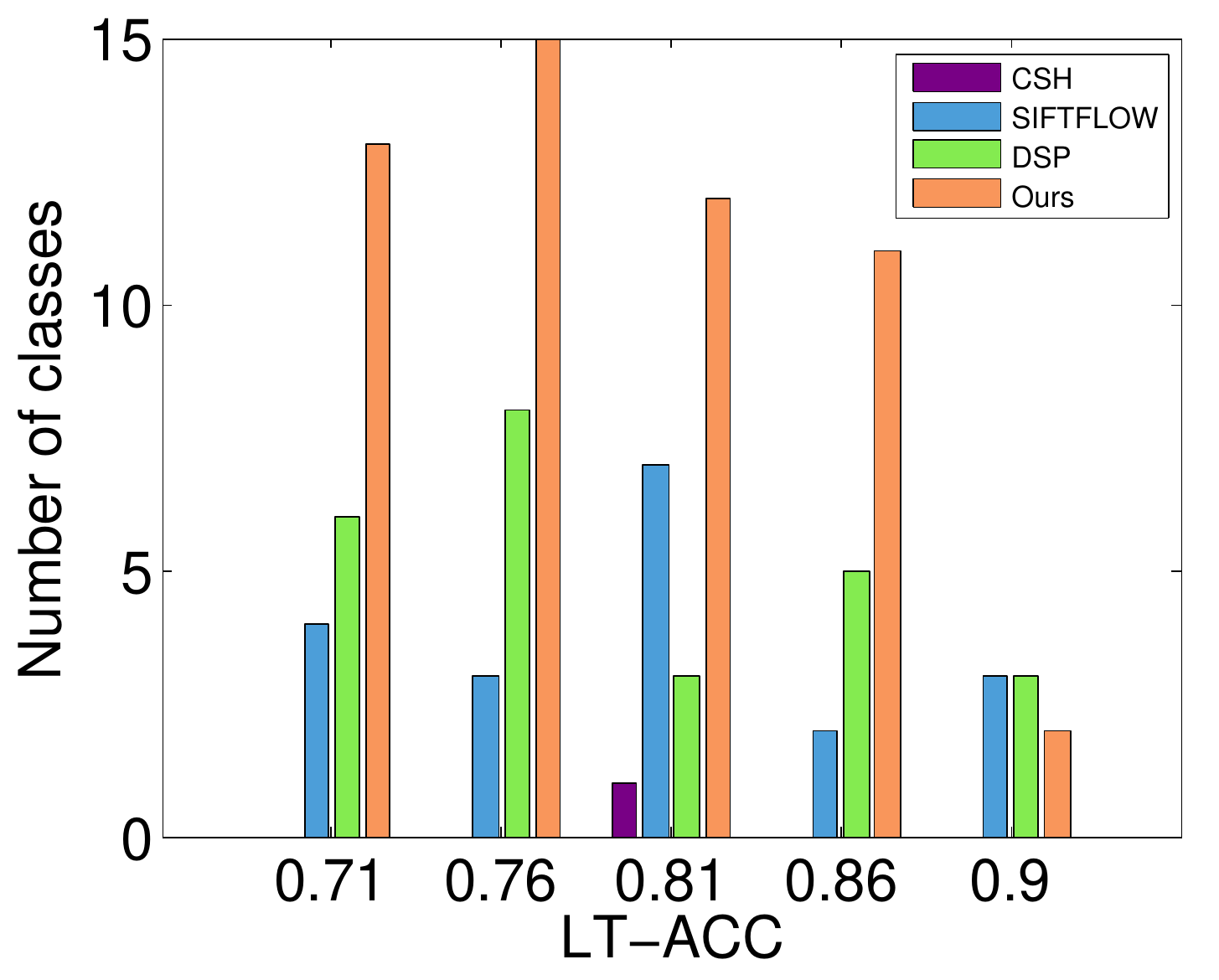} }%
\vspace{\vspacedistmid}
     \caption{
      \subref{fig:extensive_caltech:1} shows the percentage of each
	 method achieving the best performance in all of the 101 classes. Our method achieves the best
	 matching accuracy in 55 classes (54\% of 101 classes).
     \subref{fig:extensive_caltech:2} shows the histogram of
     each methods' achievements
     over matching accuracy (LT-ACC) in all classes.
     }
\label{fig:extensive_caltech}
\vspace{\vspacedist}
\end{figure}

Table \ref{tab:caltech_matching_2} shows the matching accuracy and
CPU time of our method against those state-of-the-art methods.
As can be seen, our method achieves the highest label transfer accuracy
and is more than 10 times faster than the DSP in the matching process.
In this experiment, there are only two labels for each test image
and the intra-class variability is very large. It is hard to achieve
improvement of 0.1.

We can find that the intra-class variability
differs by classes.
Such as in the `Face' class, objects in images are similar.
The highest matching accuracy for the `Face' class can reach  0.952 (obtained by the SIFT Flow method).
However, in the `Beaver' class, images vary much more than the `Face' class and are hard to match.
Therefore, the best accuracy among four compared methods of the `Beaver' class
is lower than the `Face' class, as expected.
In the `Beaver' class, our method outperforms other methods
and obtains a matching accuracy of 0.687.

As the intra-class variability is hard to measure, we use the best matching accuracy
to reflect the variability of each class.
Higher matching accuracy means smaller intra-class variability.
Figure \ref{fig:extensive_caltech:2} shows the histogram of
best matching accuracy in all classes.
For each bin, we group the data by matching methods.

Figure \ref{fig:extensive_caltech:2} shows that our method outperforms other methods in most cases.
SIFT based methods achieve better results in those `easy' high-accuracy classes such as `Face'.
Our method can achieve better
matching results for large intra-class variability classes.
SIFT based methods can well handle the similar appearance objects matching.
This conclusion is consistent with the case of sparse point matching applications.
Our proposed method is more suitable to
handle the matching problem of {\em large intra-class variability} than compared methods.

The pie chart in Figure \ref{fig:extensive_caltech:1} shows the percentage of each
	 method achieving the best accuracy in the 101 classes. Our method achieves the best
   matching accuracy in 55  out of 101 classes.
Our method outperforms the compared methods by a large margin.

\newcommand{\figobjectmatchingw}{0.0995\textwidth}
\newcommand{\figobjectmatchingsw}{0.075\textwidth}
\begin{figure*}
\vspace{\vspacedistfront}
\centering
{{Input\ \ \ \ \ \ \ GT label \ \ \ \ \ Ours \ \ \ \ \ \ \ DSP\ \ \ \ \ \ \ \ \ \ SIFT Flow\ \ \ \ CSH}} \\
	 \includegraphics[width=\figobjectmatchingw, height=\figobjectmatchingsw]{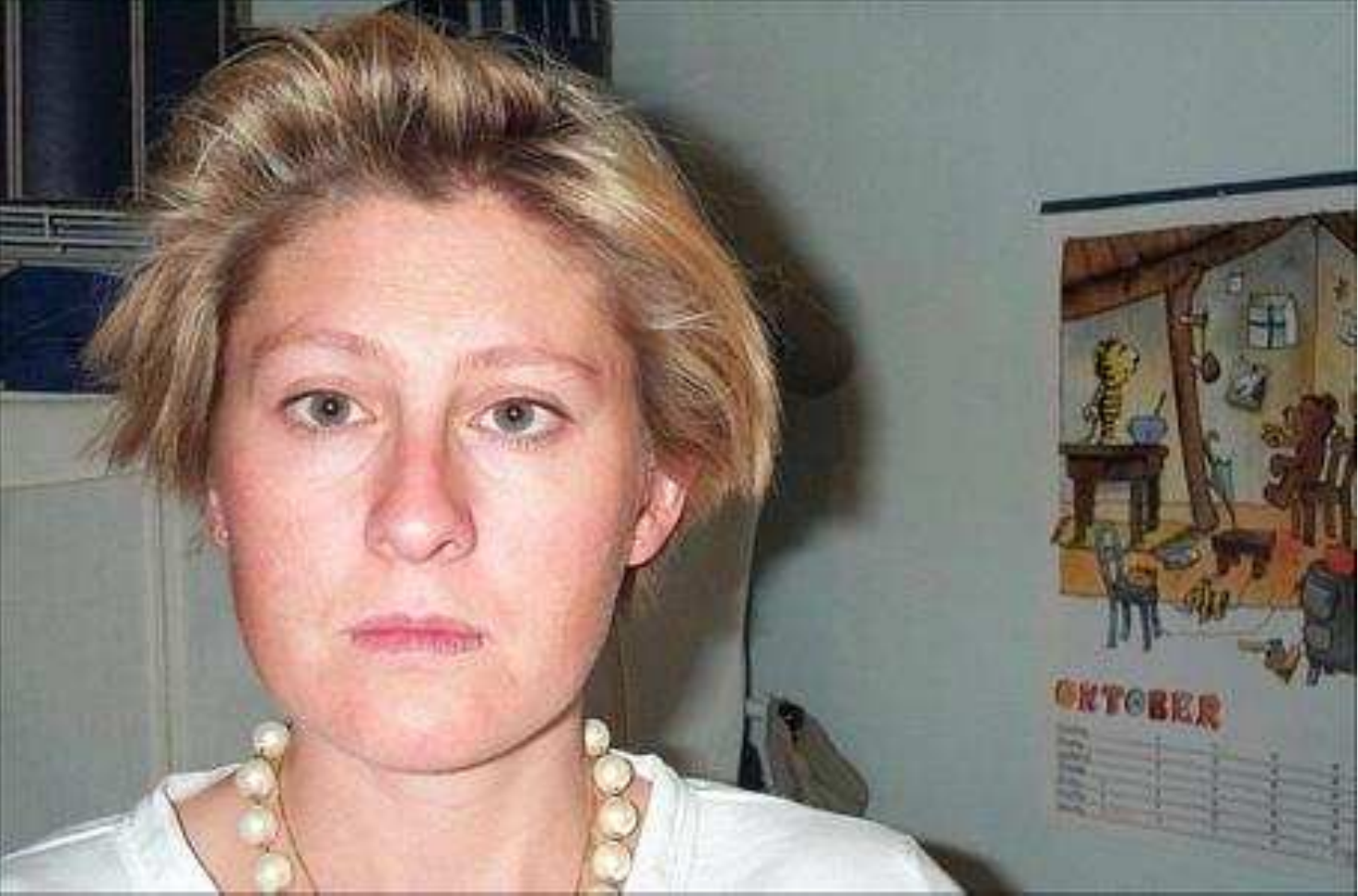}
	 \includegraphics[width=\figobjectmatchingw, height=\figobjectmatchingsw]{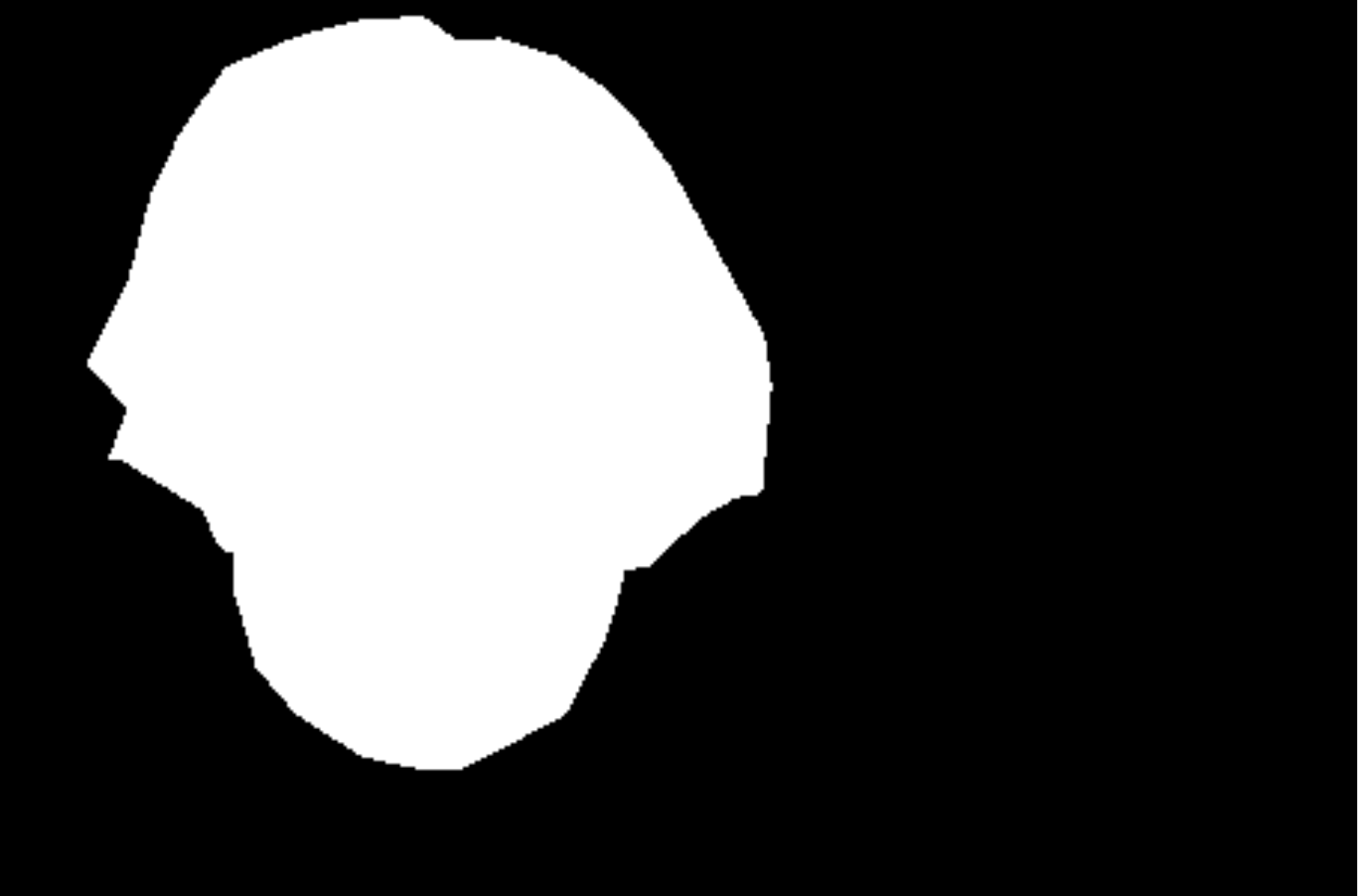}
	 \includegraphics[width=\figobjectmatchingw, height=\figobjectmatchingsw]{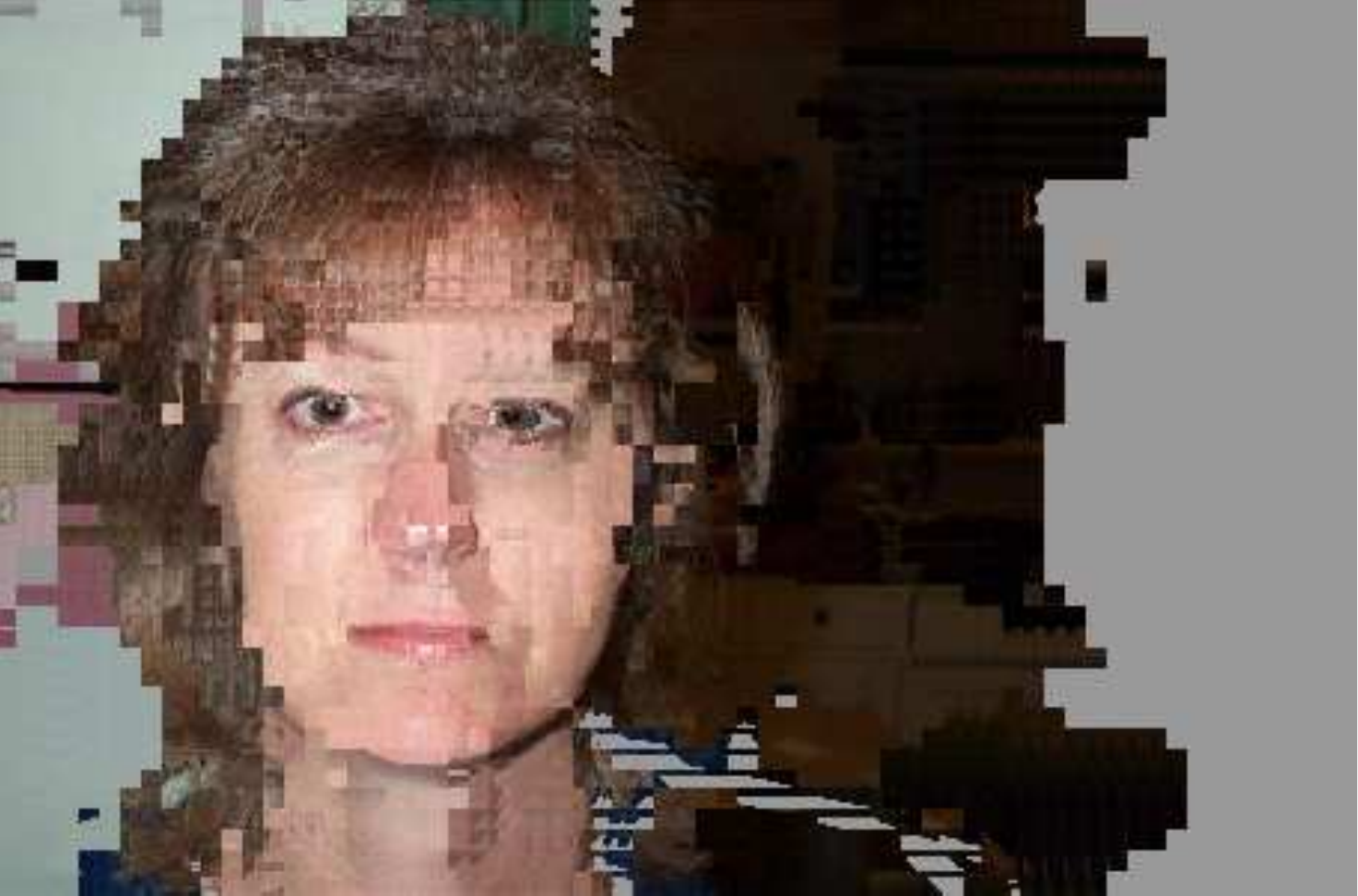}
     \includegraphics[width=\figobjectmatchingw, height=\figobjectmatchingsw]{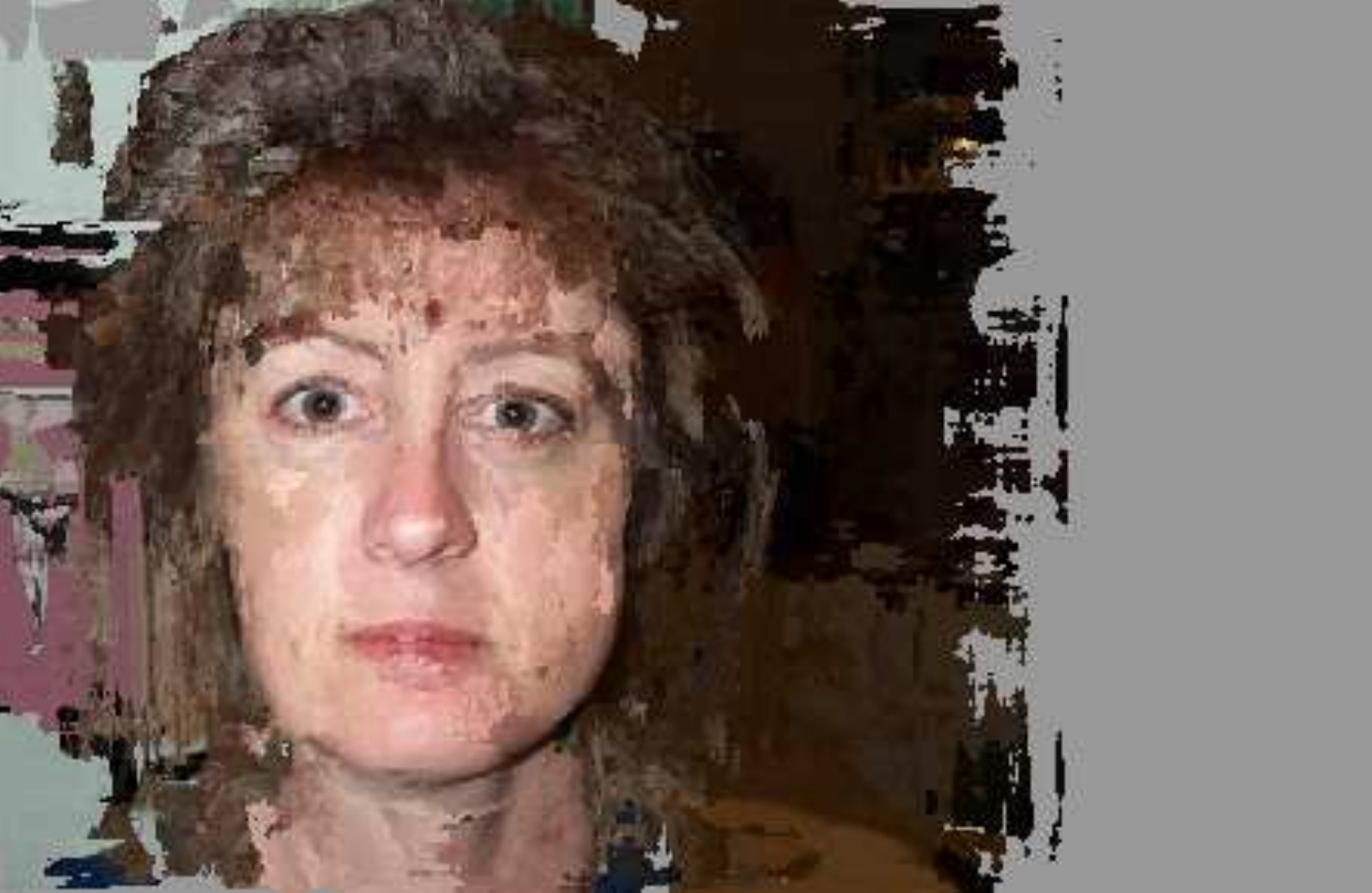}
    \includegraphics[width=\figobjectmatchingw, height=\figobjectmatchingsw]{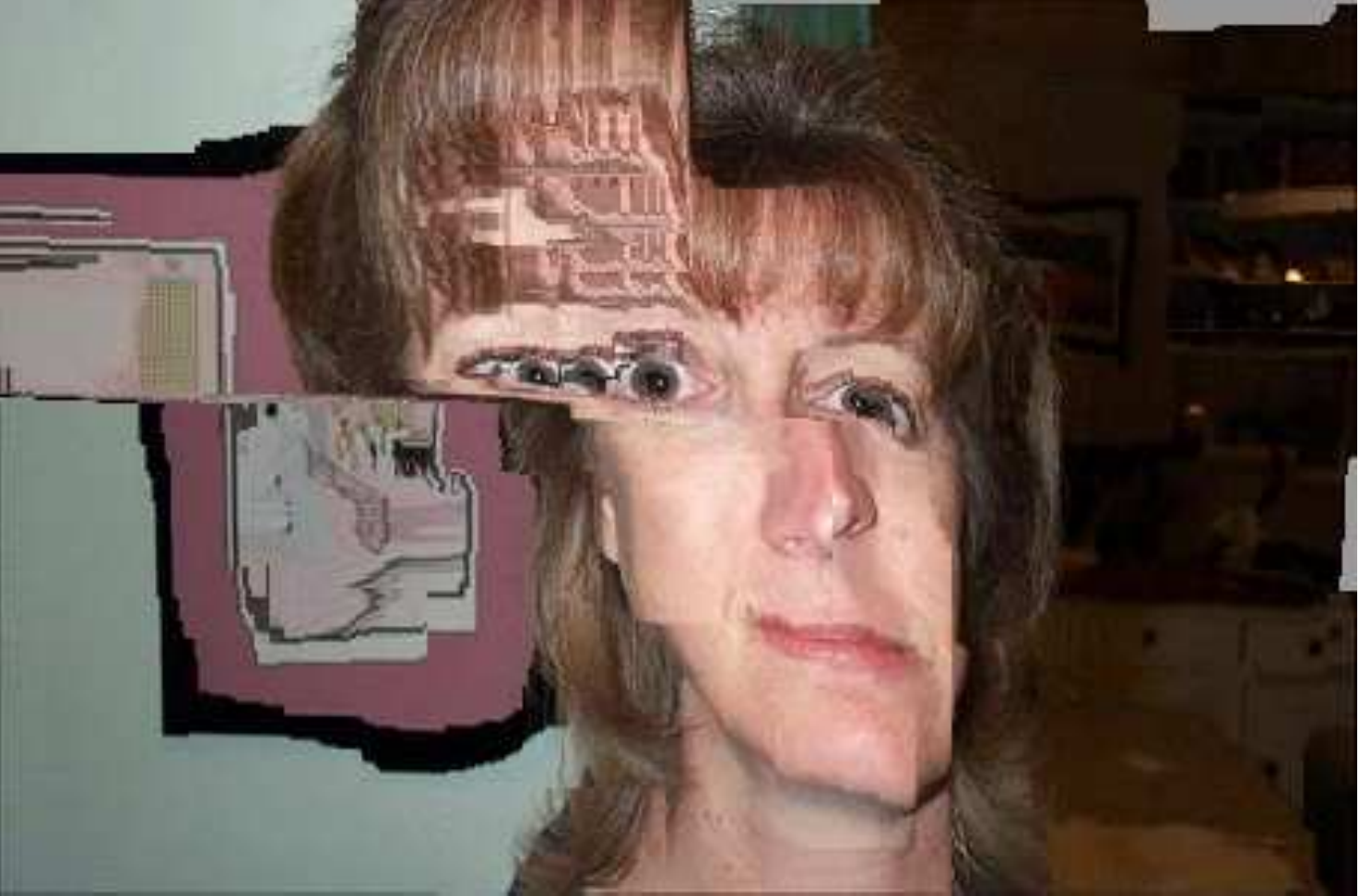}
     \includegraphics[width=\figobjectmatchingw, height=\figobjectmatchingsw]{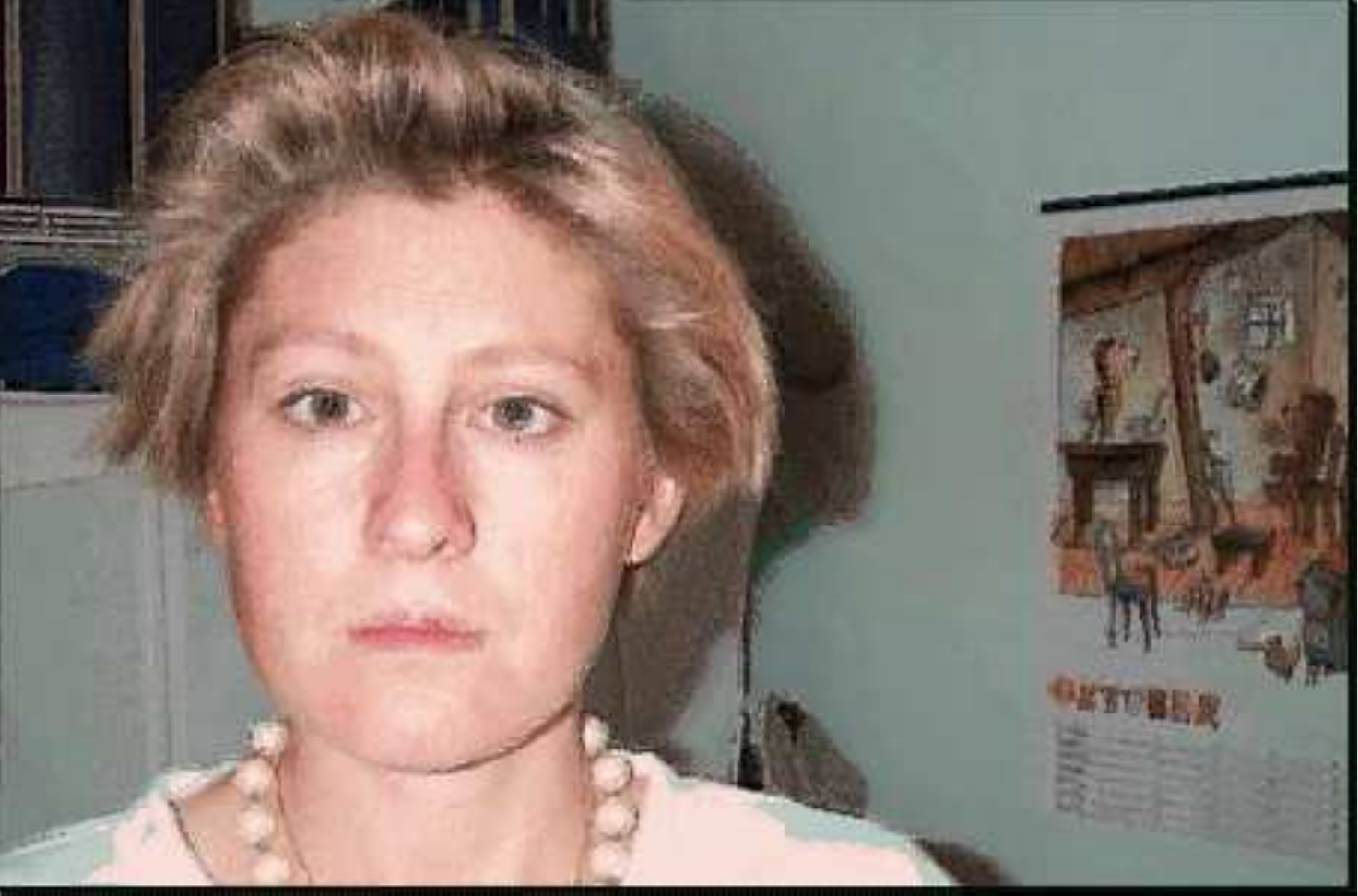} \\

 	 \includegraphics[width=\figobjectmatchingw, height=\figobjectmatchingsw]{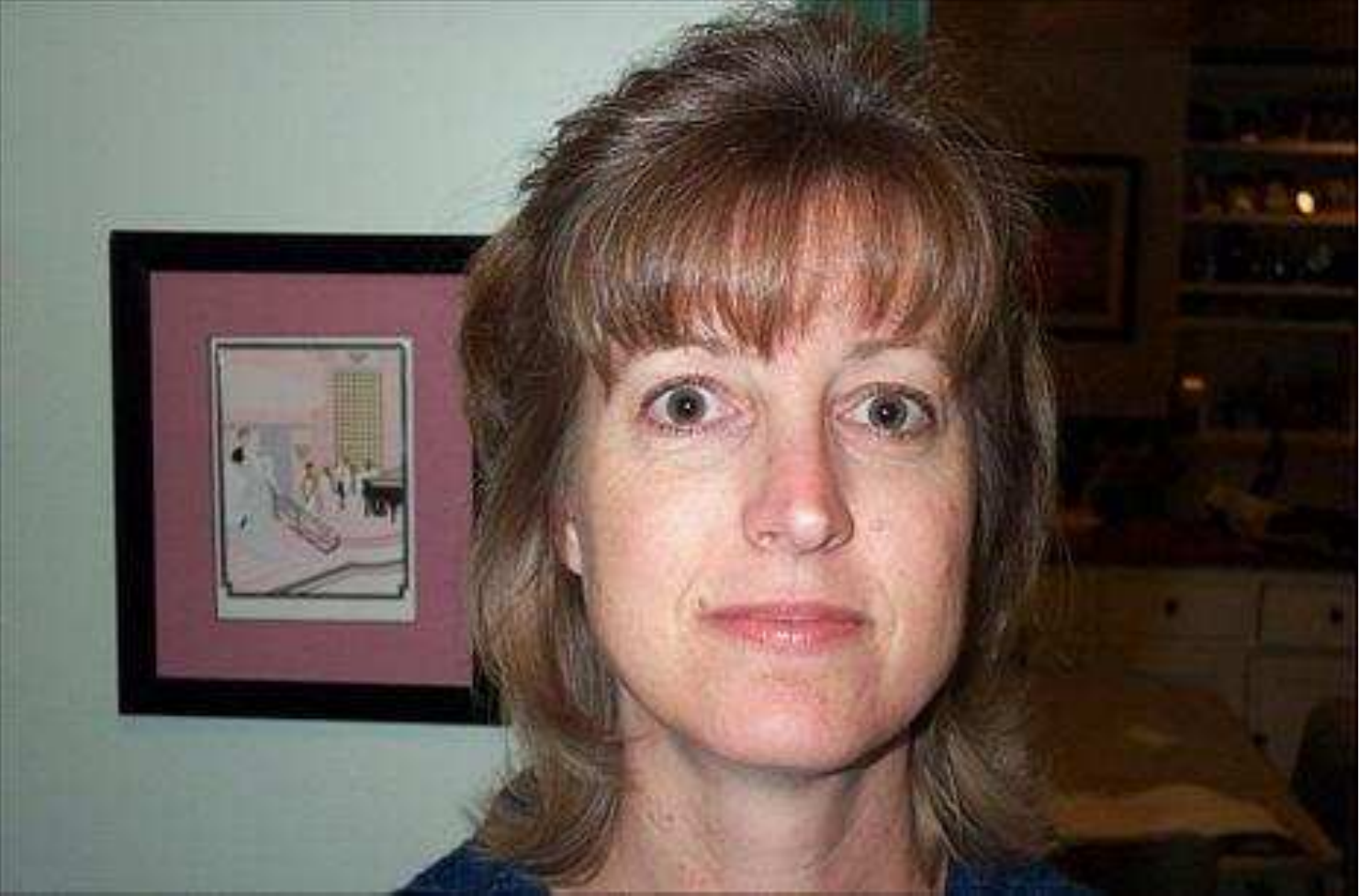}
	 \includegraphics[width=\figobjectmatchingw, height=\figobjectmatchingsw]{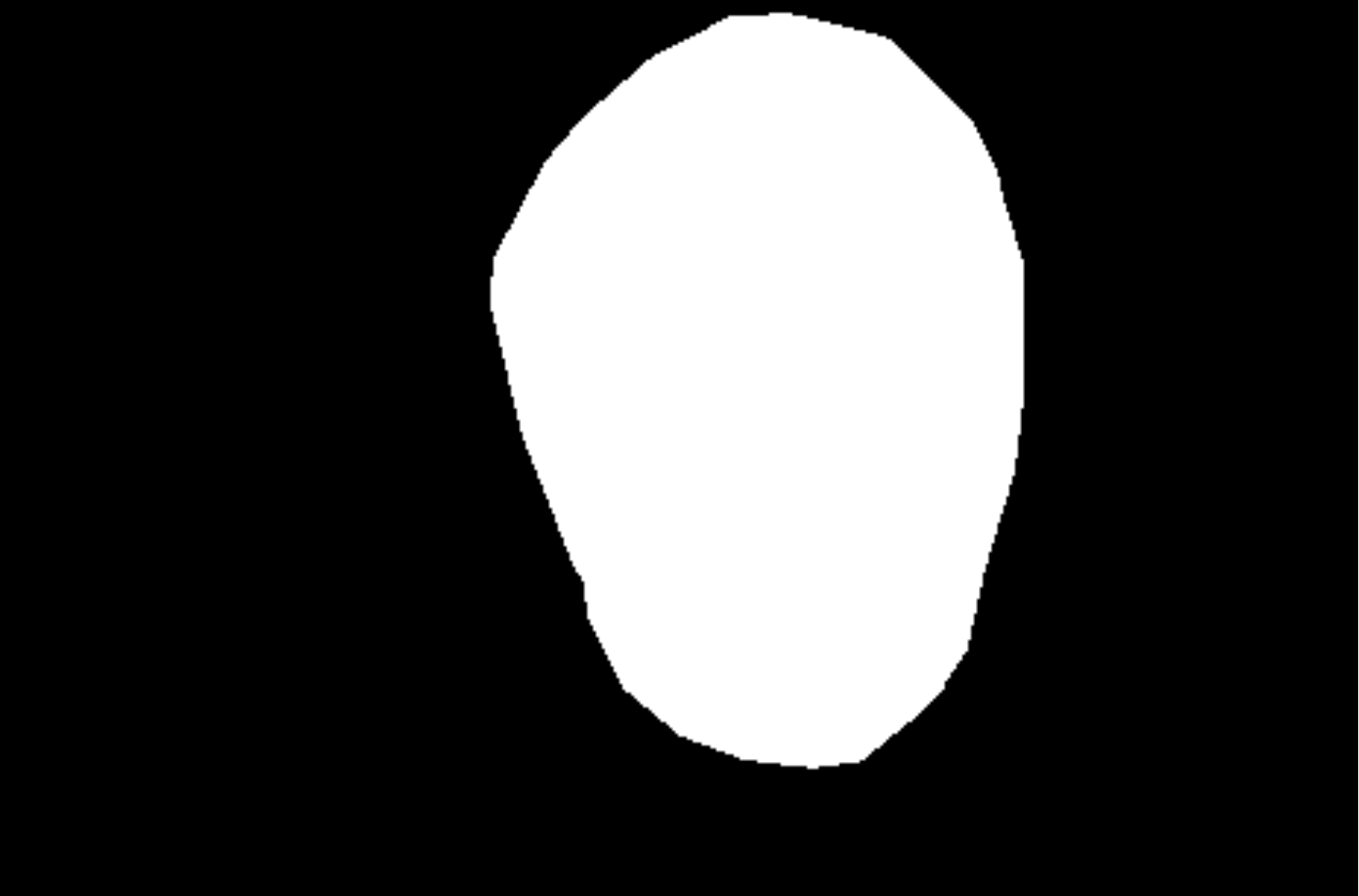}
	 \includegraphics[width=\figobjectmatchingw, height=\figobjectmatchingsw]{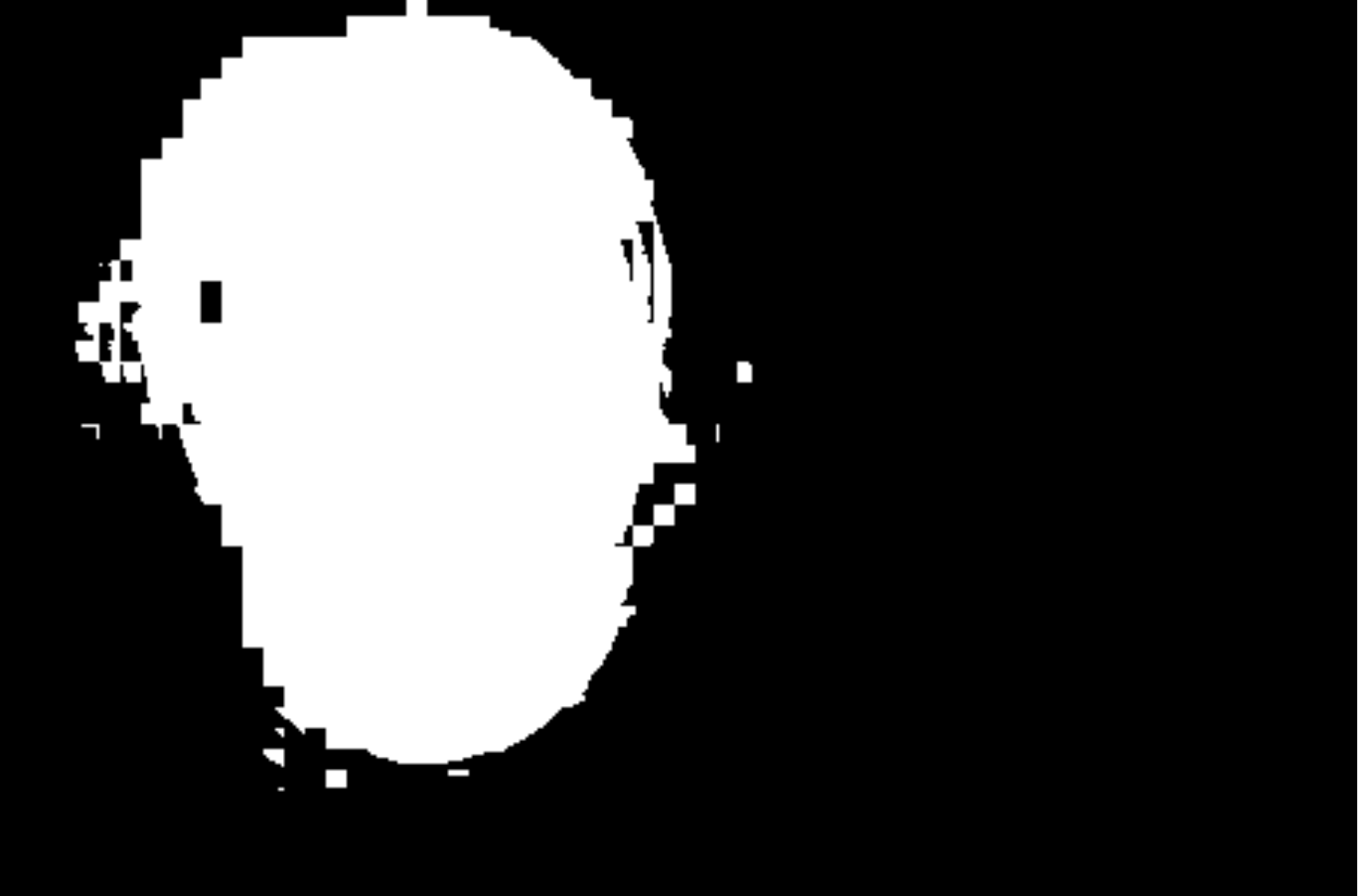}
	 \includegraphics[width=\figobjectmatchingw, height=\figobjectmatchingsw]{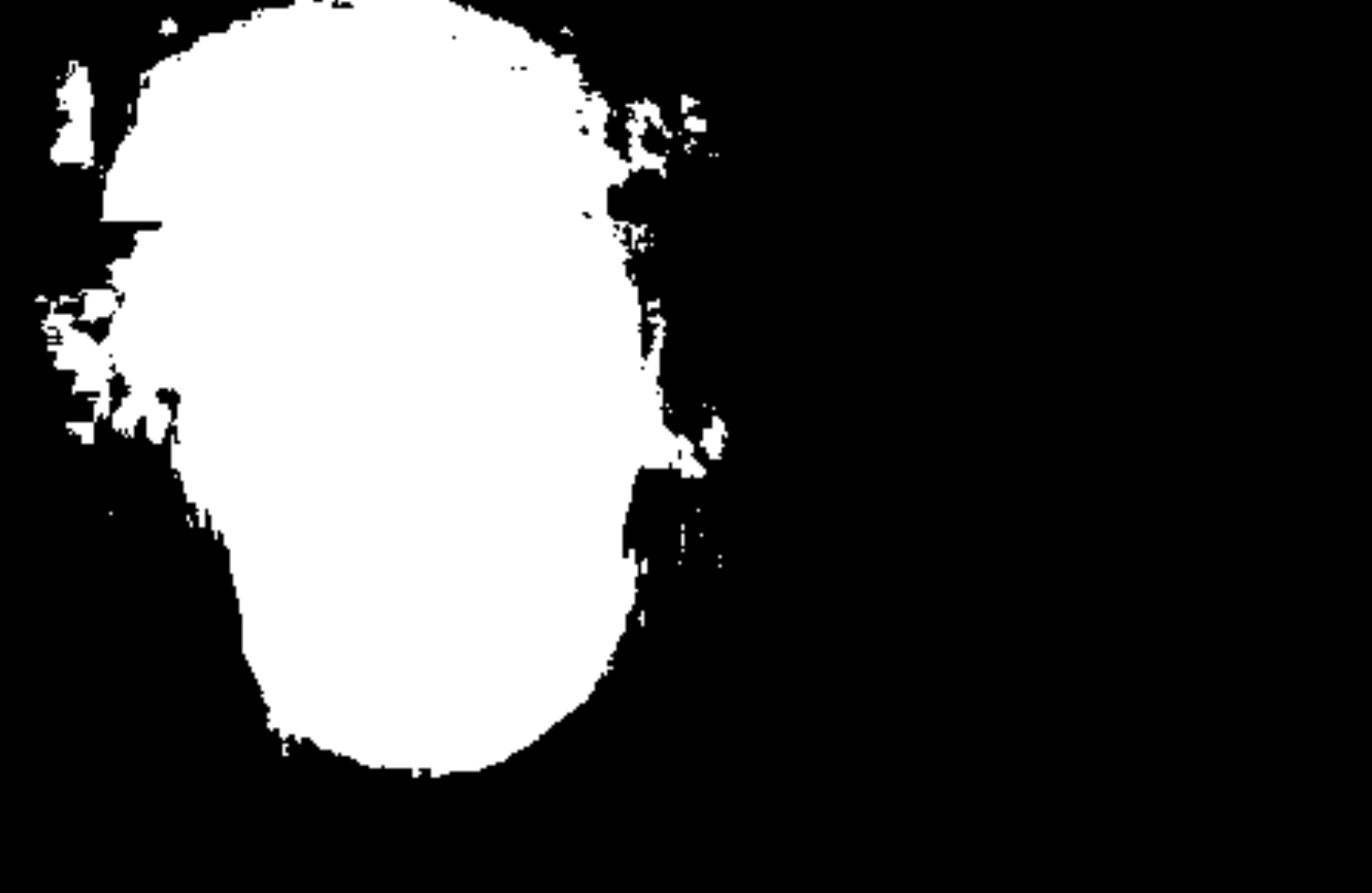}
	 \includegraphics[width=\figobjectmatchingw, height=\figobjectmatchingsw]{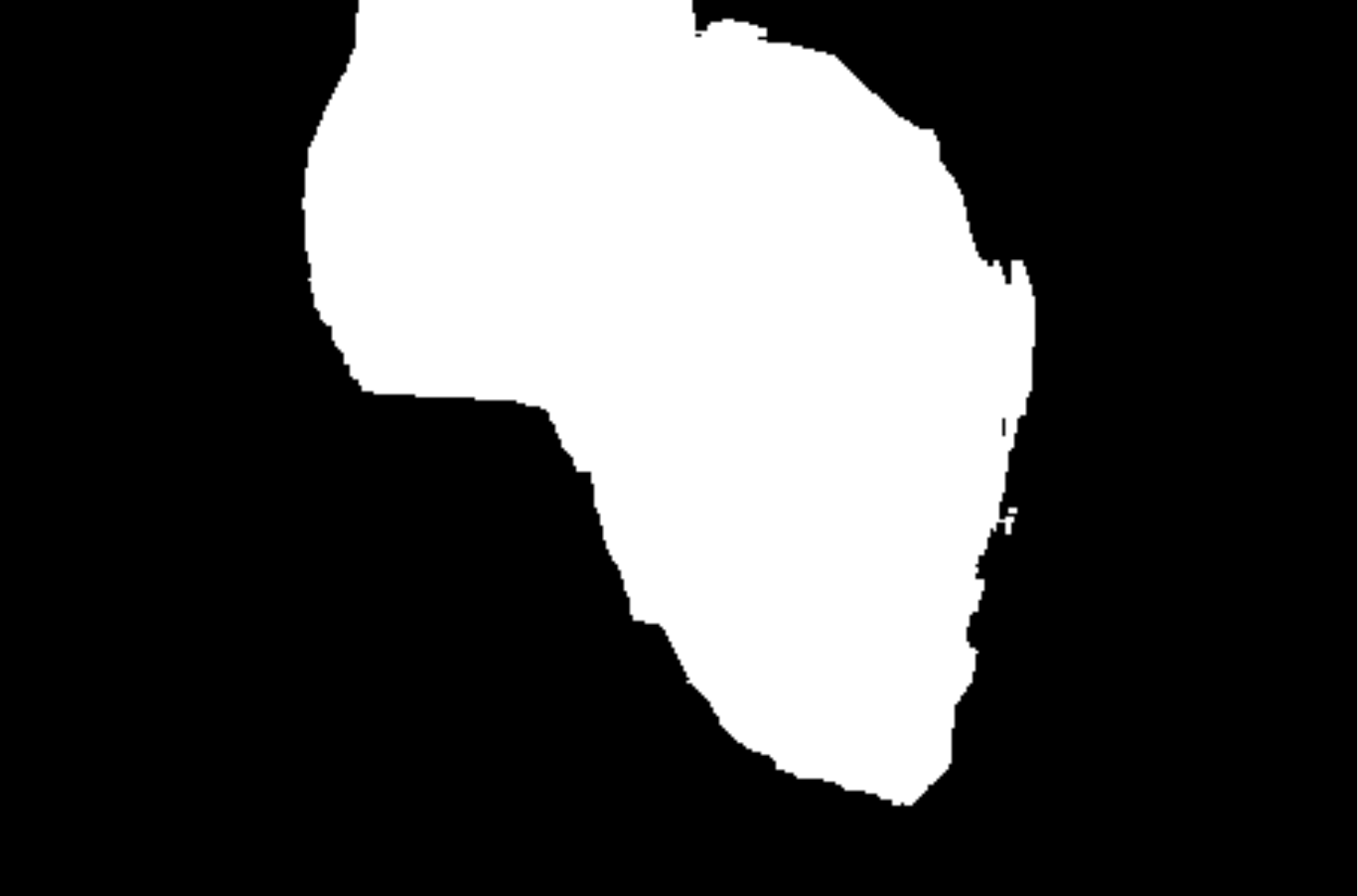}
	 \includegraphics[width=\figobjectmatchingw, height=\figobjectmatchingsw]{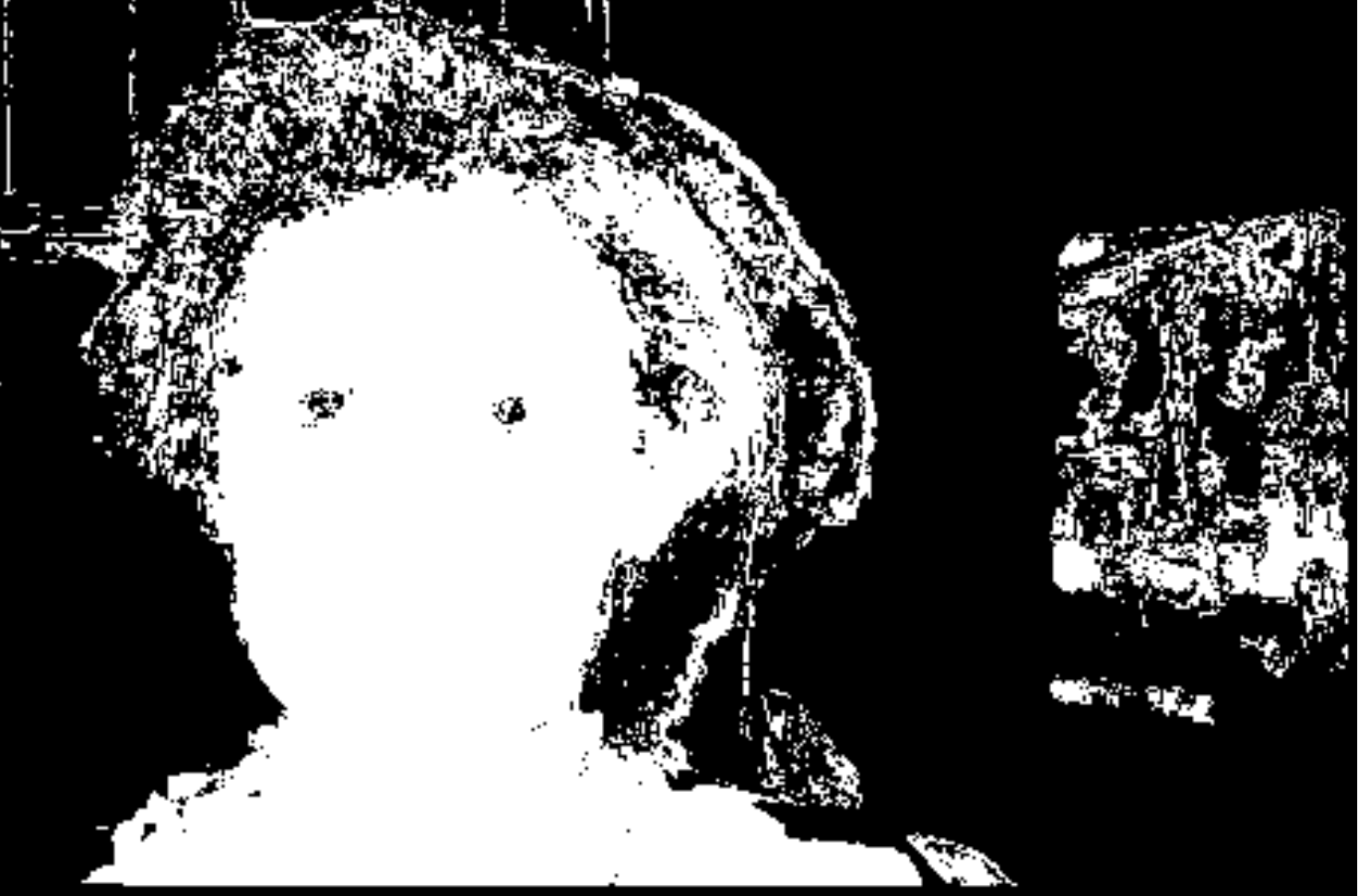} \\

	 \includegraphics[width=\figobjectmatchingw, height=\figobjectmatchingsw]{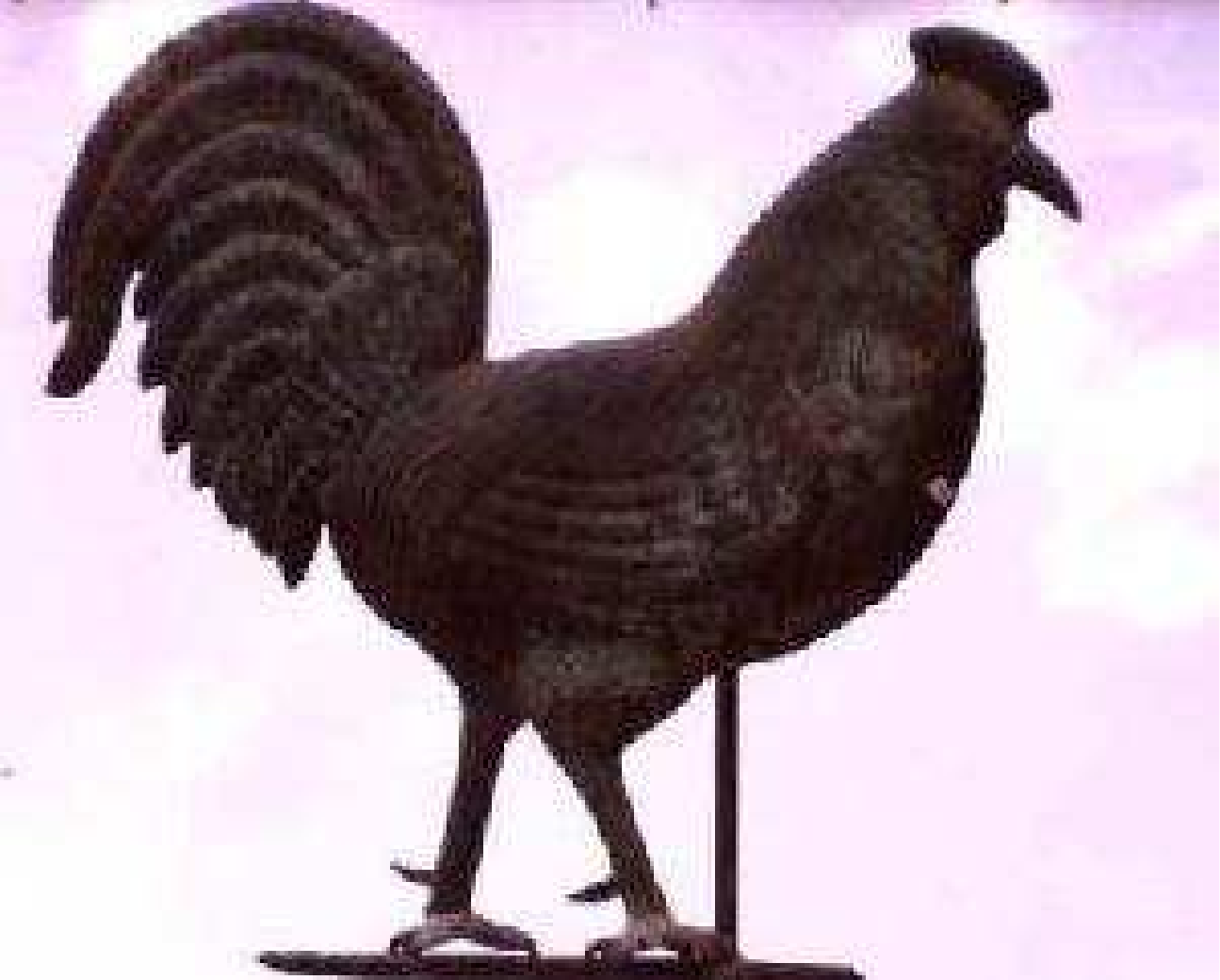}
	 \includegraphics[width=\figobjectmatchingw, height=\figobjectmatchingsw]{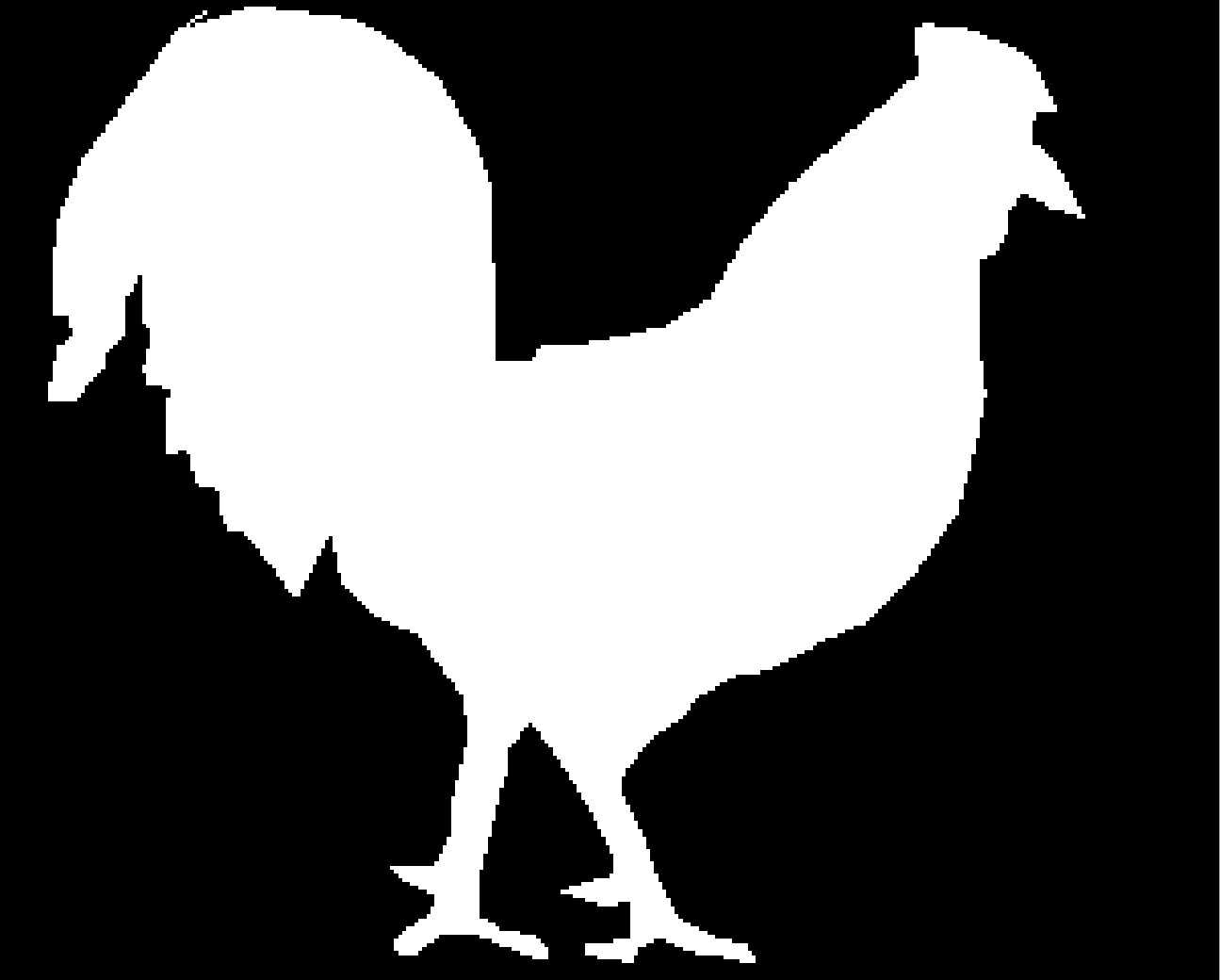}
	 \includegraphics[width=\figobjectmatchingw, height=\figobjectmatchingsw]{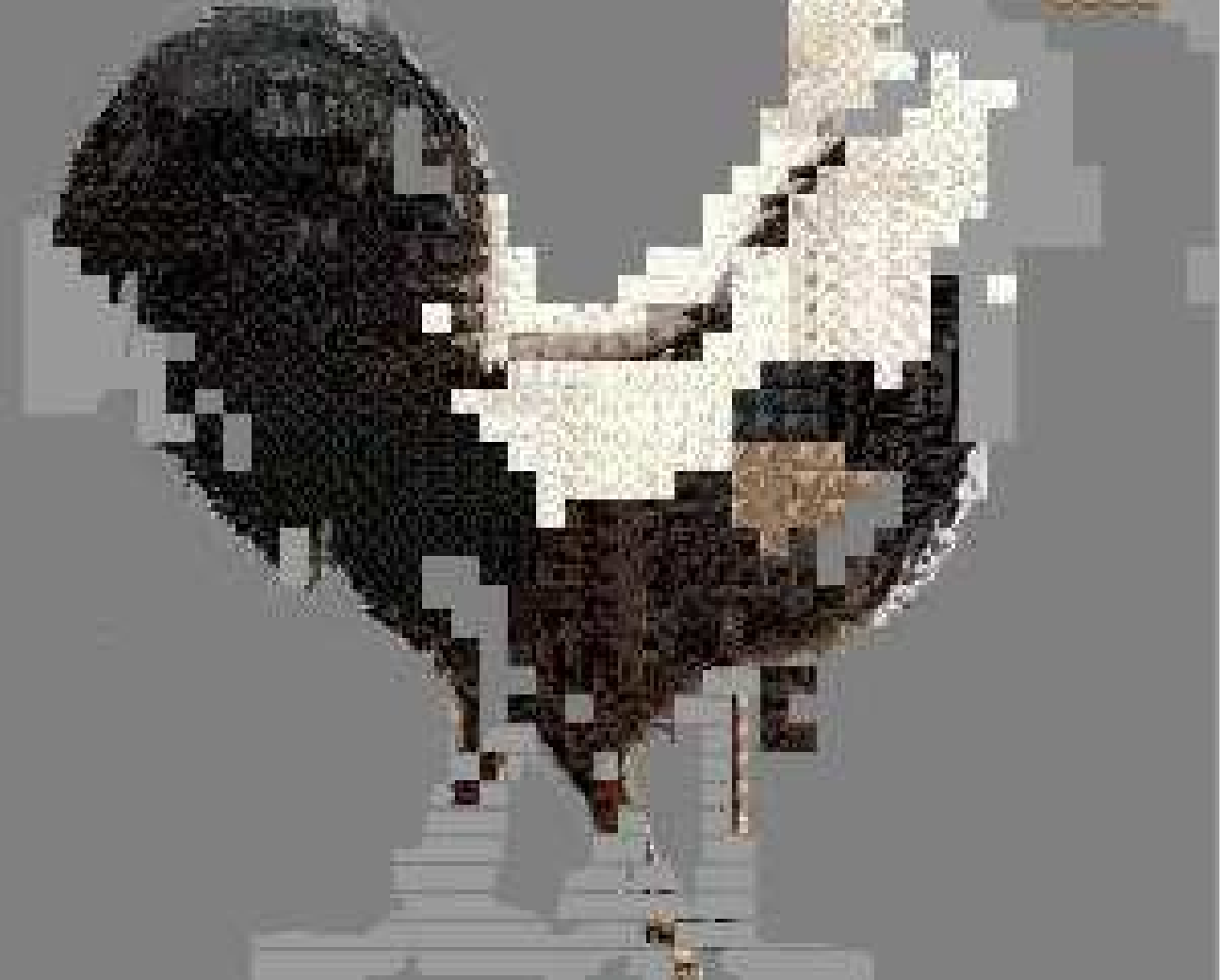}
	 \includegraphics[width=\figobjectmatchingw, height=\figobjectmatchingsw]{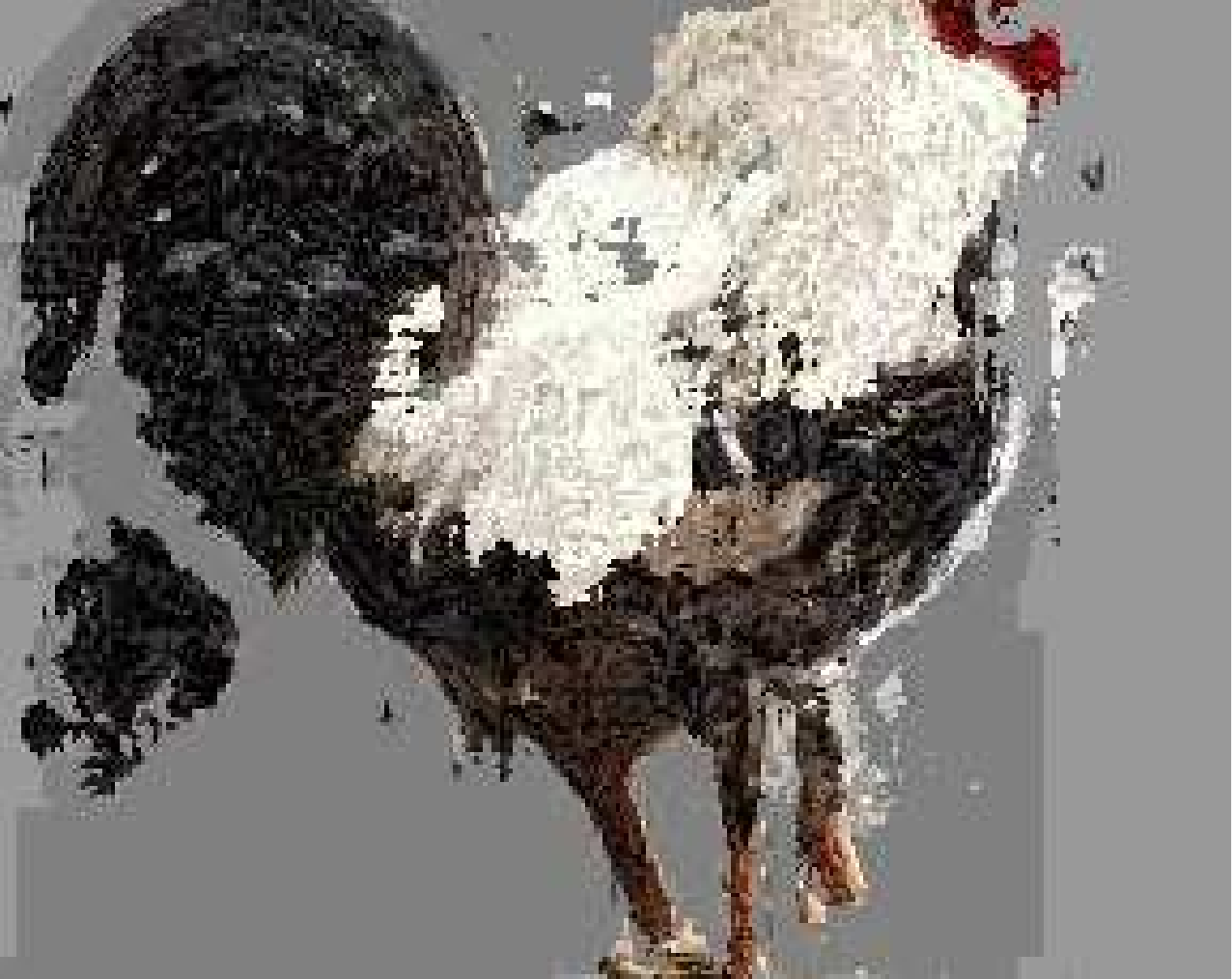}
	 \includegraphics[width=\figobjectmatchingw, height=\figobjectmatchingsw]{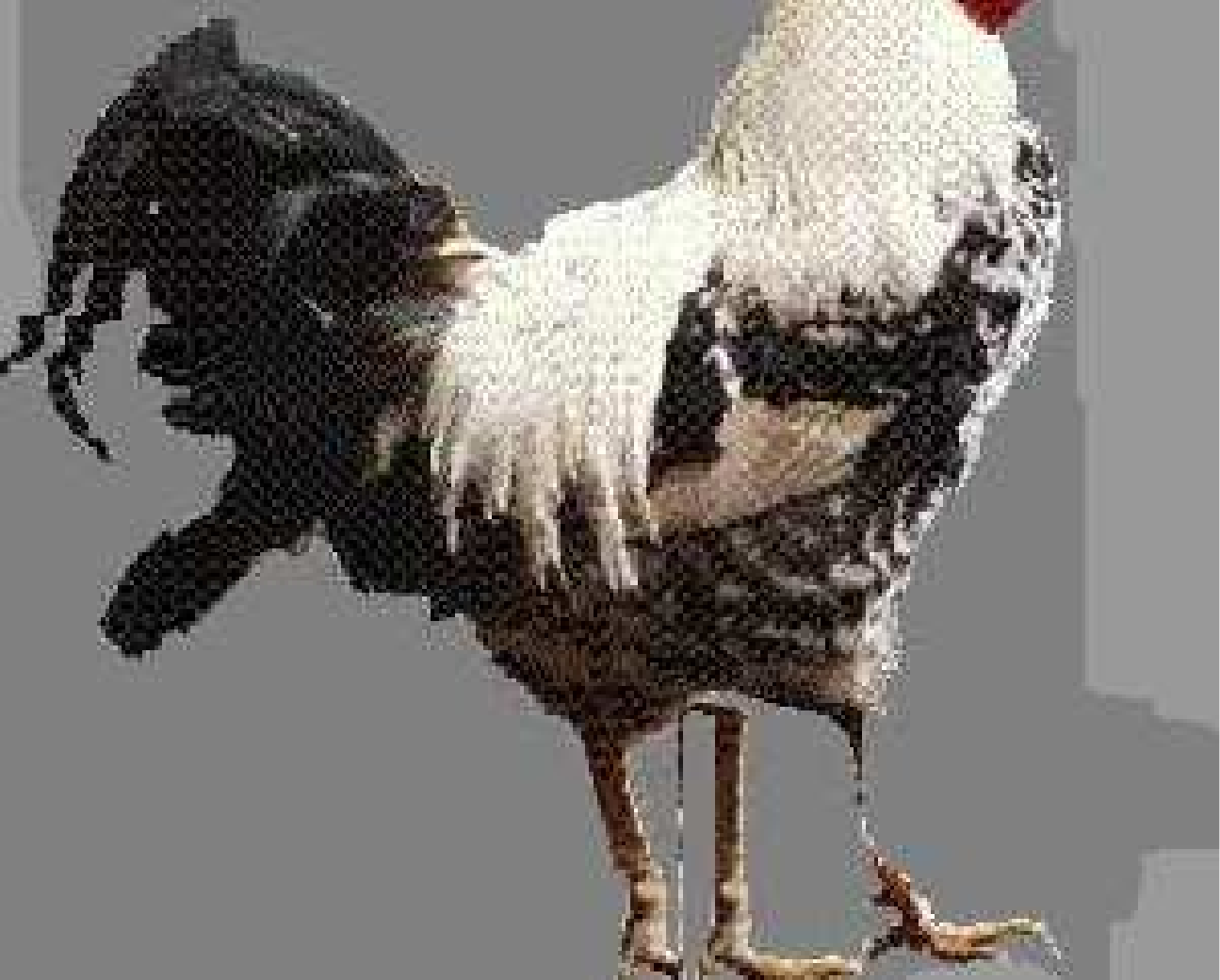}
	 \includegraphics[width=\figobjectmatchingw, height=\figobjectmatchingsw]{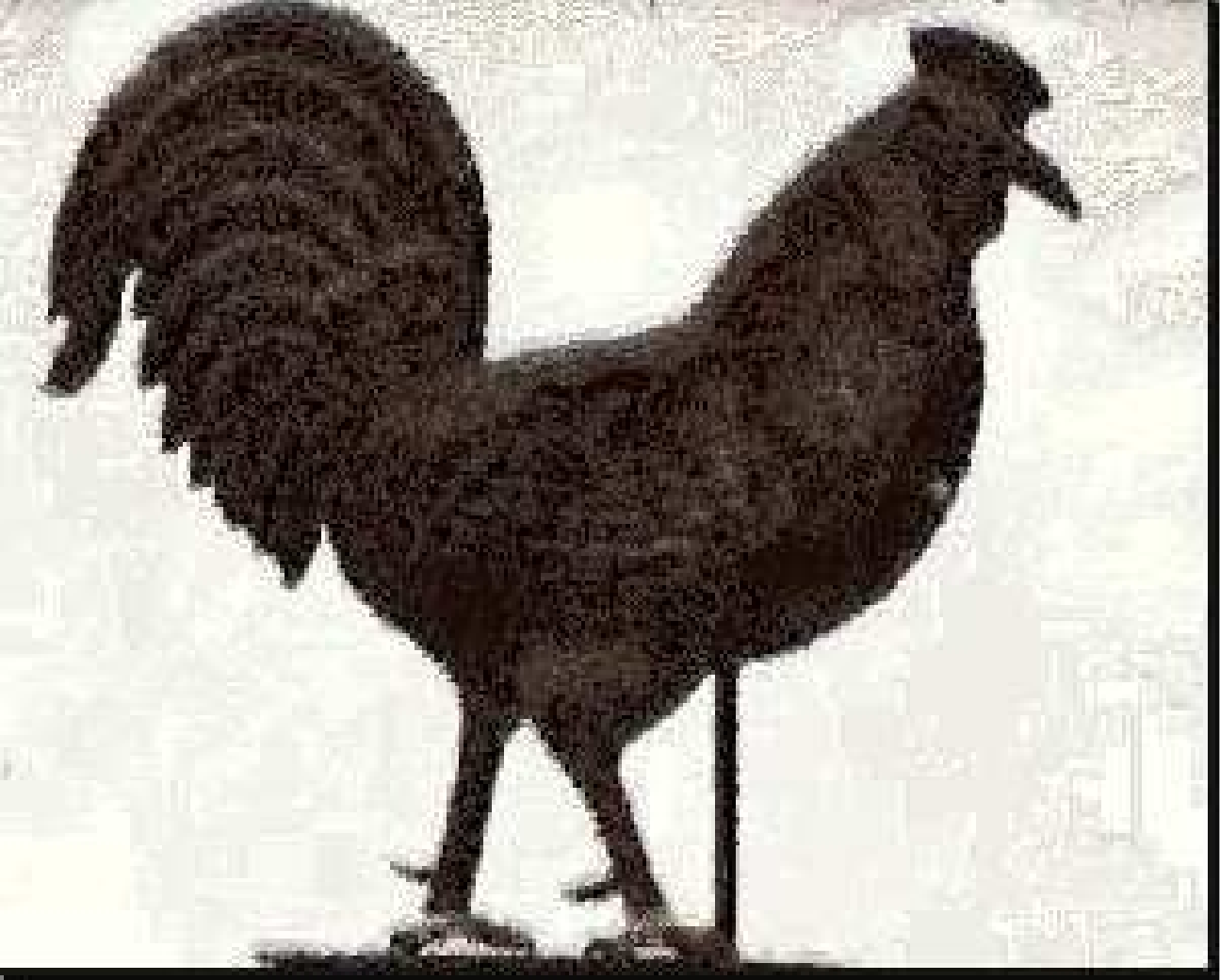} \\

	 \includegraphics[width=\figobjectmatchingw, height=\figobjectmatchingsw]{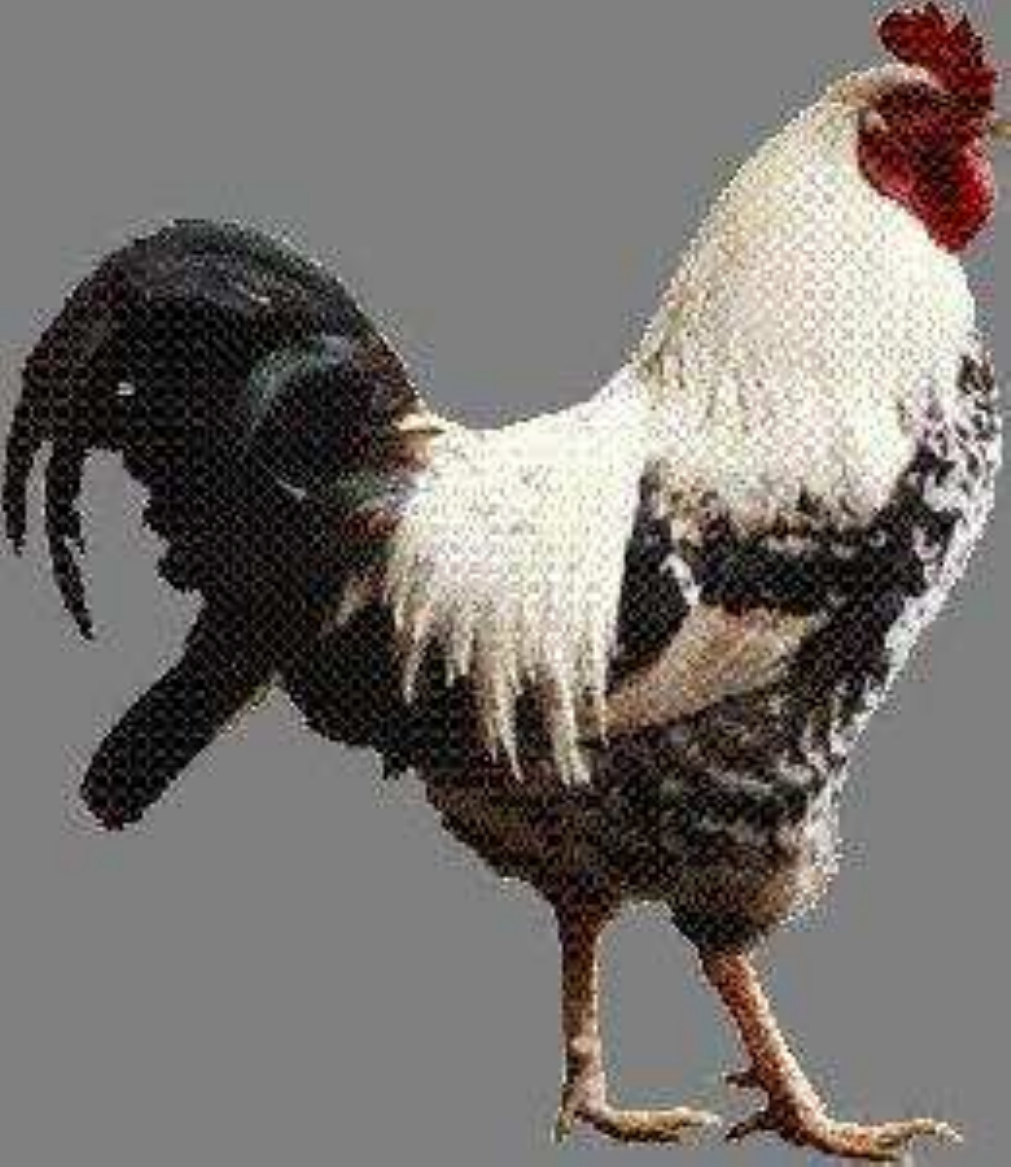}
	 \includegraphics[width=\figobjectmatchingw, height=\figobjectmatchingsw]{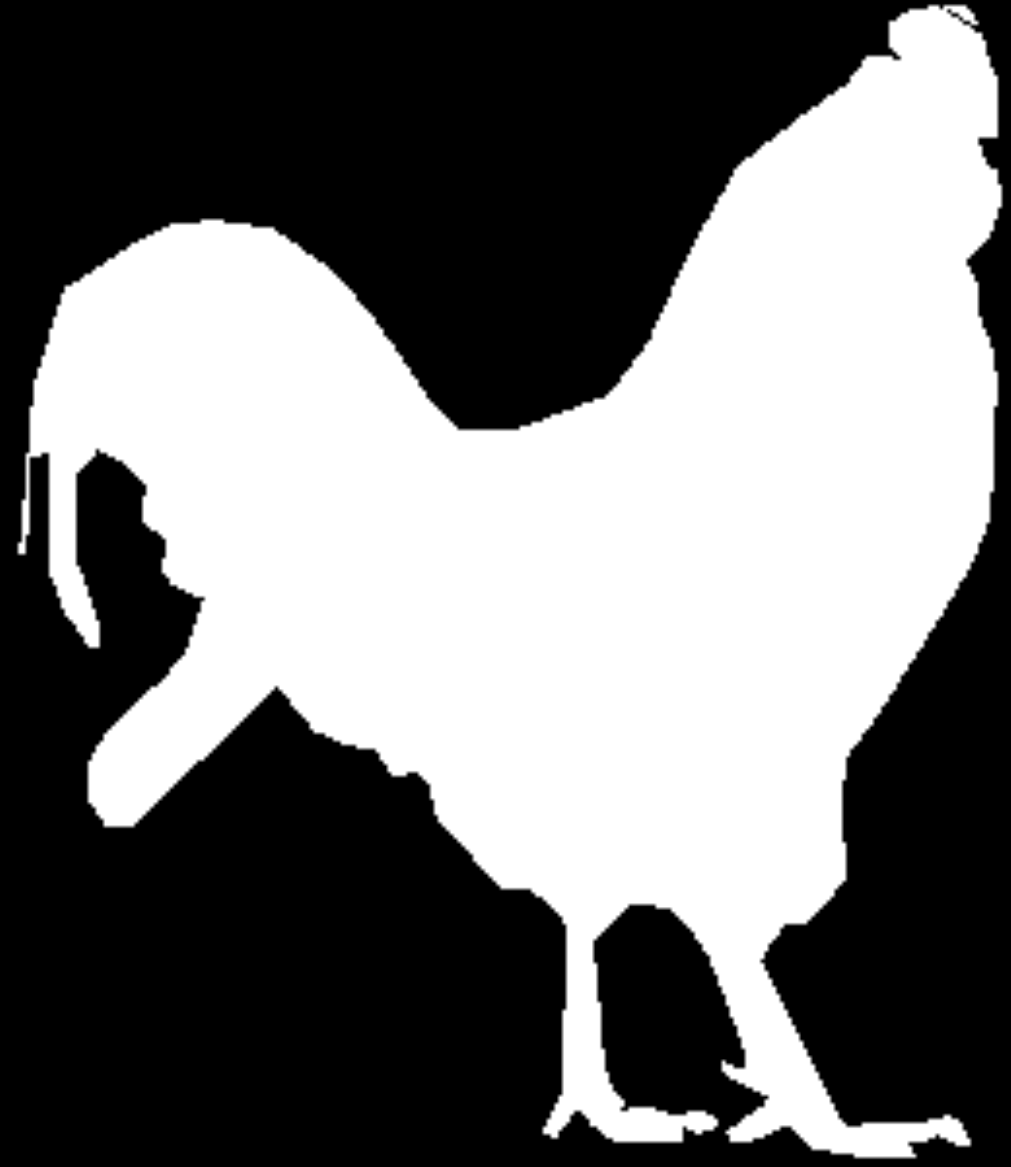}
	 \includegraphics[width=\figobjectmatchingw, height=\figobjectmatchingsw]{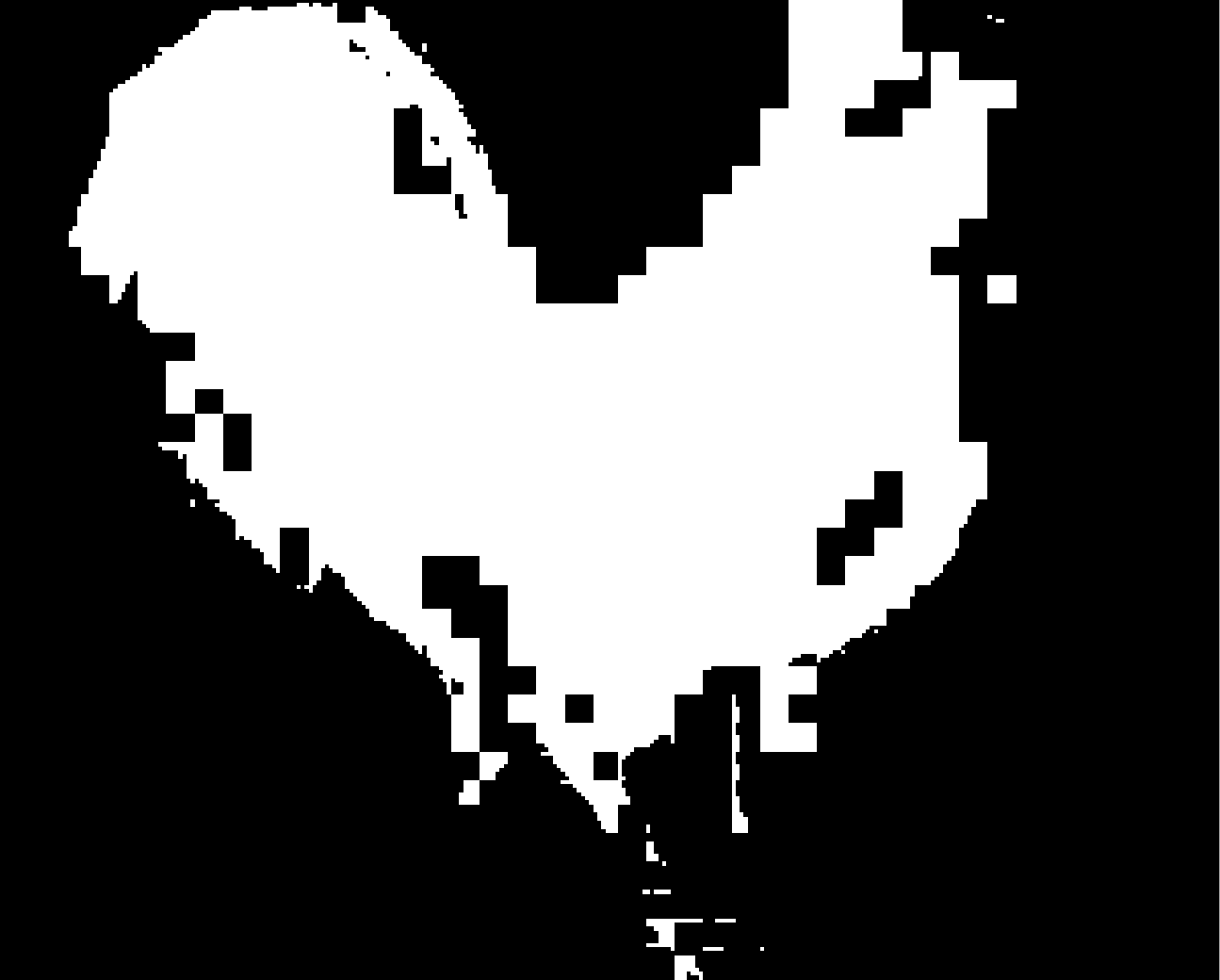}
	 \includegraphics[width=\figobjectmatchingw, height=\figobjectmatchingsw]{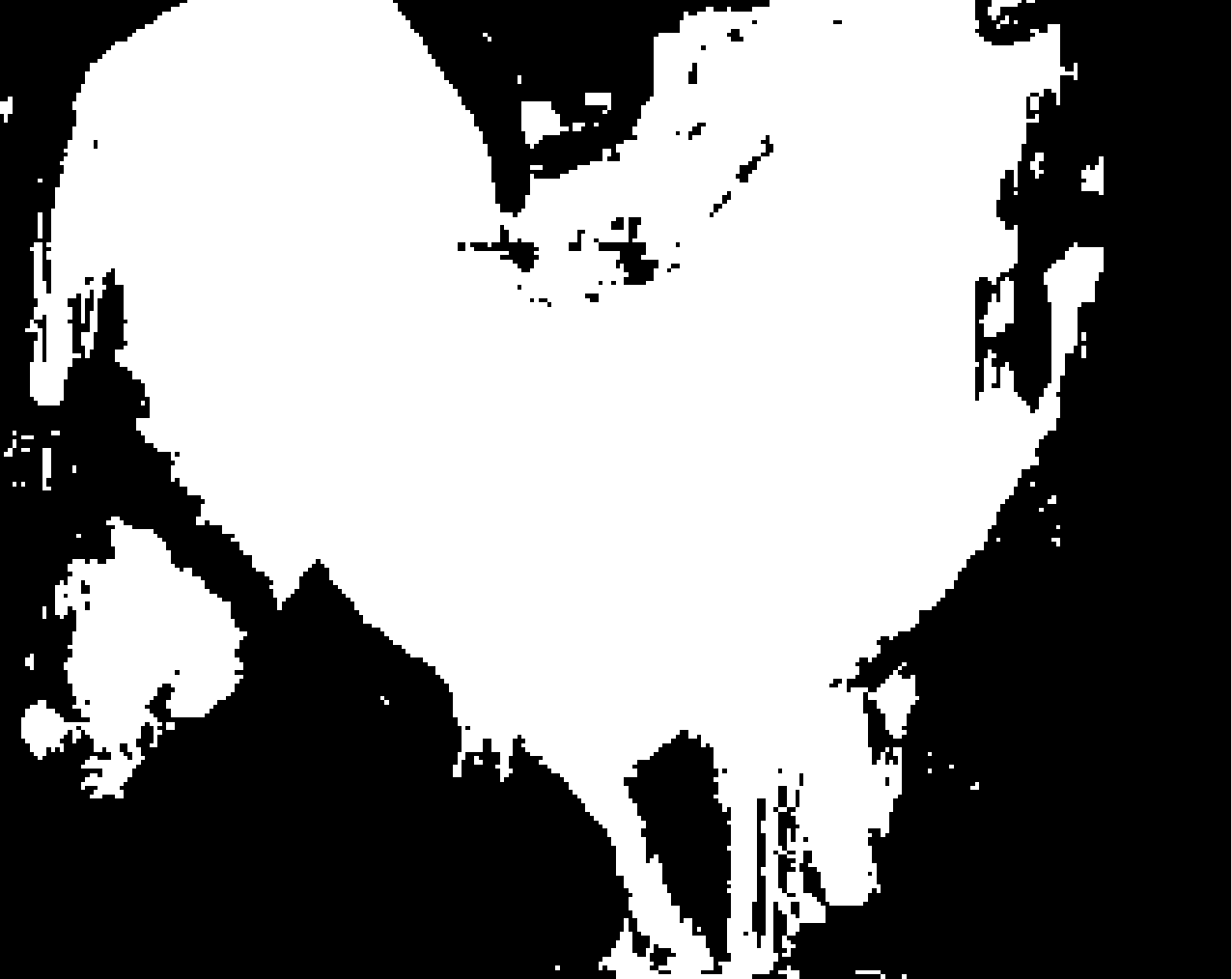}
	 \includegraphics[width=\figobjectmatchingw, height=\figobjectmatchingsw]{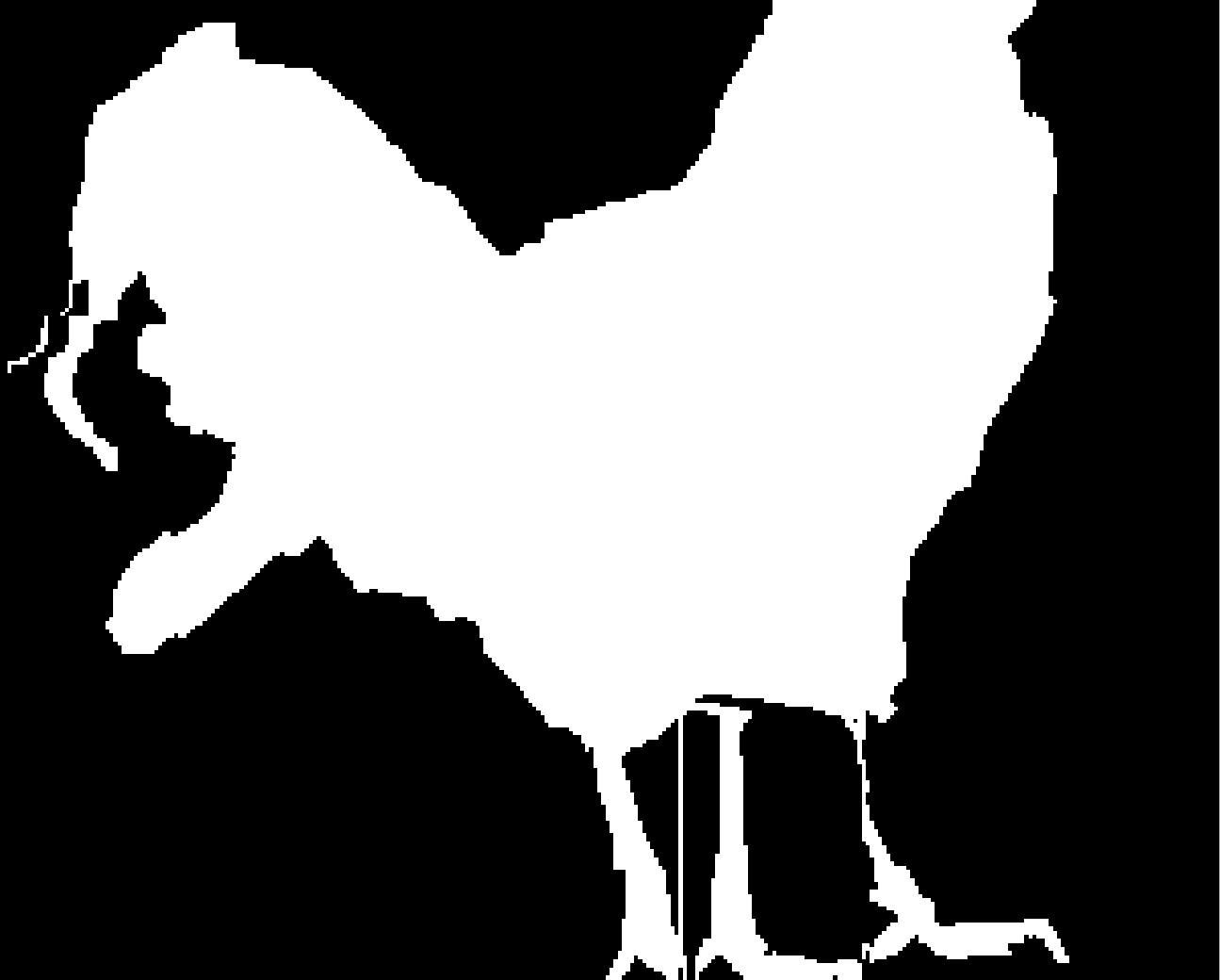}
	 \includegraphics[width=\figobjectmatchingw, height=\figobjectmatchingsw]{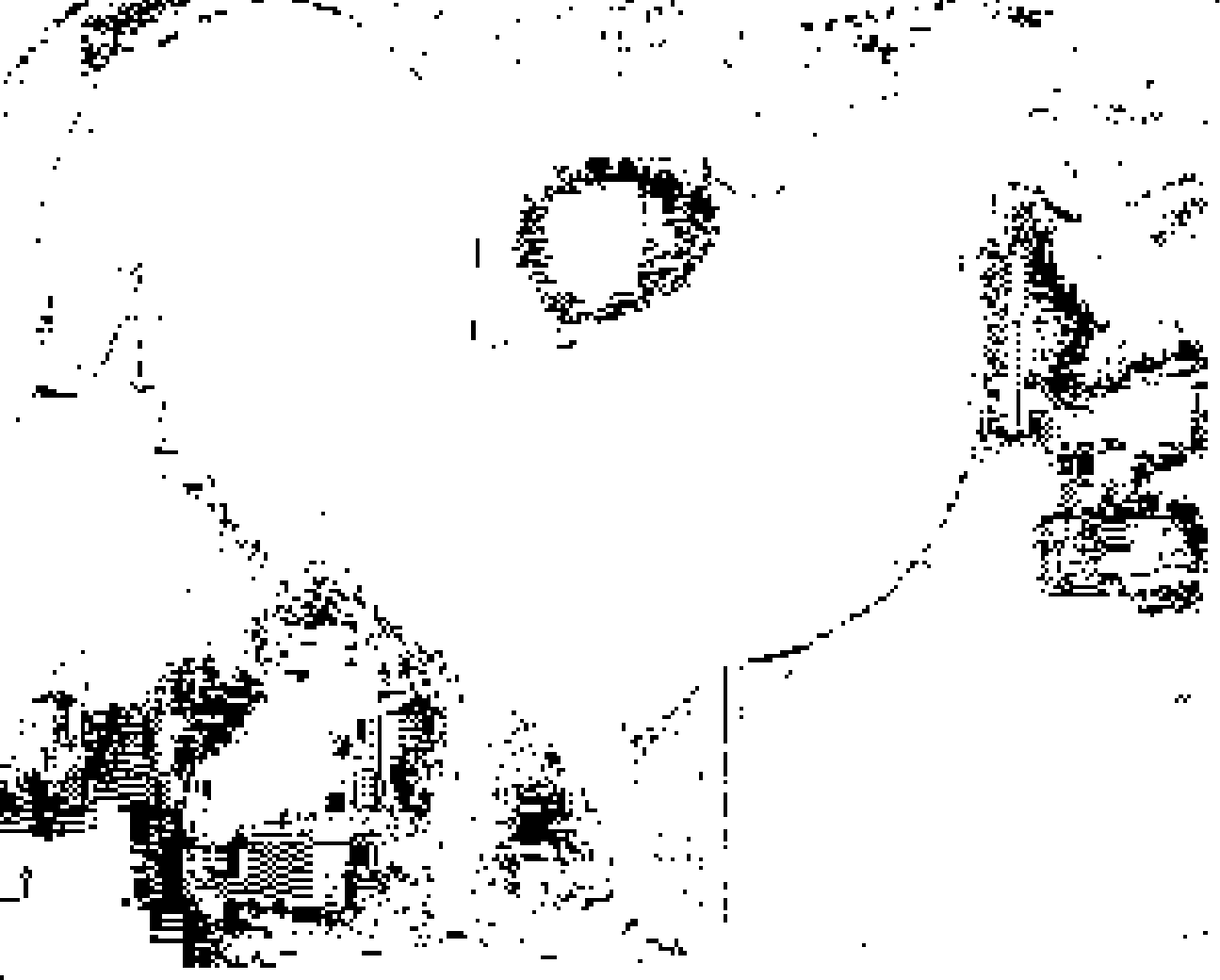}  \\

	 \includegraphics[width=\figobjectmatchingw, height=\figobjectmatchingsw]{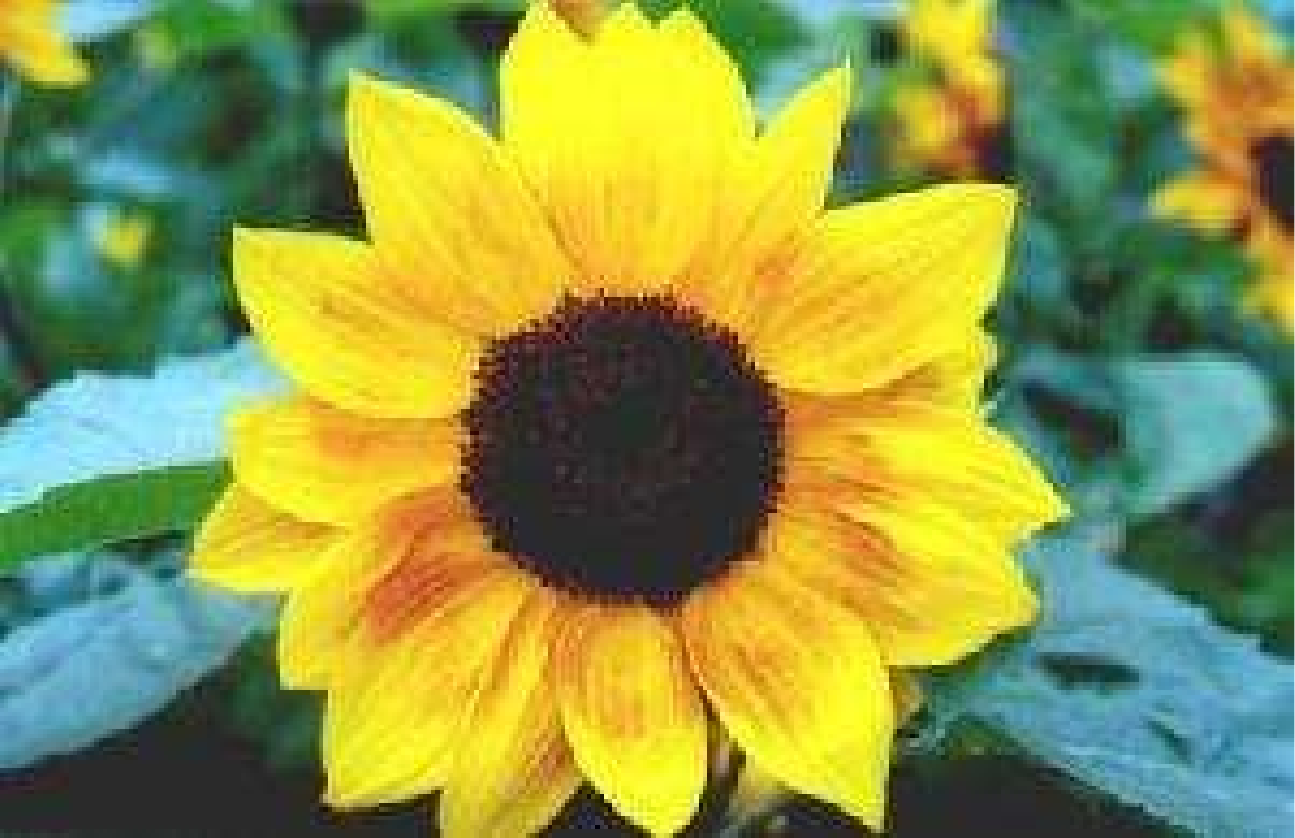}
	 \includegraphics[width=\figobjectmatchingw, height=\figobjectmatchingsw]{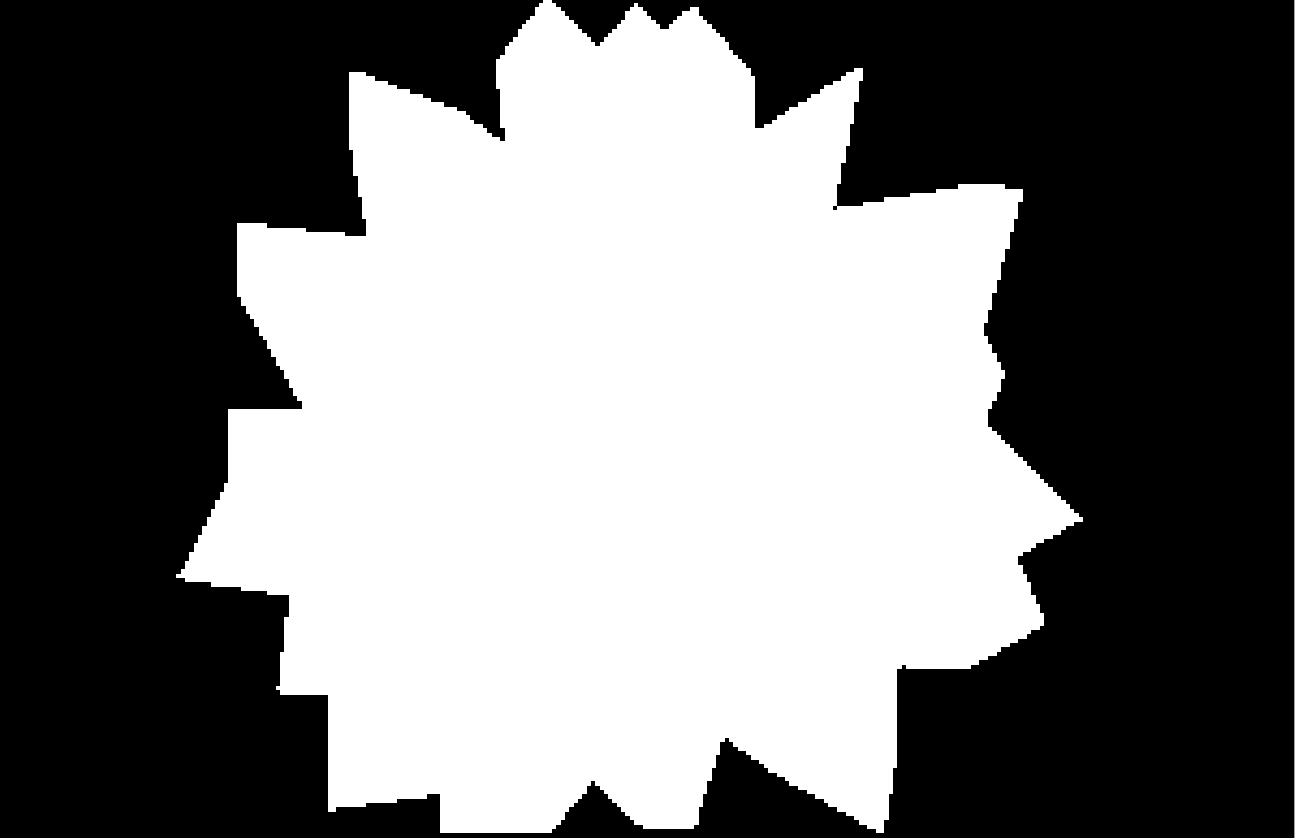}
	 \includegraphics[width=\figobjectmatchingw, height=\figobjectmatchingsw]{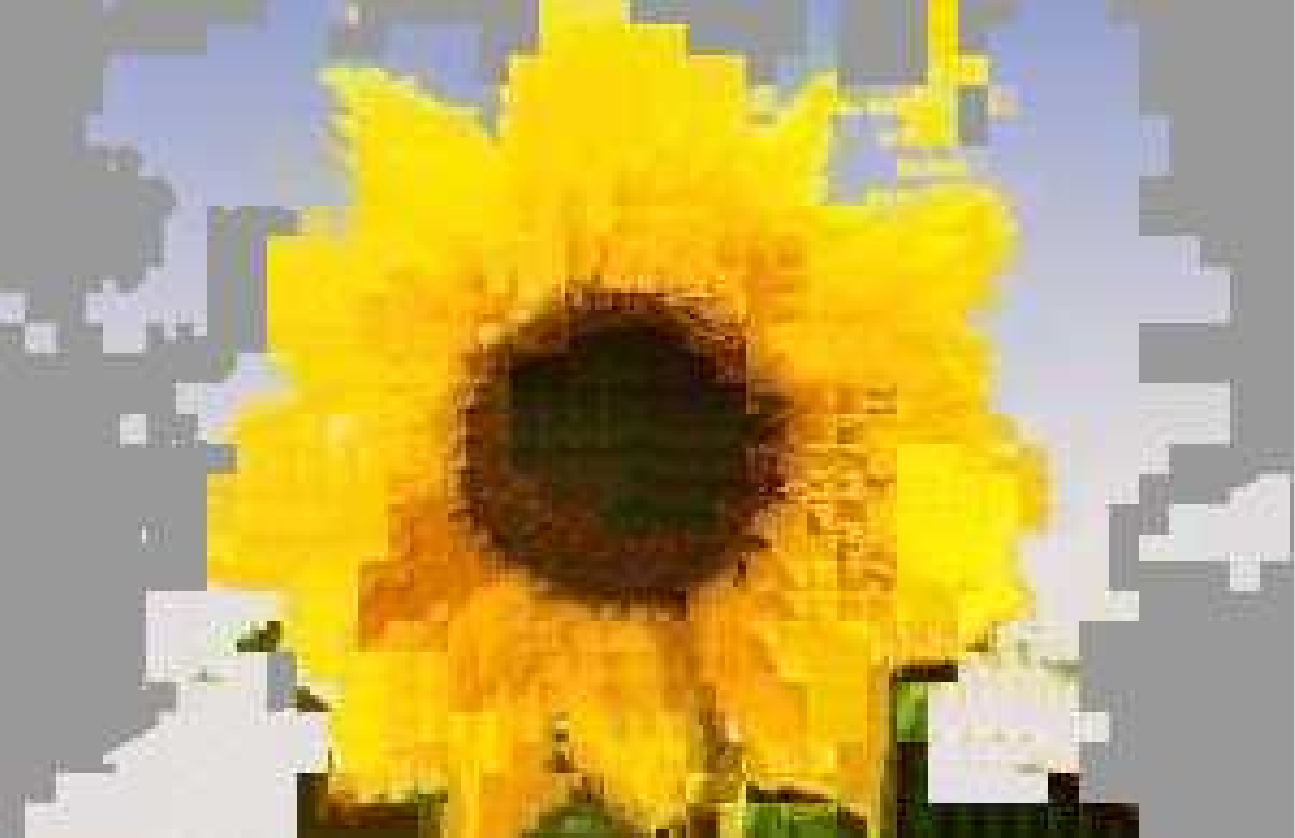}
     \includegraphics[width=\figobjectmatchingw, height=\figobjectmatchingsw]{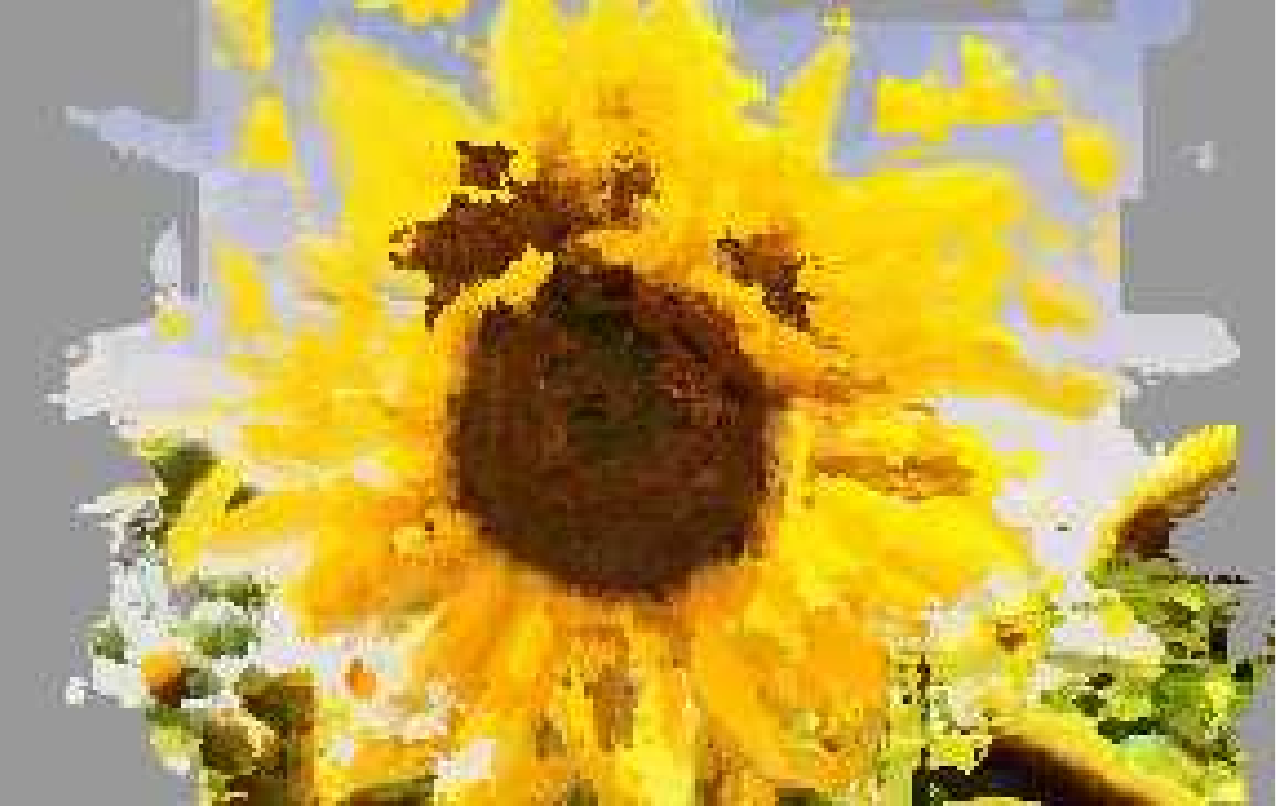}
     \includegraphics[width=\figobjectmatchingw, height=\figobjectmatchingsw]{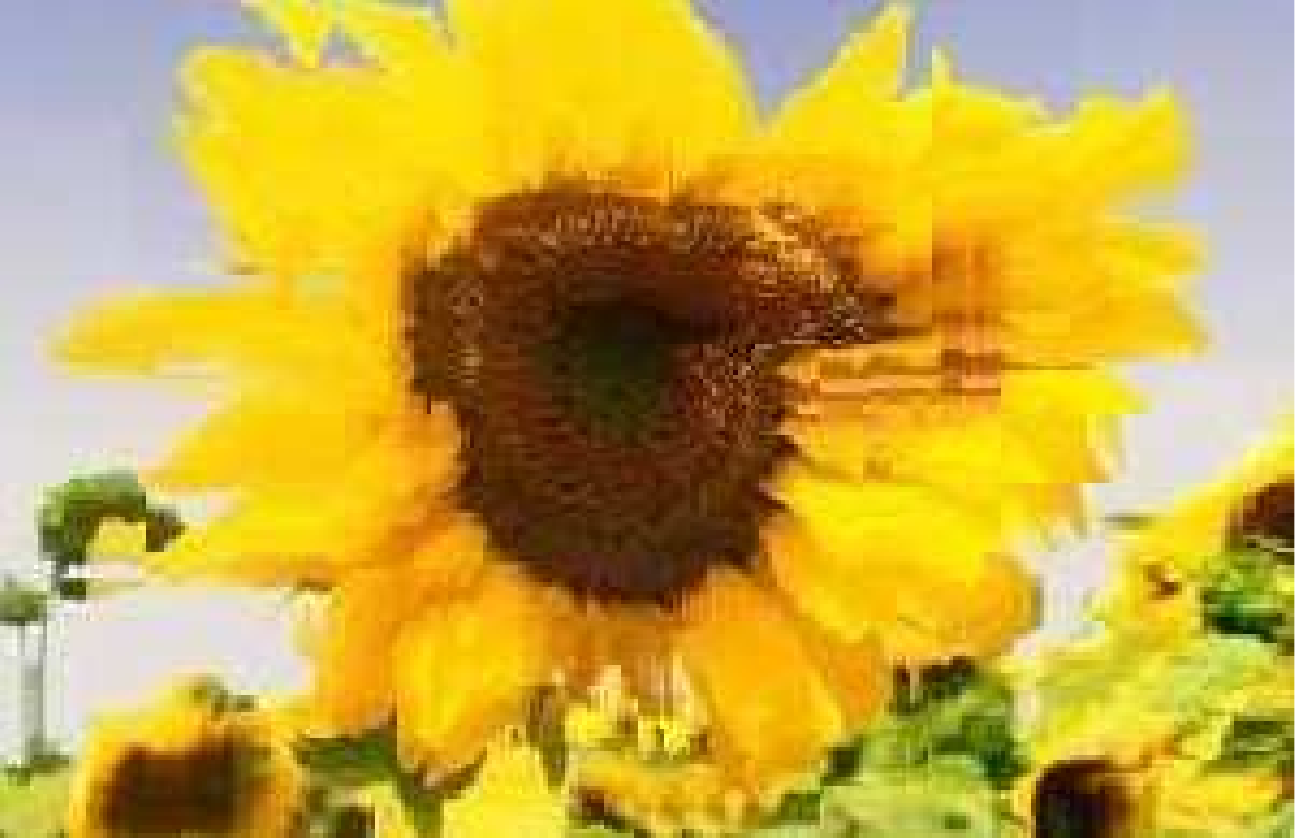}
     \includegraphics[width=\figobjectmatchingw, height=\figobjectmatchingsw]{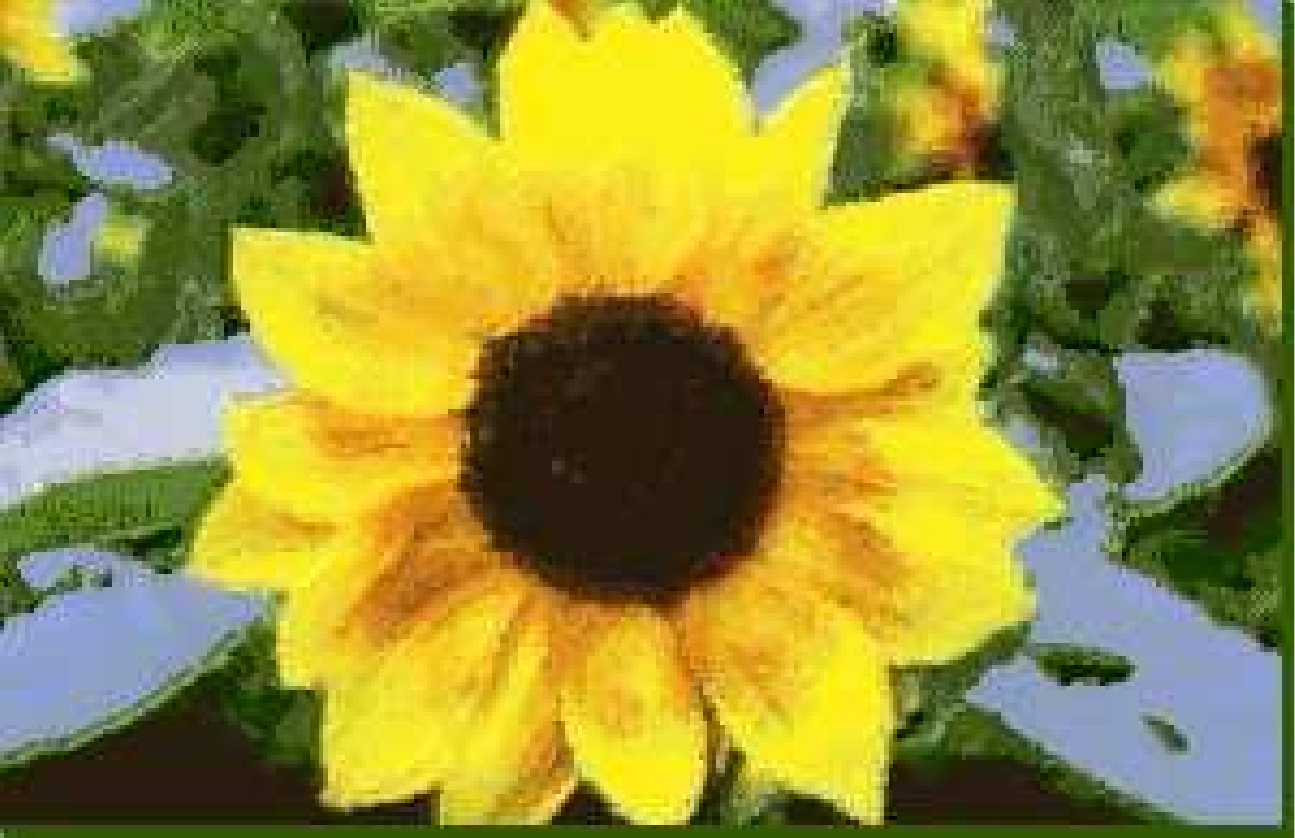} \\

	 \includegraphics[width=\figobjectmatchingw, height=\figobjectmatchingsw]{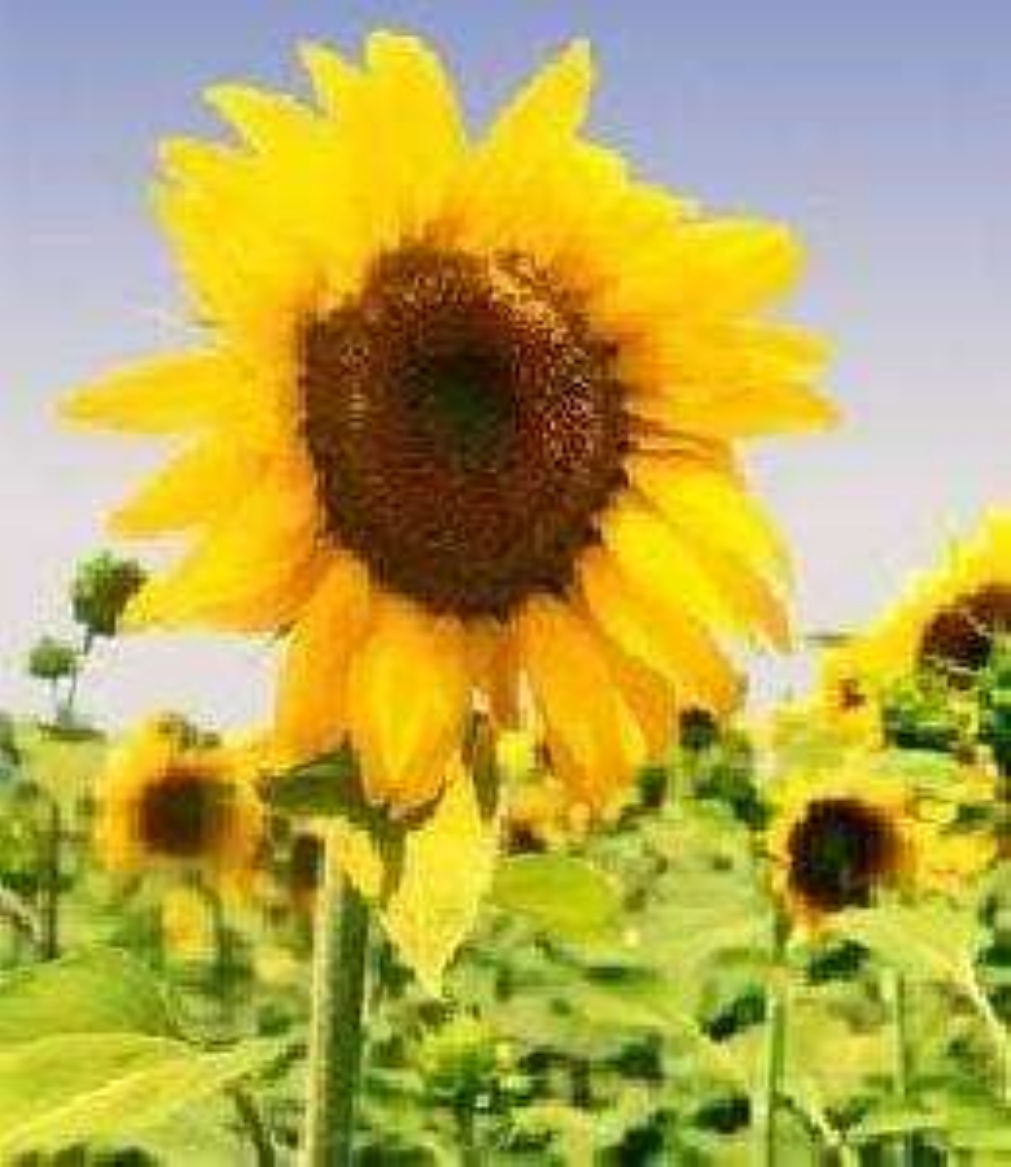}
	 \includegraphics[width=\figobjectmatchingw, height=\figobjectmatchingsw]{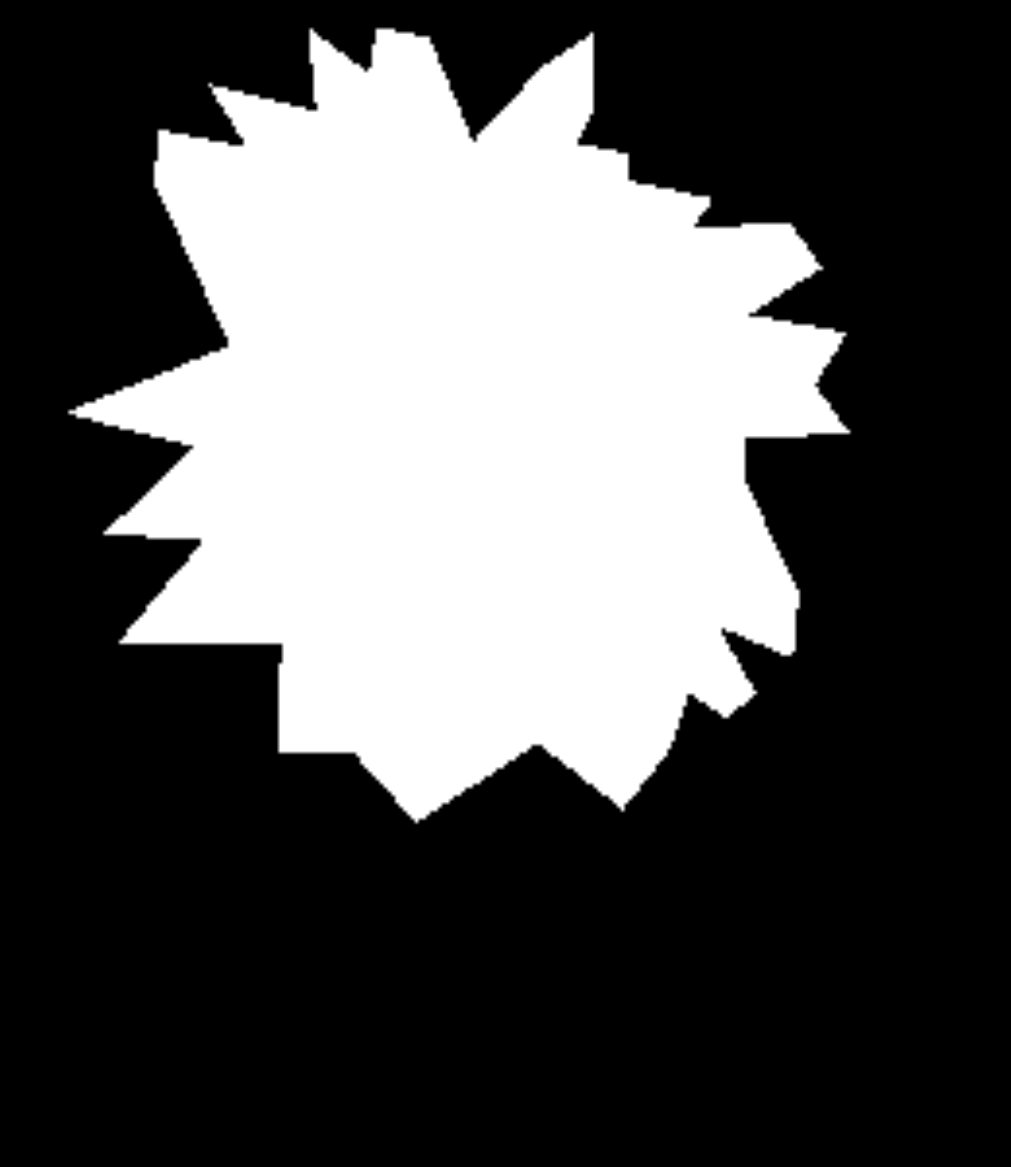}
	 \includegraphics[width=\figobjectmatchingw, height=\figobjectmatchingsw]{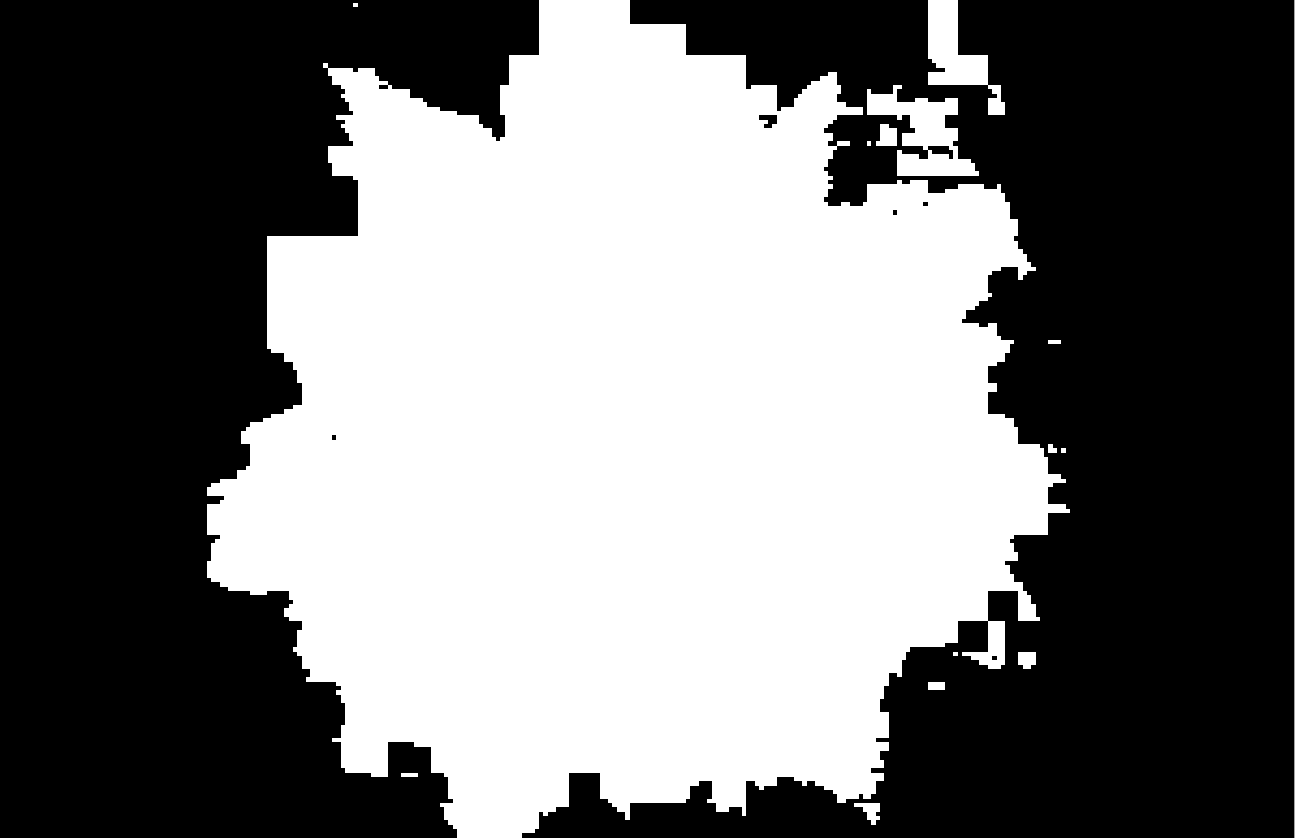}
	 \includegraphics[width=\figobjectmatchingw, height=\figobjectmatchingsw]{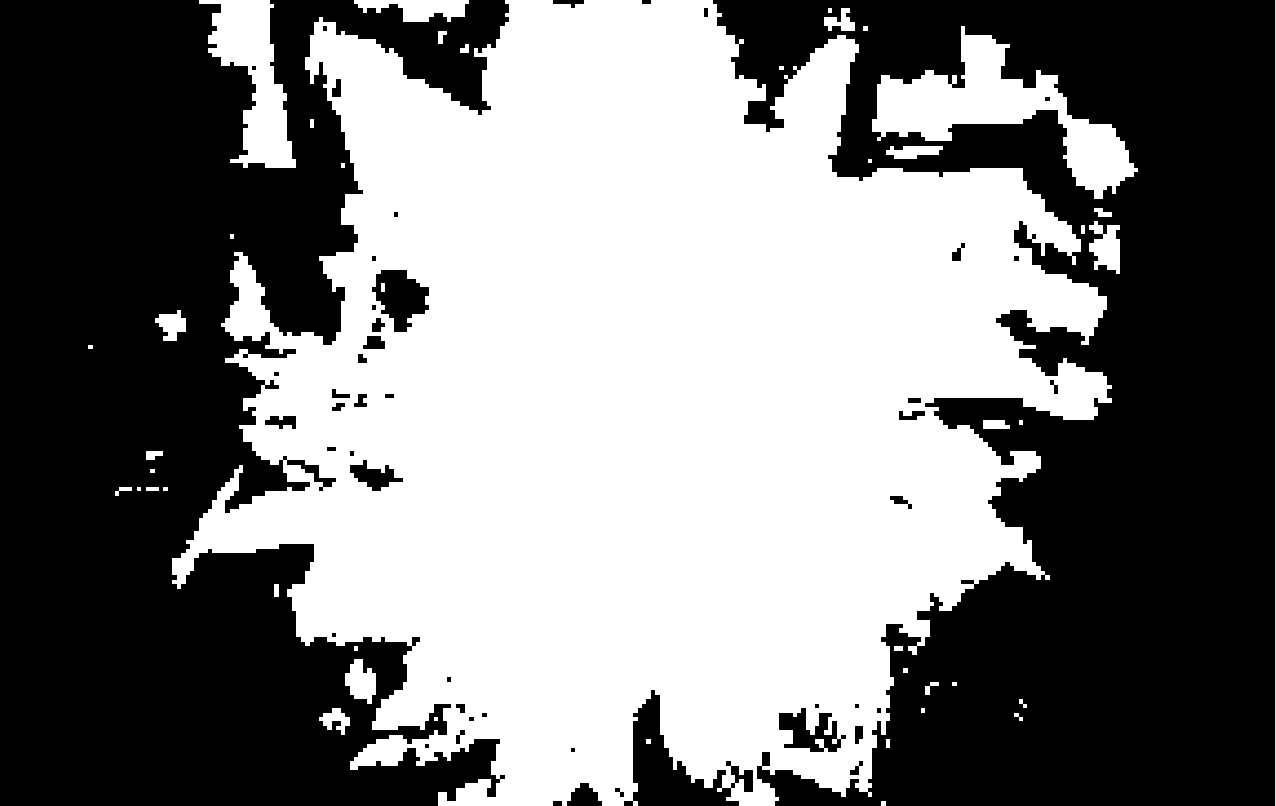}
	 \includegraphics[width=\figobjectmatchingw, height=\figobjectmatchingsw]{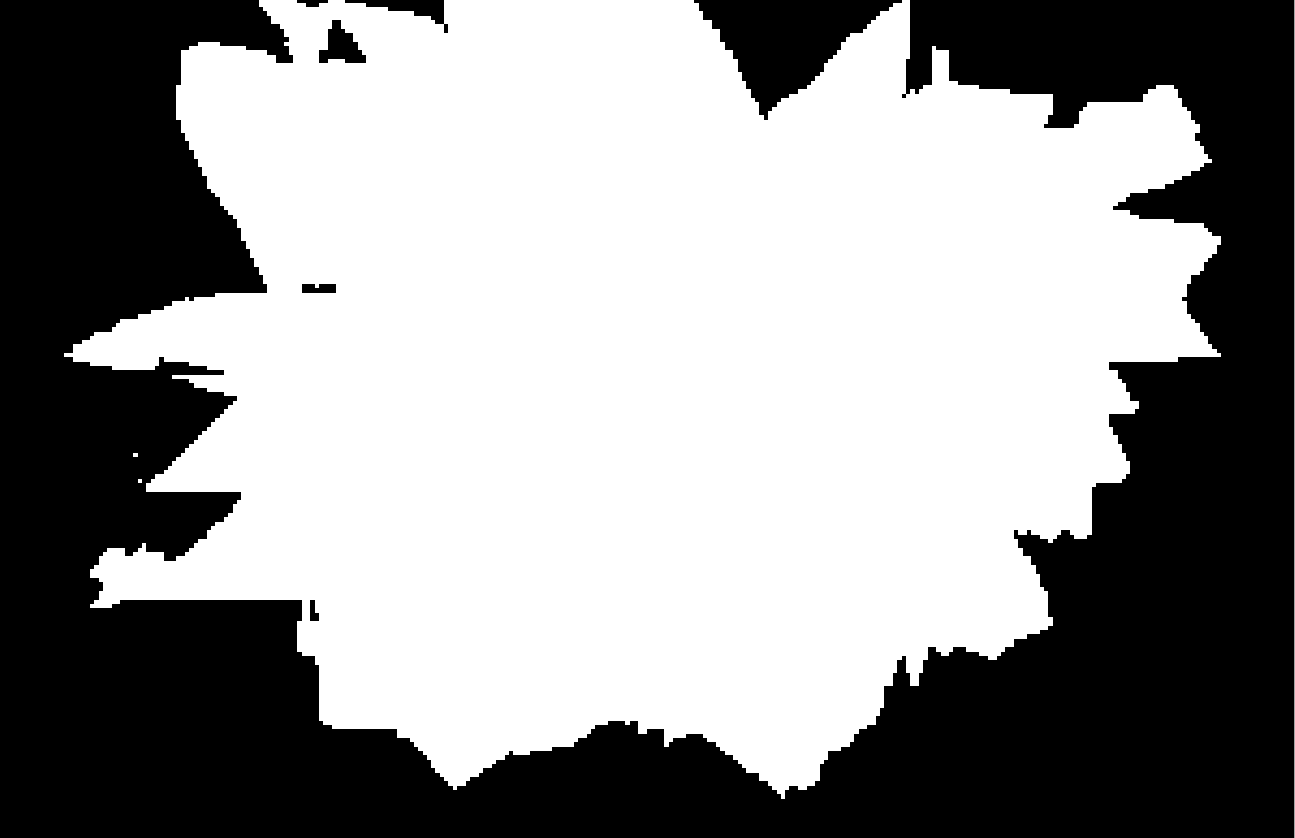}
	 \includegraphics[width=\figobjectmatchingw, height=\figobjectmatchingsw]{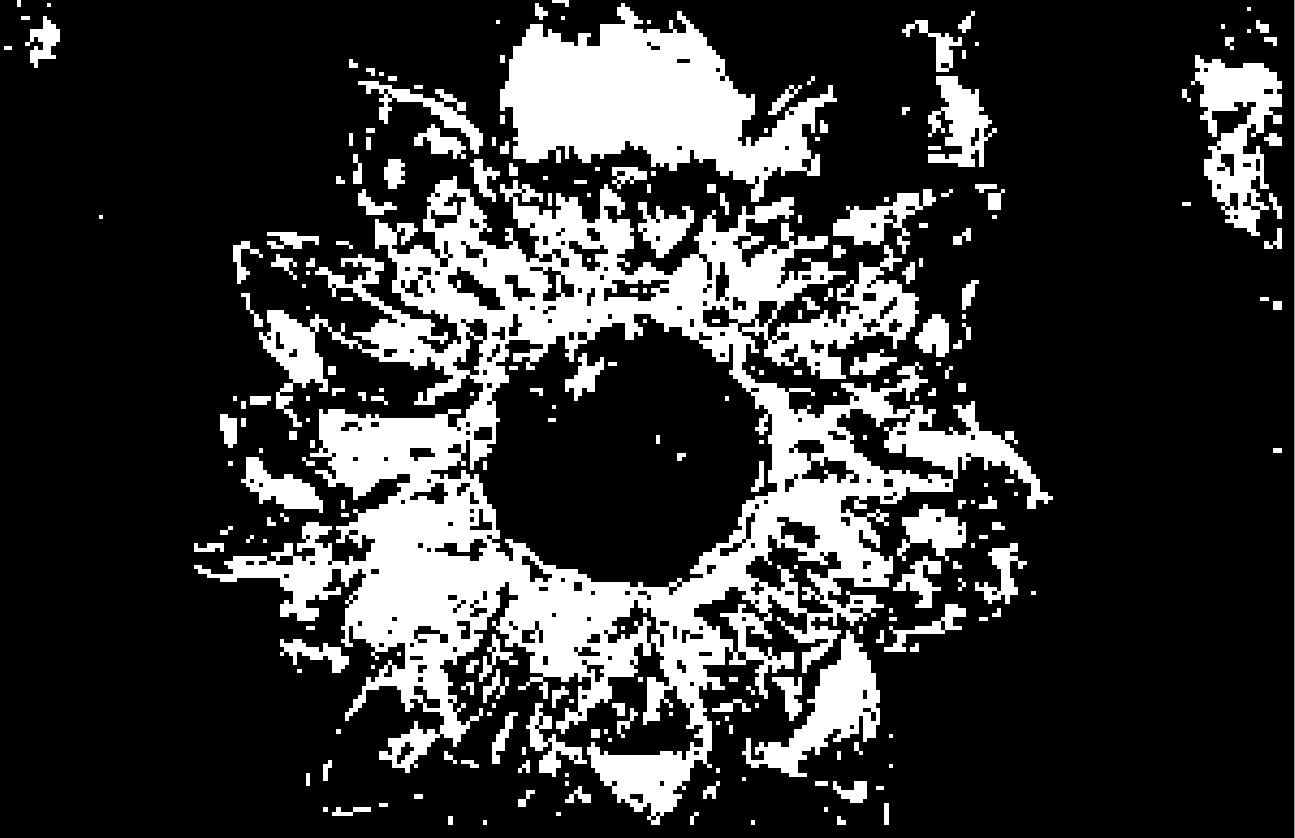} \\

	 \includegraphics[width=\figobjectmatchingw, height=\figobjectmatchingsw]{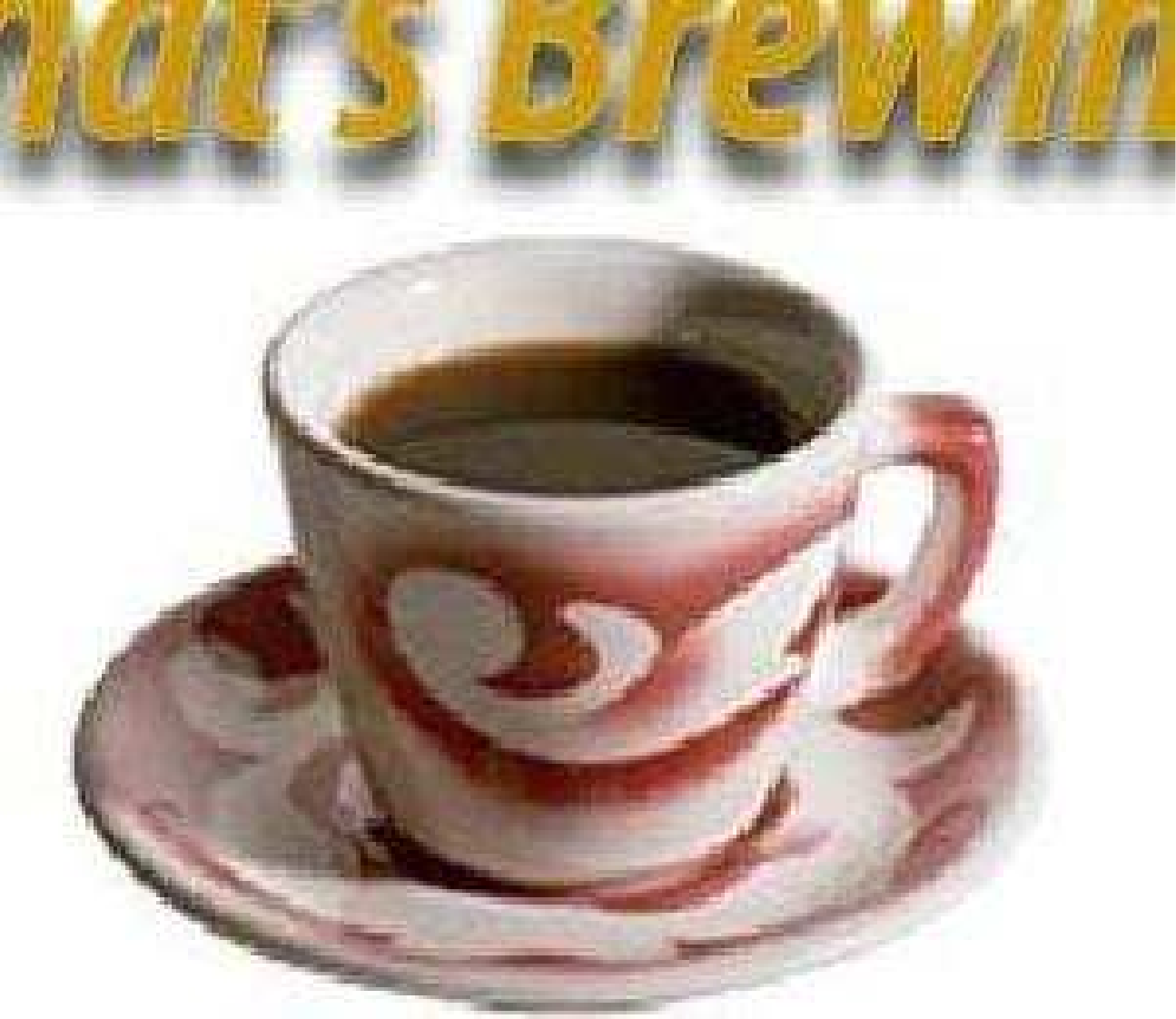}
	 \includegraphics[width=\figobjectmatchingw, height=\figobjectmatchingsw]{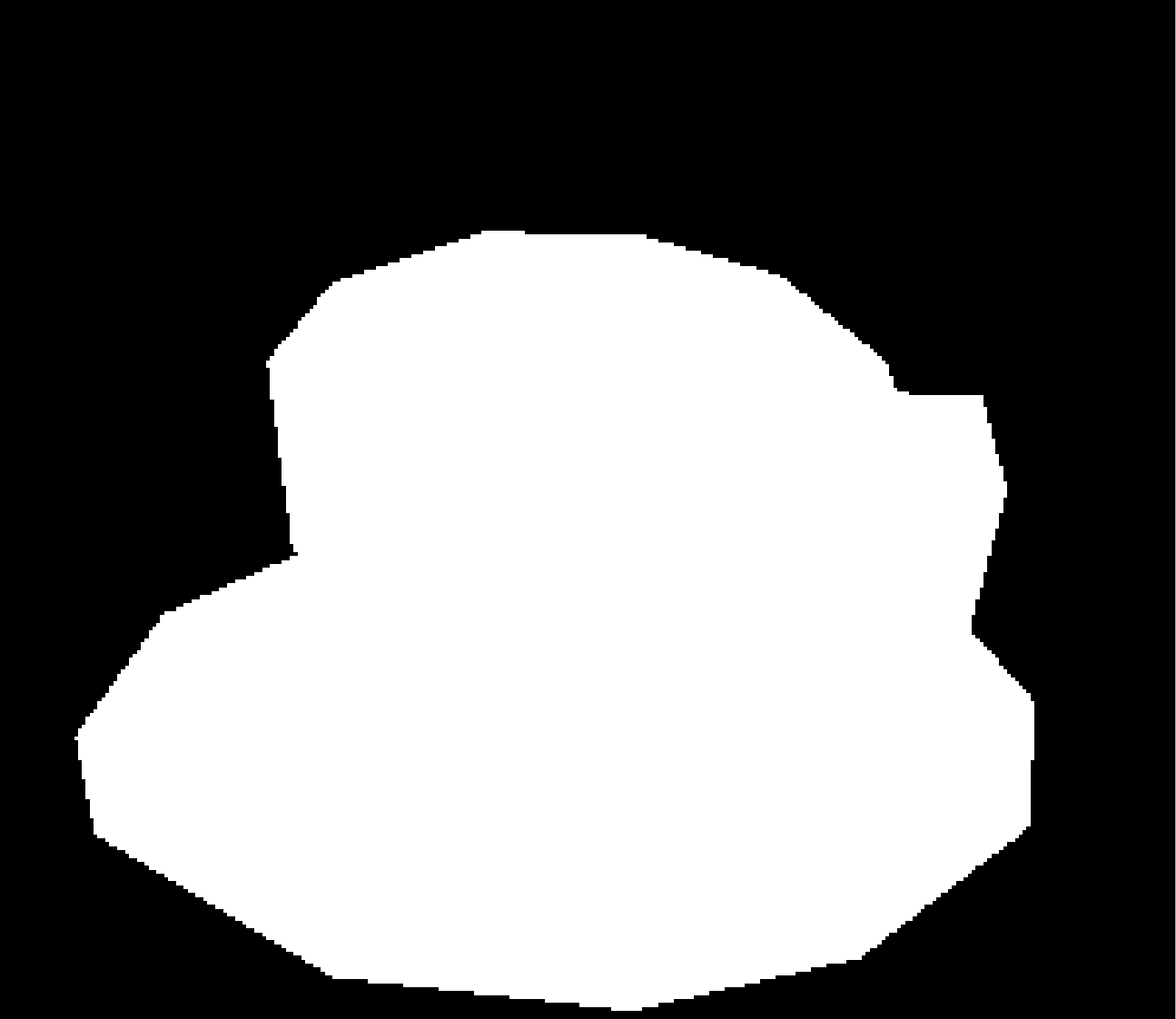}
	 \includegraphics[width=\figobjectmatchingw, height=\figobjectmatchingsw]{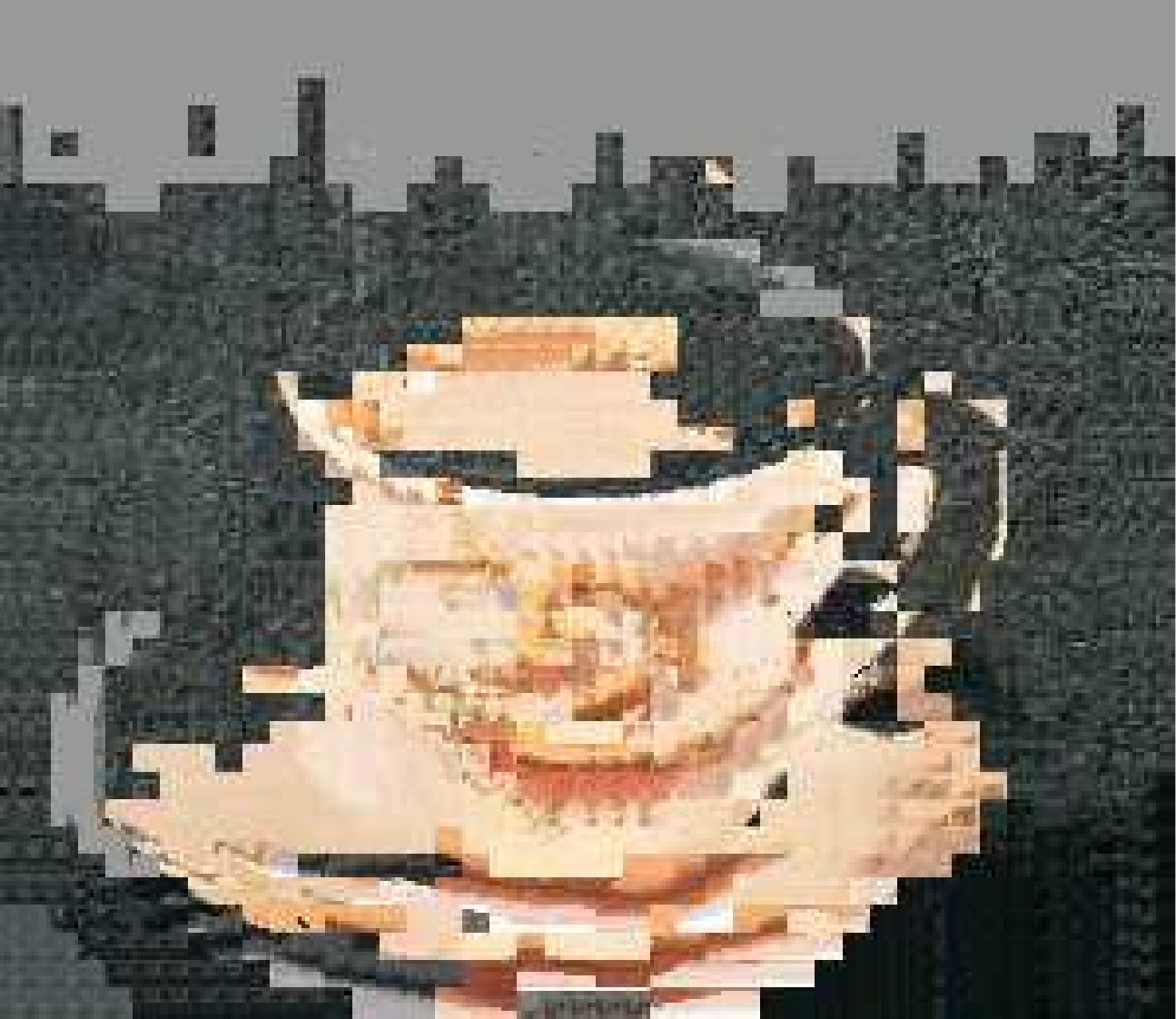}
     \includegraphics[width=\figobjectmatchingw, height=\figobjectmatchingsw]{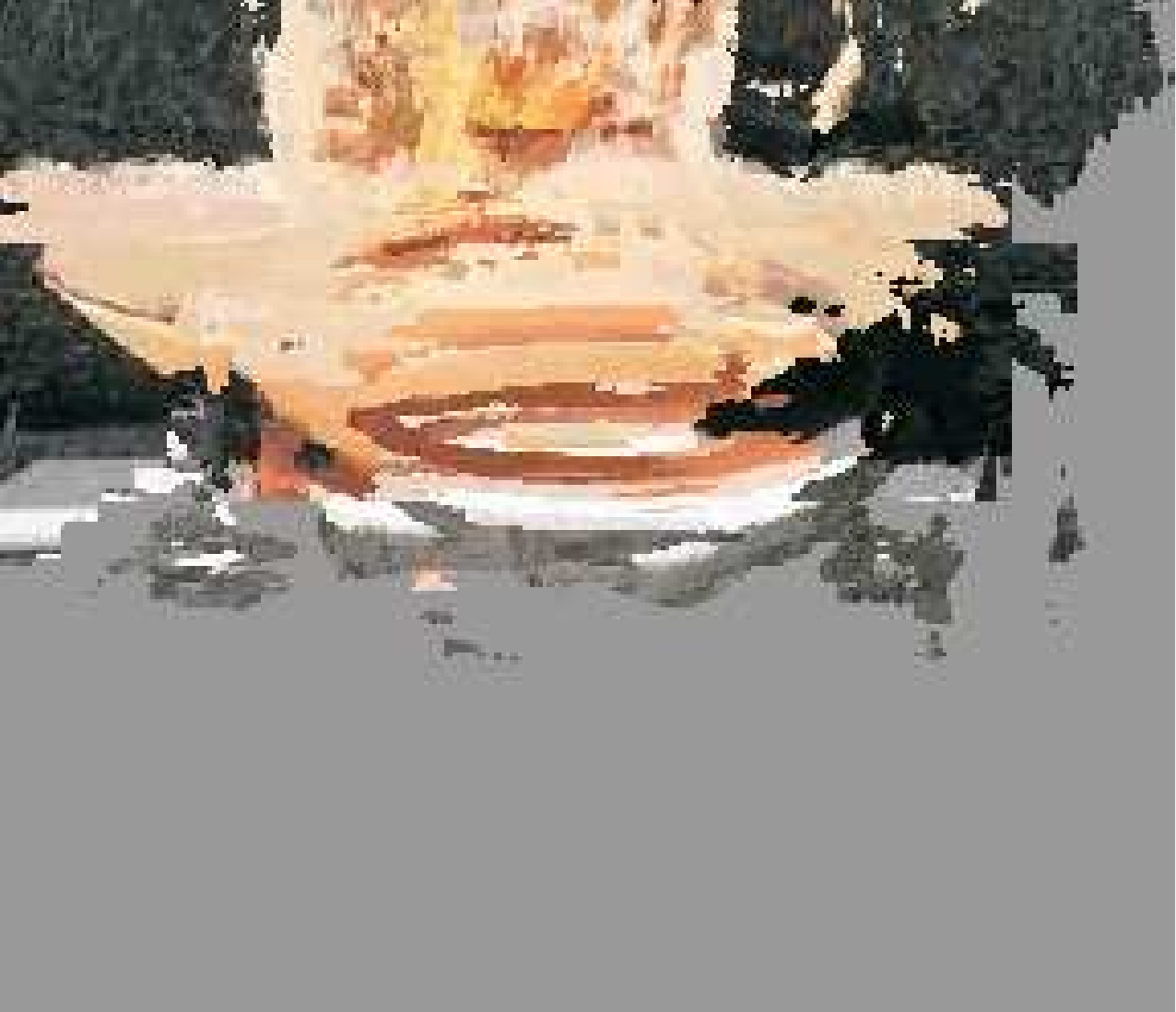}
     \includegraphics[width=\figobjectmatchingw, height=\figobjectmatchingsw]{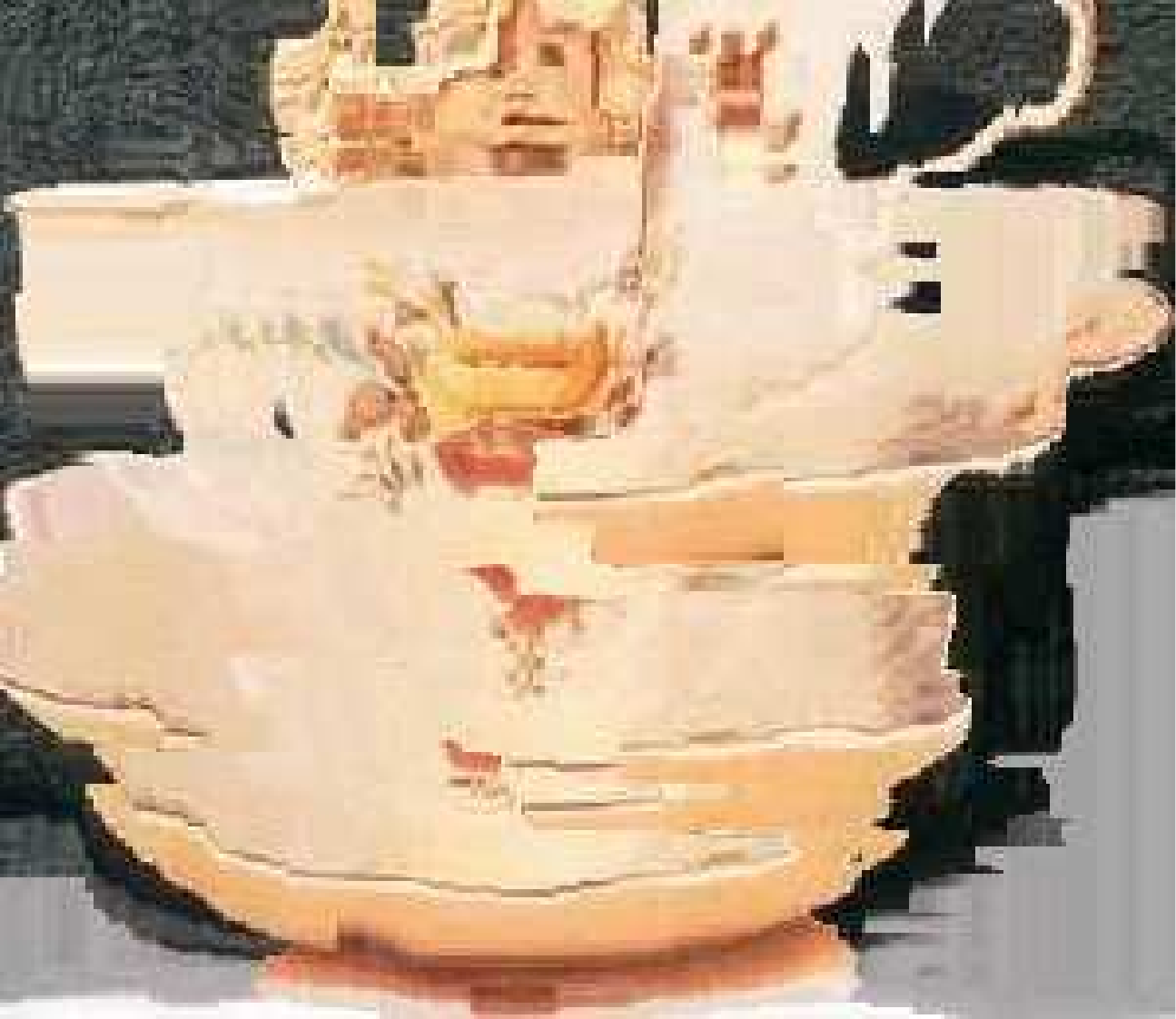}
     \includegraphics[width=\figobjectmatchingw, height=\figobjectmatchingsw]{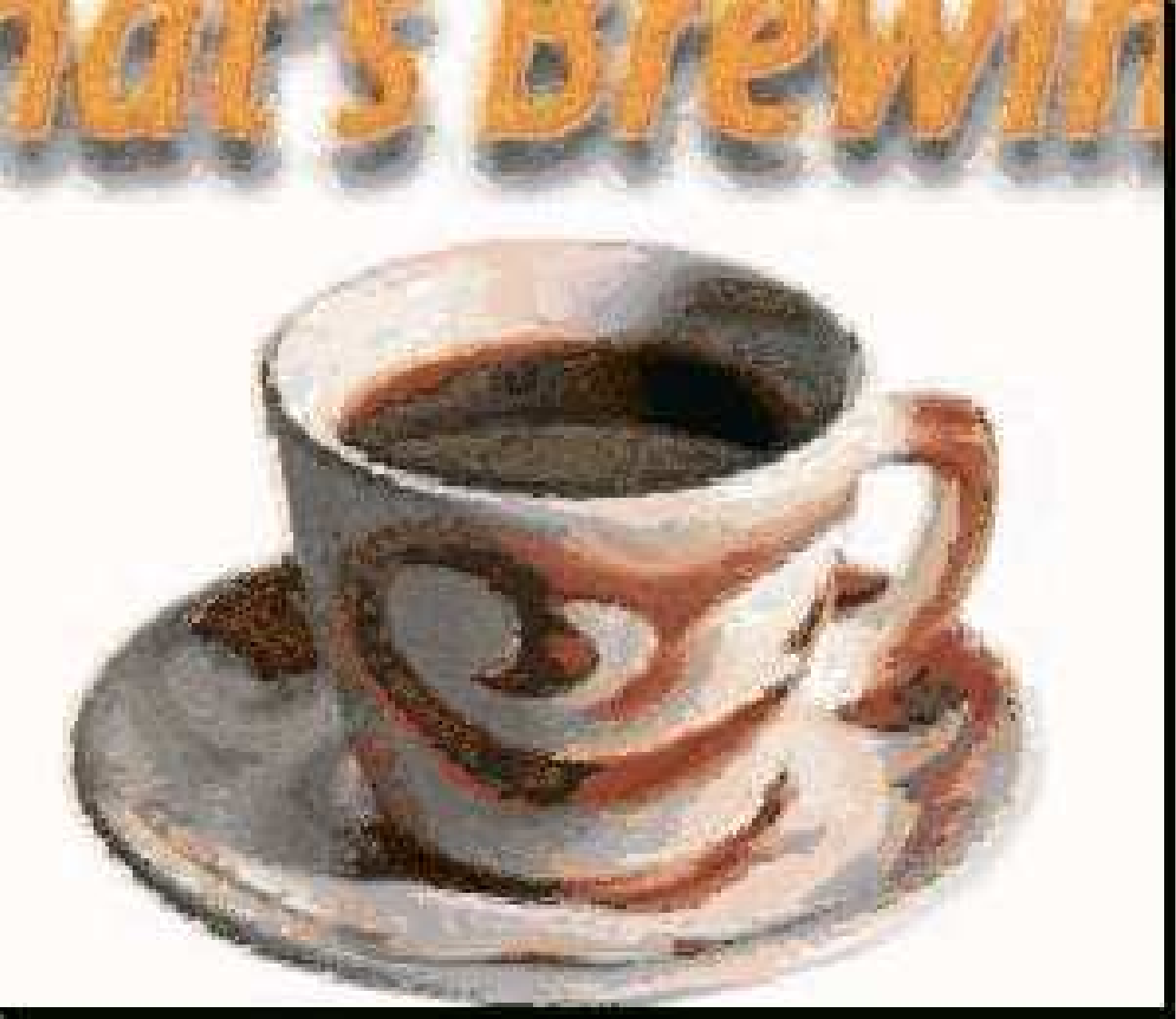} \\

	 \includegraphics[width=\figobjectmatchingw, height=\figobjectmatchingsw]{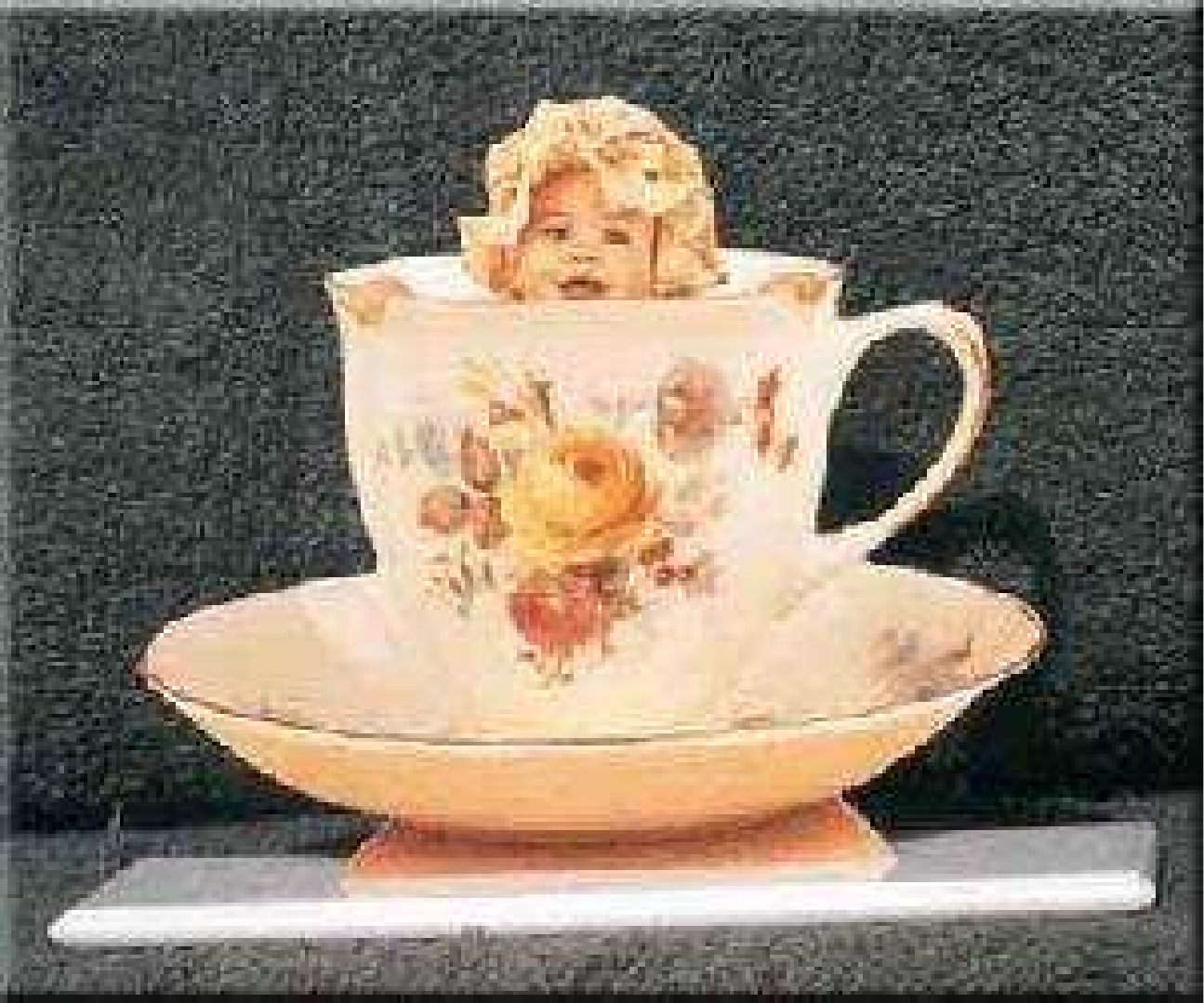}
	 \includegraphics[width=\figobjectmatchingw, height=\figobjectmatchingsw]{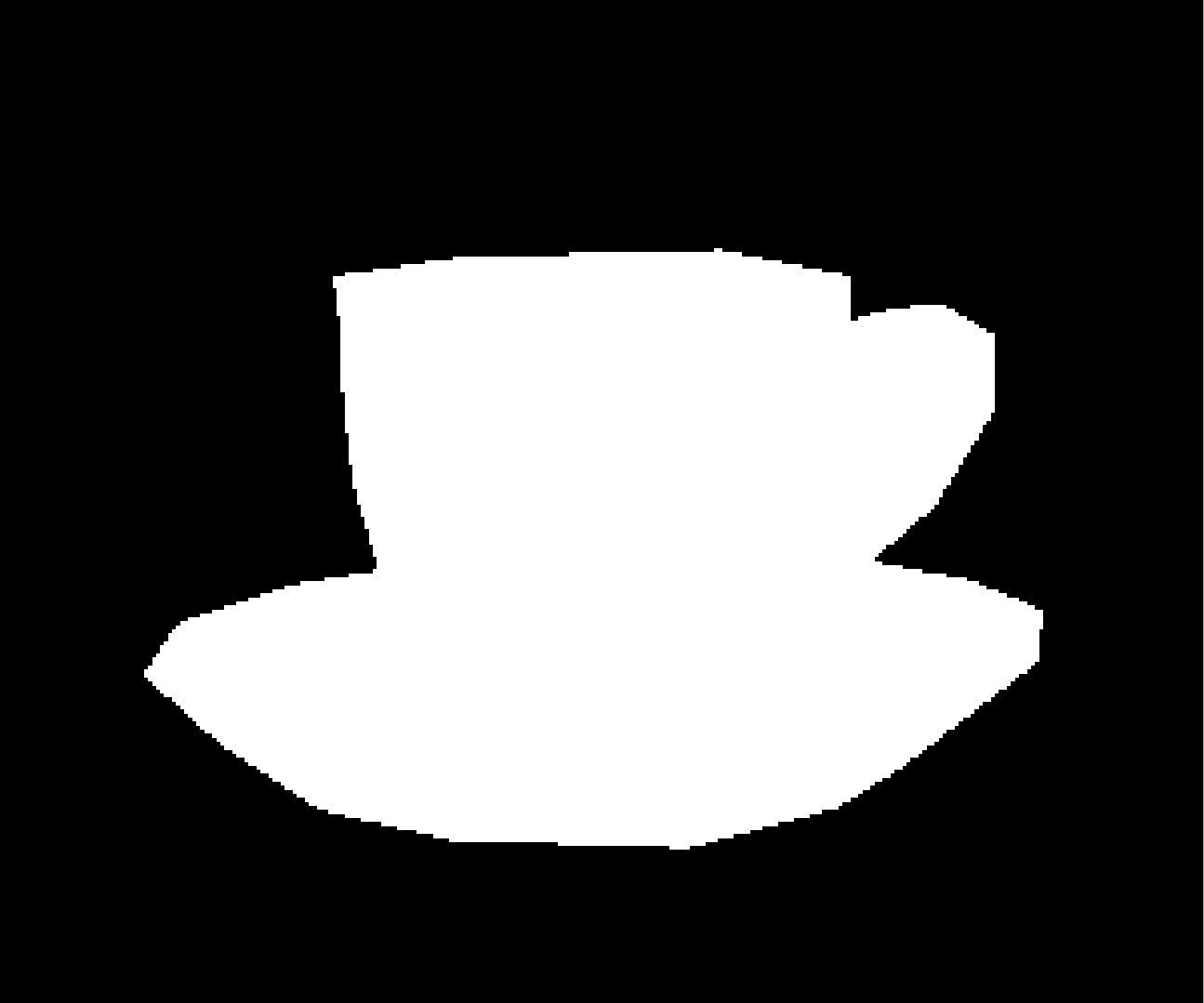}
	 \includegraphics[width=\figobjectmatchingw, height=\figobjectmatchingsw]{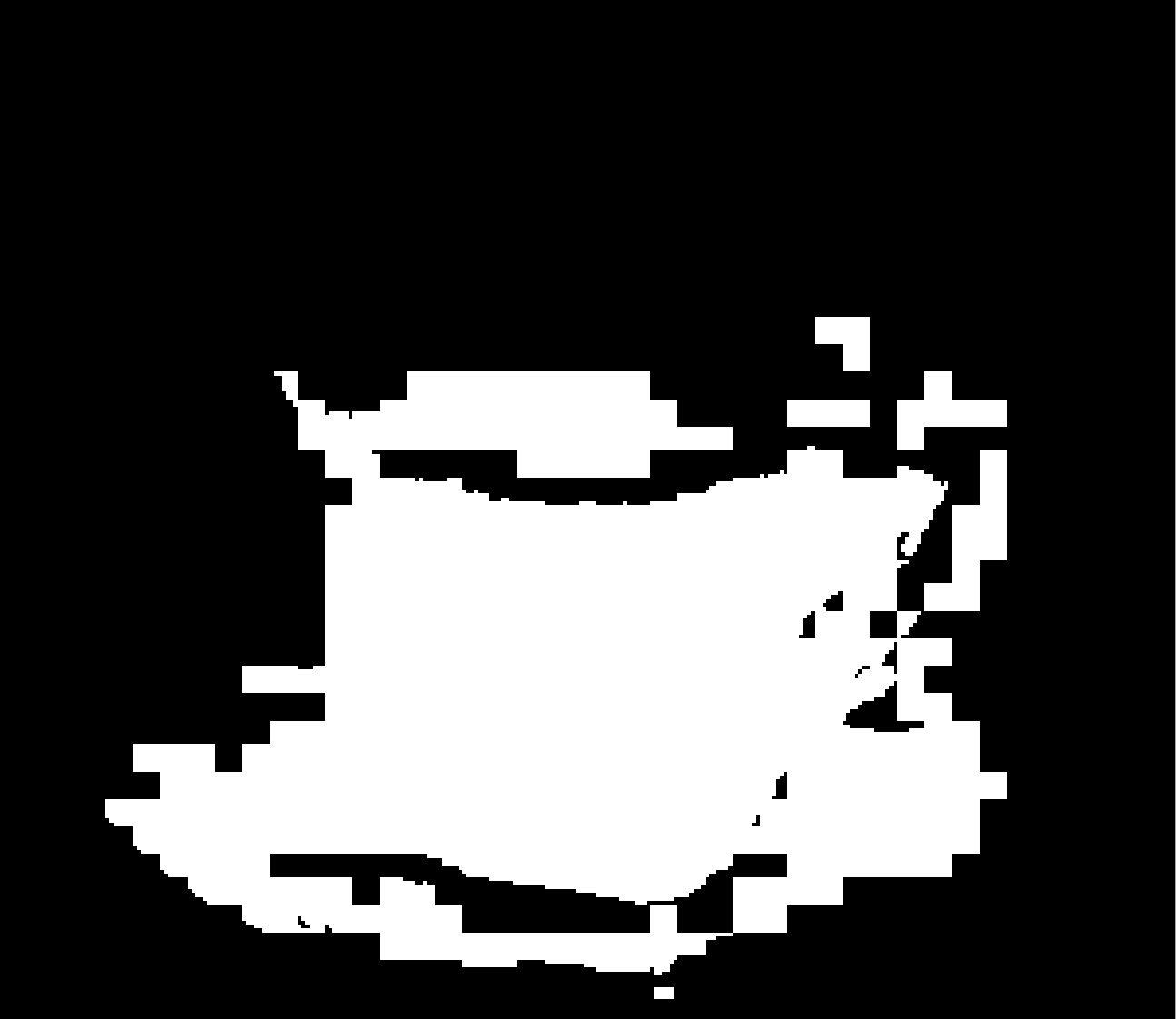}
     \includegraphics[width=\figobjectmatchingw, height=\figobjectmatchingsw]{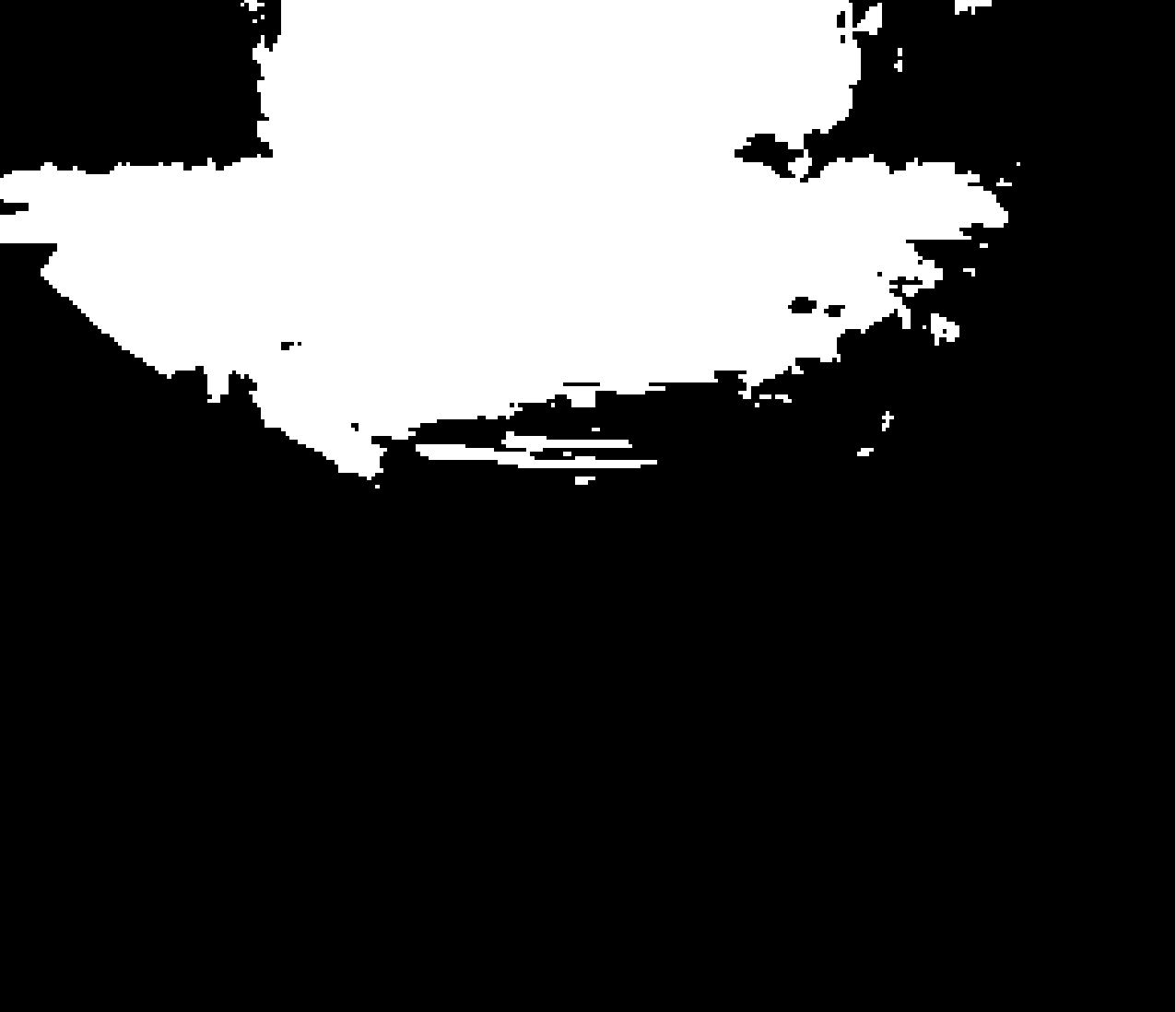}
     \includegraphics[width=\figobjectmatchingw, height=\figobjectmatchingsw]{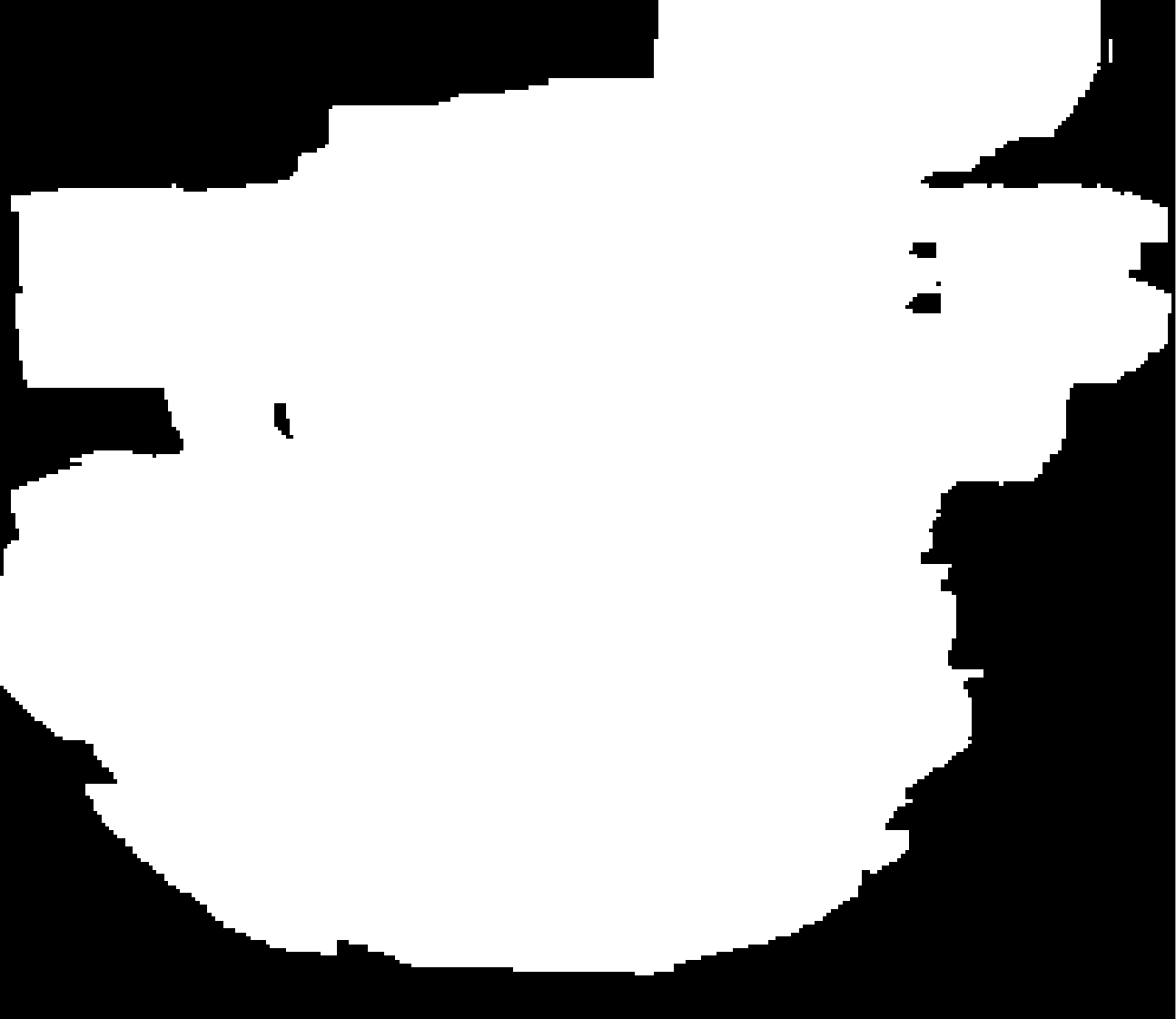}
     \includegraphics[width=\figobjectmatchingw, height=\figobjectmatchingsw]{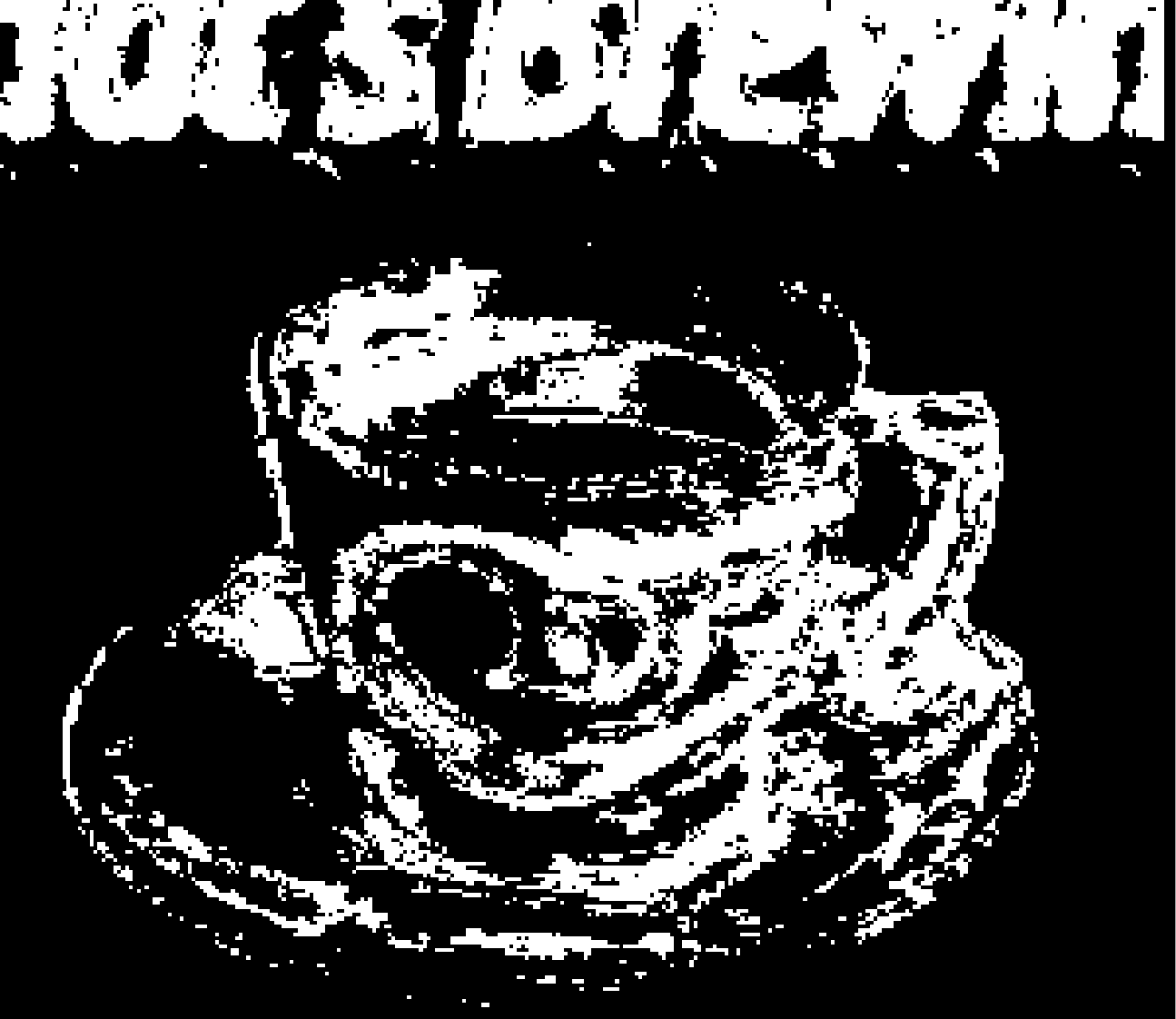} \\
\vspace{\vspacedistmid}
\caption{Qualitative comparison.
		 We show some example results of compared methods.
         The first column stacks matching image pairs.
         The second column shows ground-truth labels of input images.
         We establish the pixel correspondences for the top image
         of each image pair.
         Columns 3--6 show the warping results
		from the exemplar image (the bottom one in the pair) to
		the test image (top one) via pixel correspondences
		by our method, DSP, SIFT flow and CSH, respectively.
  Here   black and white colors indicate the labels of background and target.
}  \label{fig:object_matching_seg}
\vspace{\vspacedist}
\end{figure*}

Figure \ref{fig:object_matching_seg} shows some qualitative results of the compared methods.
The results show that our method is more robust than other methods under large object
appearance variations (e.g., the second example)
and cluttered backgrounds (e.g., the first and third examples).

The DSP and SIFT Flow methods are based on SIFT features which are robust to local geometric
and local affine distortions.
The SIFT Flow method enforces the matching smoothness between
the pixel and its neighboring pixels.
Due to the enforcement of pixels' connections,
the SIFT Flow matching results may exhibit large distortions.

The DSP method takes advantages of pyramid graph matching and
focuses on  the pixel level optimization efficiency.
The results show that the DSP method can achieve better results compared with SIFT Flow,
which is consistent with the results presented in \cite{kim2013deformable}.
Since DSP removes the neighboring-pixel connection at the pixel-level matching
optimization, pixels tend to match dispersedly beyond the object boundary.

The CSH method finds the patch matches mainly on patch appearance similarity
and does not rely on the image coherence assumption.
Their results are visually more pleasant (the warping
result is highly similar to the test image).
However, object pixels in a test image can easily incorrectly match
the background pixels in the exemplar image.
This causes the low label-transfer accuracy.

Our method takes advantages of these three methods.
Grid-cell layer matching in our method considers the matching cost and geometric regularization
in the pyramid of cells with different sizes,
and cell matching guides the patch-level matching.
Then the pixel matching refines the results of patch-level matching.
Figure \ref{fig:object_matching_seg} shows that our method not only
achieves accurate deformable matching
but also keeps the object's main shape.
The matching accuracy by using our patch-level matching
is  better than that of DSP, while being much faster.
As shown above, our patch-level results already outperform state-of-the-art results.
Our pixel-level matching in general provides even better matching accuracy with
more costly computation.

\newcommand{\figscenesw}{0.12981\textwidth}
\begin{figure*}
\vspace{\vspacedistfront}
\center
 {{Input \ \ \ \ \ \ \ \ \ \ \ GT label\ \ \ \ \ \ \ \ \ \  \ \ \ \ \ \
 Ours\ \ \ \ \ \ \ \ \ \ \ \ \ \ \ \ \ \  DSP\ \ \ \ \ \ \ \ \ \ \ \ \
 SF \ \ \ \ \ \ \ \ \ \ \ \ \  \ CSH \ \ }} \\
     \includegraphics[width=\figscenesw, height=\figscenesw]{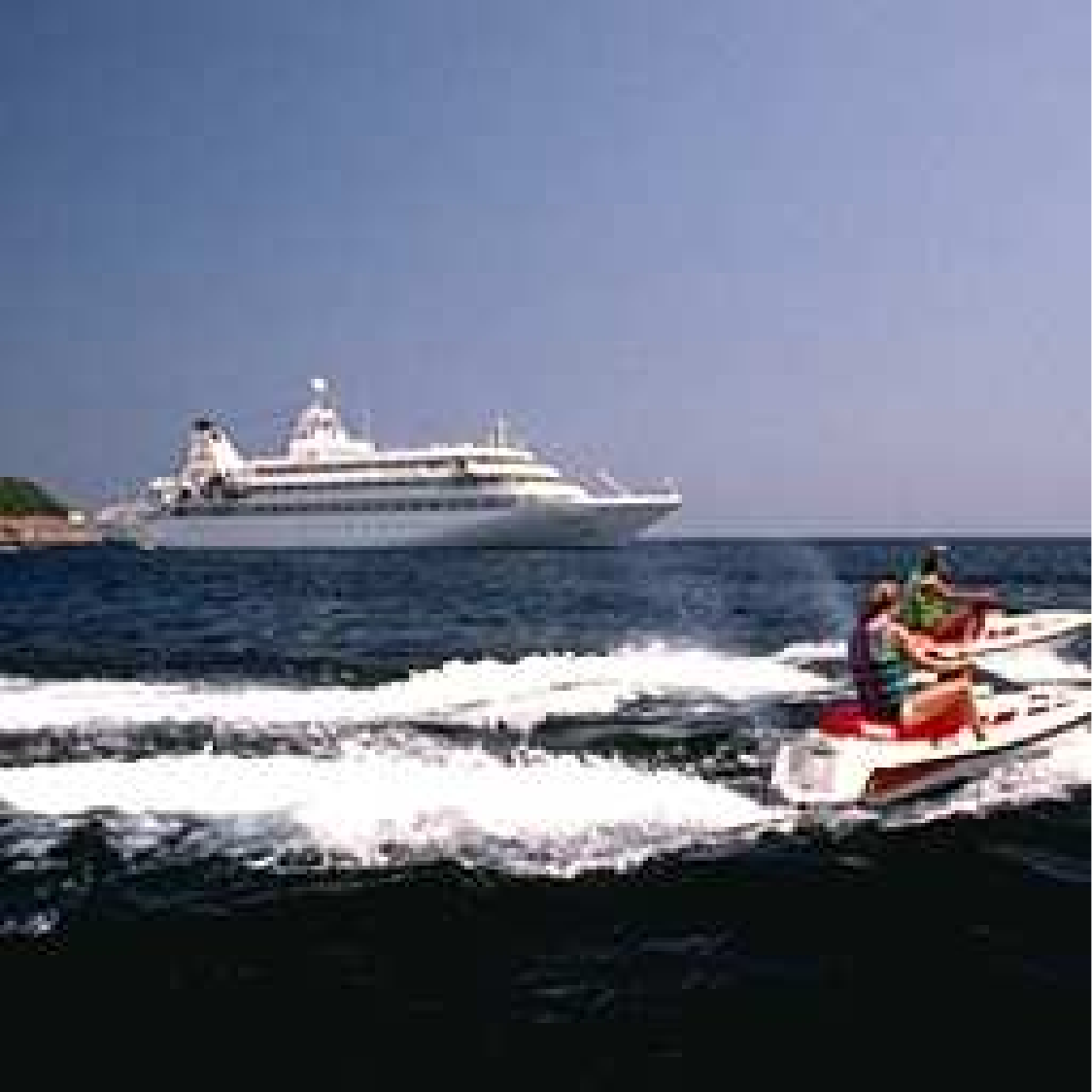}
	 \includegraphics[width=\figscenesw, height=\figscenesw]{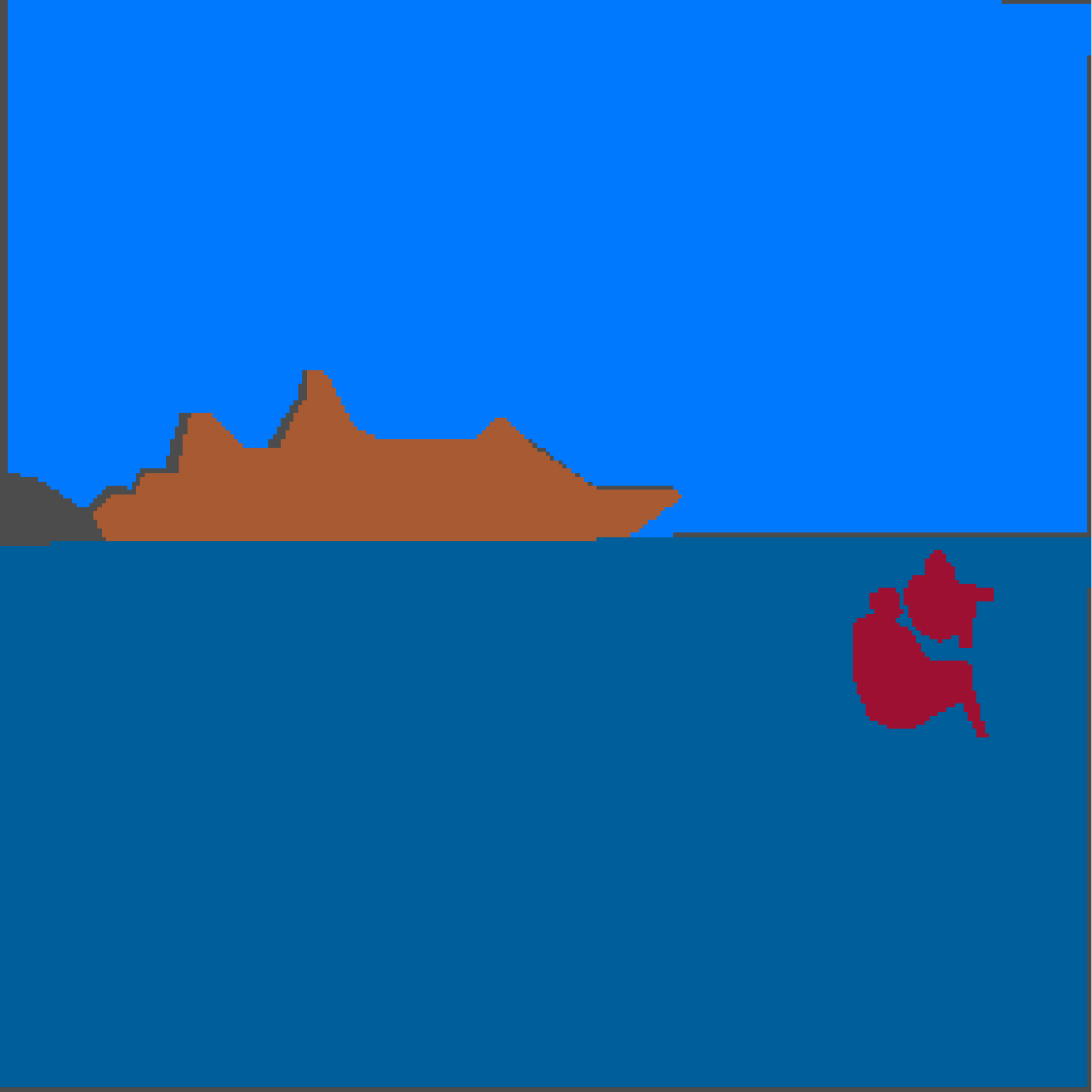}
	 \includegraphics[width=\figscenesw, height=\figscenesw]{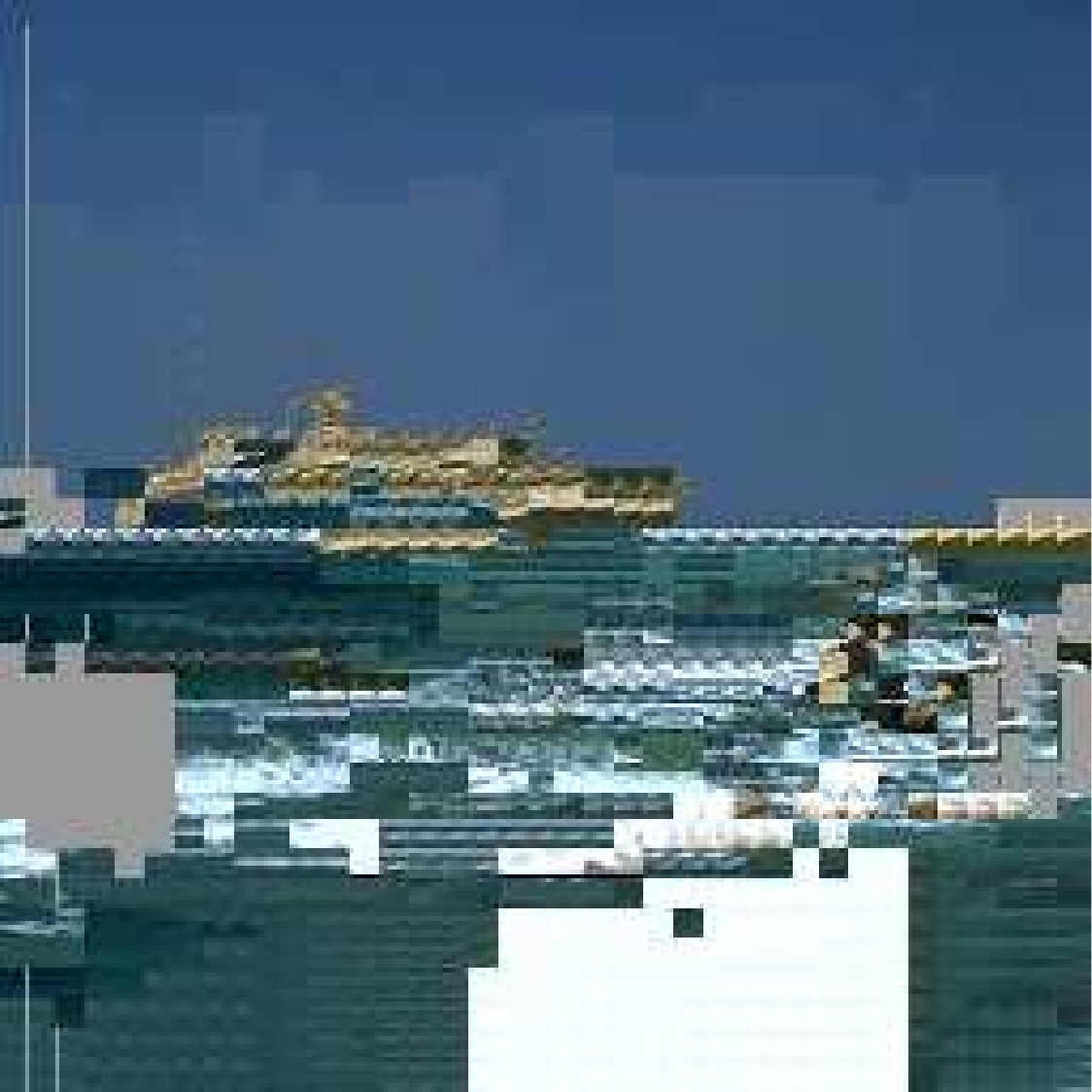}
	 \includegraphics[width=\figscenesw, height=\figscenesw]{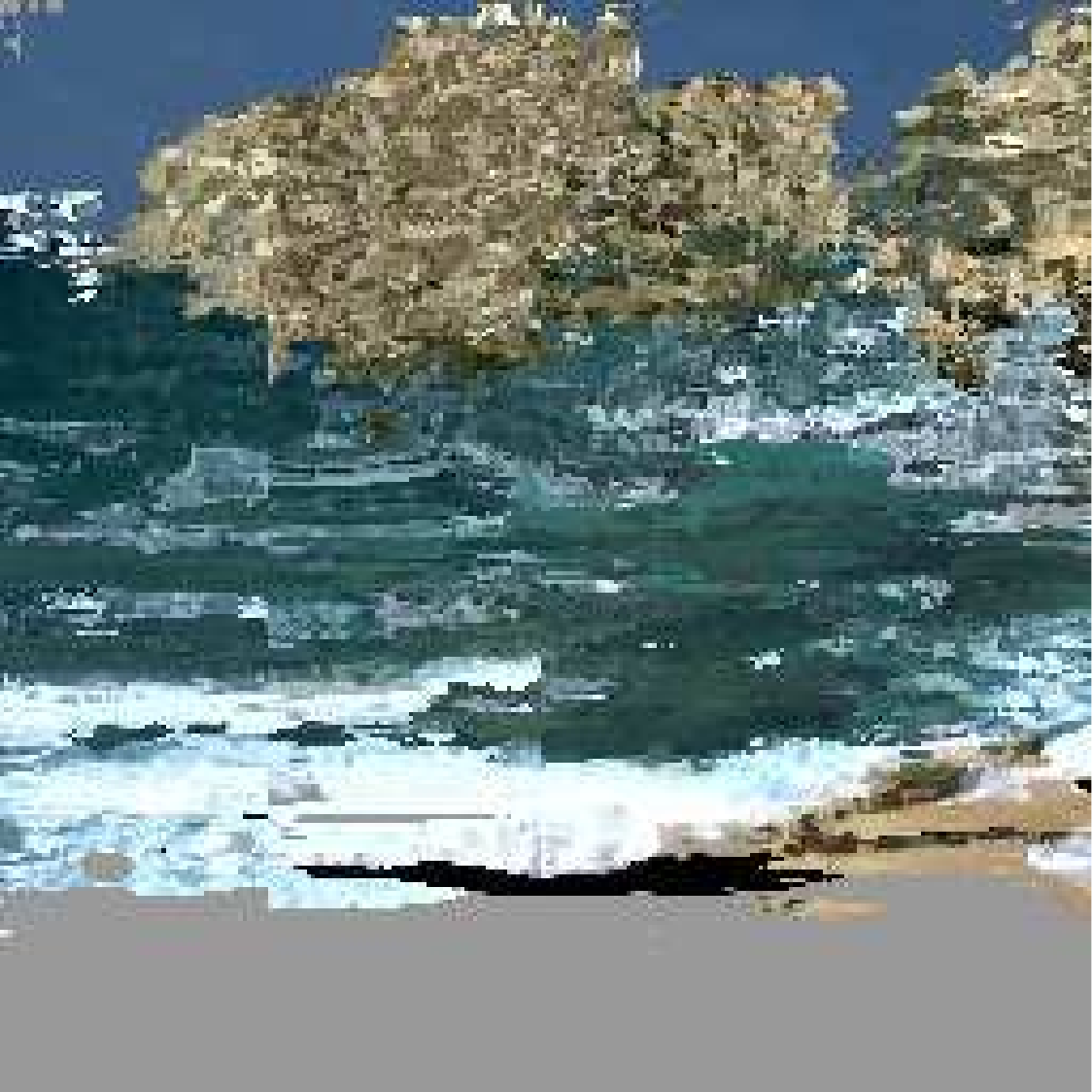}
     \includegraphics[width=\figscenesw, height=\figscenesw]{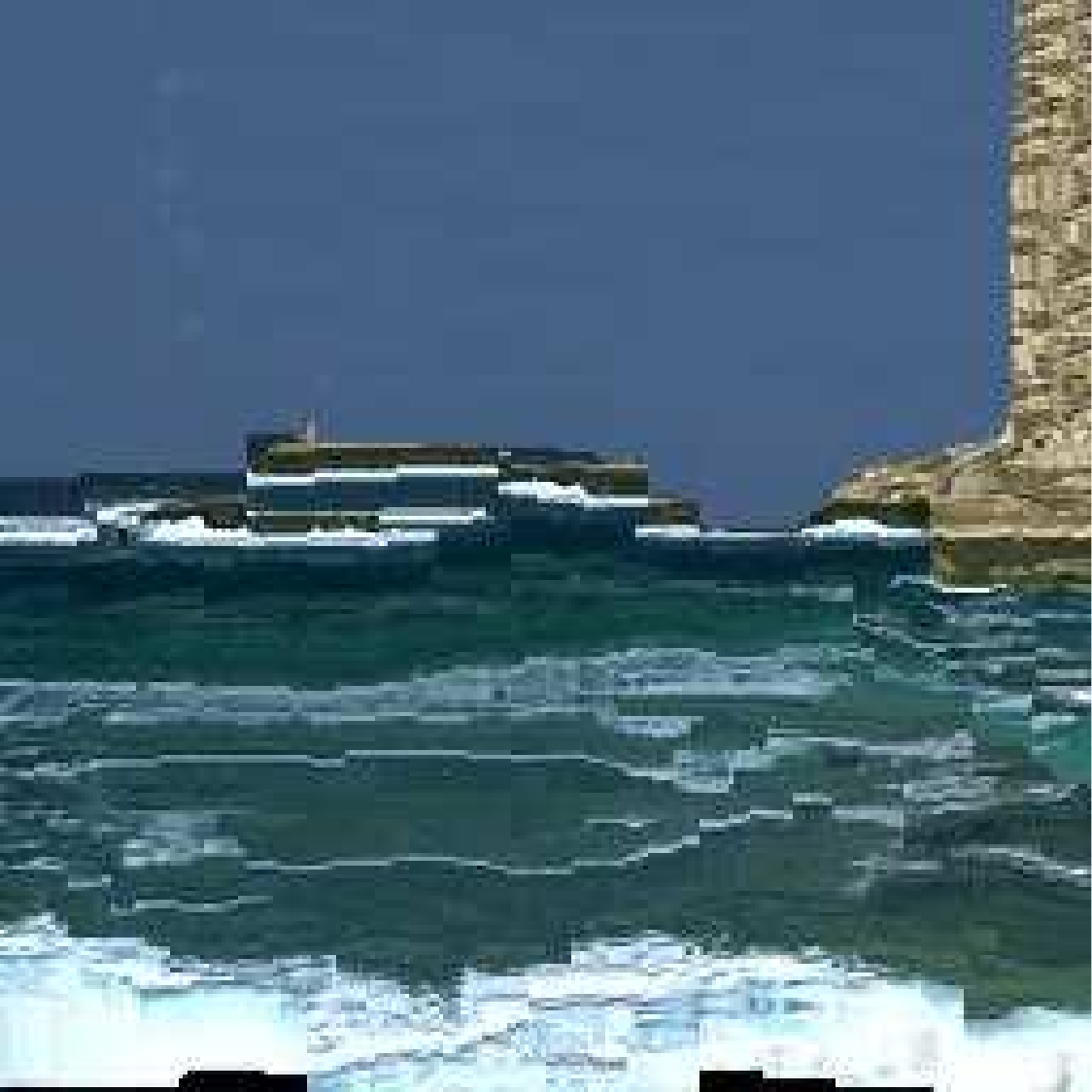}
     \includegraphics[width=\figscenesw, height=\figscenesw]{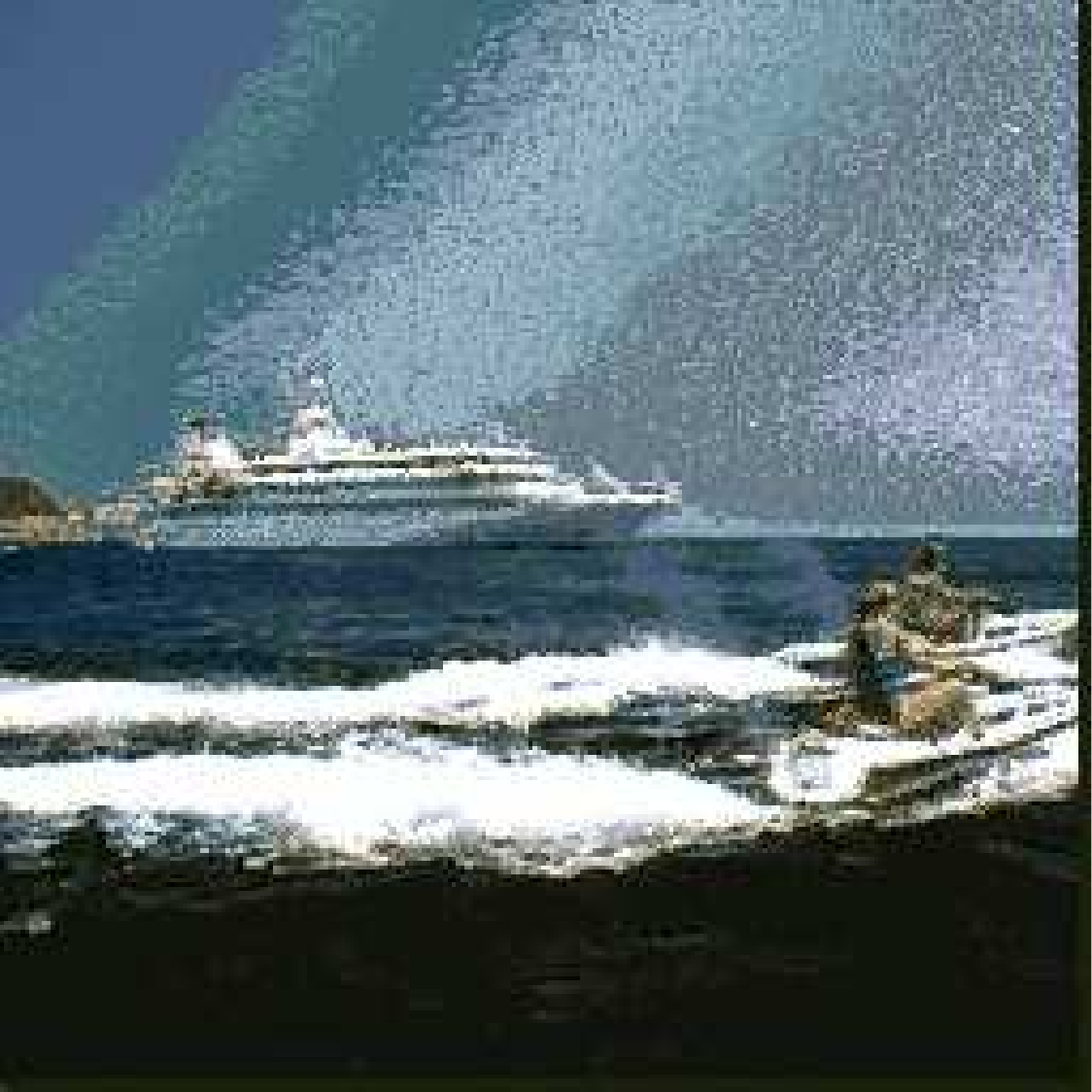} \\
     \includegraphics[width=\figscenesw, height=\figscenesw]{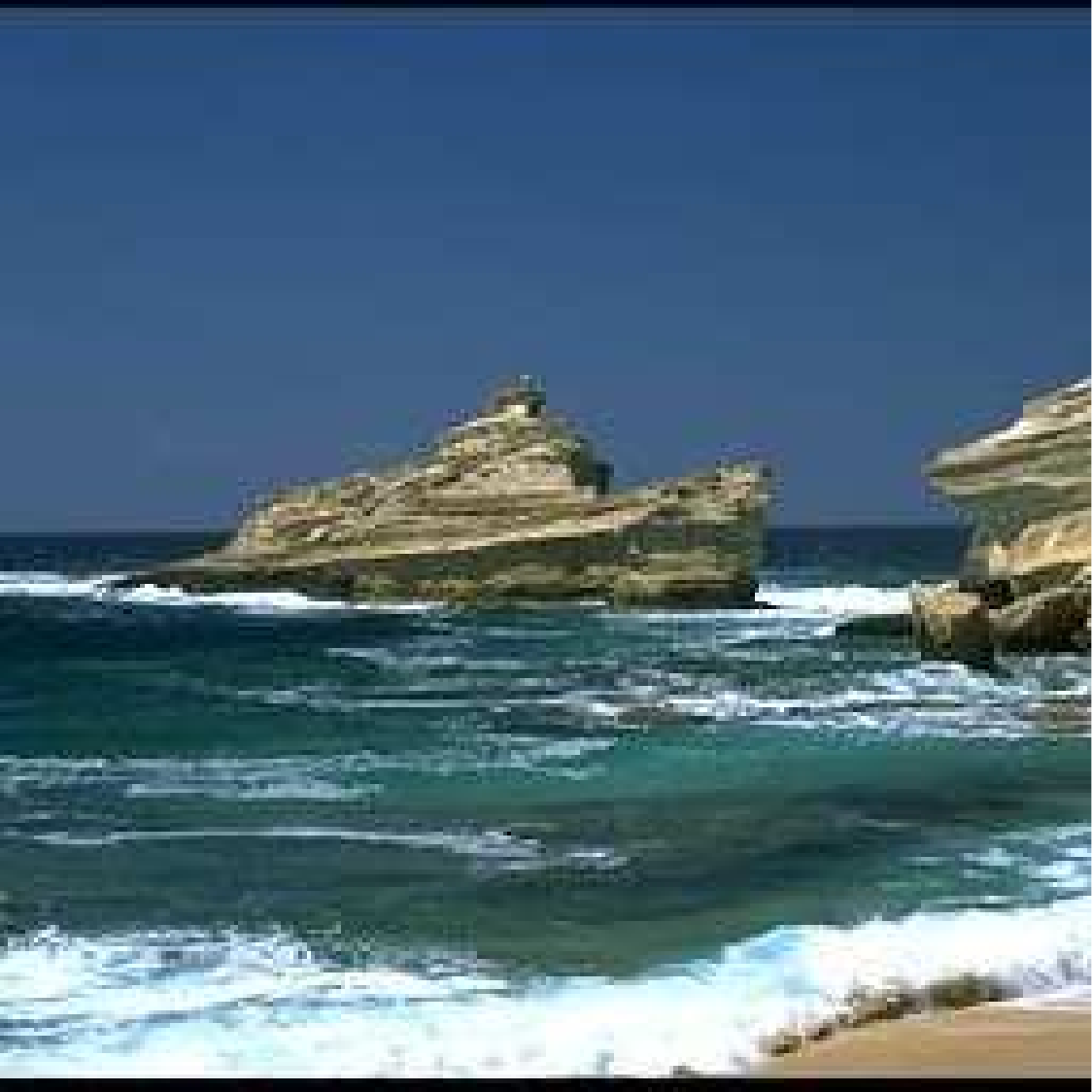}
	 \includegraphics[width=\figscenesw, height=\figscenesw]{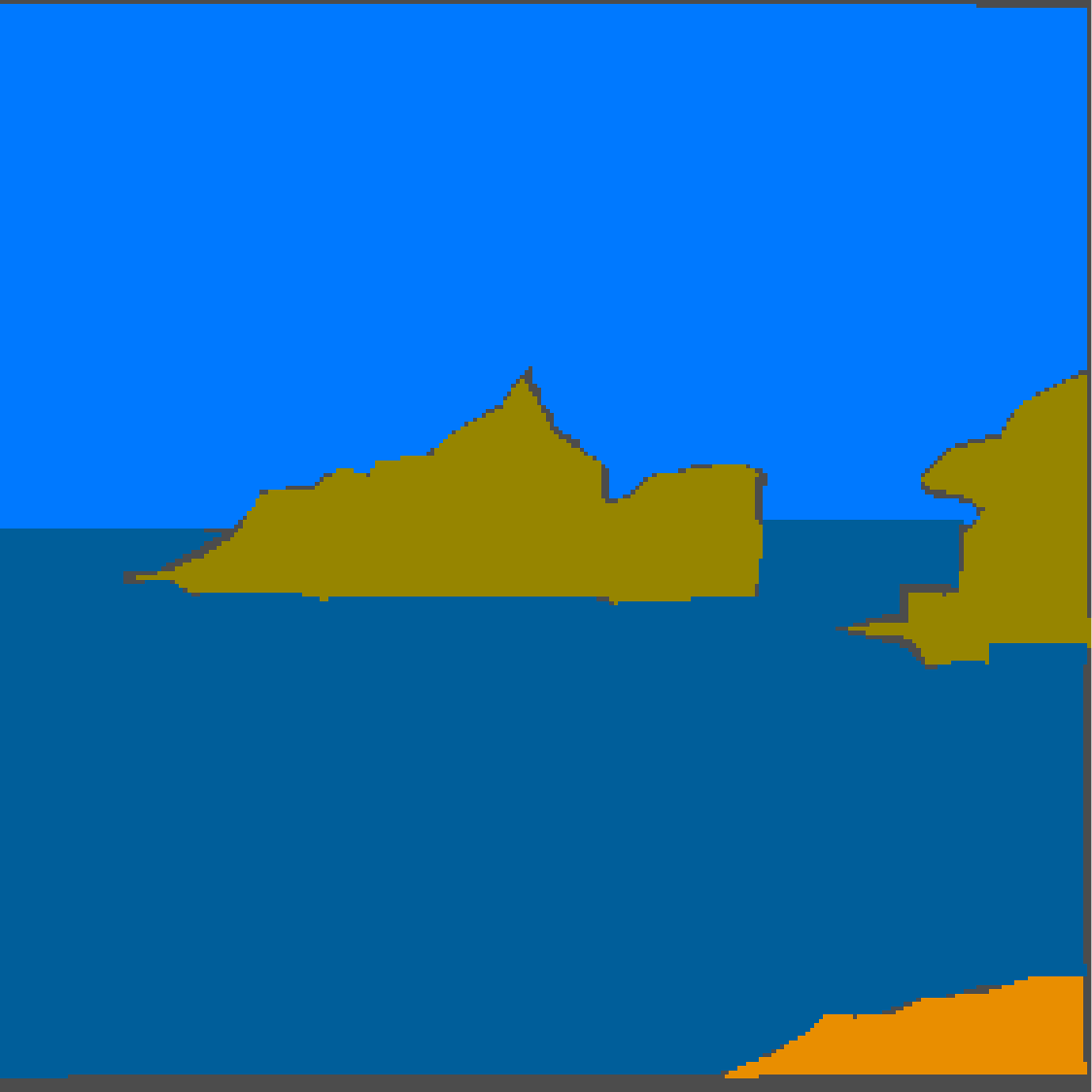}
	 \includegraphics[width=\figscenesw, height=\figscenesw]{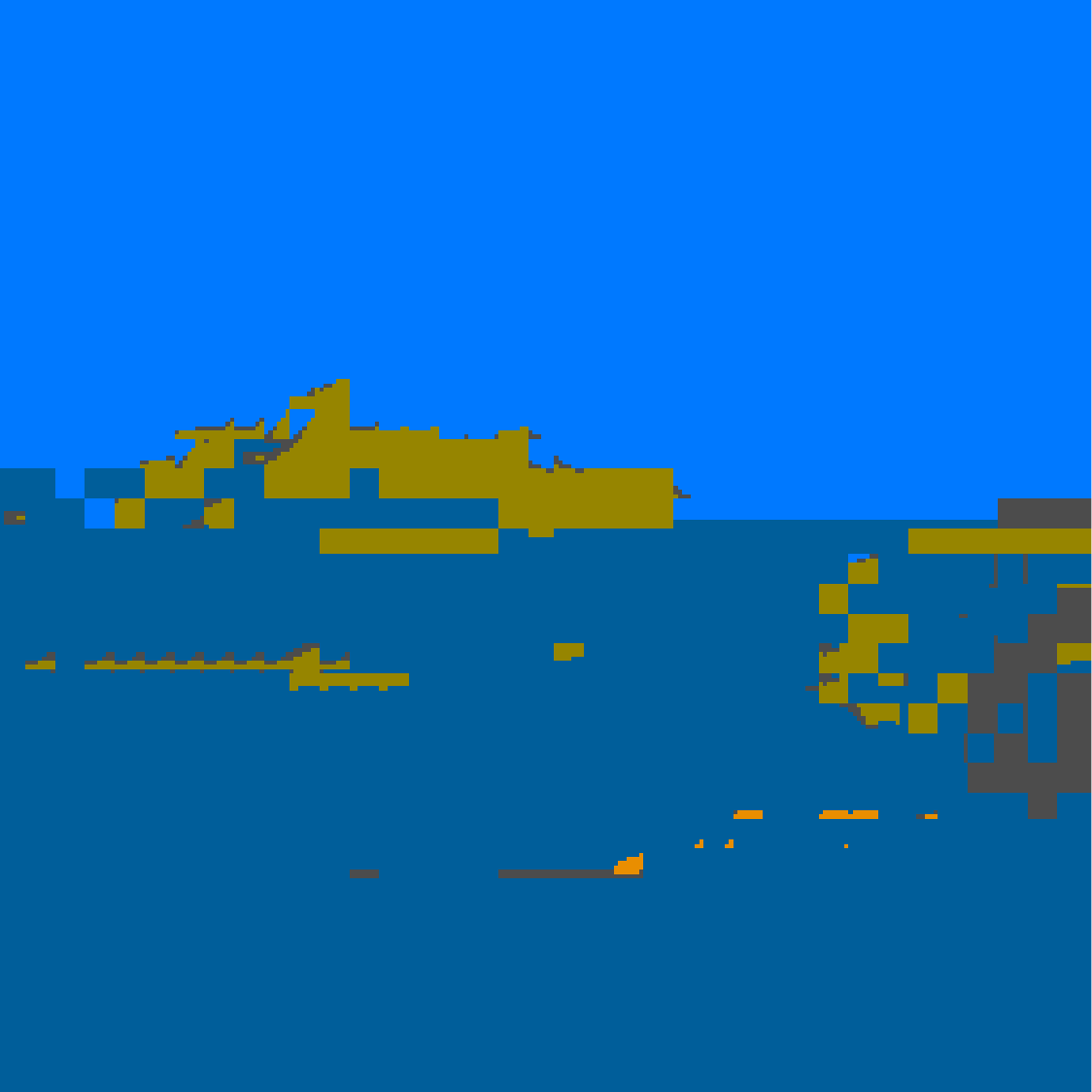}
	 \includegraphics[width=\figscenesw, height=\figscenesw]{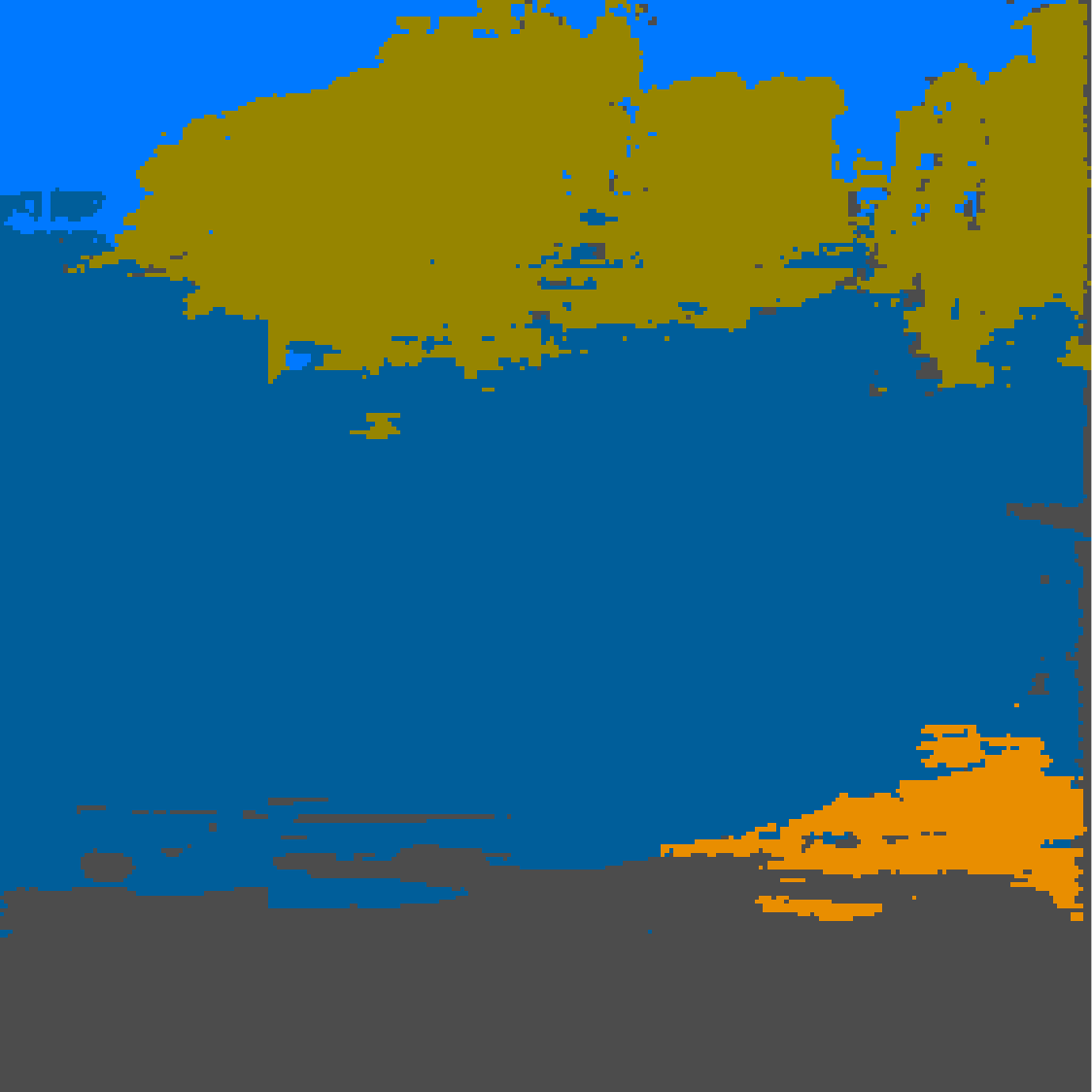}
     \includegraphics[width=\figscenesw, height=\figscenesw]{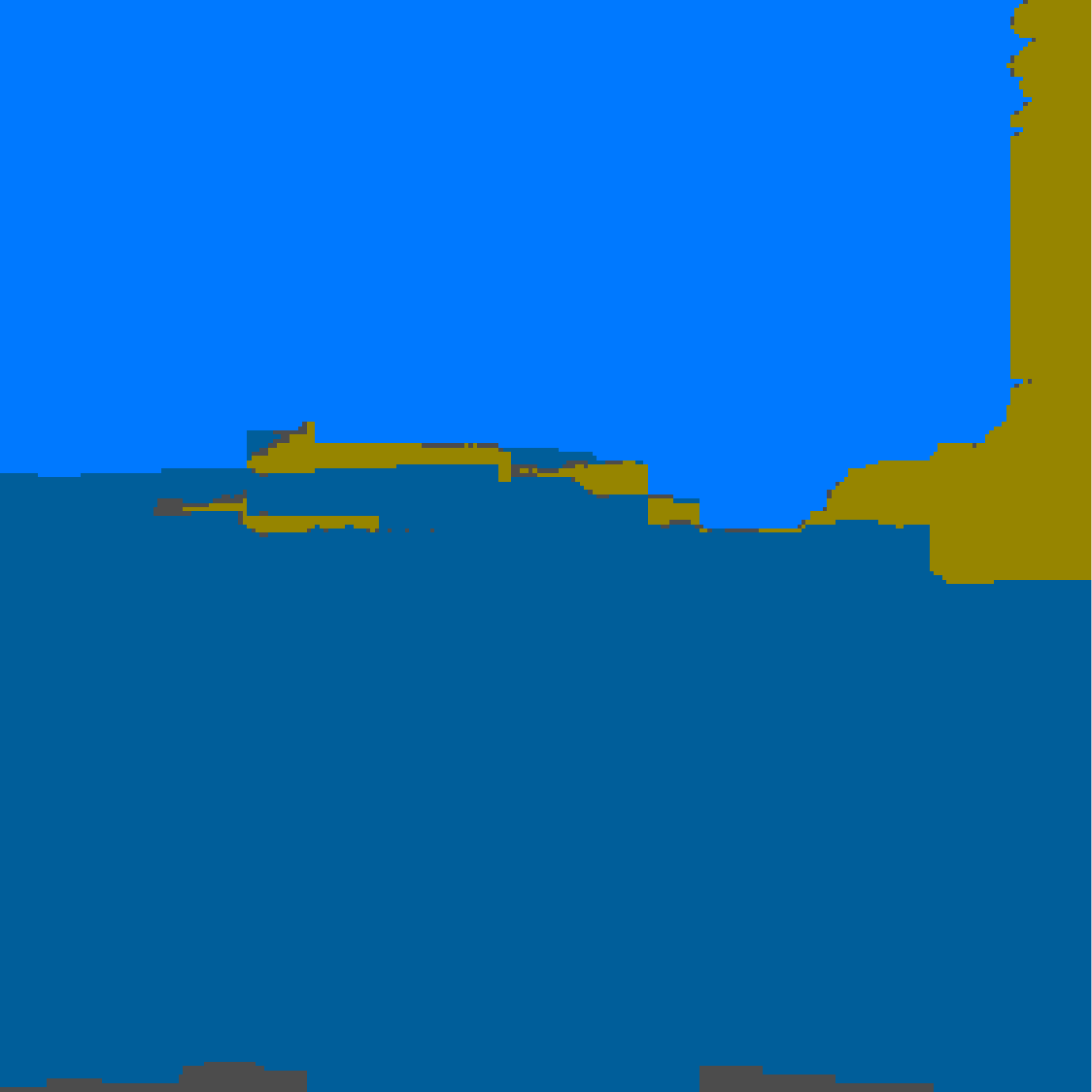}
     \includegraphics[width=\figscenesw, height=\figscenesw]{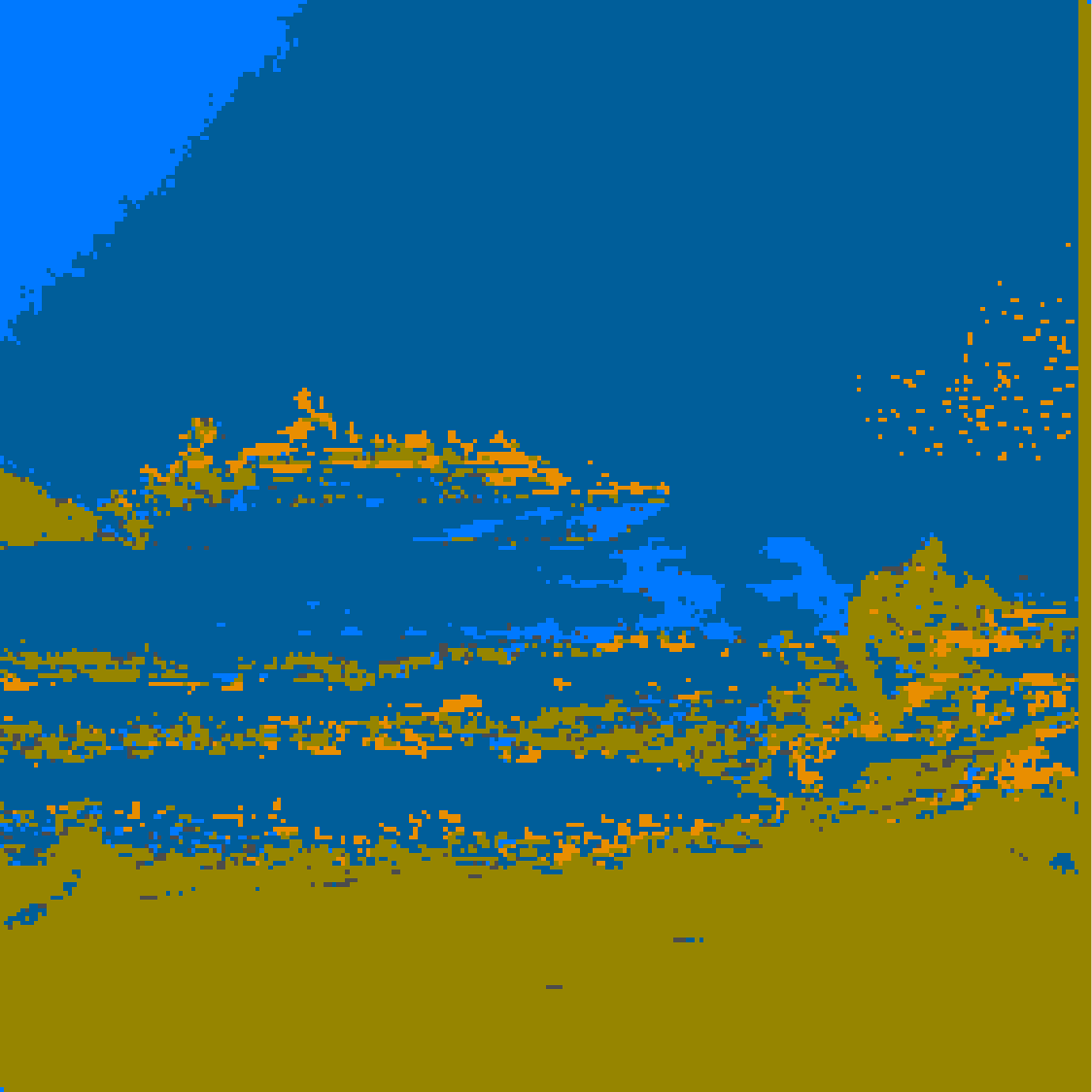} \\
     \includegraphics[width=\figscenesw, height=\figscenesw]{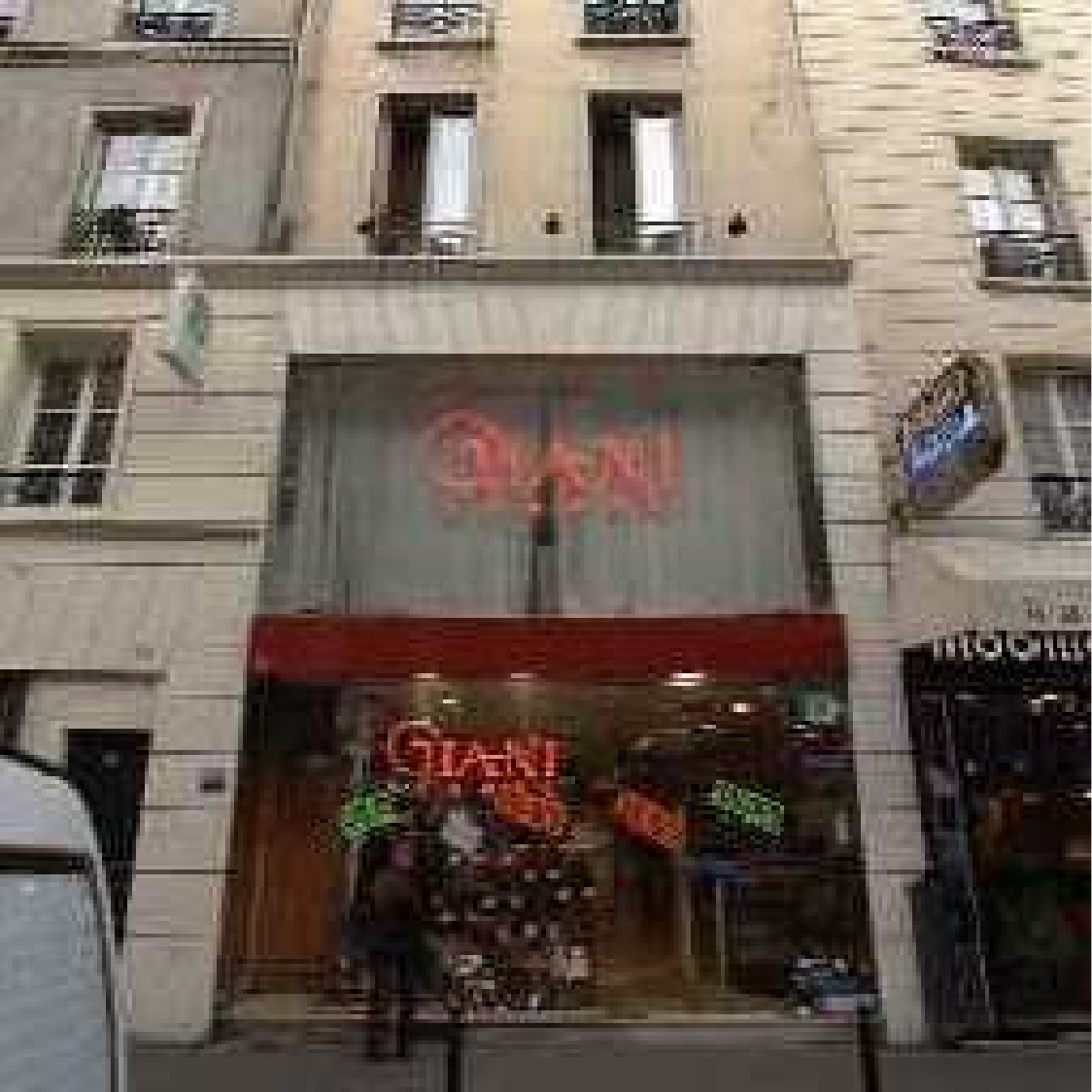}
	 \includegraphics[width=\figscenesw, height=\figscenesw]{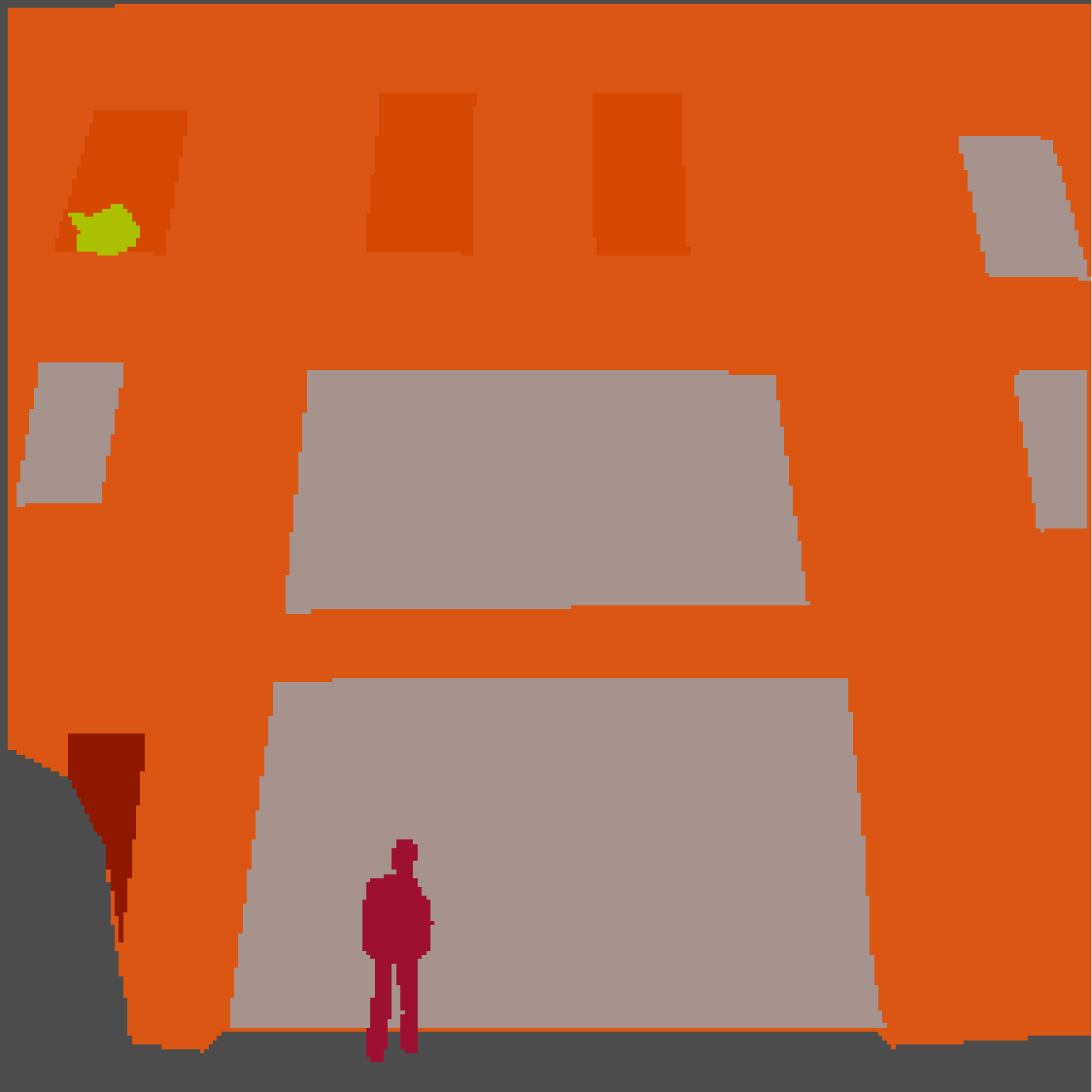}
	 \includegraphics[width=\figscenesw, height=\figscenesw]{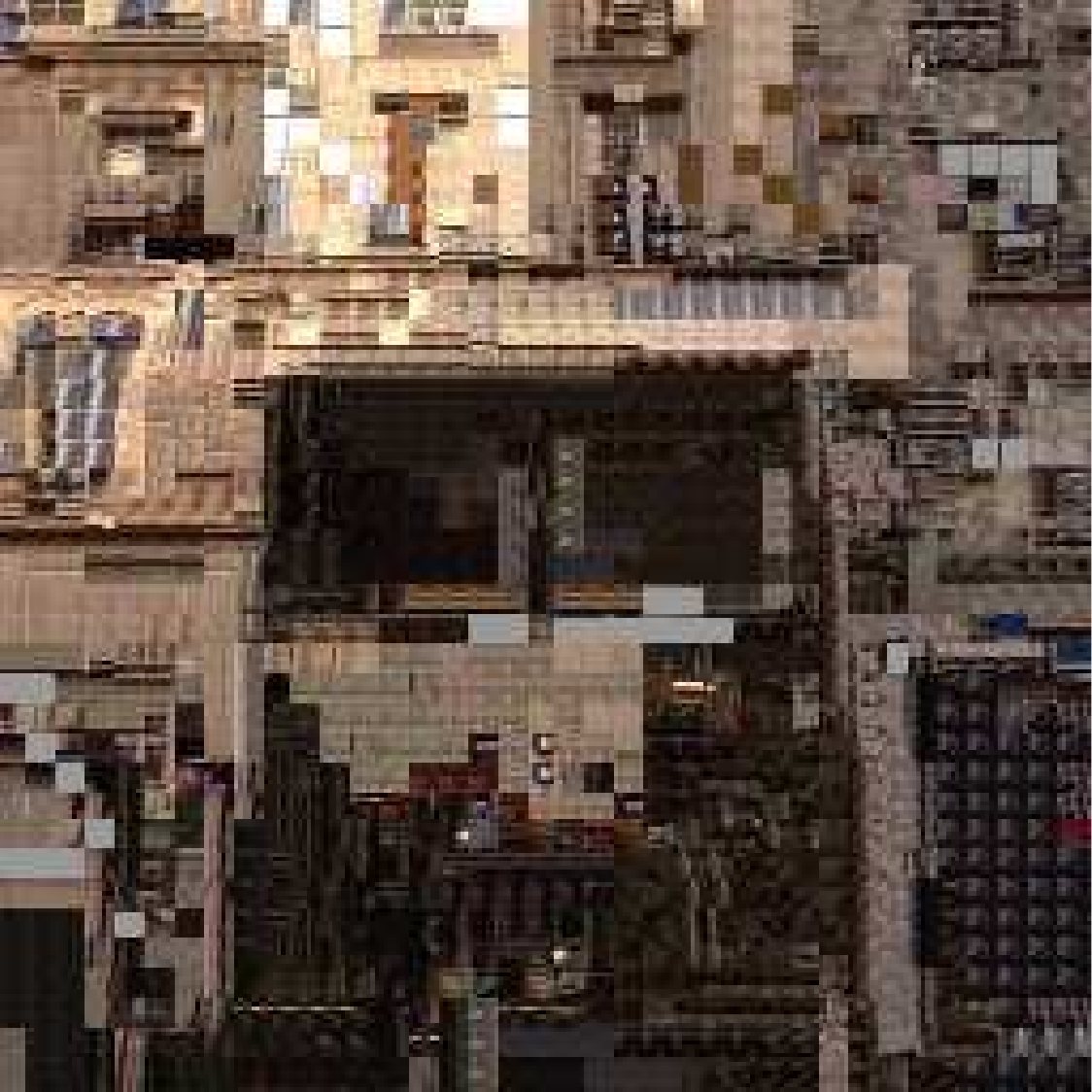}
	 \includegraphics[width=\figscenesw, height=\figscenesw]{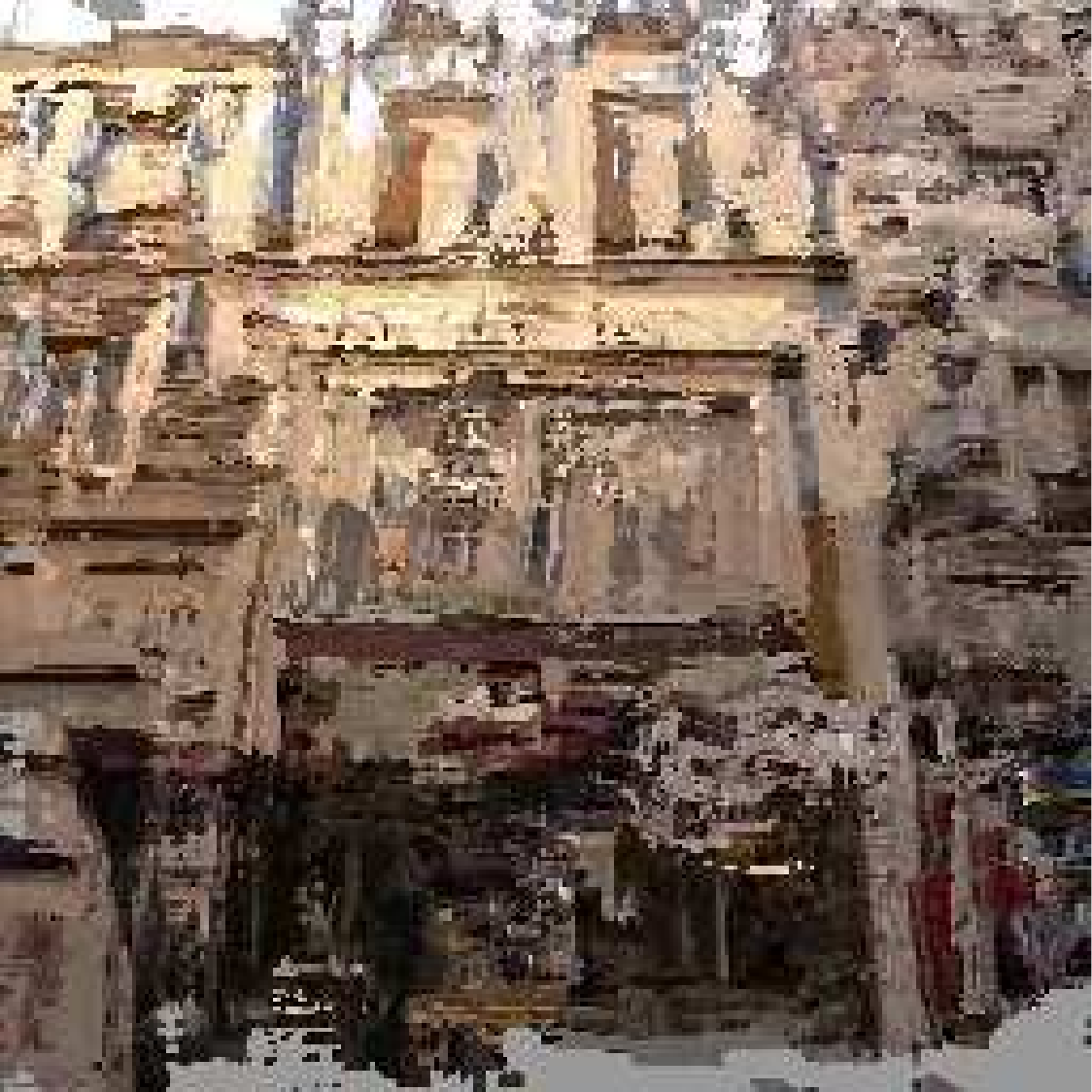}
     \includegraphics[width=\figscenesw, height=\figscenesw]{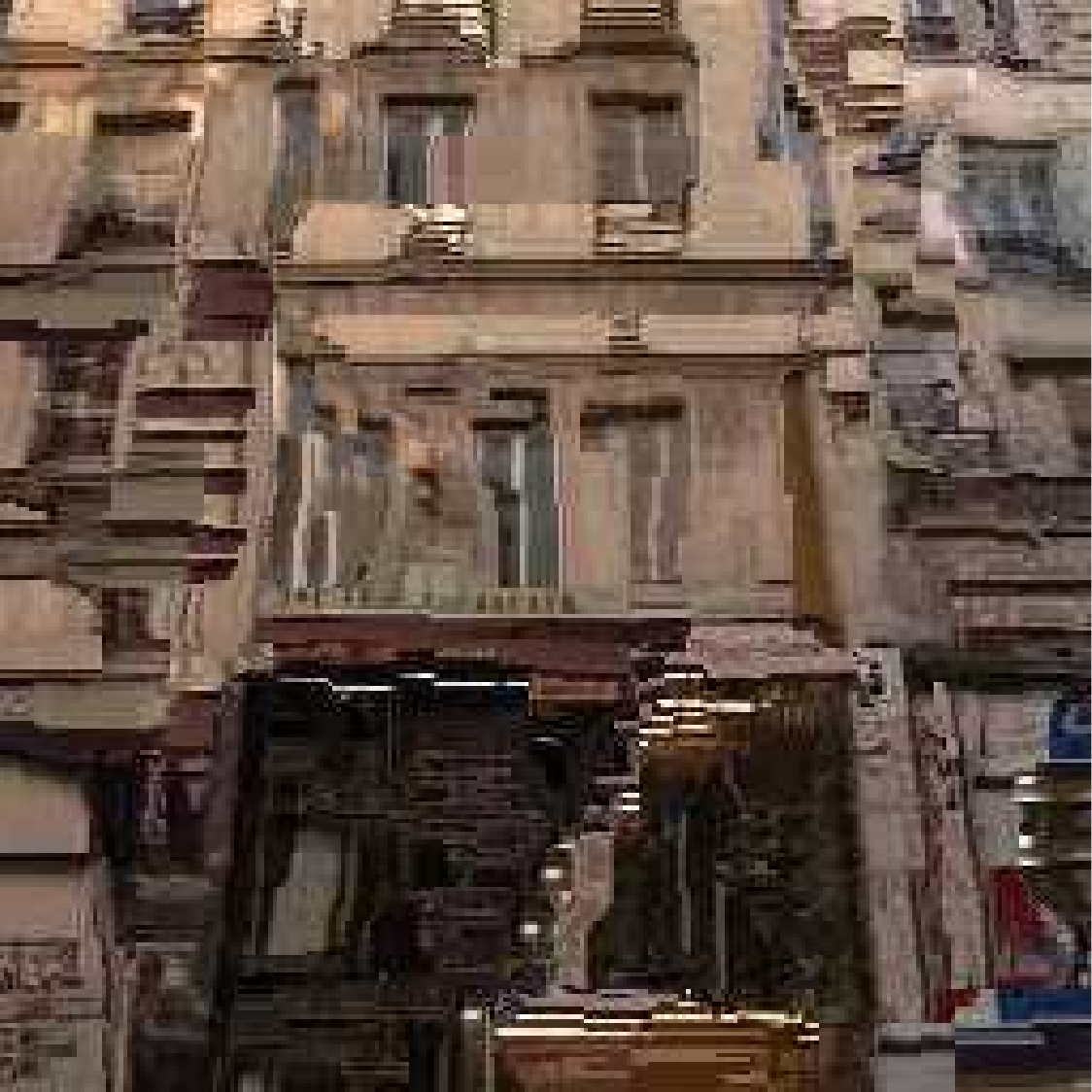}
     \includegraphics[width=\figscenesw, height=\figscenesw]{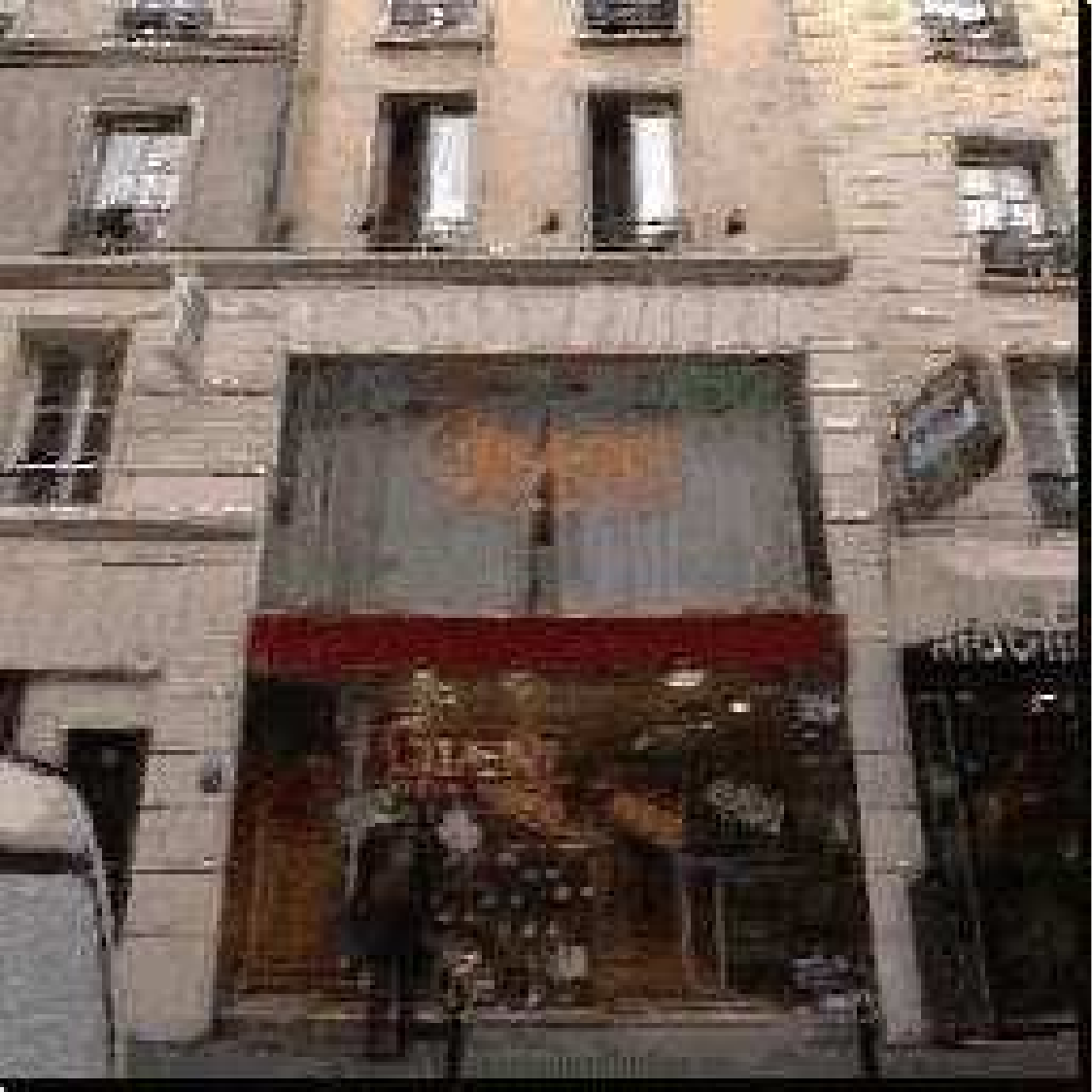} \\
     \includegraphics[width=\figscenesw, height=\figscenesw]{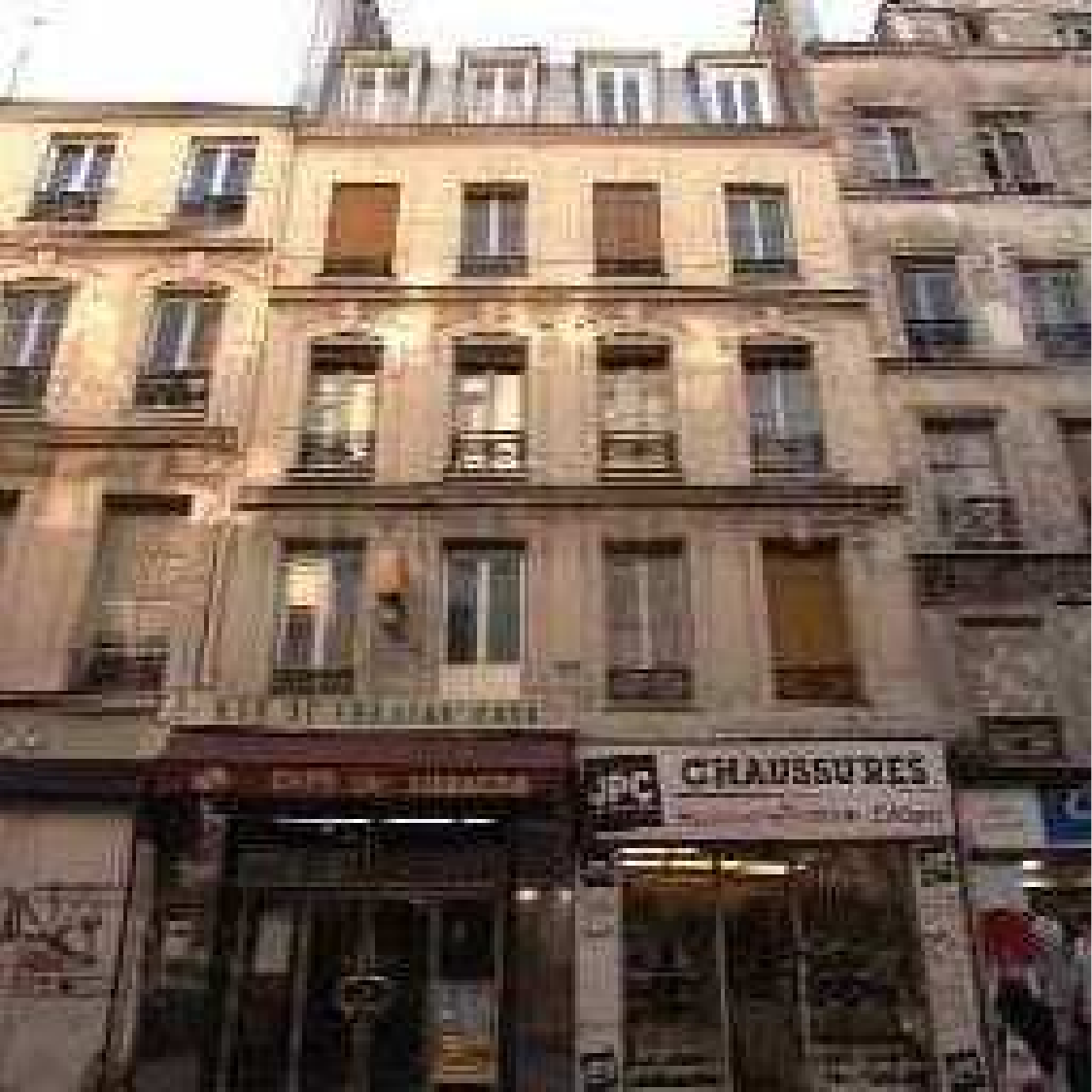}
	 \includegraphics[width=\figscenesw, height=\figscenesw]{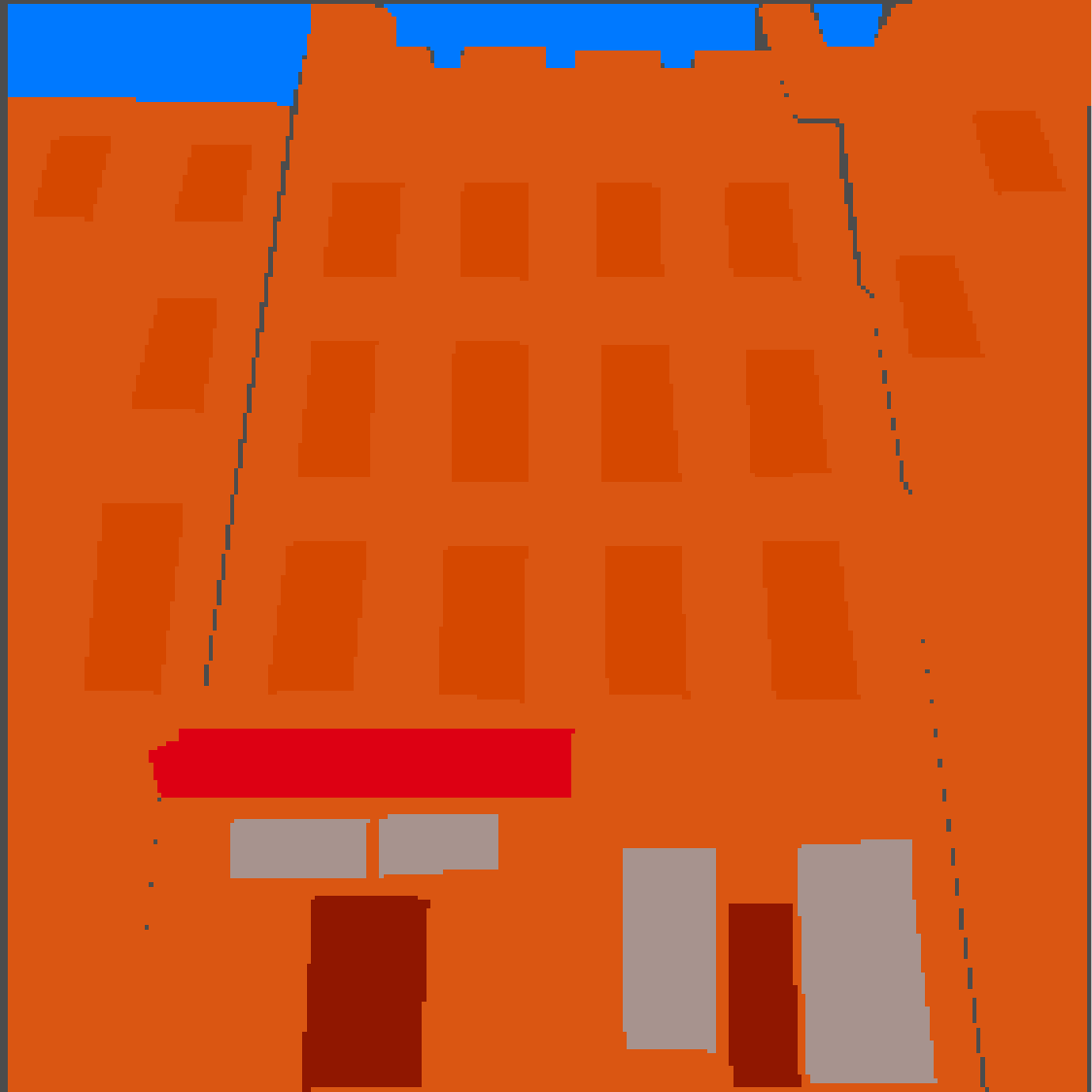}
	 \includegraphics[width=\figscenesw, height=\figscenesw]{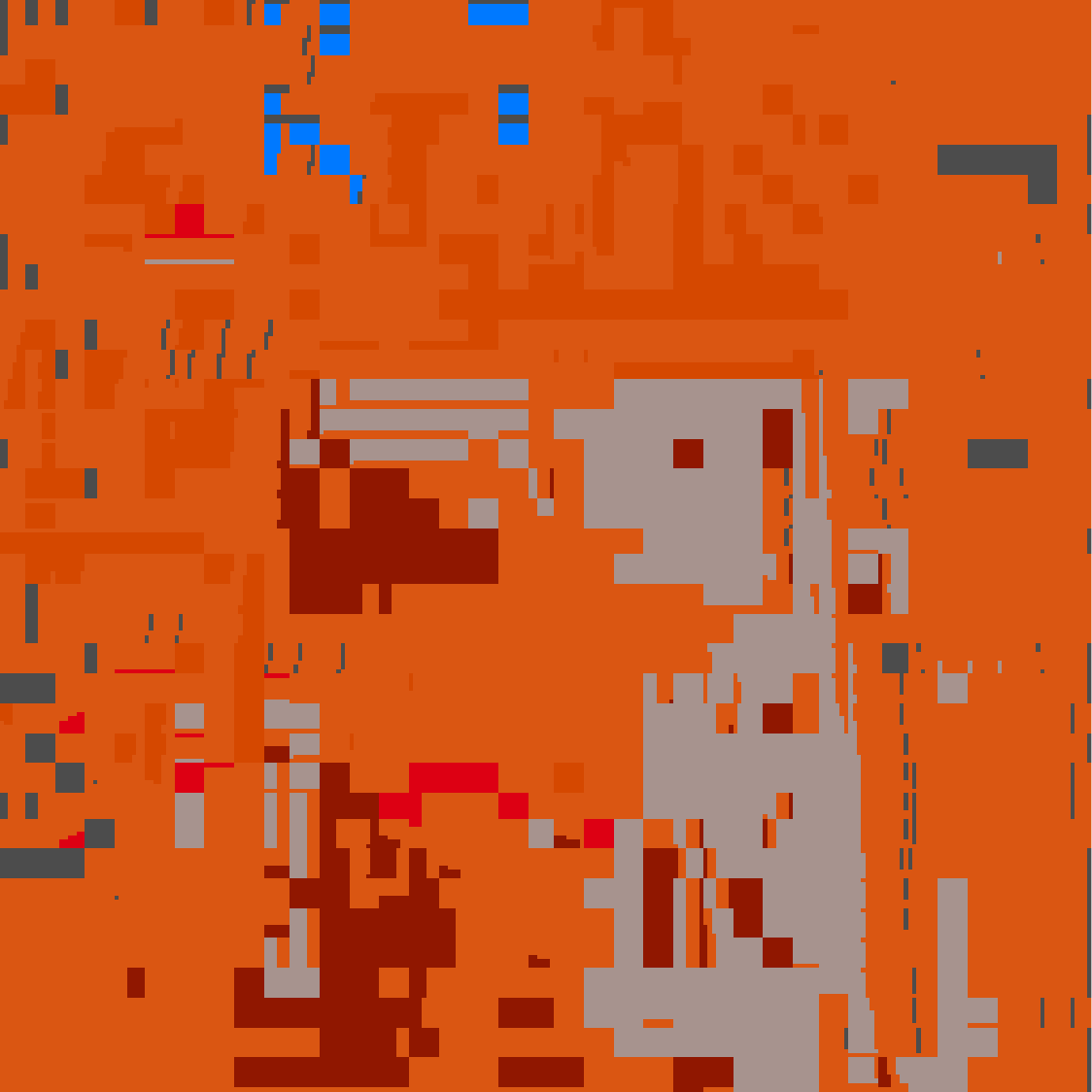}
	 \includegraphics[width=\figscenesw, height=\figscenesw]{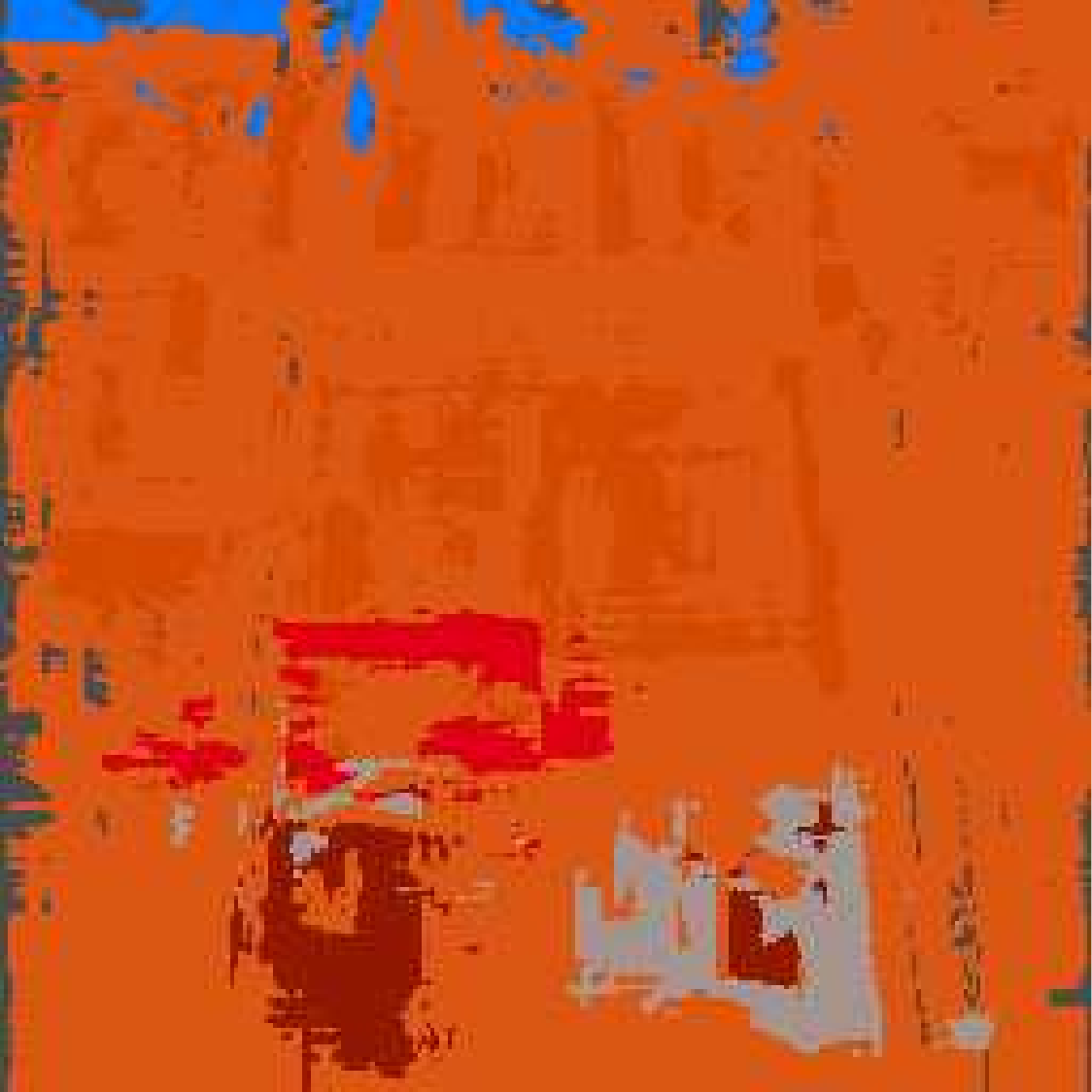}
     \includegraphics[width=\figscenesw, height=\figscenesw]{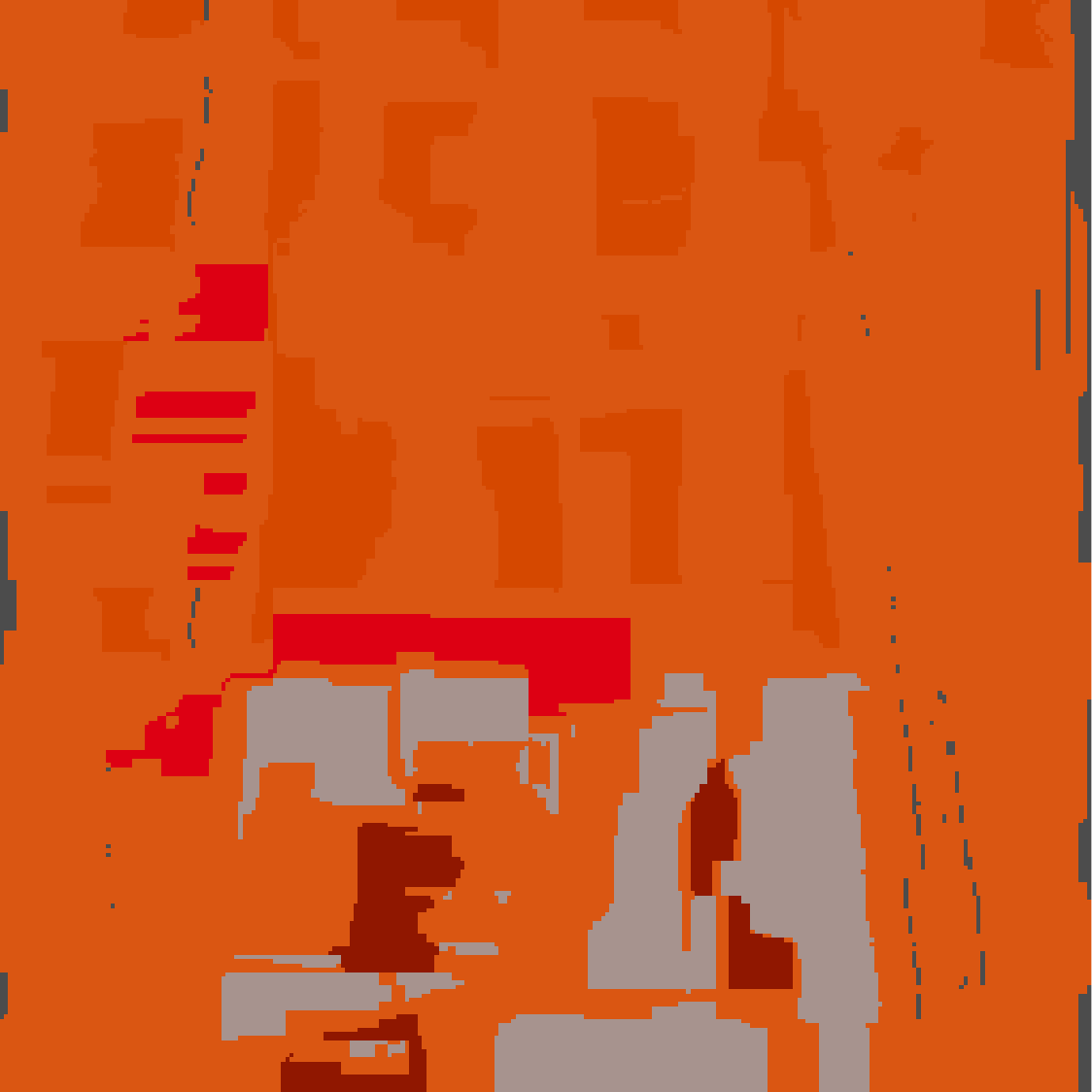}
     \includegraphics[width=\figscenesw, height=\figscenesw]{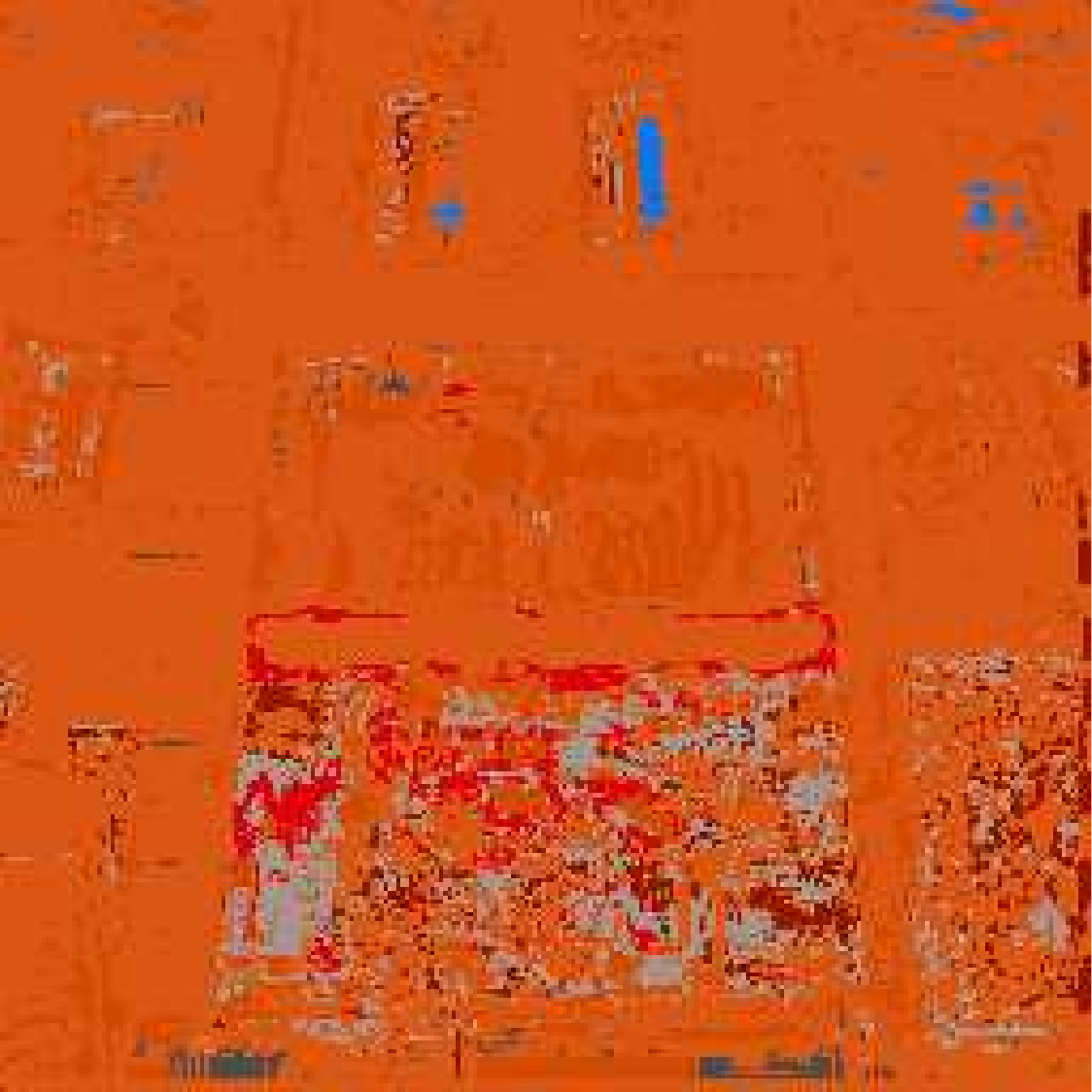} \\

     \includegraphics[width=\figscenesw, height=\figscenesw]{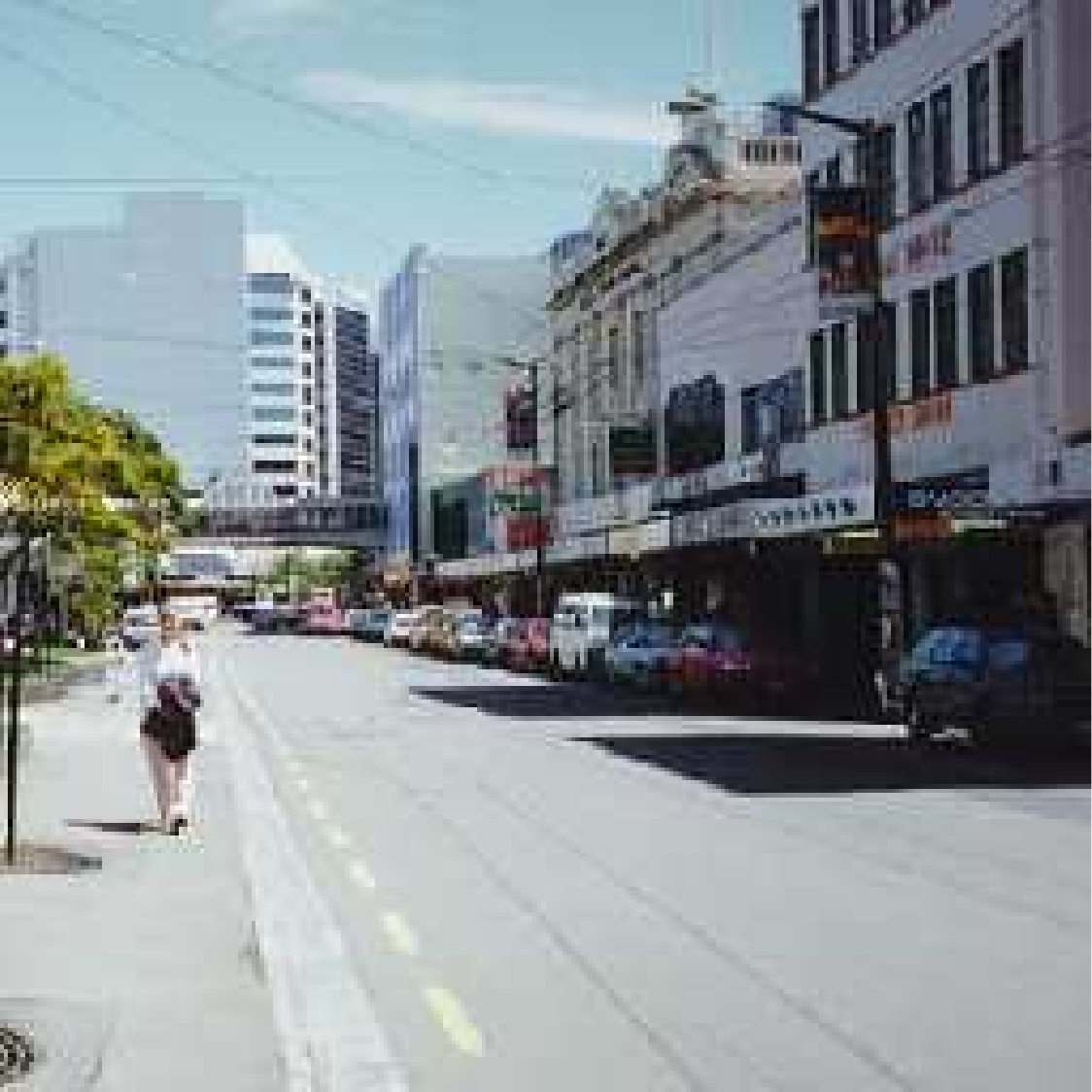}
	 \includegraphics[width=\figscenesw, height=\figscenesw]{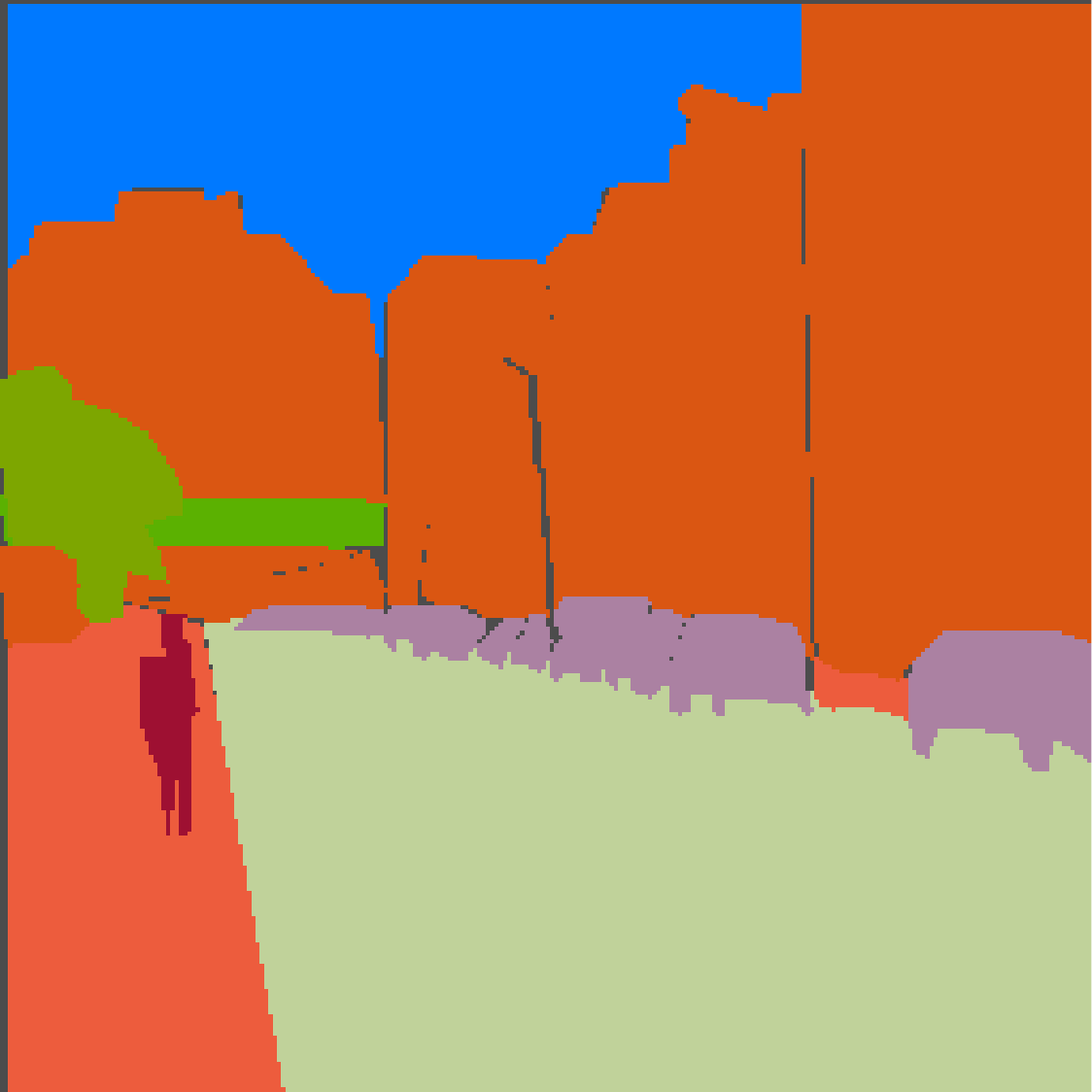}
	 \includegraphics[width=\figscenesw, height=\figscenesw]{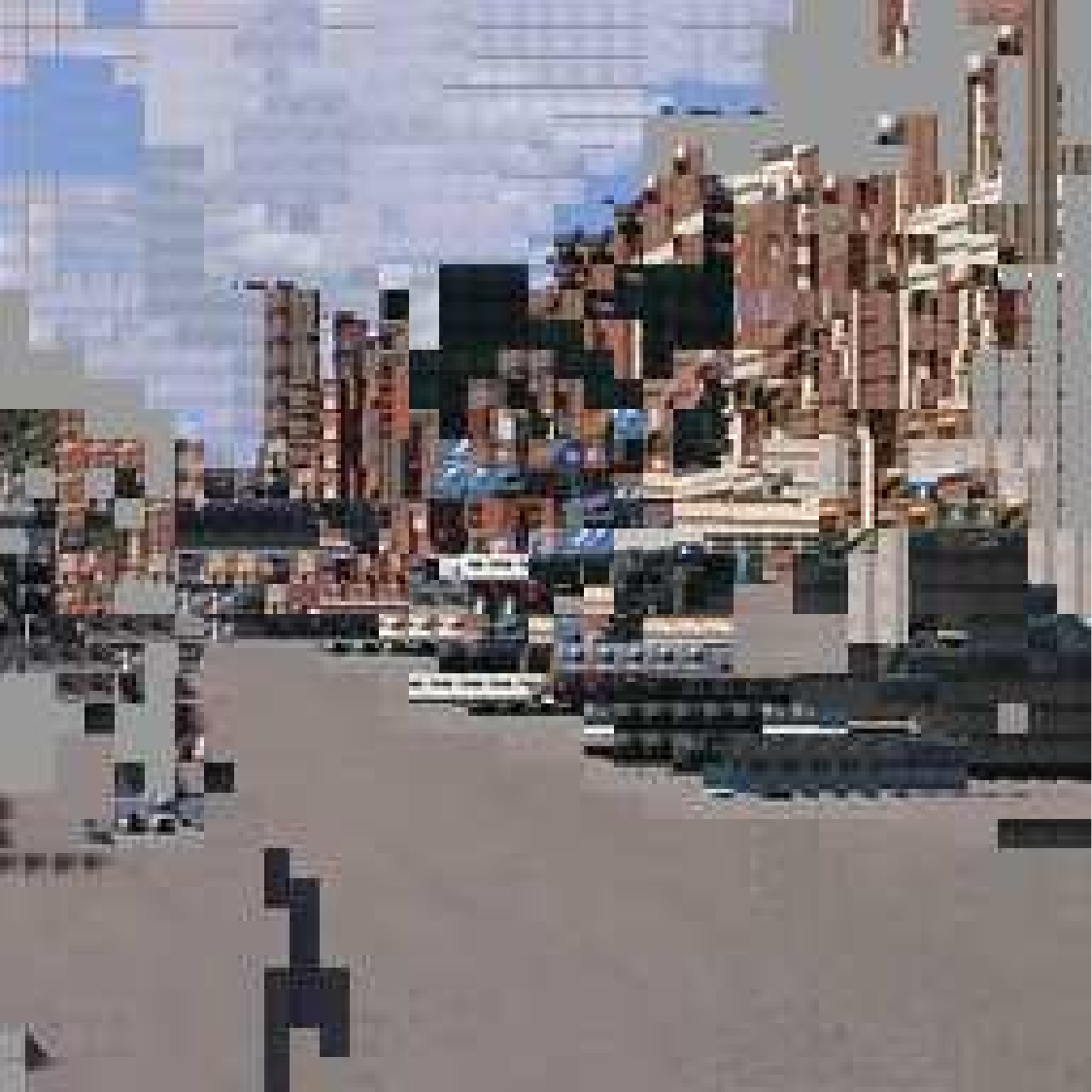}
	 \includegraphics[width=\figscenesw, height=\figscenesw]{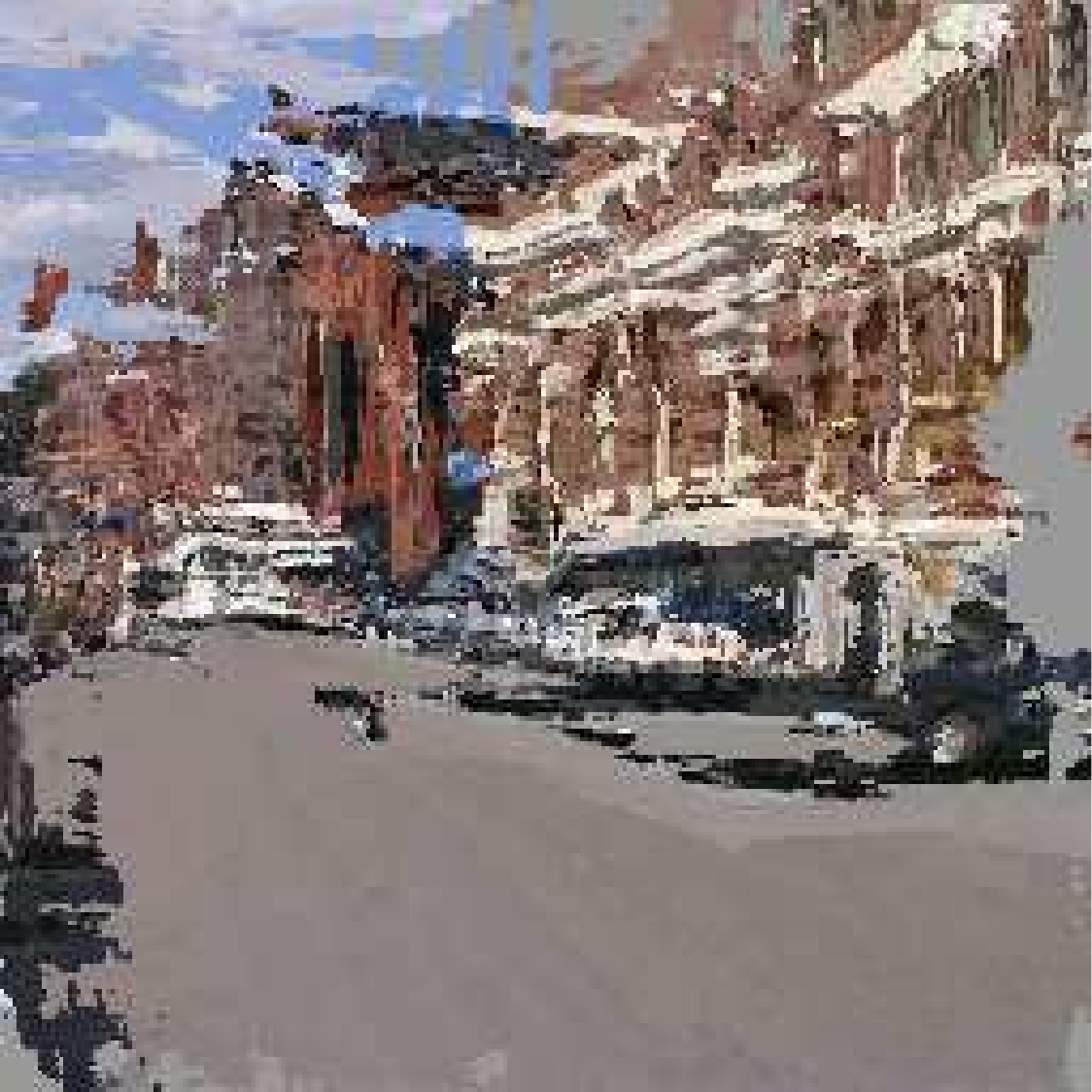}
     \includegraphics[width=\figscenesw, height=\figscenesw]{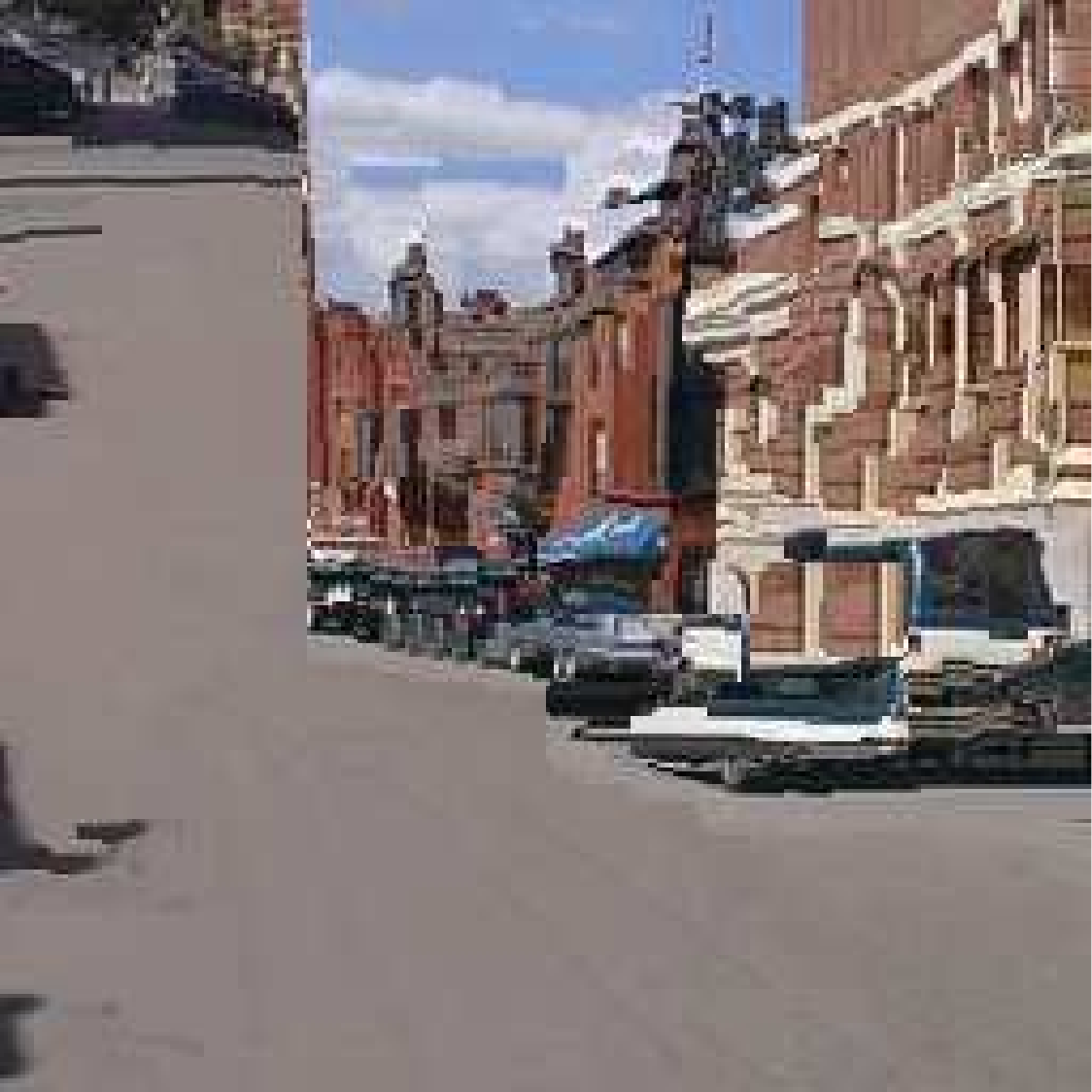}
     \includegraphics[width=\figscenesw, height=\figscenesw]{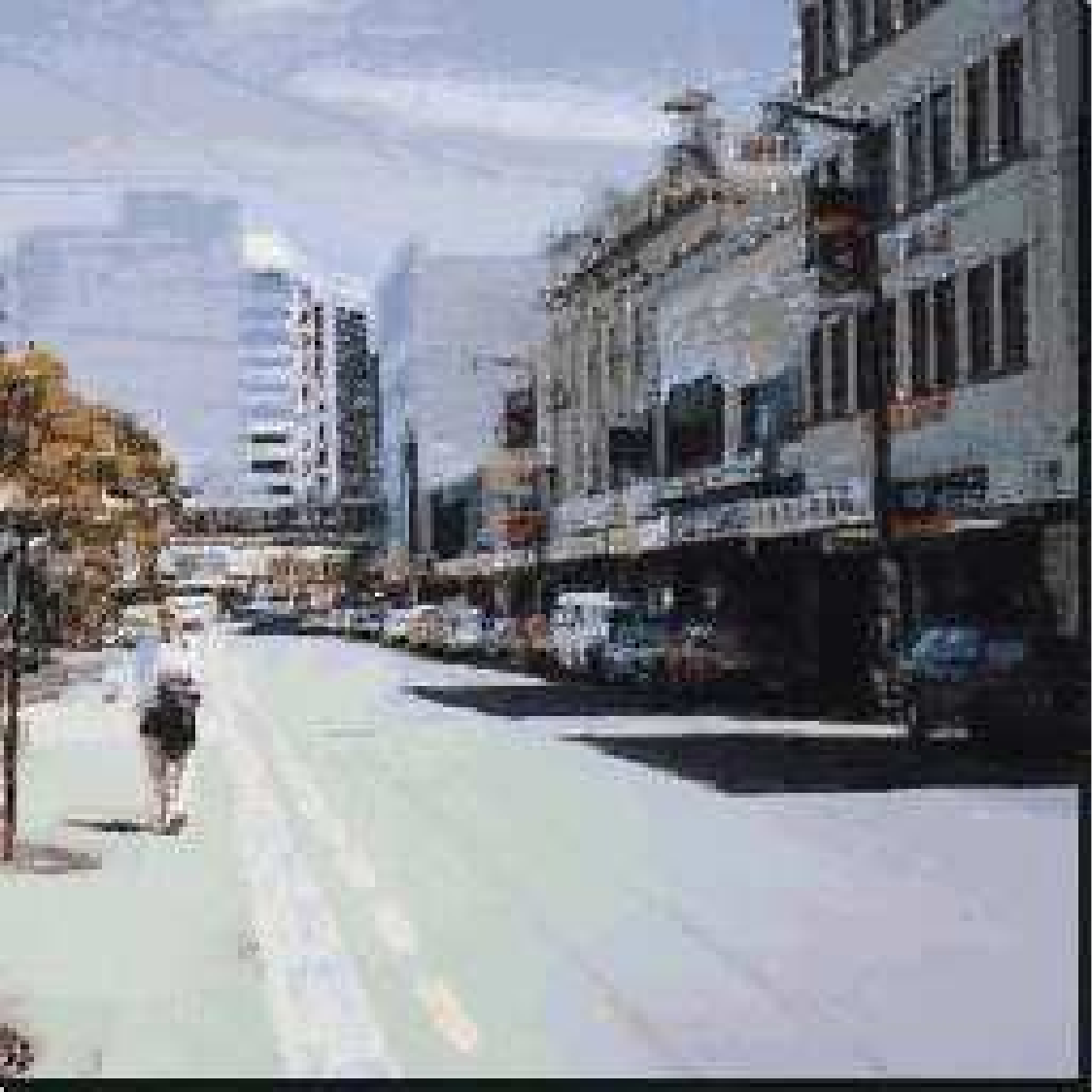} \\
     \includegraphics[width=\figscenesw, height=\figscenesw]{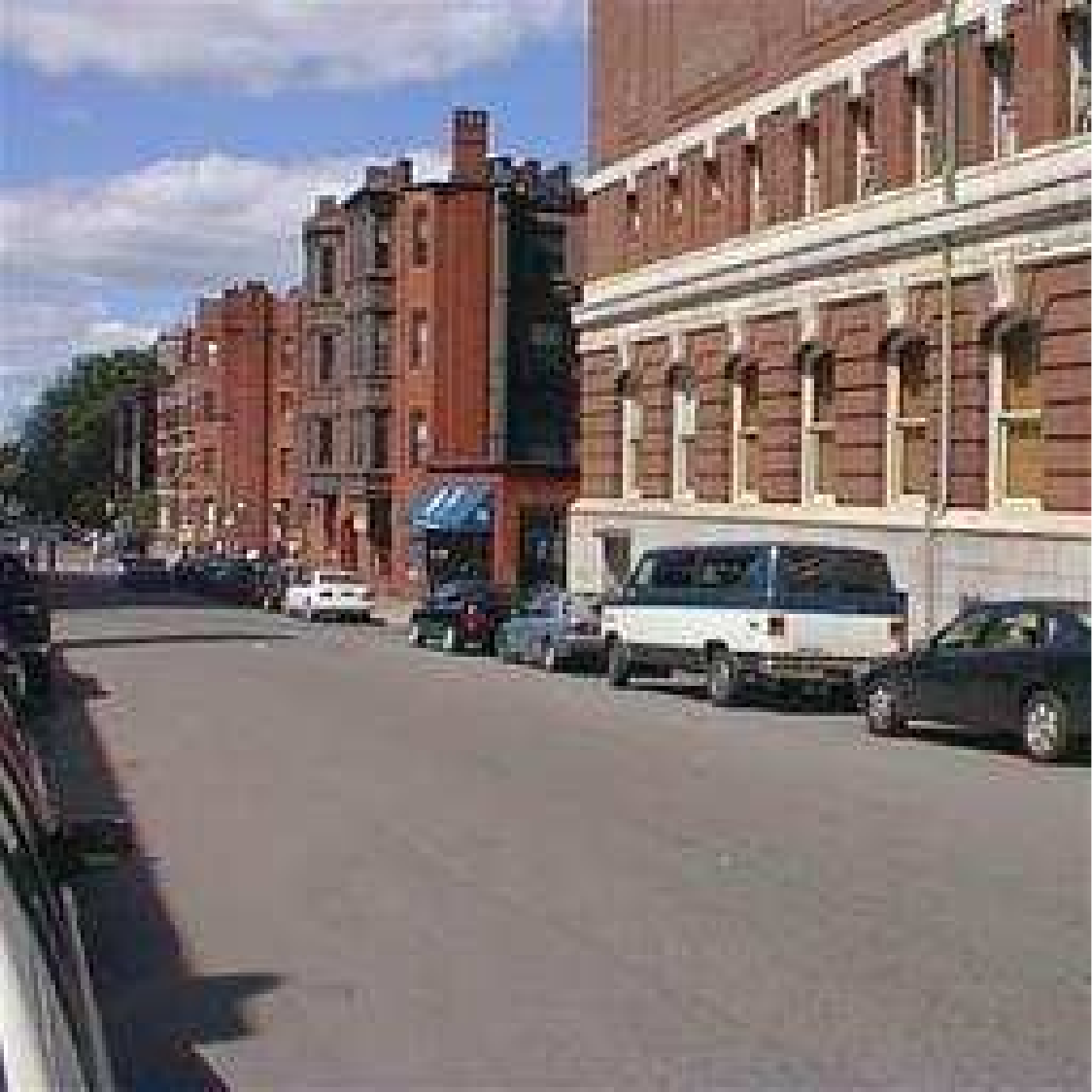}
	 \includegraphics[width=\figscenesw, height=\figscenesw]{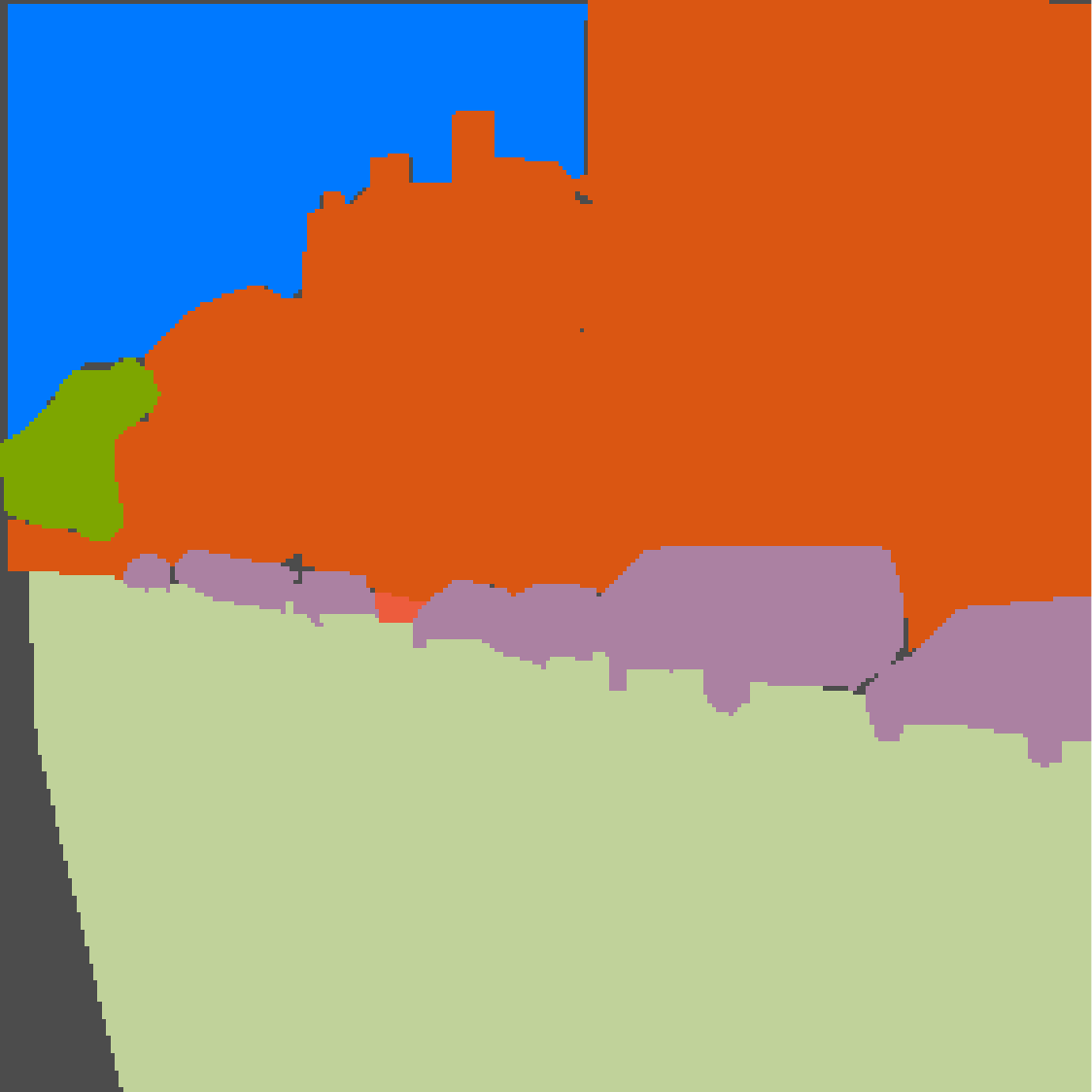}
	 \includegraphics[width=\figscenesw, height=\figscenesw]{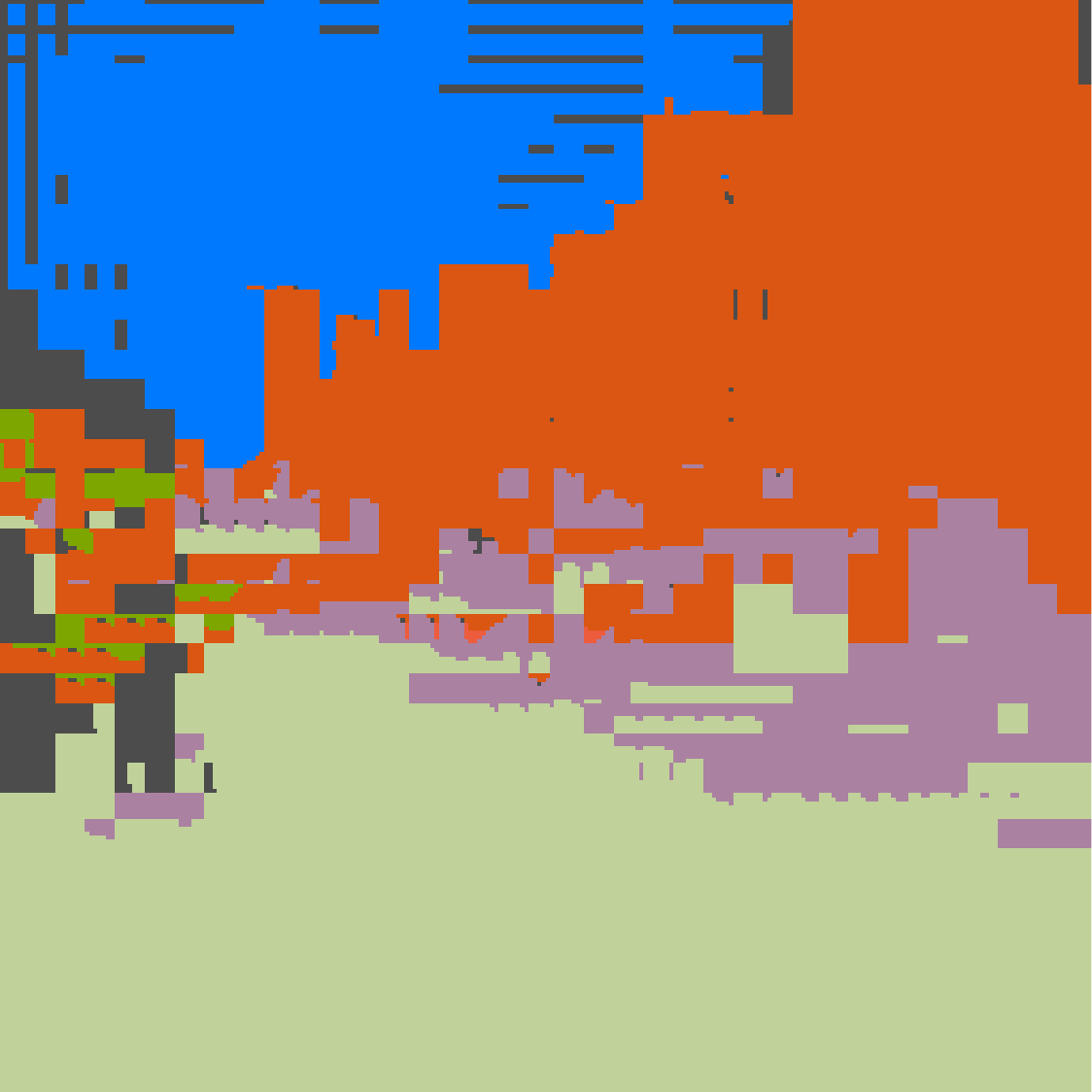}
	 \includegraphics[width=\figscenesw, height=\figscenesw]{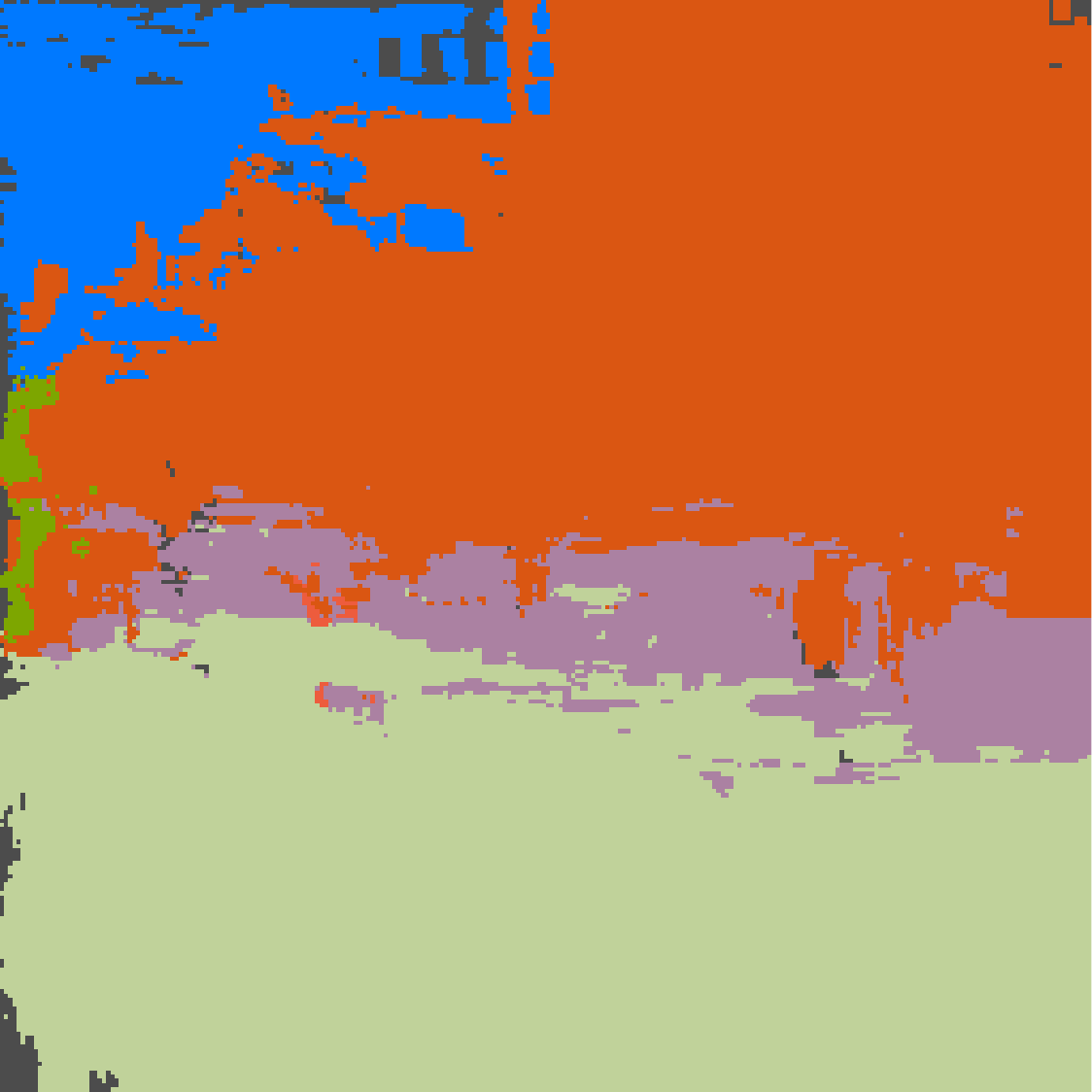}
     \includegraphics[width=\figscenesw, height=\figscenesw]{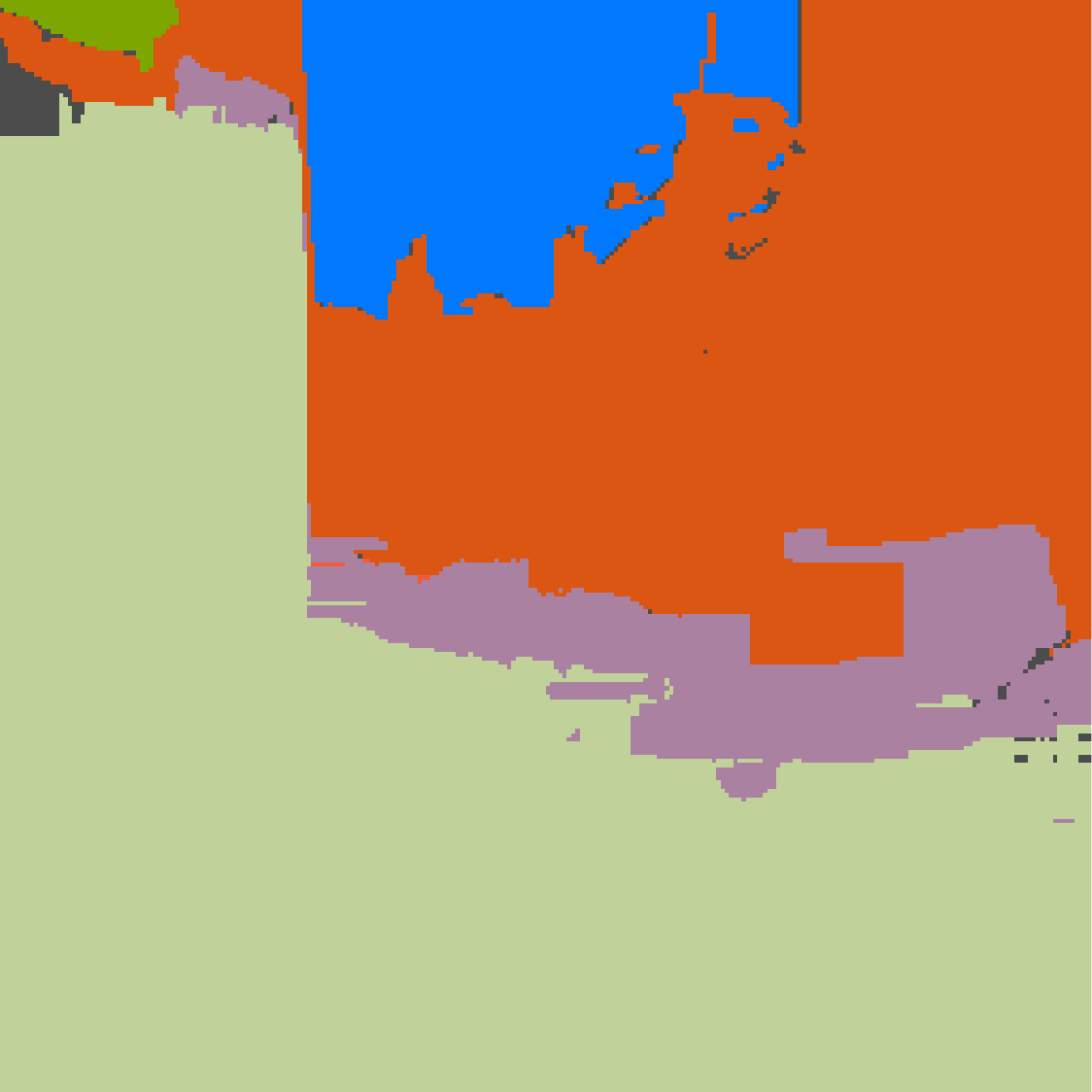}
     \includegraphics[width=\figscenesw, height=\figscenesw]{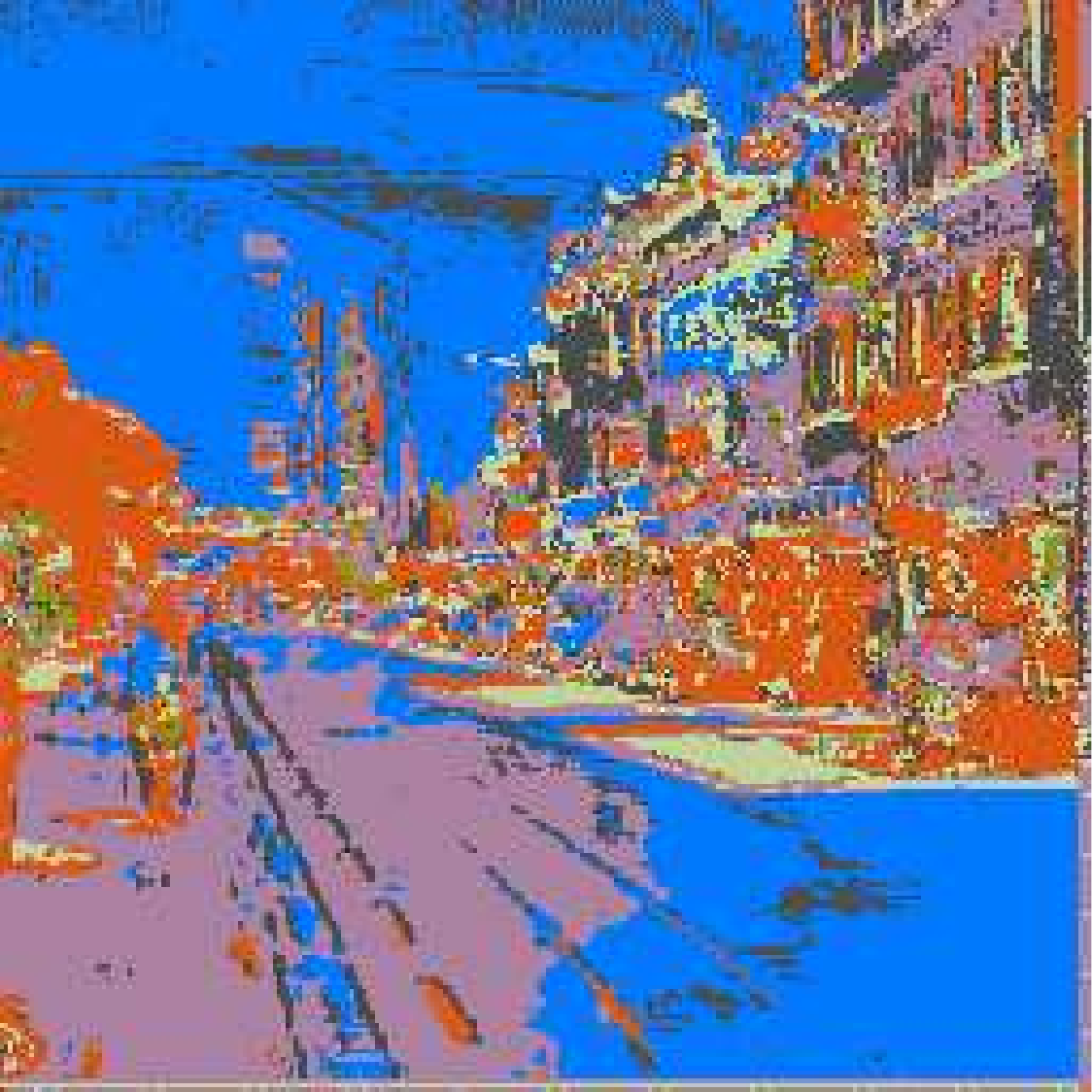} \\

\vspace{\vspacedistmid}
\caption{Example scene matching results of compared methods.
This plot is displayed  in accordance with
Figure \ref{fig:object_matching_seg}.
The only difference here is that the scenes have multiple labels.
Different pixel colors encode the 33 classes contained in the LMO dataset (e.g., tree, building, car, sky).
}
\label{fig:scene_matching_seg}
\vspace{\vspacedist}
\end{figure*}

\paragraph{Scene matching and segmentation}

In this scene matching and segmentation experiment,
we report the results on the LMO dataset\cite{liu2009nonparametric}.
Most of the LMO images are outdoor scenes including streets, beaches, mountains and buildings.
The dataset names top 33 object categories with the most labeled pixels.
Other object categories are considered as the 34th category `unlabeled'
\cite{liu2009nonparametric}.
This experiment is more complex than the experiment on the Caltech-101 dataset,
which contains only two labels (foreground or background).

For each test image, we select 9 most similar exemplar images
by Euclidean distances of GIST as in
\cite{liu2009nonparametric,kim2013deformable}.
Through the matching process, we obtain dense correspondences between
 the test image and the selected exemplar images.
 Similar to the Caltech-101 matching experiment,
 we transfer the pixel labels to the test image pixels
 from the corresponding exemplar pixels.

For the scene matching, some example results are shown in Figure \ref{fig:scene_matching_seg}.
Again, our method is more robust to the image variations
(scene appearance changes and structure changes)
not only in scene warping but also in label transferring.
Our labeling results appear more similar to the
test image ground truth labels.

For the scene segmentation, we follow the method described in \cite{liu2009nonparametric}.
After the matching and warping process, each pixel in the test image may have multiple
labels by matching different exemplars.
To obtain the final image segmentation, we reconcile multiple labels and impose spatial
smoothness under a Markov random field (MRF) model. The label likelihood is defined by the
feature distance between test image pixel and its corresponding exemplar image pixel.
In this experiment, we randomly pick 40 images as test images.
We report the patch-level results of our method on this dataset.

\begin{table} \center
\vspace{\vspacedistfront}
{
\begin{tabular}{ r | c     c    c  c  }
\hline
\multirow{1}{*}{{{method}}} &\multicolumn{1}{c}{{Ours}} &\multicolumn{1}{c}{{DSP}}
&\multicolumn{1}{c}{{SIFT Flow}} &\multicolumn{1}{c}{{CSH}} \\
\hline
	{LT-ACC}     &\textbf{0.702} & 0.677 &0.687 &0.612\\
	{IOU}     &\textbf{0.505} & 0.498 &0.479  &0.365\\
	\hline

\end{tabular}
}
\caption{Scene segmentation performance of different methods on the LMO dataset
		in matching accuracy. For our method, the dictionary is learned by using K-means
		and the encoding schemes is K-means triangle. The dictionary size is set to 100.
		The best results are boldfaced.} \label{tab:lmo_segmentation}
\vspace{\vspacedist}
\end{table}

Table \ref{tab:lmo_segmentation} shows the segmentation accuracy of our method compared with
state-of-the-art methods.
Our method outperforms
 state-of-the-art methods in the segmentation accuracy.
In this experiment, we notice that SIFT Flow outperforms the DSP method in GIST neighbors,
which is consistent with the results in \cite{kim2013deformable}.
The segmentation accuracy of DSP relies on the exemplar list\cite{kim2013deformable}.
Our method does not have this problem.
In the multi-class pixel labelling experiment, our method outperforms
the compared methods by 0.02 of matching accuracy.
Our IOU score is also better.
The experimental results show that our method provides higher matching accuracy.
The reason is two-fold: (a) the learned features provide higher discriminability between classes,
and (b) our matching model is more suitable for the learned features to carry out  dense matching
tasks.

\paragraph{Matching results on the Pascal VOC dataset}
This part of experiments is carried out on the Pascal Visual Object Classes (VOC) 2012 dataset
\cite{everingham2014pascal}.
There are 2913 images in the segmentation task, which have
ground-truth annotations for each image pixel.
There are 20 specified object classes in the Pascal 2012 dataset.
The objects which do not belong to one of these classes are given a `background' label.

In our experiment, we random choose 30 image pairs for
each class (600 image pairs in total)
from those images which only contain one object class.
Each pair of images come from the same class.
We consider the objects as `foreground' and others as `background'.
The parameter setting is the same as our experiments on the
Caltech-101 dataset.
We use the same dictionary to obtain our pixel features as in the other experiments.

\begin{figure}
\vspace{\vspacedistfront}
\centering
\subfigure[]{
    \label{fig:extensive_pascal:1}
	\includegraphics[width=0.28\textwidth]{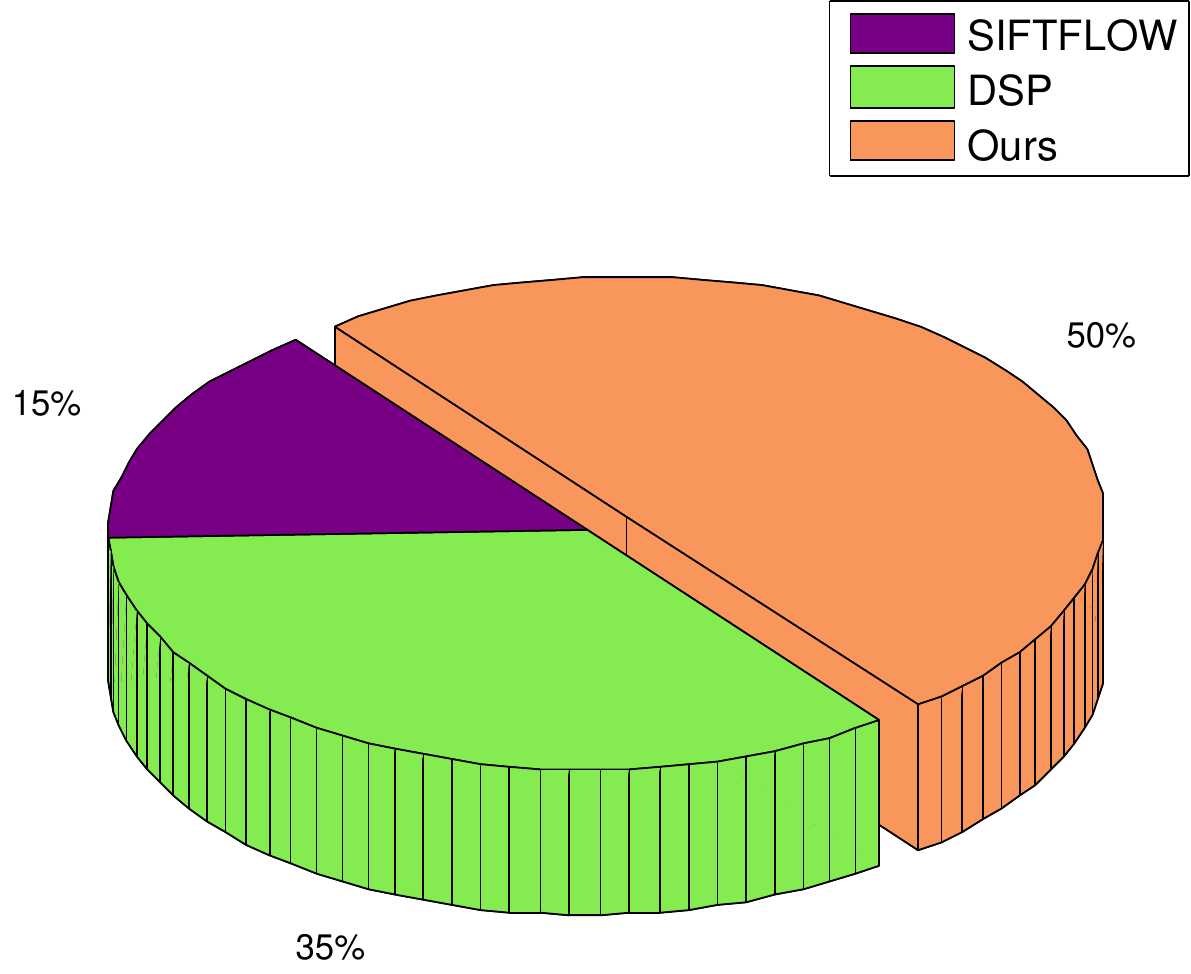} } \\%
	\vspace{\vspacedist}
\subfigure[]{
    \label{fig:extensive_pascal:2}
	\includegraphics[width=0.28\textwidth]{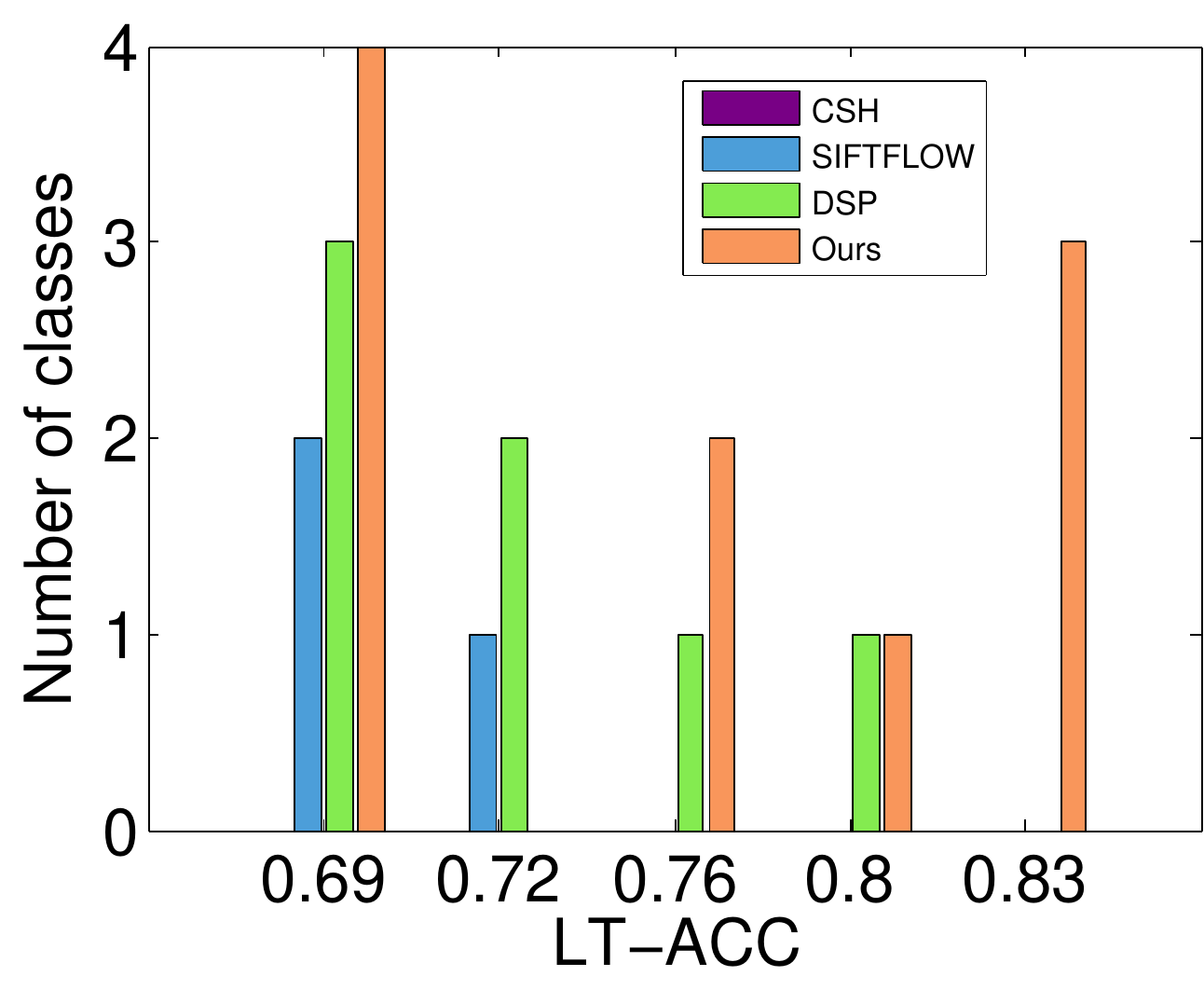} }%
\vspace{\vspacedistmid}
     \caption{
      \subref{fig:extensive_pascal:1} shows the percentage of each
	 method achieving the best accuracy on the Pascal dataset. Our method achieves the best
	 matching accuracy in 10 classes out of 20.
     \subref{fig:extensive_pascal:2} shows the histogram of
     each methods' achievements
     in matching accuracy (LT-ACC) in all classes.
     }
\label{fig:extensive_pascal}
\vspace{\vspacedist}
\end{figure}

Table \ref{tab:pascal_matching_acc} shows the matching accuracy and
CPU time of our method as well as those compared methods.
Again, we can see that our method achieves the highest label transfer accuracies
and is more than 8 times faster than the DSP method.
The pie chart in Figure \ref{fig:extensive_pascal:1} shows the percentage of each
	 method achieving the best accuracy in 20 classes.
   We see that the proposed method achieves the best matching accuracy in 10 classes,
	 which outperforms the compared methods.
Figure \ref{fig:extensive_pascal:2} shows the histogram of
best matching accuracy in all classes.

The  results on the Pascal dataset are slightly different from those on the Caltech-101 dataset.
As we can see from  Figure \ref{fig:extensive_pascal:2},
most of the classes in the Pascal dataset are low-accuracy classes.
It means that objects in the same class vary more than those in the Caltech-101 dataset.
In this experiment, our method still outperforms other methods for most cases.
This  confirms that our method achieves better
matching results under large object appearance variations.
It is further demonstrated that our proposed method is more suitable to
handle the matching problem of large intra-class variability than those compared methods.

\begin{figure*}
\vspace{\vspacedistfront}
\centering
{{Input\ \ \ \ \ \ \ GT label\ \ \ \ \ \ \ Ours \ \ \ \ \ \ \ DSP\ \ \ \ \ \ \ \ \ \ SIFT Flow\ \ \ \ CSH}} \\
	 \includegraphics[width=\figobjectmatchingw, height=\figobjectmatchingsw]{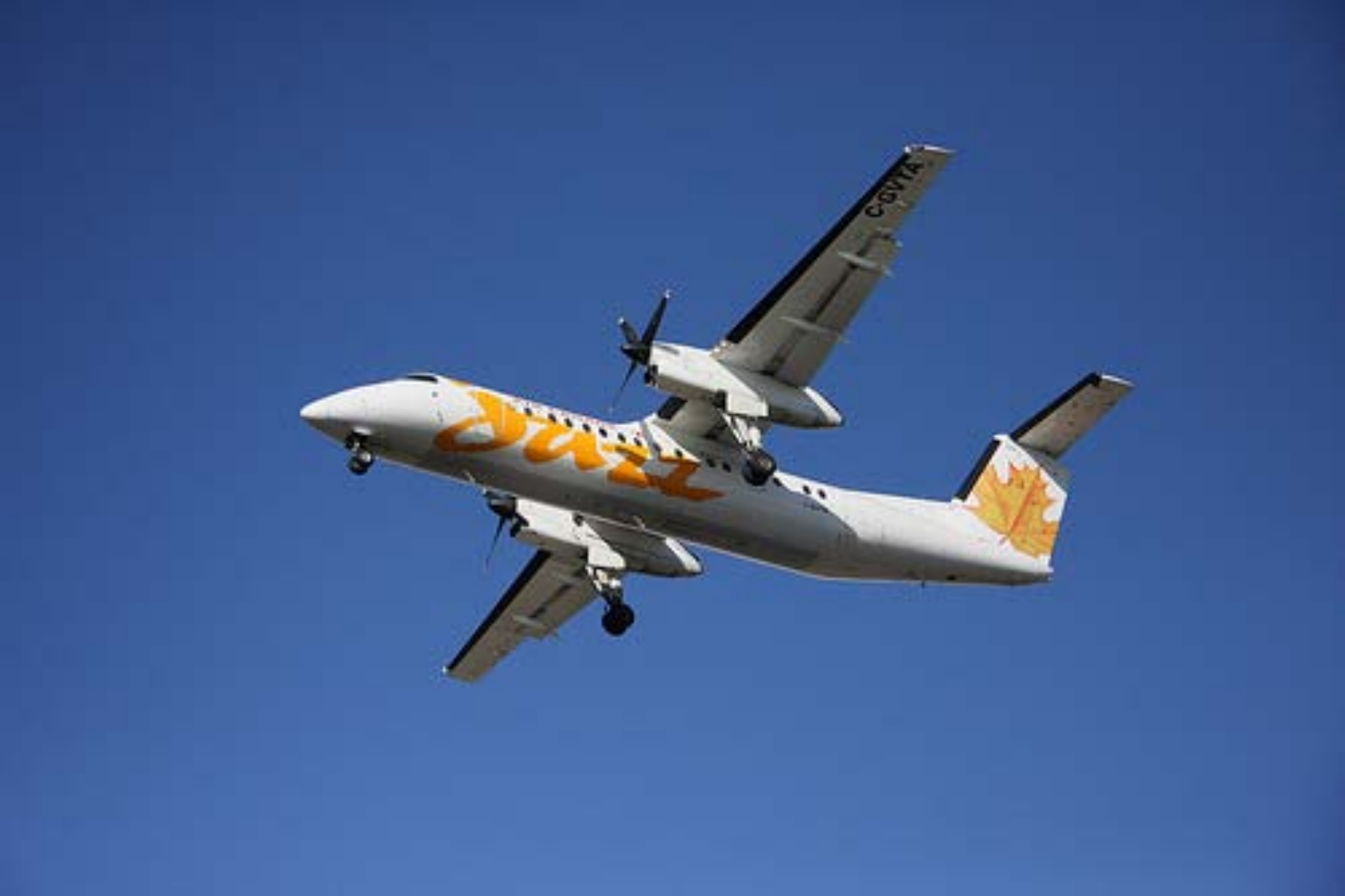}
	 \includegraphics[width=\figobjectmatchingw, height=\figobjectmatchingsw]{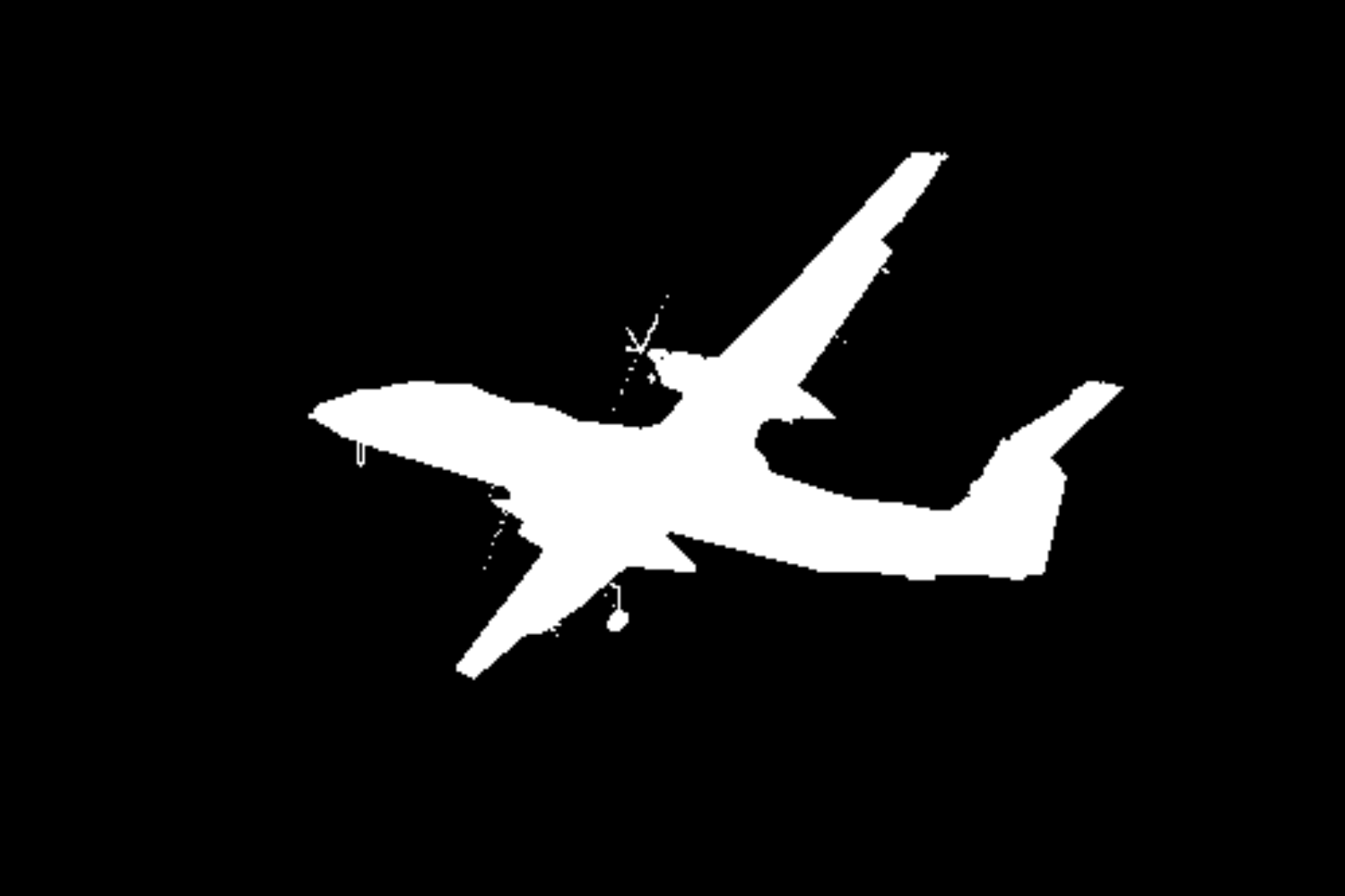}
	 \includegraphics[width=\figobjectmatchingw, height=\figobjectmatchingsw]{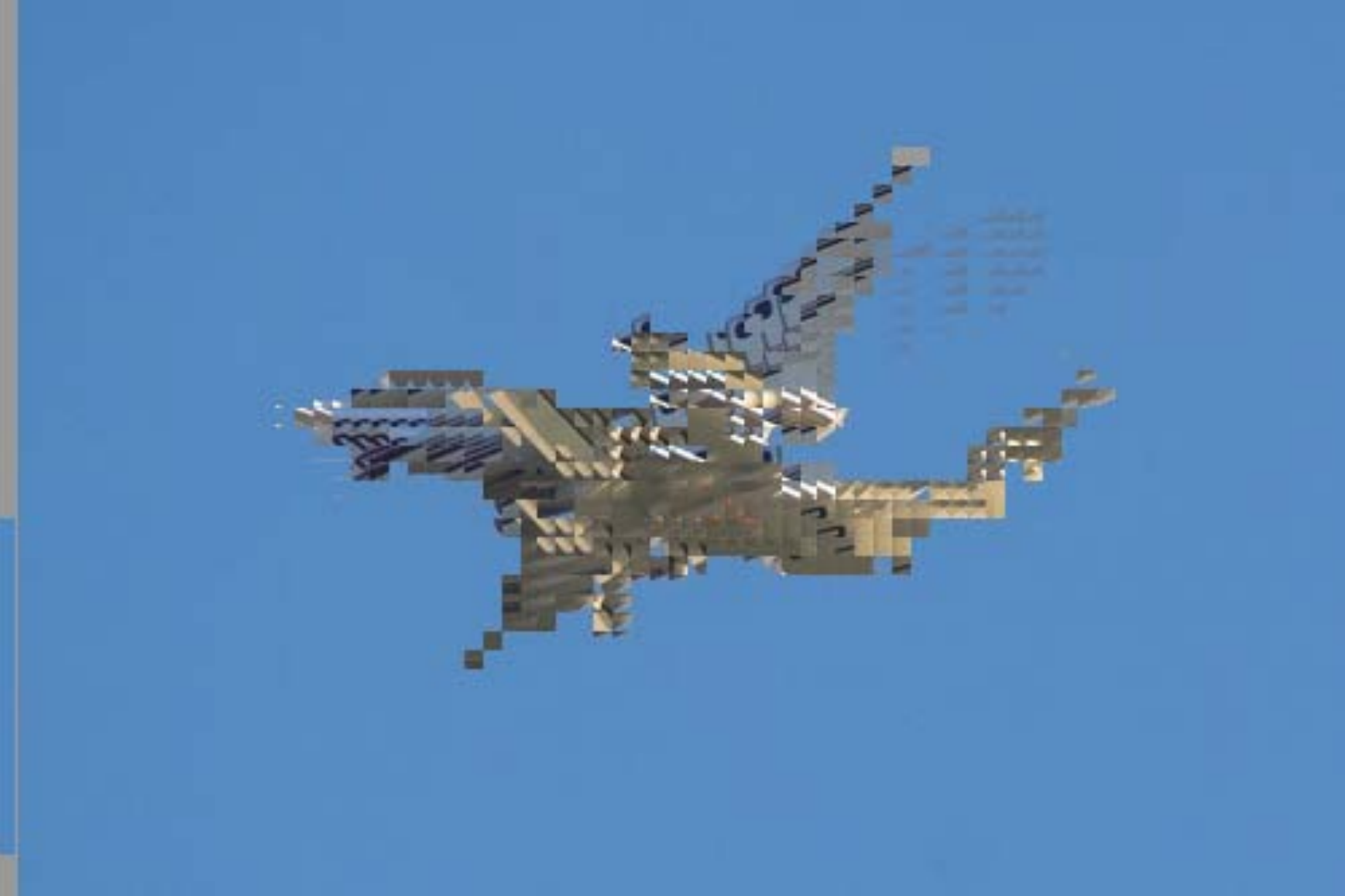}
     \includegraphics[width=\figobjectmatchingw, height=\figobjectmatchingsw]{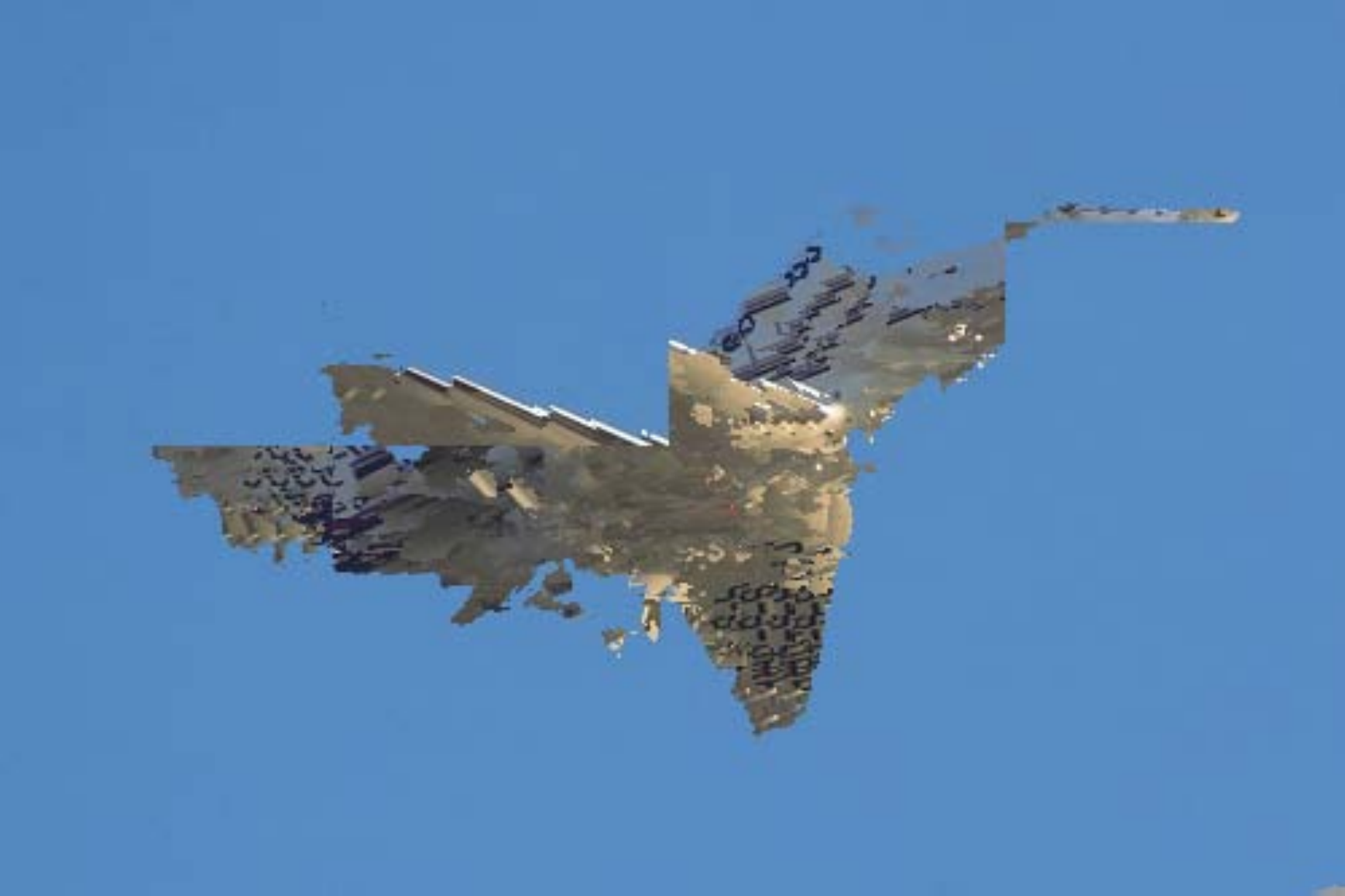}
    \includegraphics[width=\figobjectmatchingw, height=\figobjectmatchingsw]{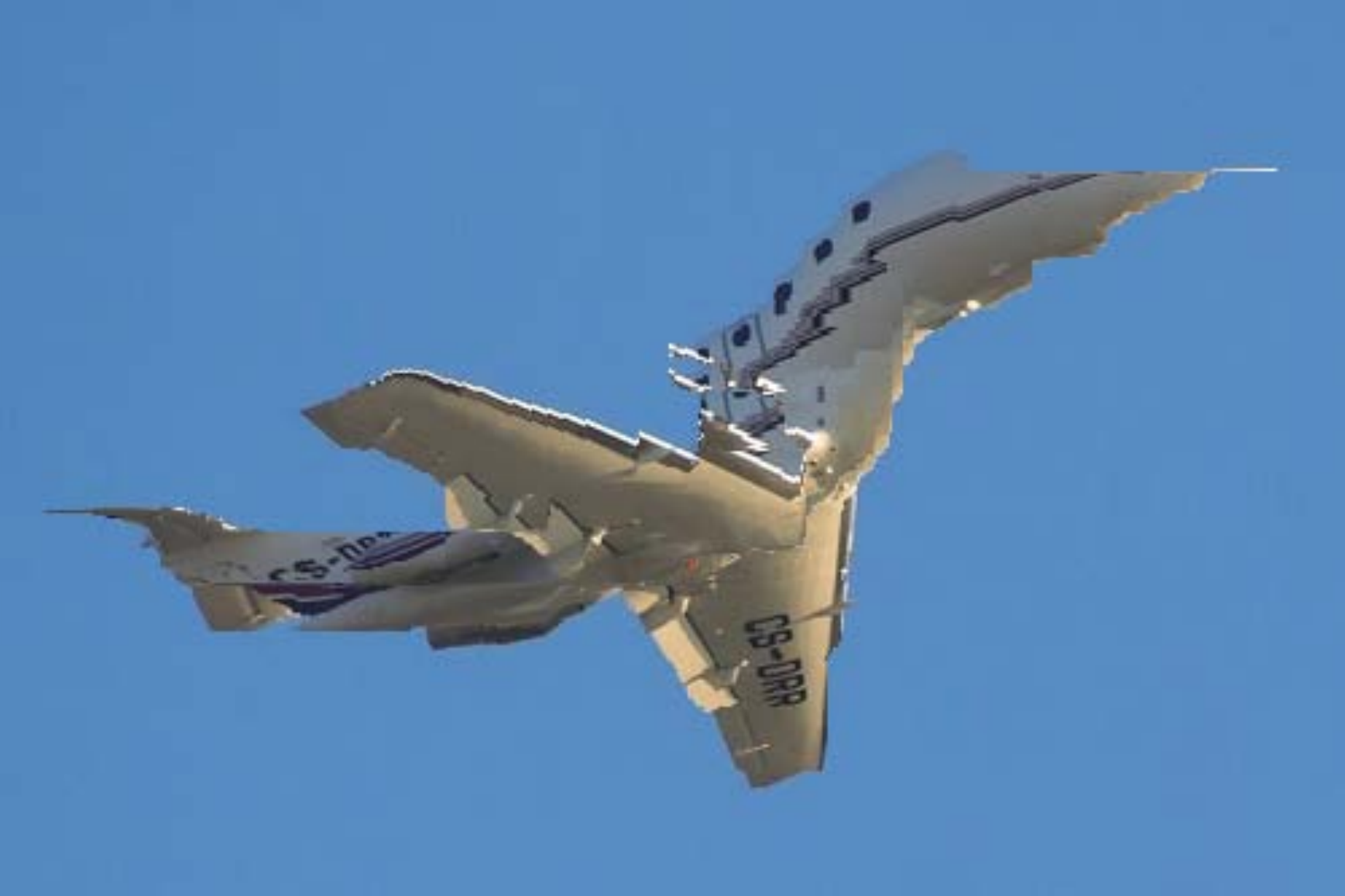}
     \includegraphics[width=\figobjectmatchingw, height=\figobjectmatchingsw]{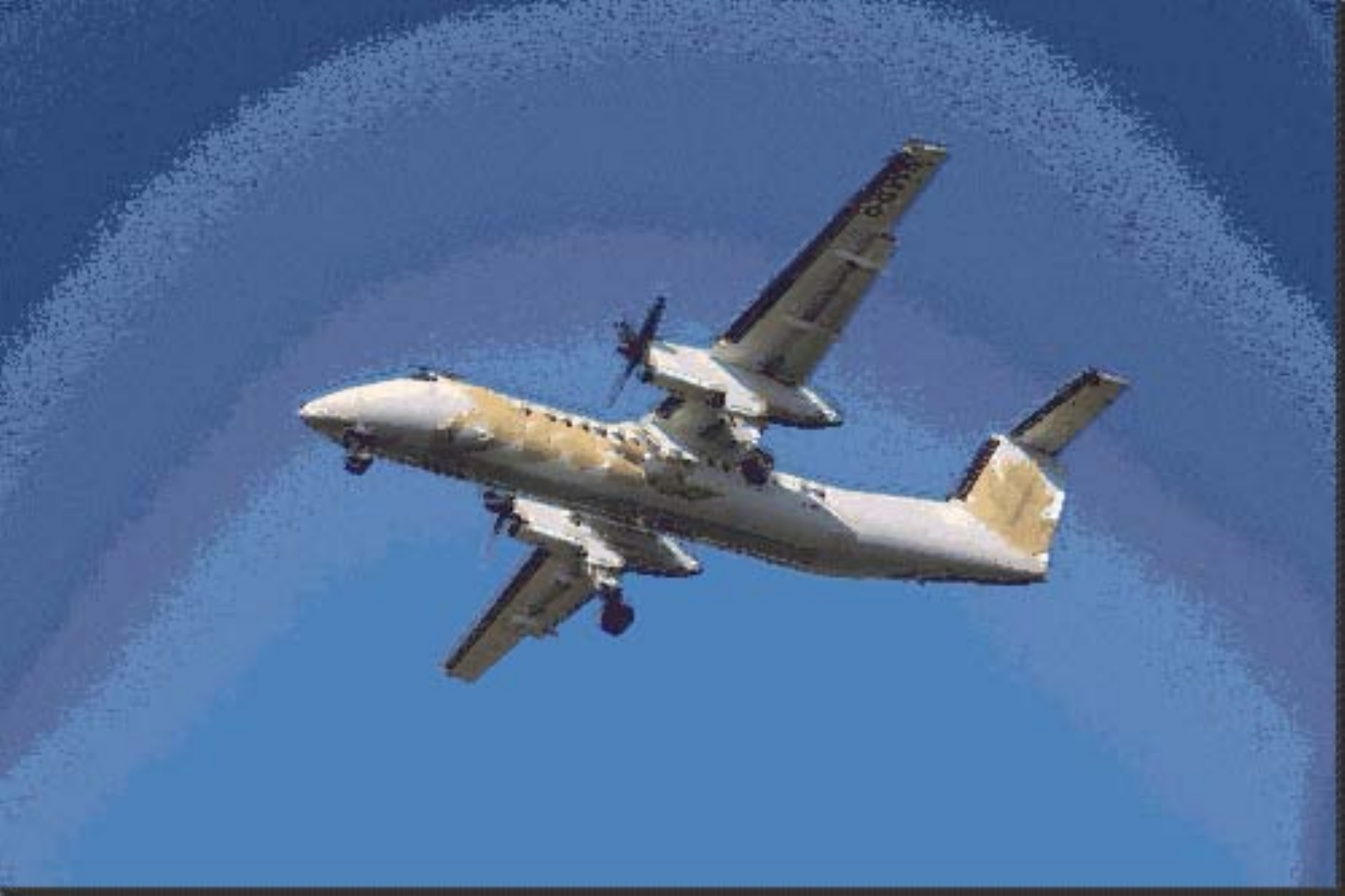} \\

 	 \includegraphics[width=\figobjectmatchingw, height=\figobjectmatchingsw]{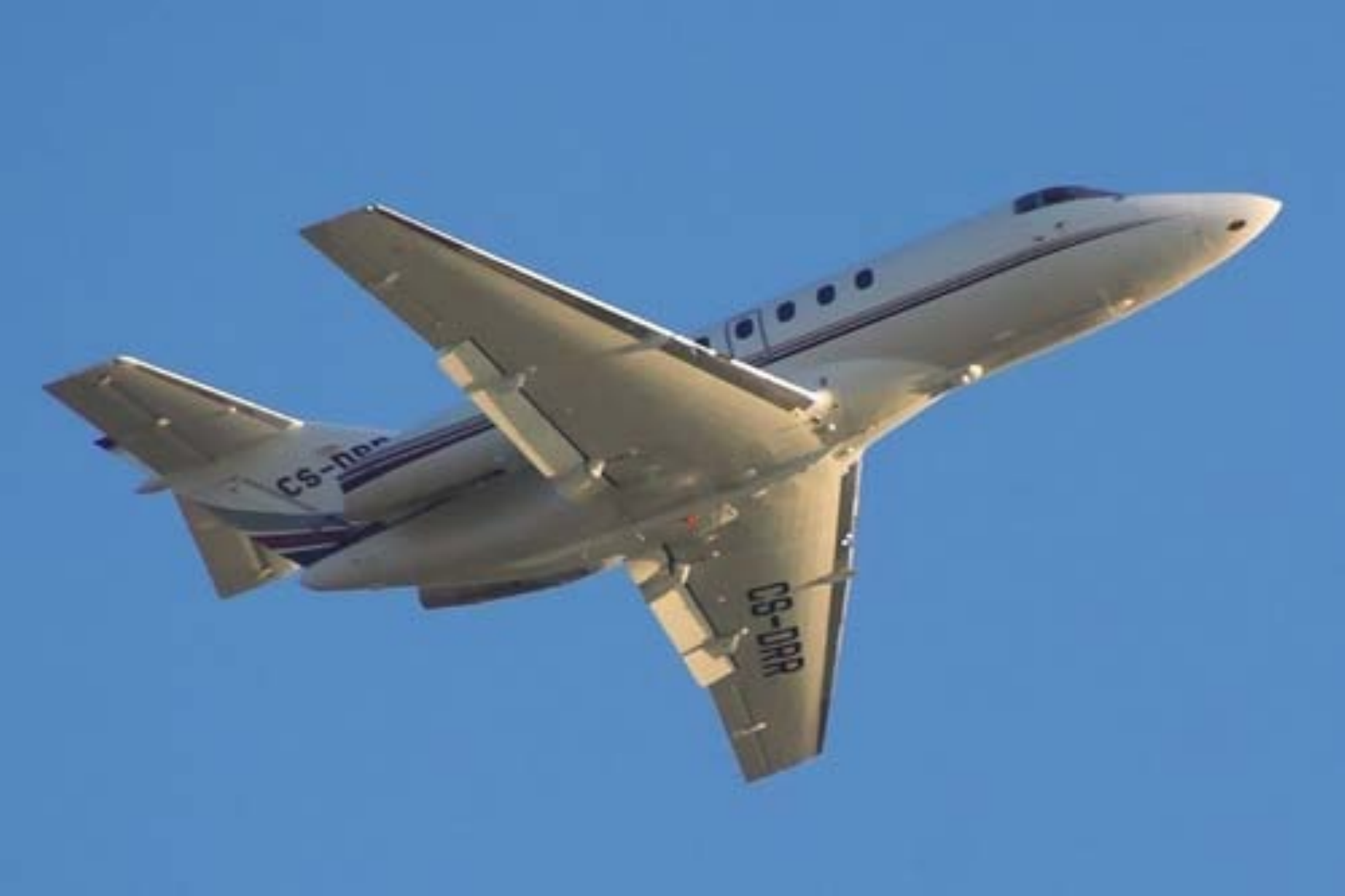}
	 \includegraphics[width=\figobjectmatchingw, height=\figobjectmatchingsw]{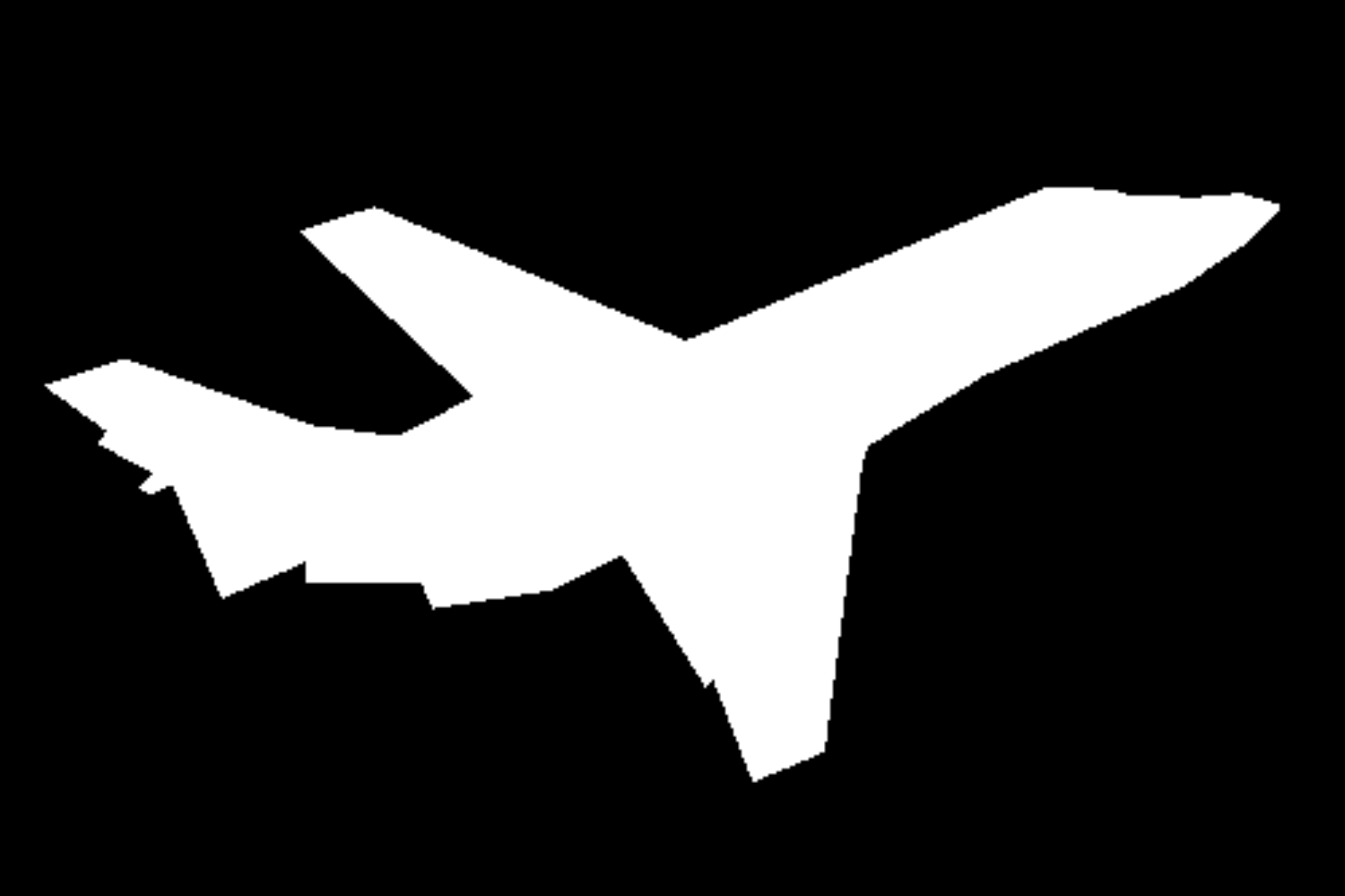}
	 \includegraphics[width=\figobjectmatchingw, height=\figobjectmatchingsw]{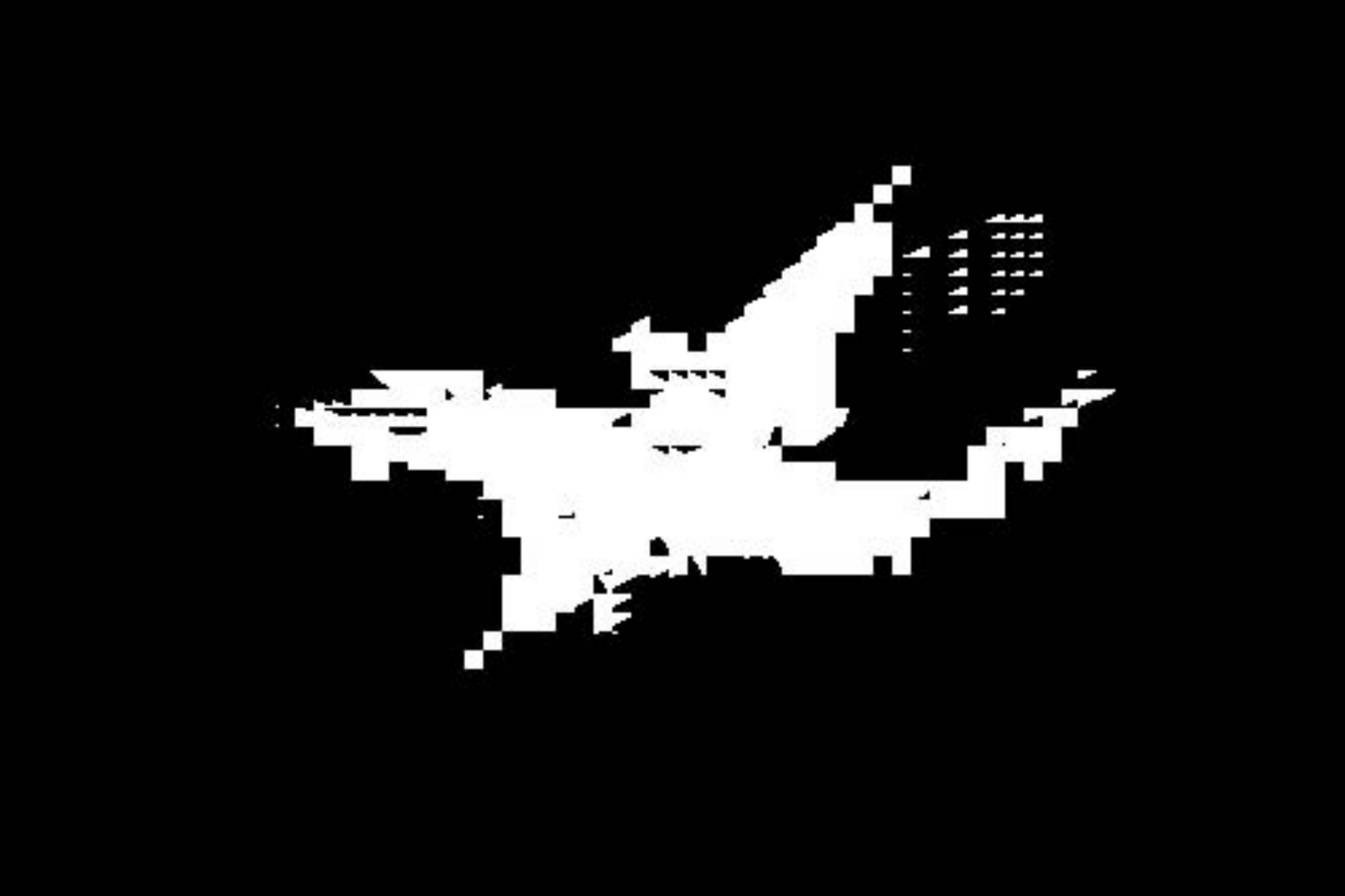}
	 \includegraphics[width=\figobjectmatchingw, height=\figobjectmatchingsw]{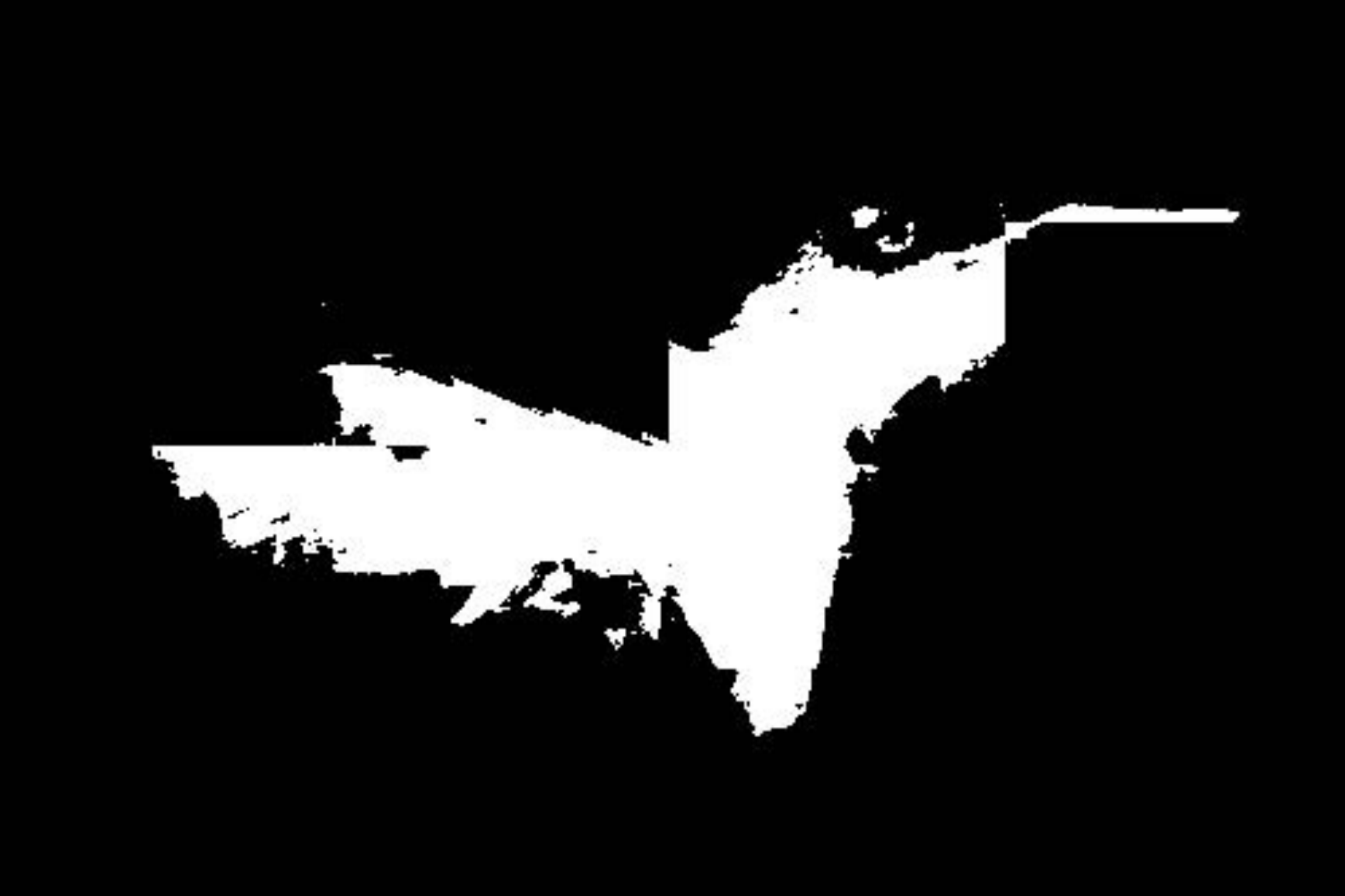}
	 \includegraphics[width=\figobjectmatchingw, height=\figobjectmatchingsw]{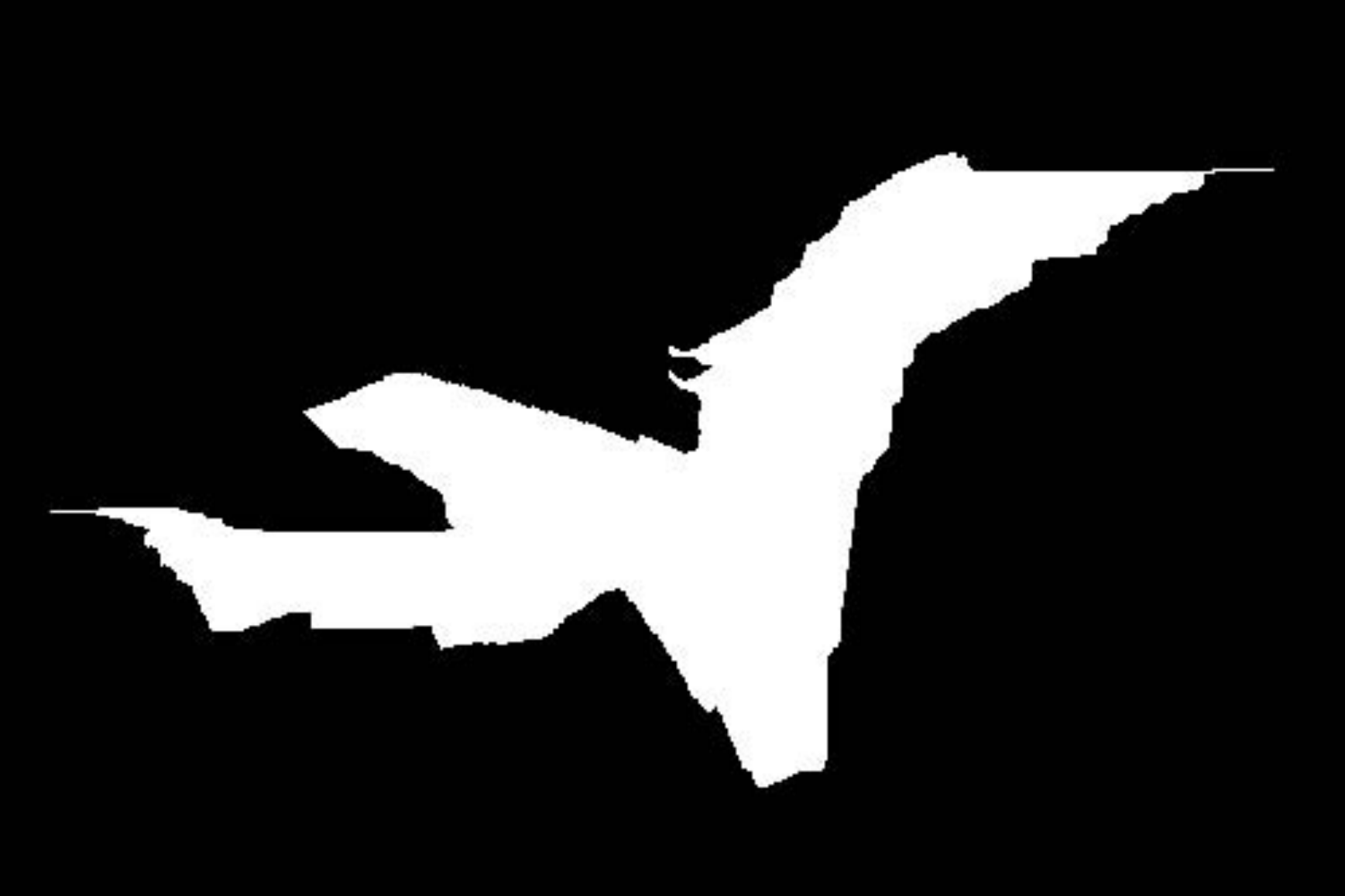}
	 \includegraphics[width=\figobjectmatchingw, height=\figobjectmatchingsw]{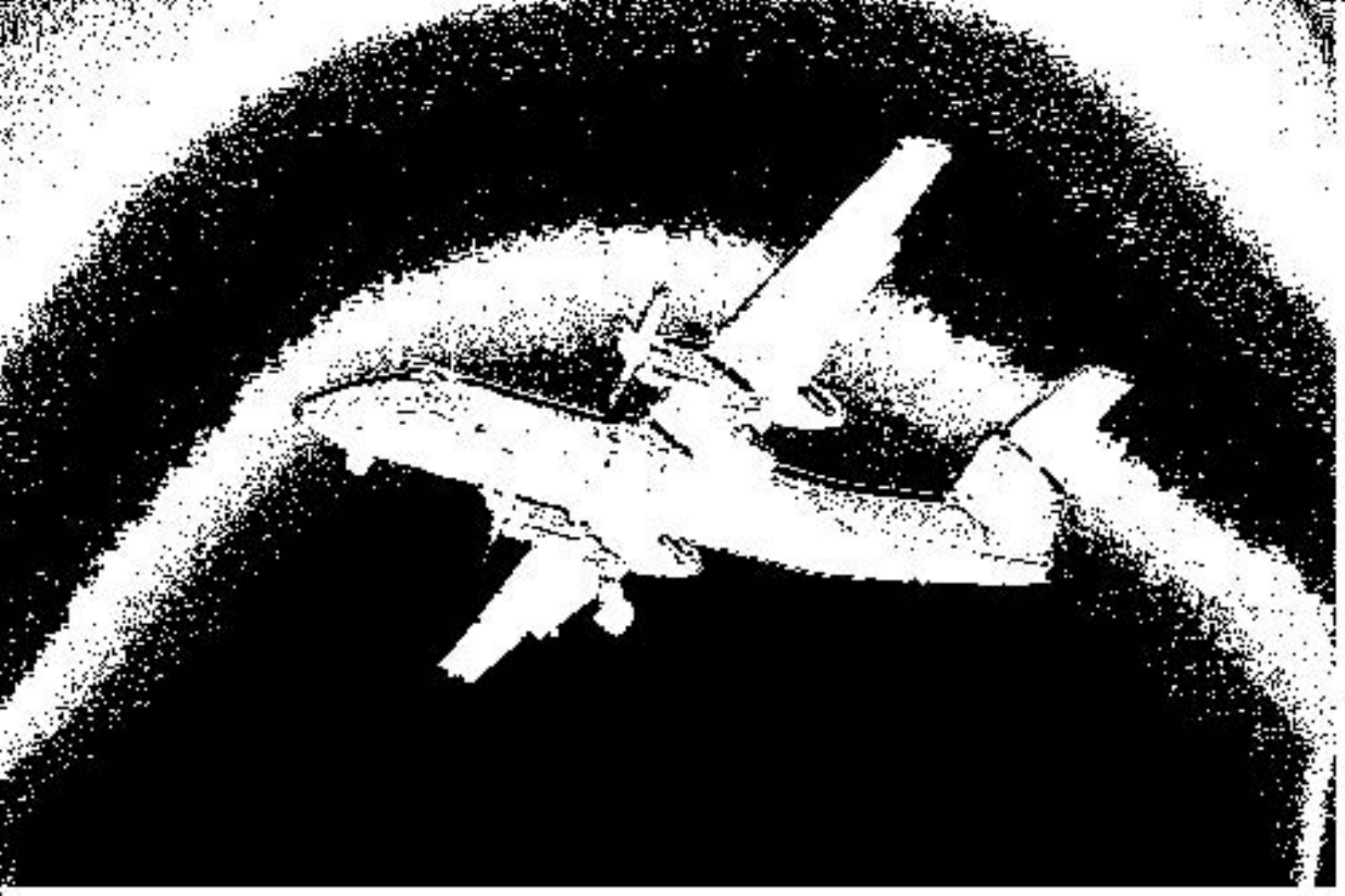} \\

	 \includegraphics[width=\figobjectmatchingw, height=\figobjectmatchingsw]{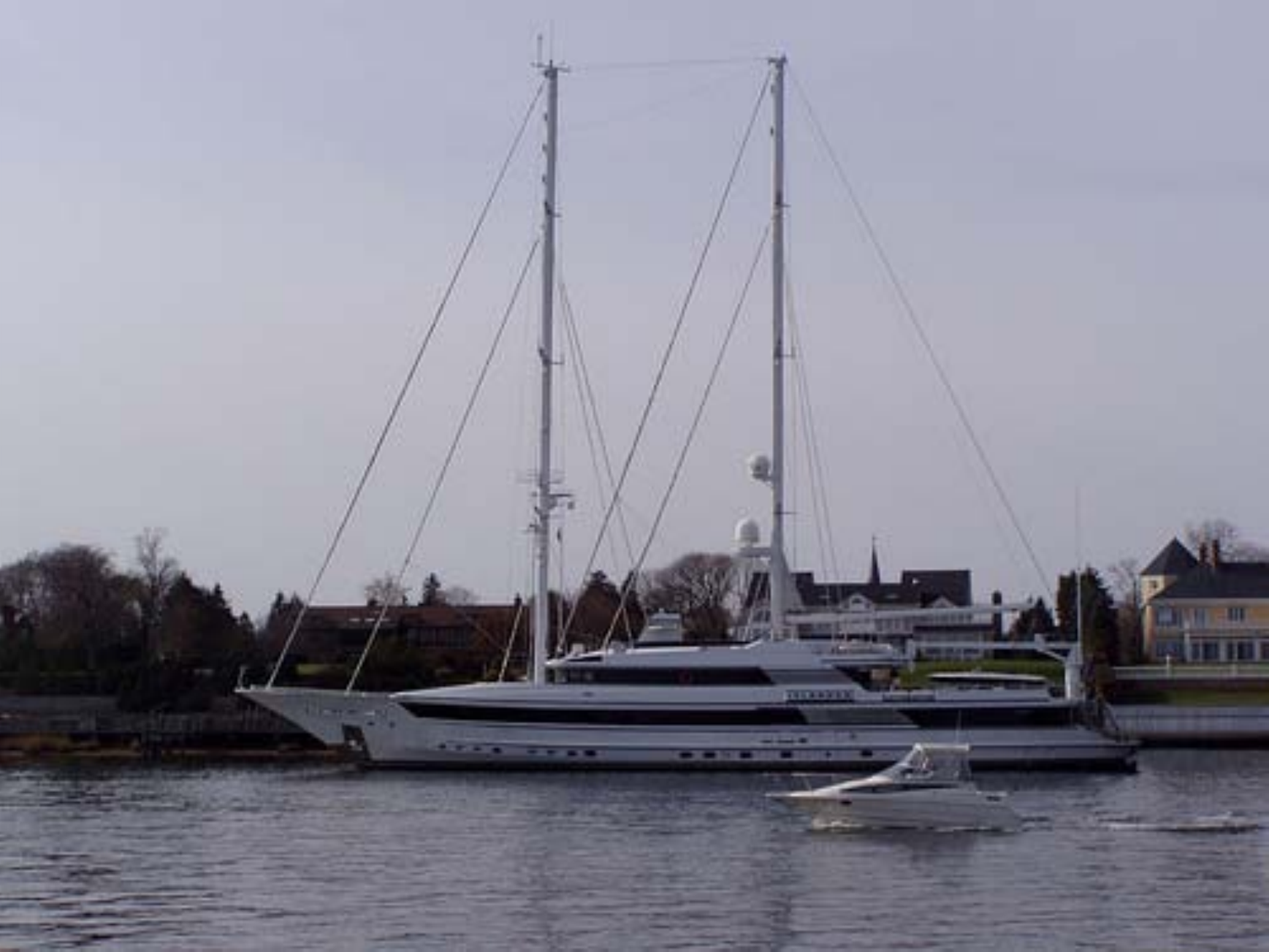}
	 \includegraphics[width=\figobjectmatchingw, height=\figobjectmatchingsw]{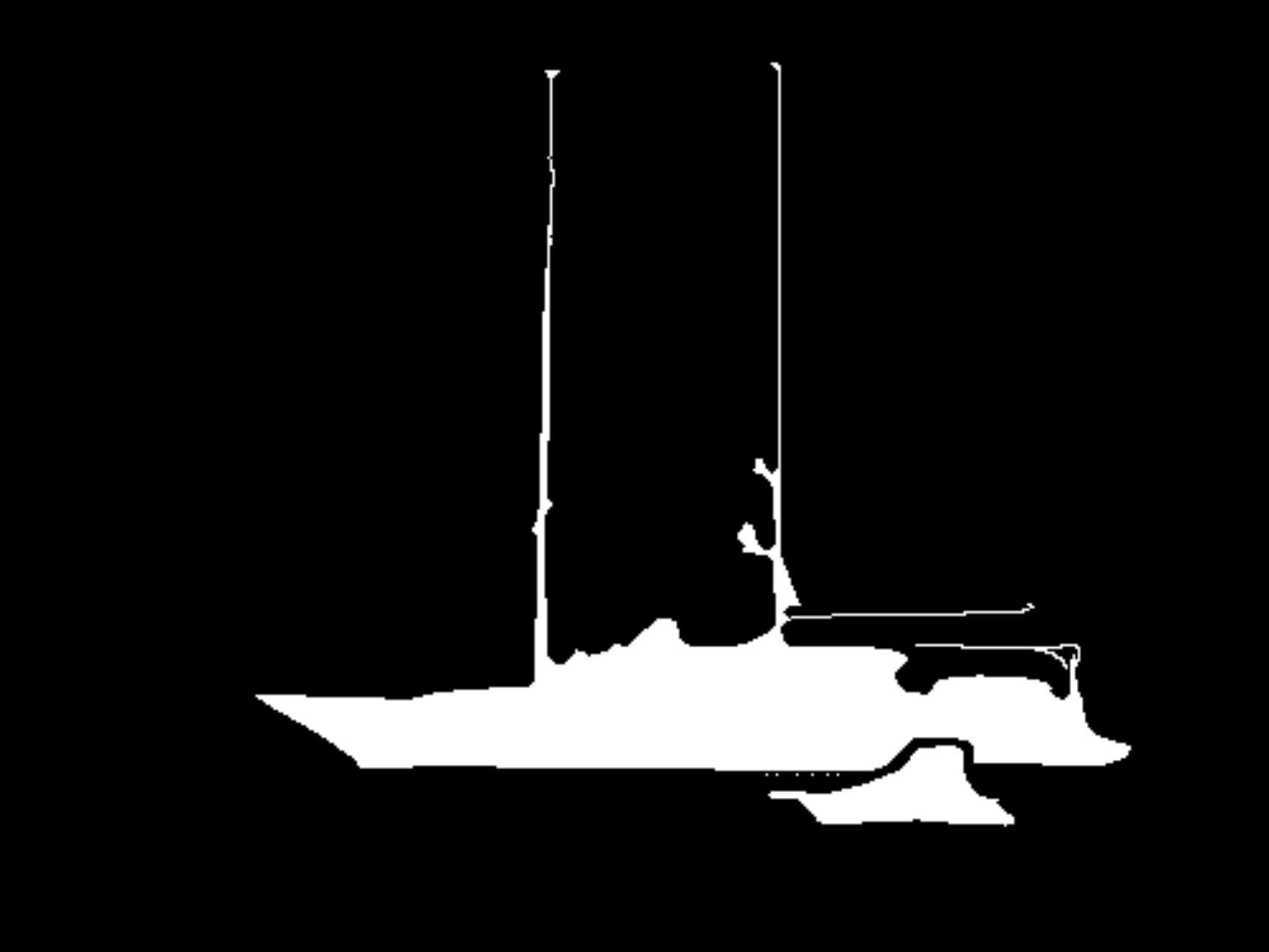}
	 \includegraphics[width=\figobjectmatchingw, height=\figobjectmatchingsw]{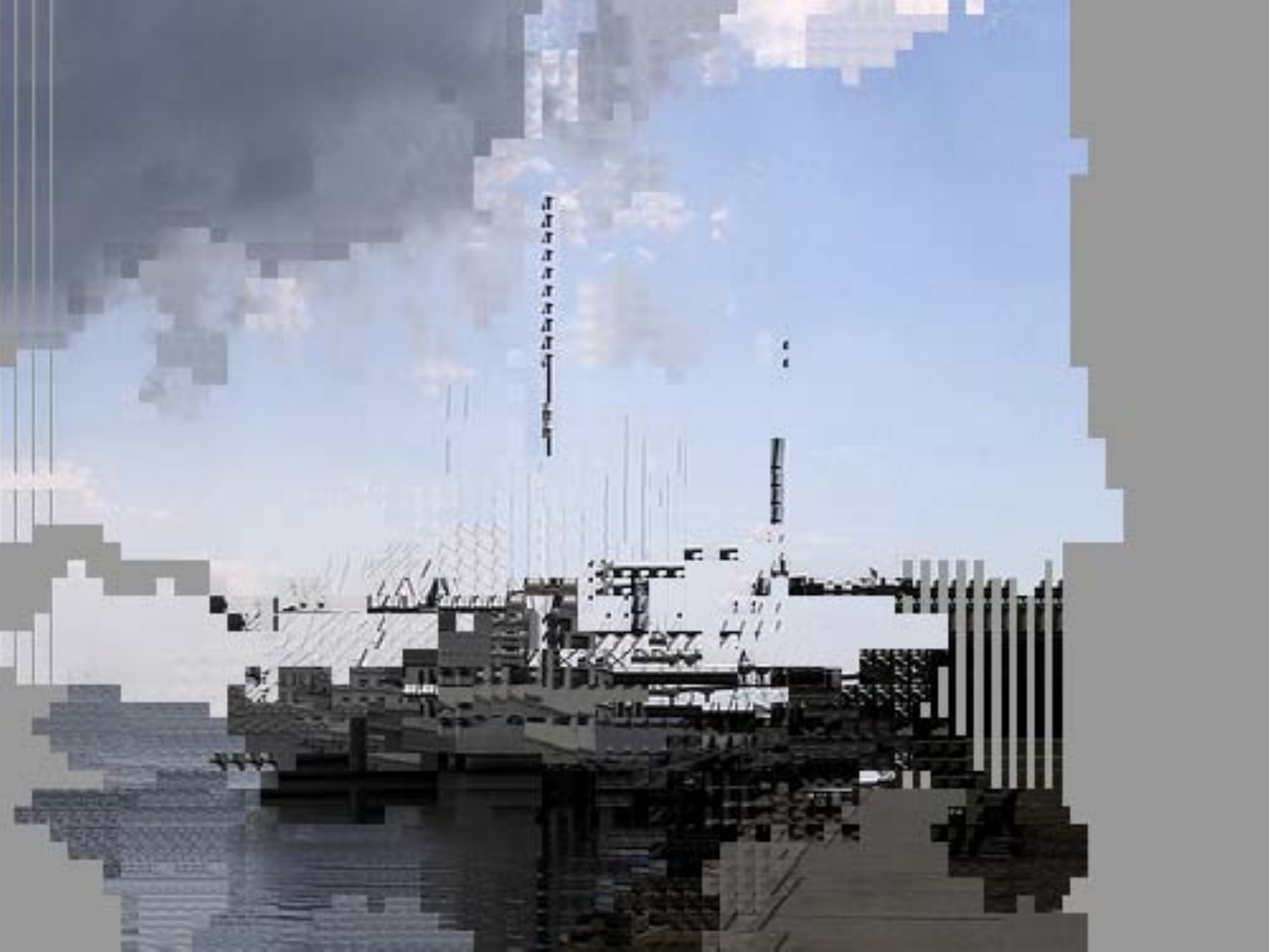}
     \includegraphics[width=\figobjectmatchingw, height=\figobjectmatchingsw]{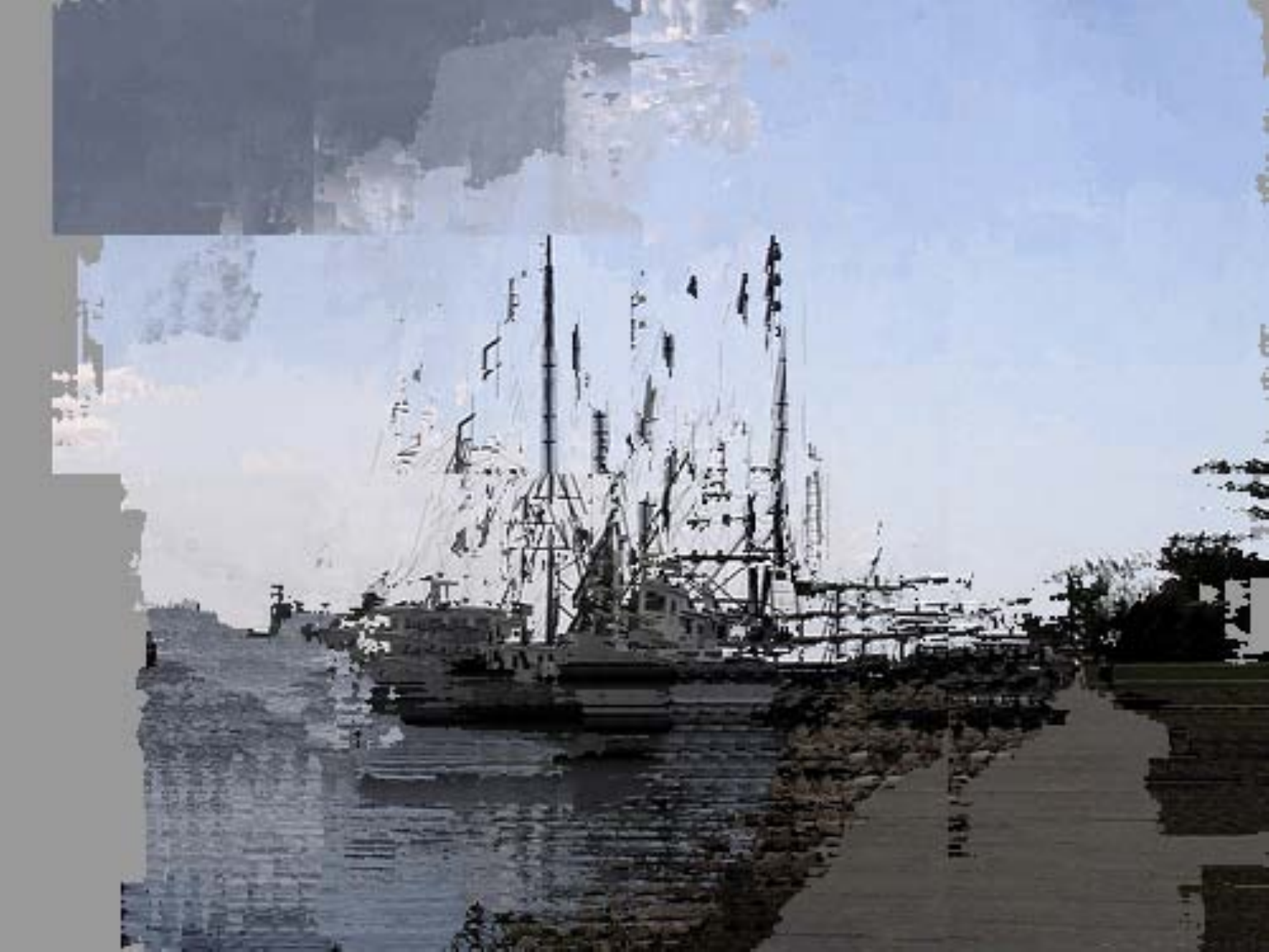}
    \includegraphics[width=\figobjectmatchingw, height=\figobjectmatchingsw]{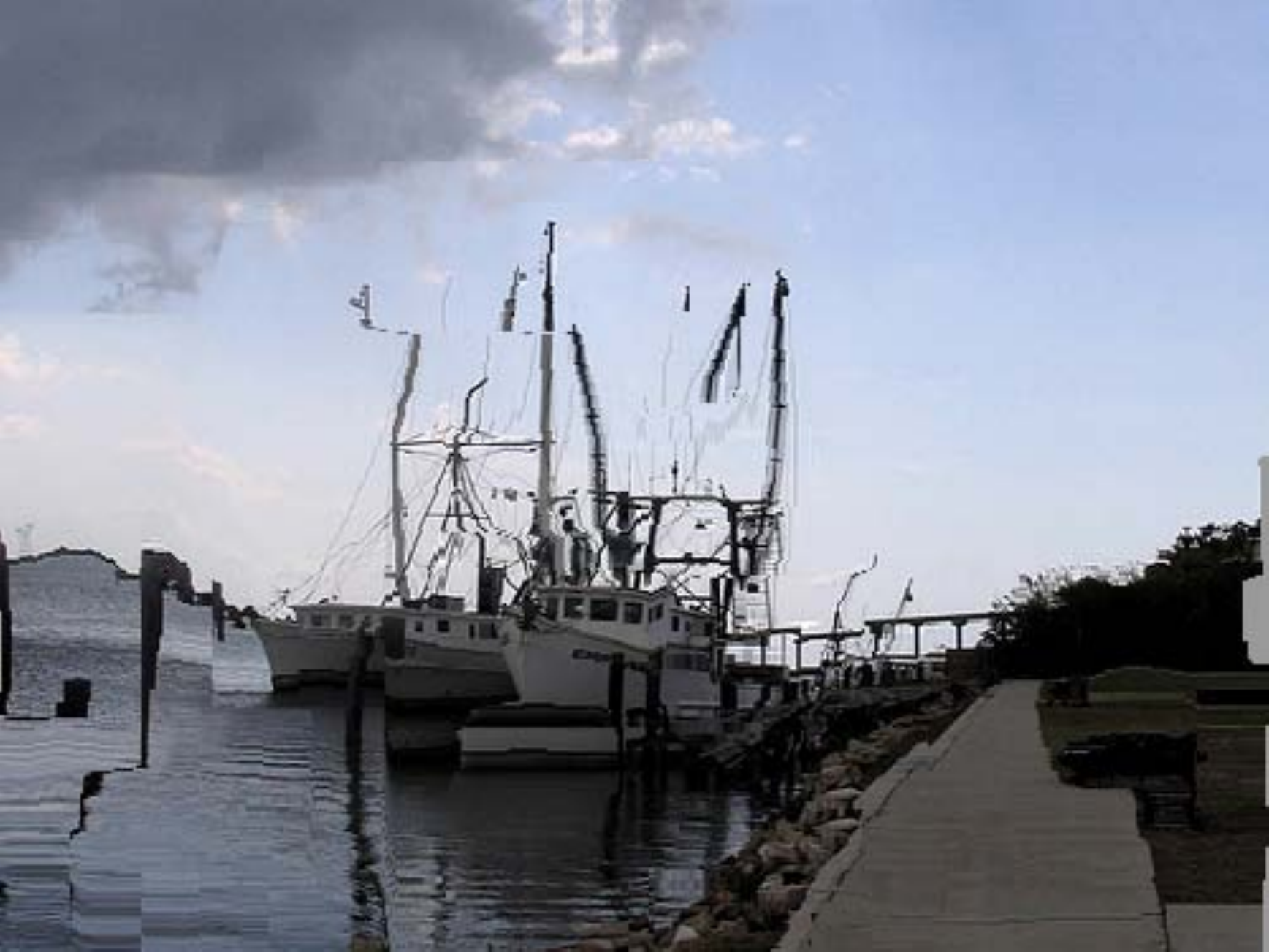}
     \includegraphics[width=\figobjectmatchingw, height=\figobjectmatchingsw]{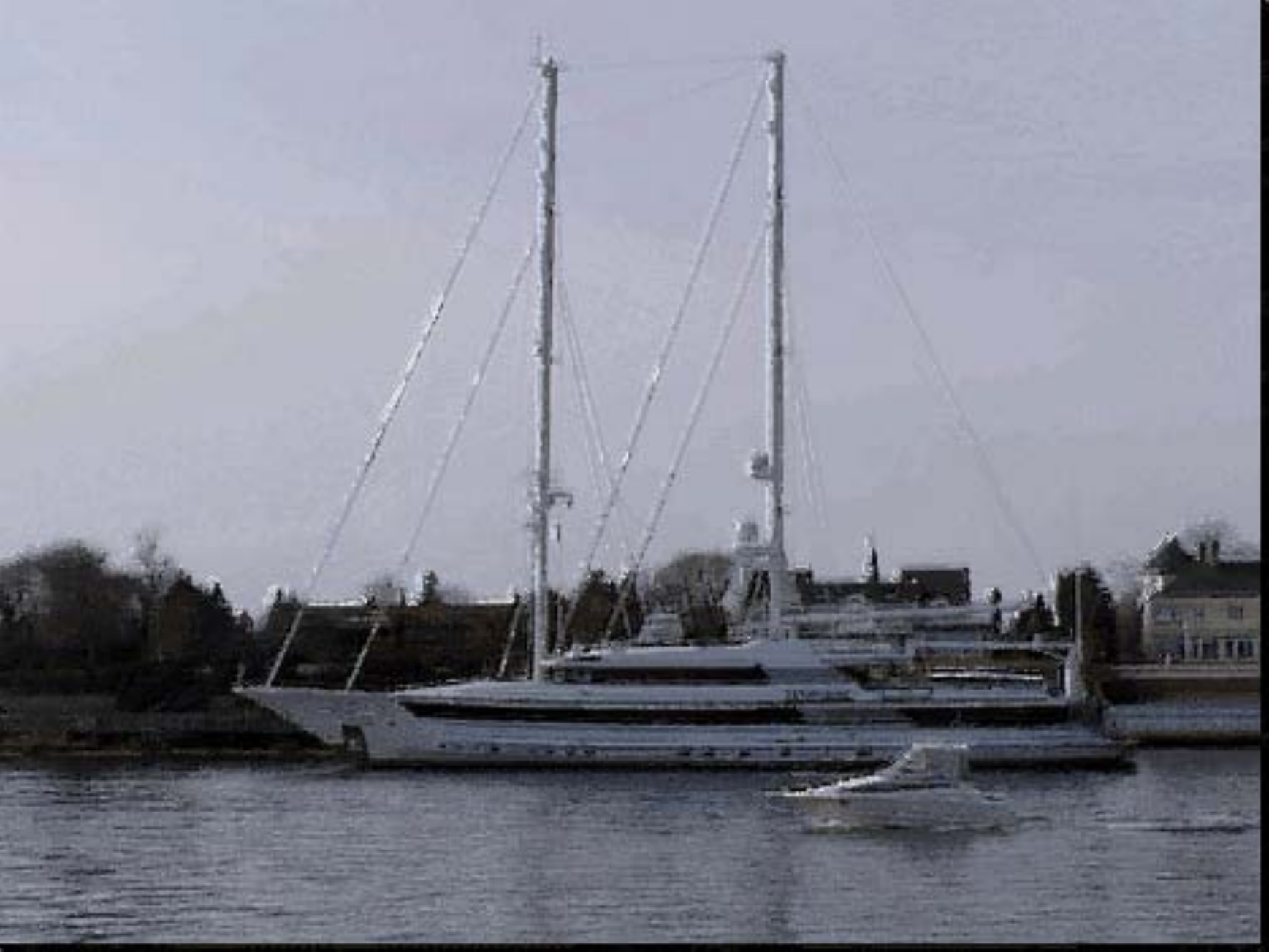} \\

 	 \includegraphics[width=\figobjectmatchingw, height=\figobjectmatchingsw]{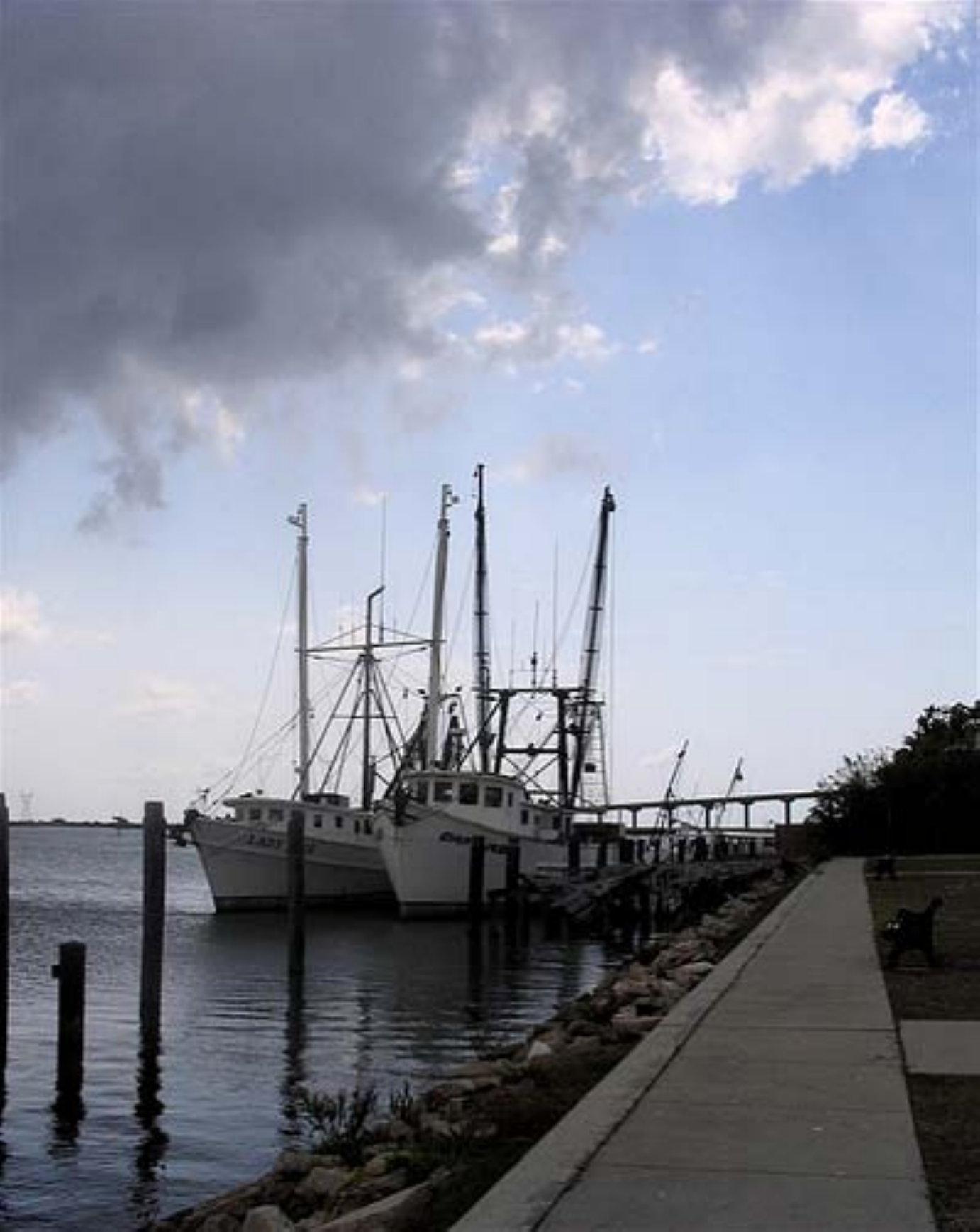}
	 \includegraphics[width=\figobjectmatchingw, height=\figobjectmatchingsw]{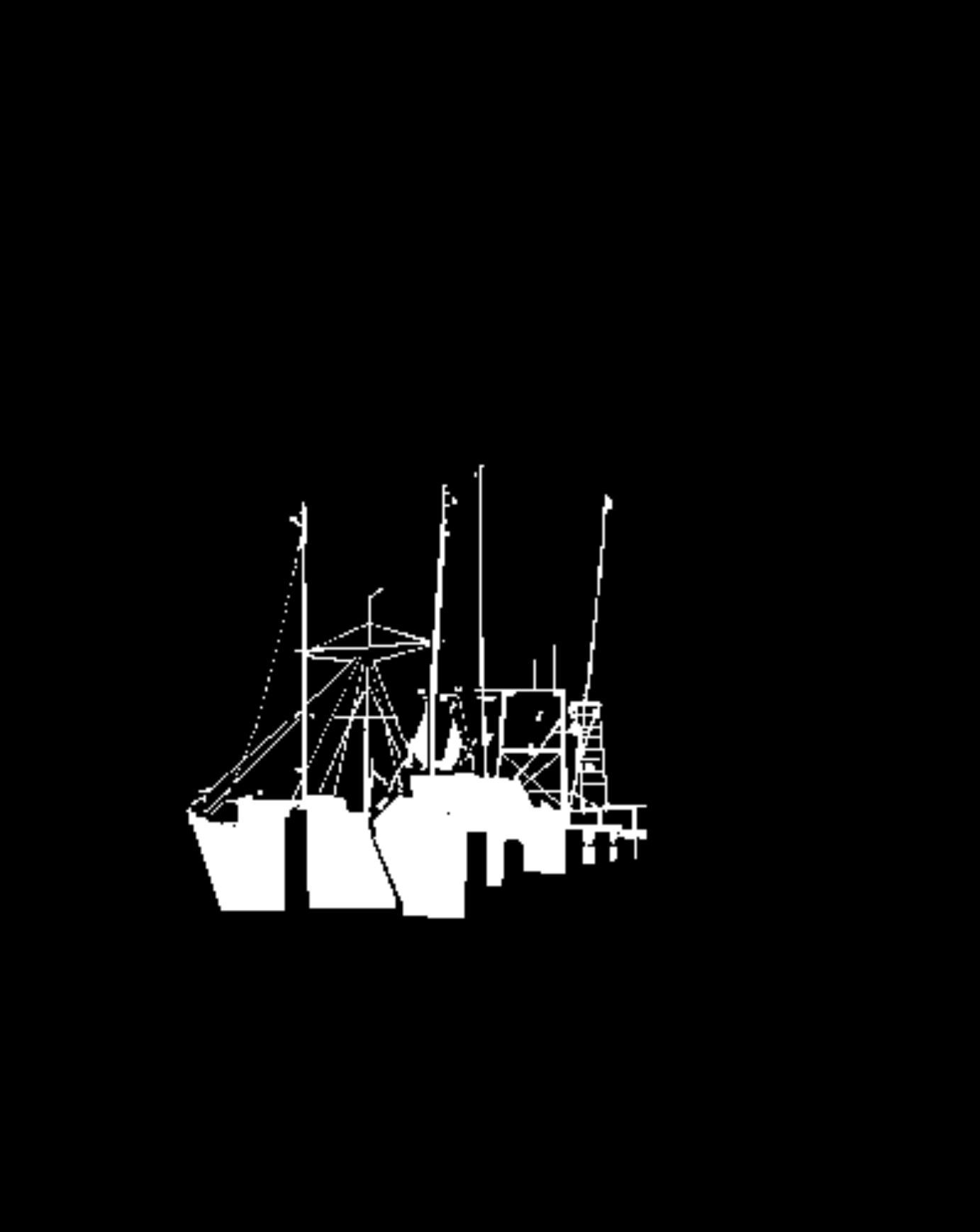}
	 \includegraphics[width=\figobjectmatchingw, height=\figobjectmatchingsw]{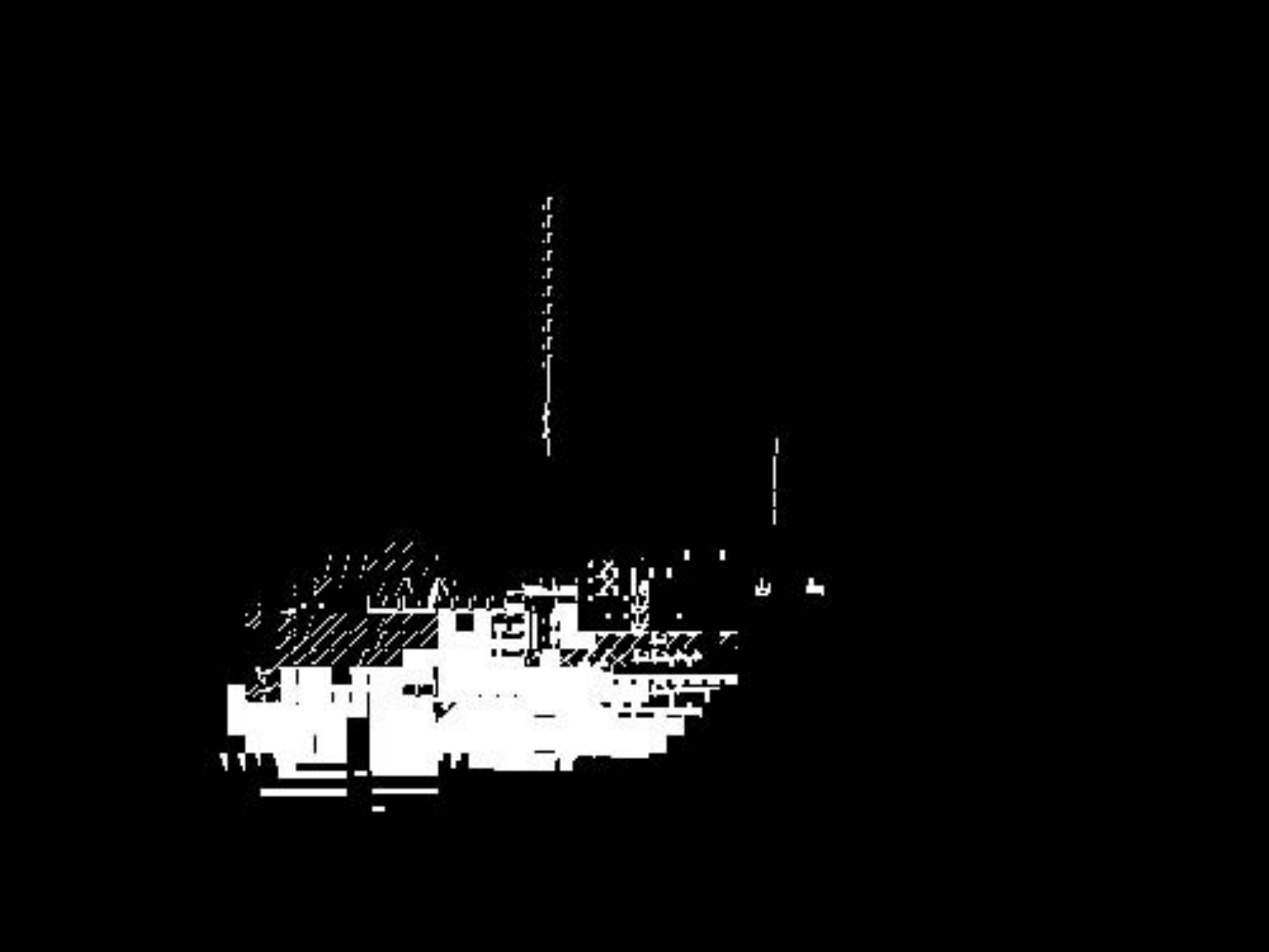}
	 \includegraphics[width=\figobjectmatchingw, height=\figobjectmatchingsw]{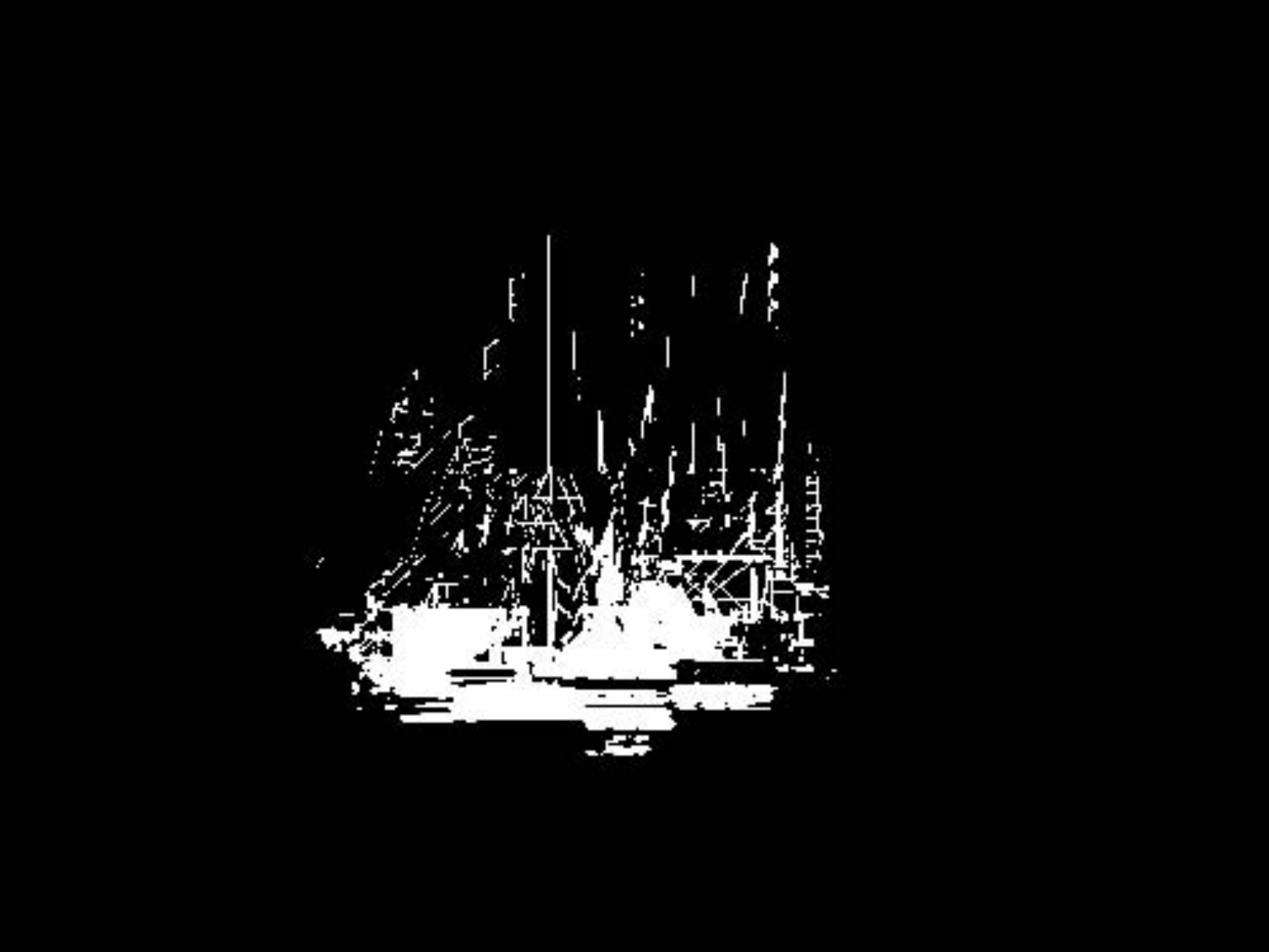}
	 \includegraphics[width=\figobjectmatchingw, height=\figobjectmatchingsw]{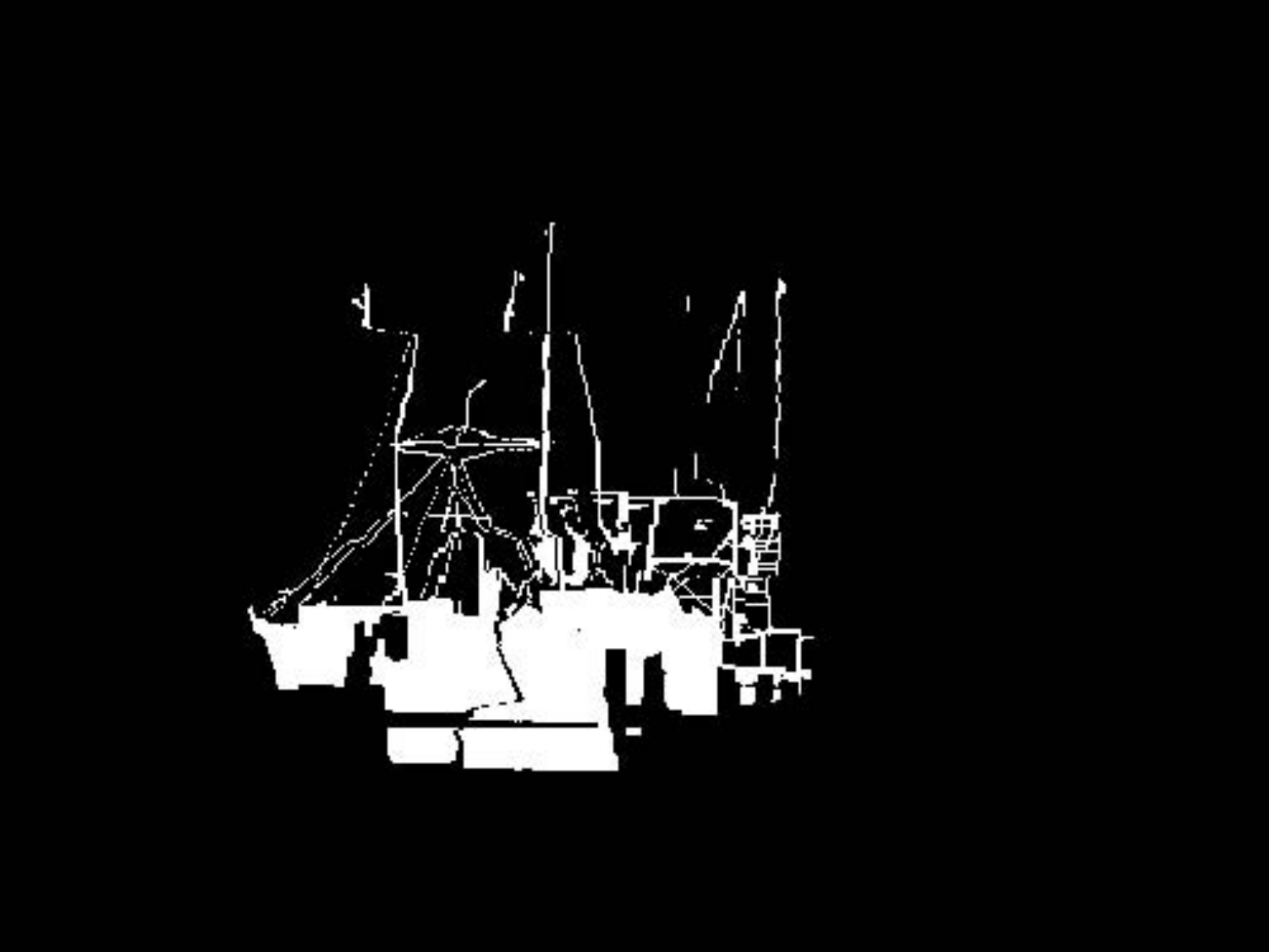}
	 \includegraphics[width=\figobjectmatchingw, height=\figobjectmatchingsw]{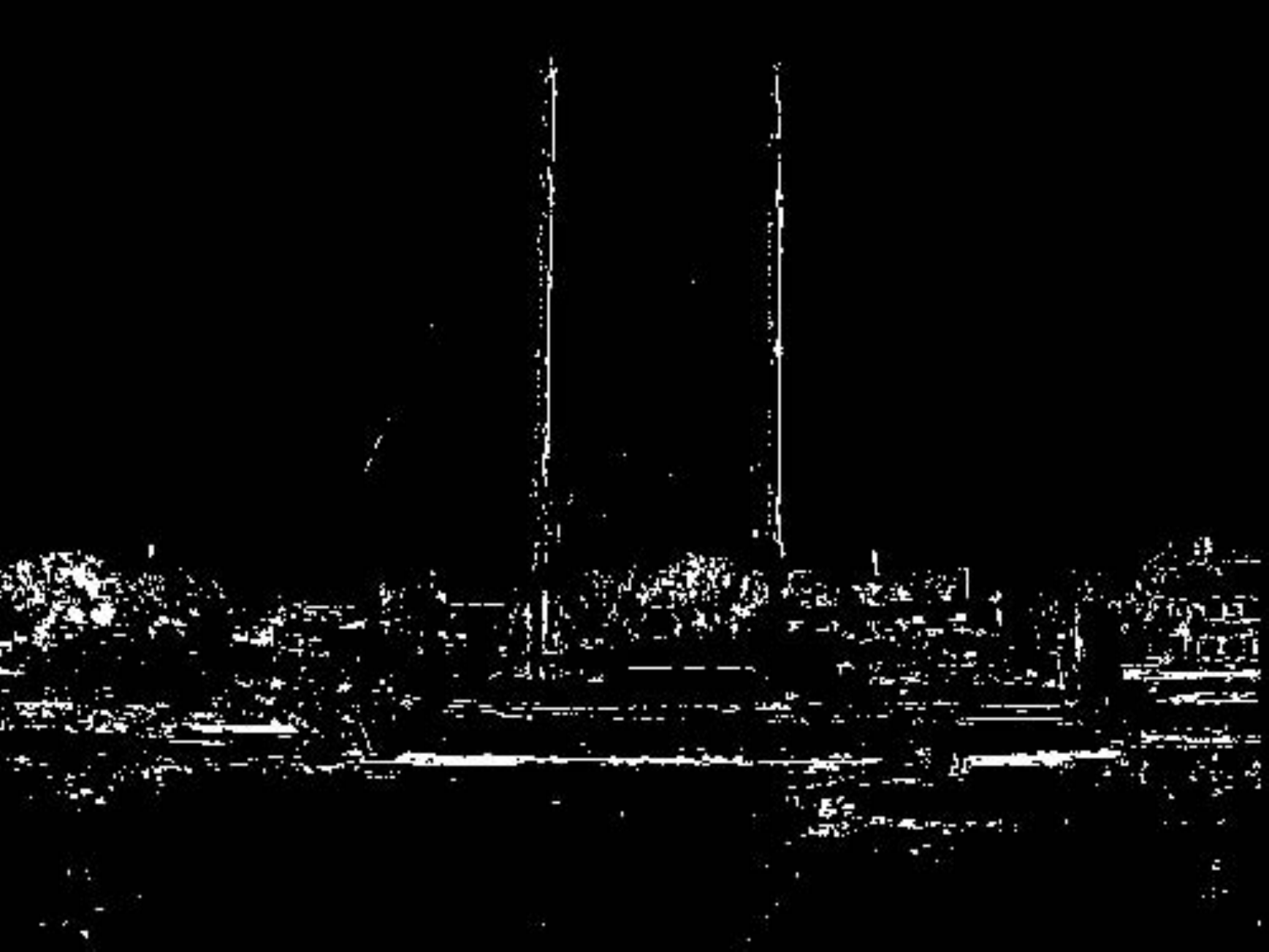} \\

	 \includegraphics[width=\figobjectmatchingw, height=\figobjectmatchingsw]{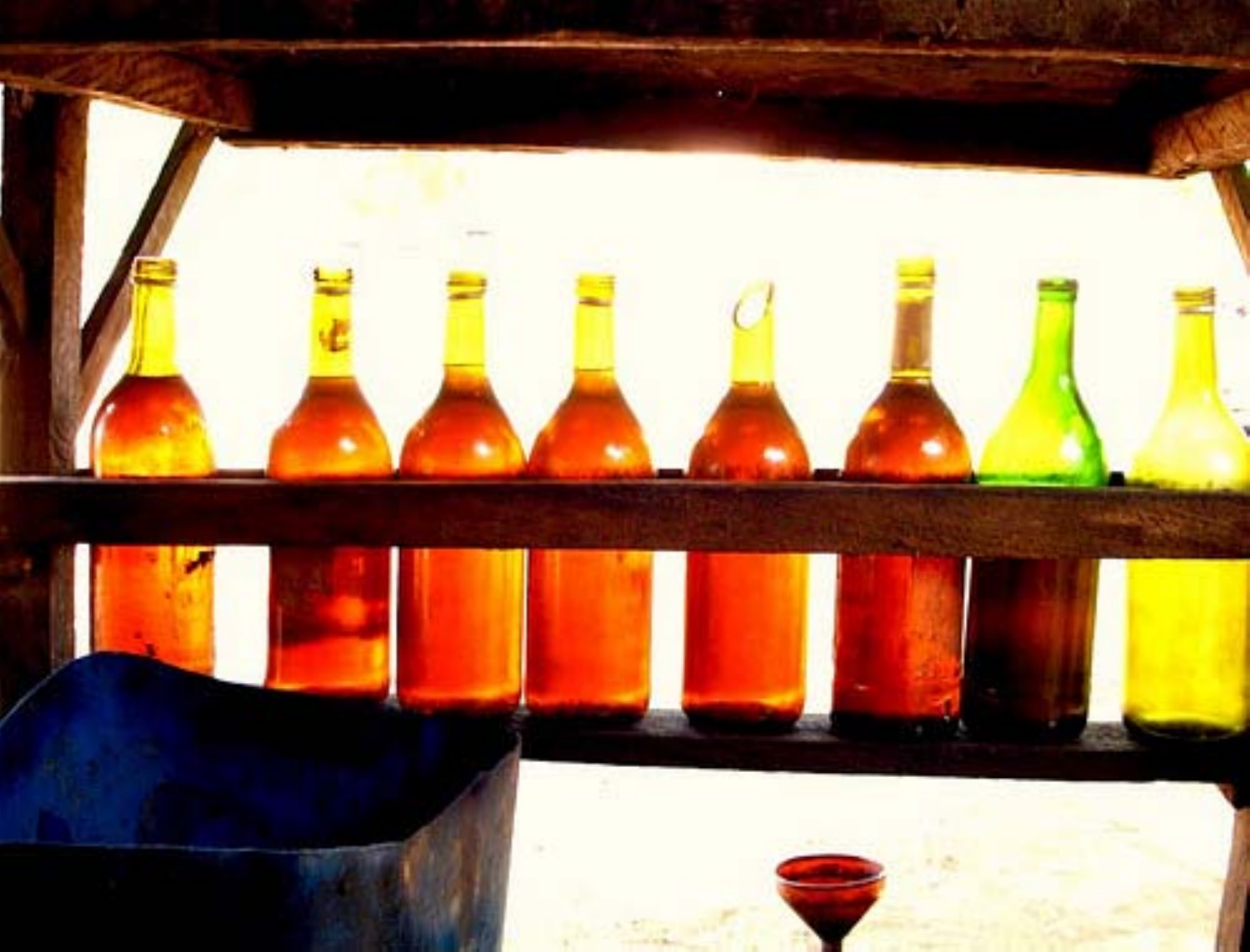}
	 \includegraphics[width=\figobjectmatchingw, height=\figobjectmatchingsw]{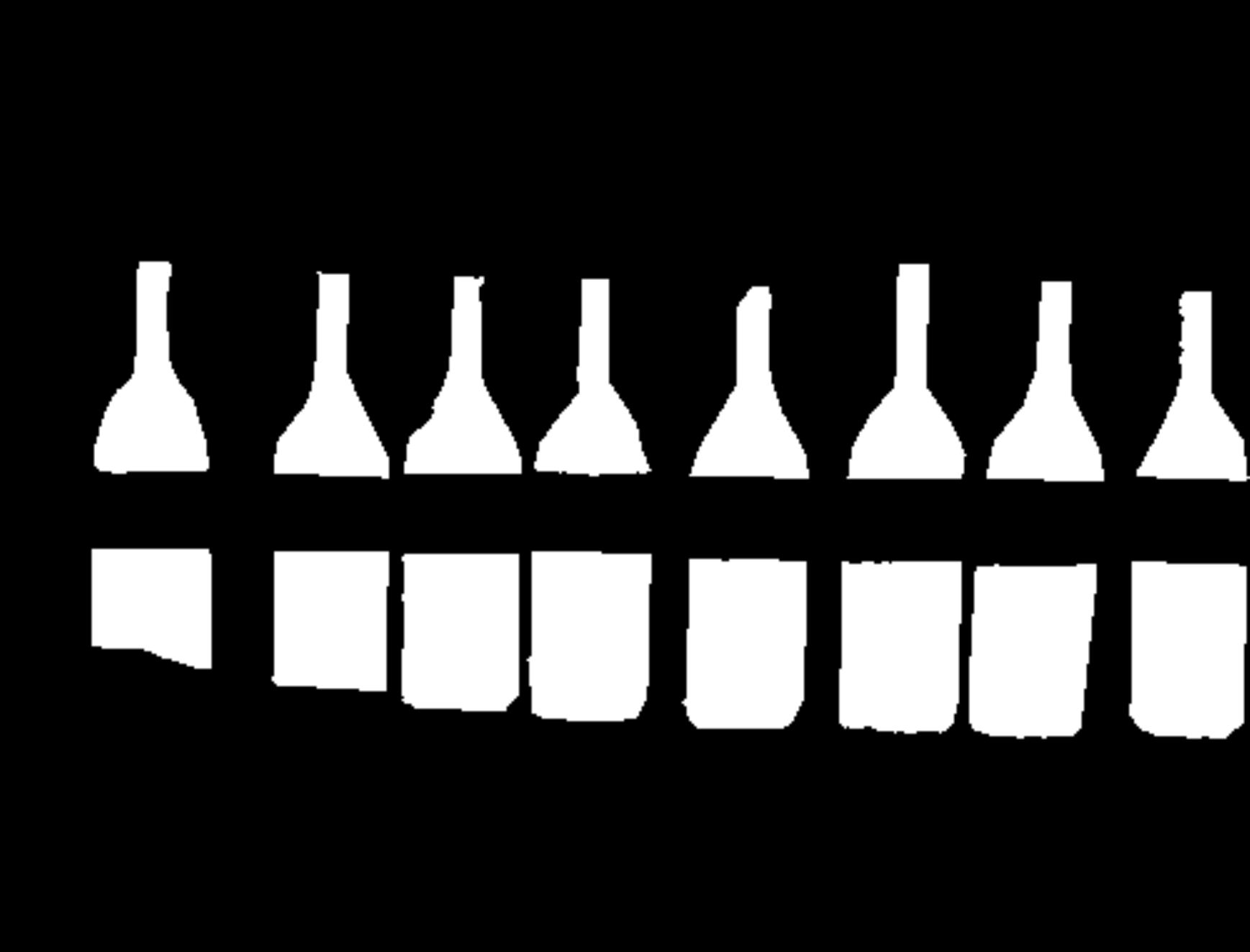}
	 \includegraphics[width=\figobjectmatchingw, height=\figobjectmatchingsw]{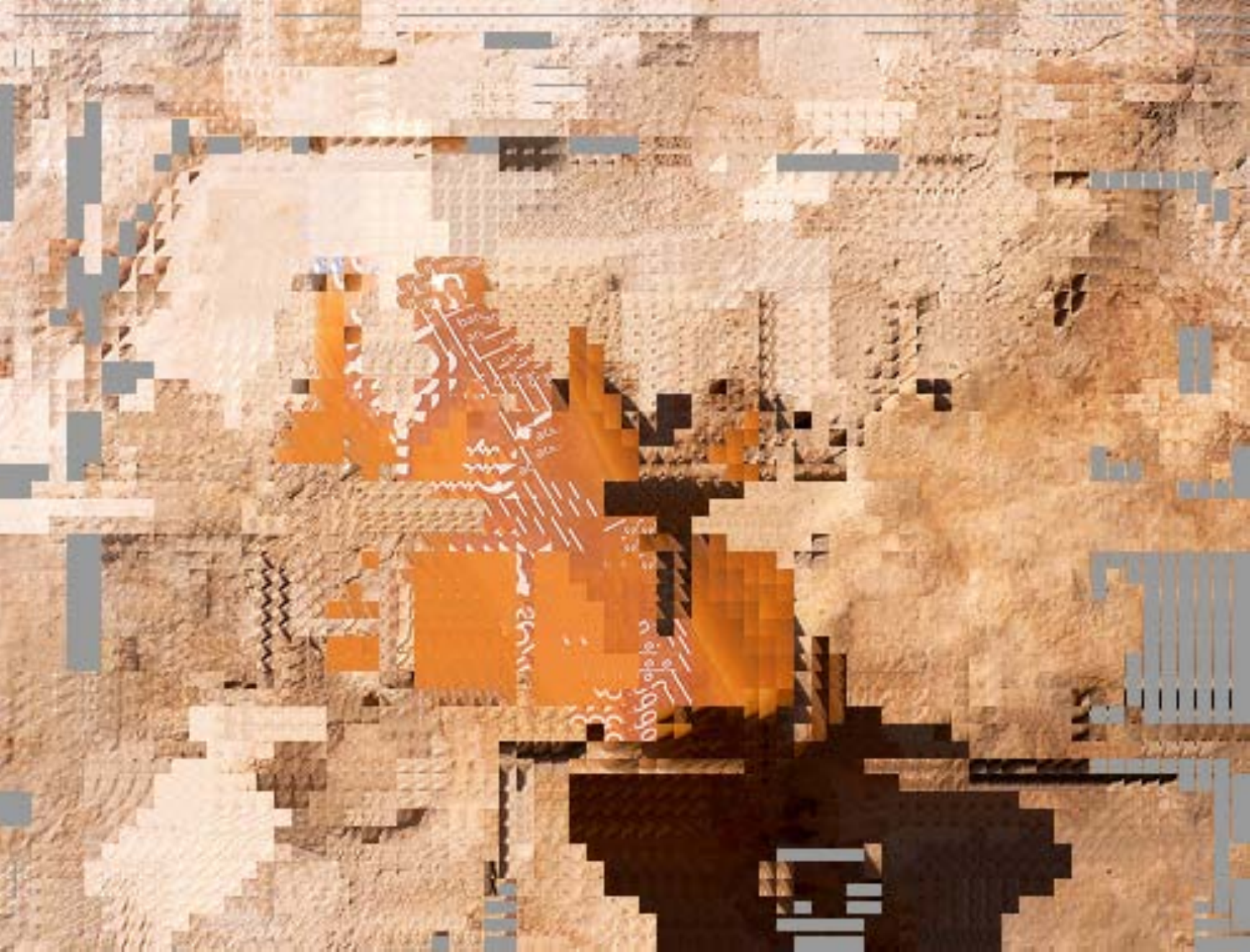}
     \includegraphics[width=\figobjectmatchingw, height=\figobjectmatchingsw]{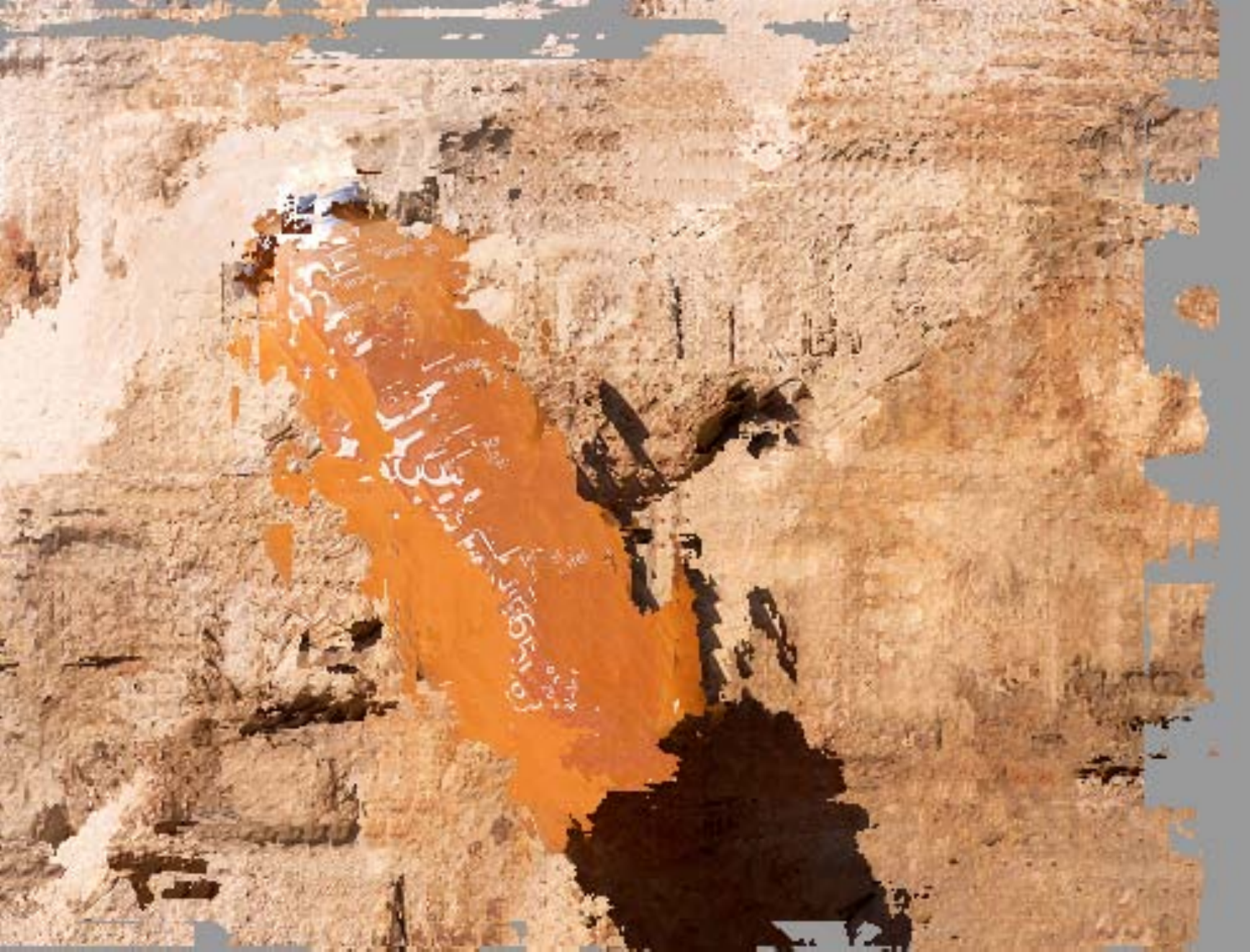}
    \includegraphics[width=\figobjectmatchingw, height=\figobjectmatchingsw]{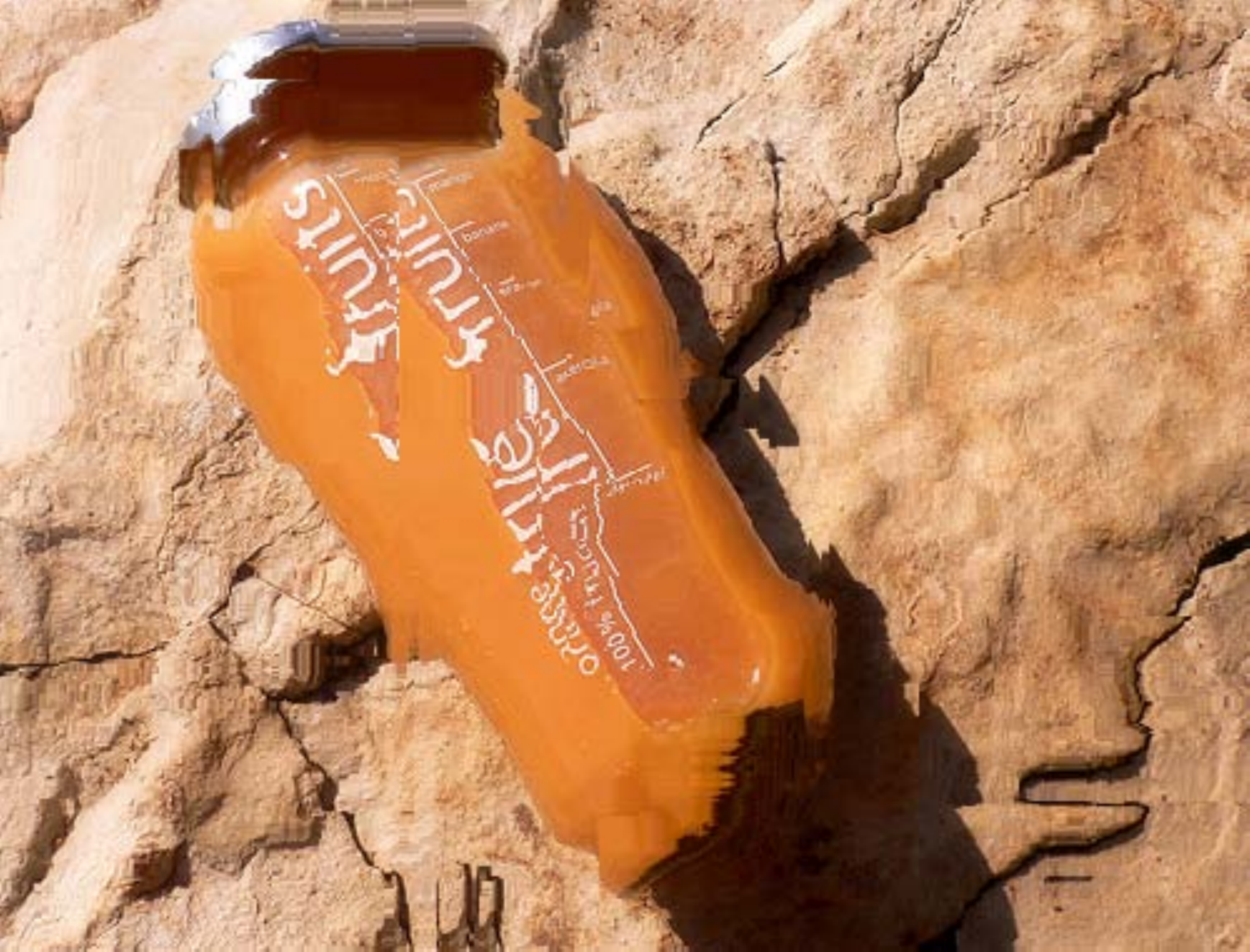}
     \includegraphics[width=\figobjectmatchingw, height=\figobjectmatchingsw]{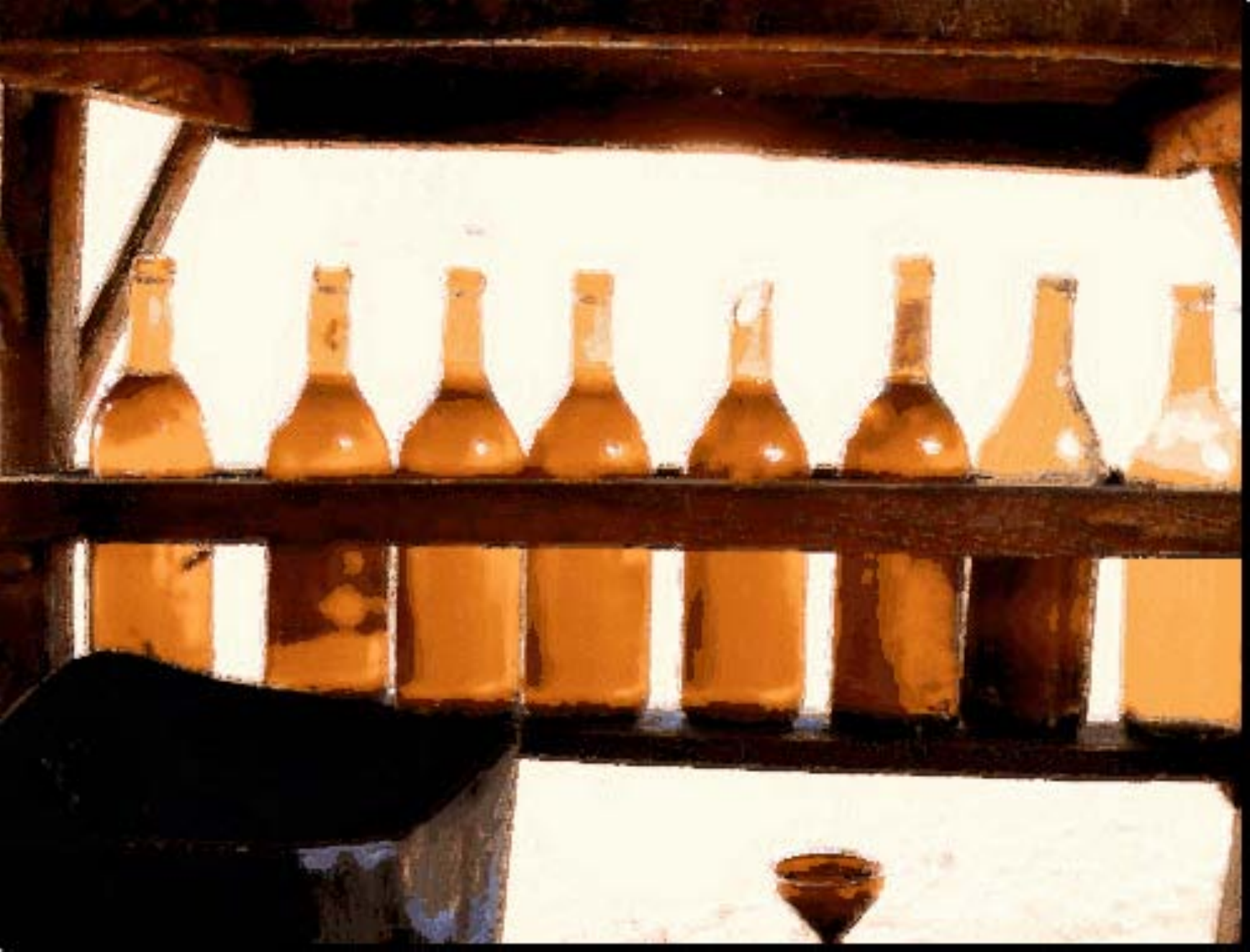} \\

 	 \includegraphics[width=\figobjectmatchingw, height=\figobjectmatchingsw]{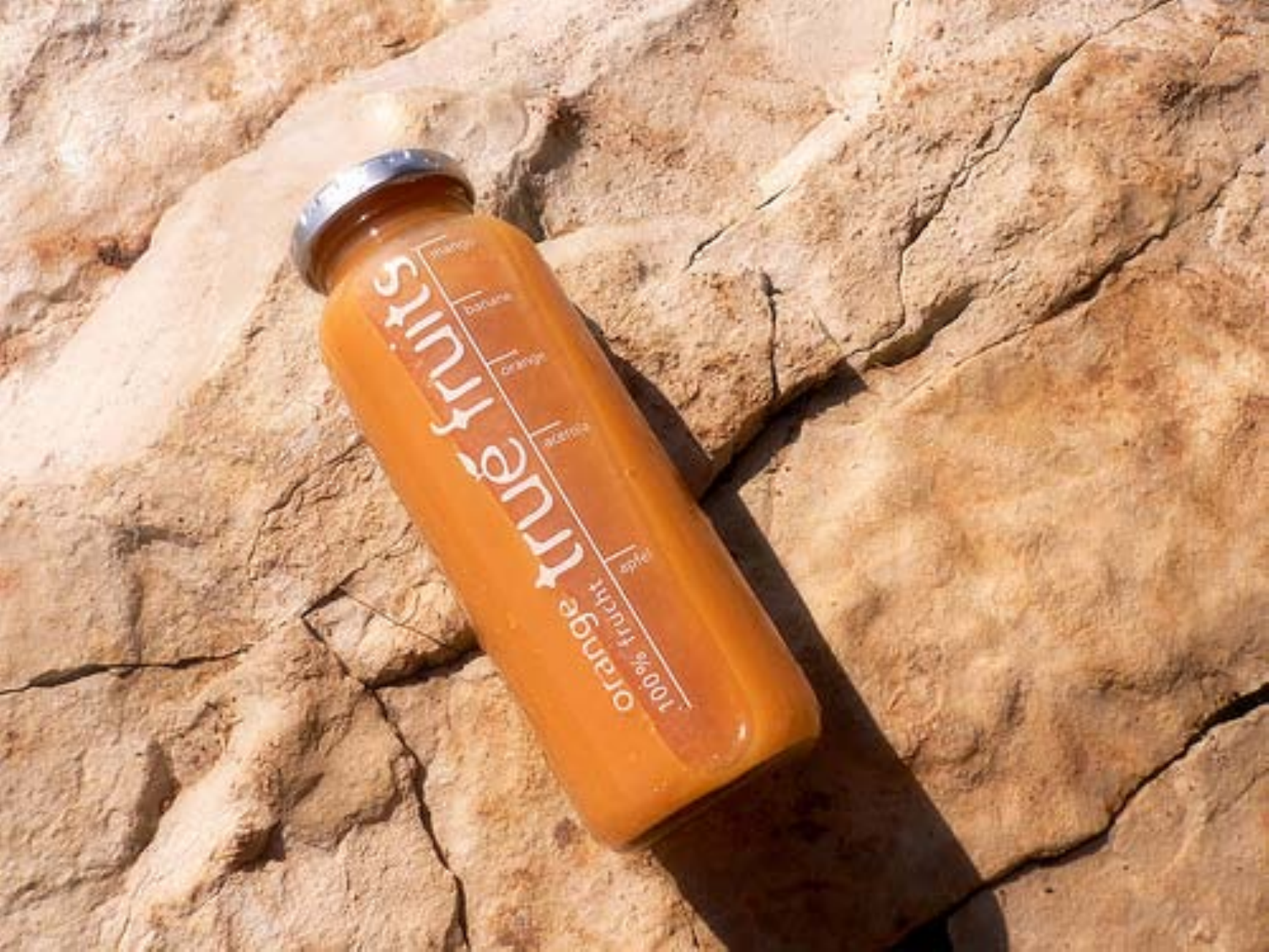}
	 \includegraphics[width=\figobjectmatchingw, height=\figobjectmatchingsw]{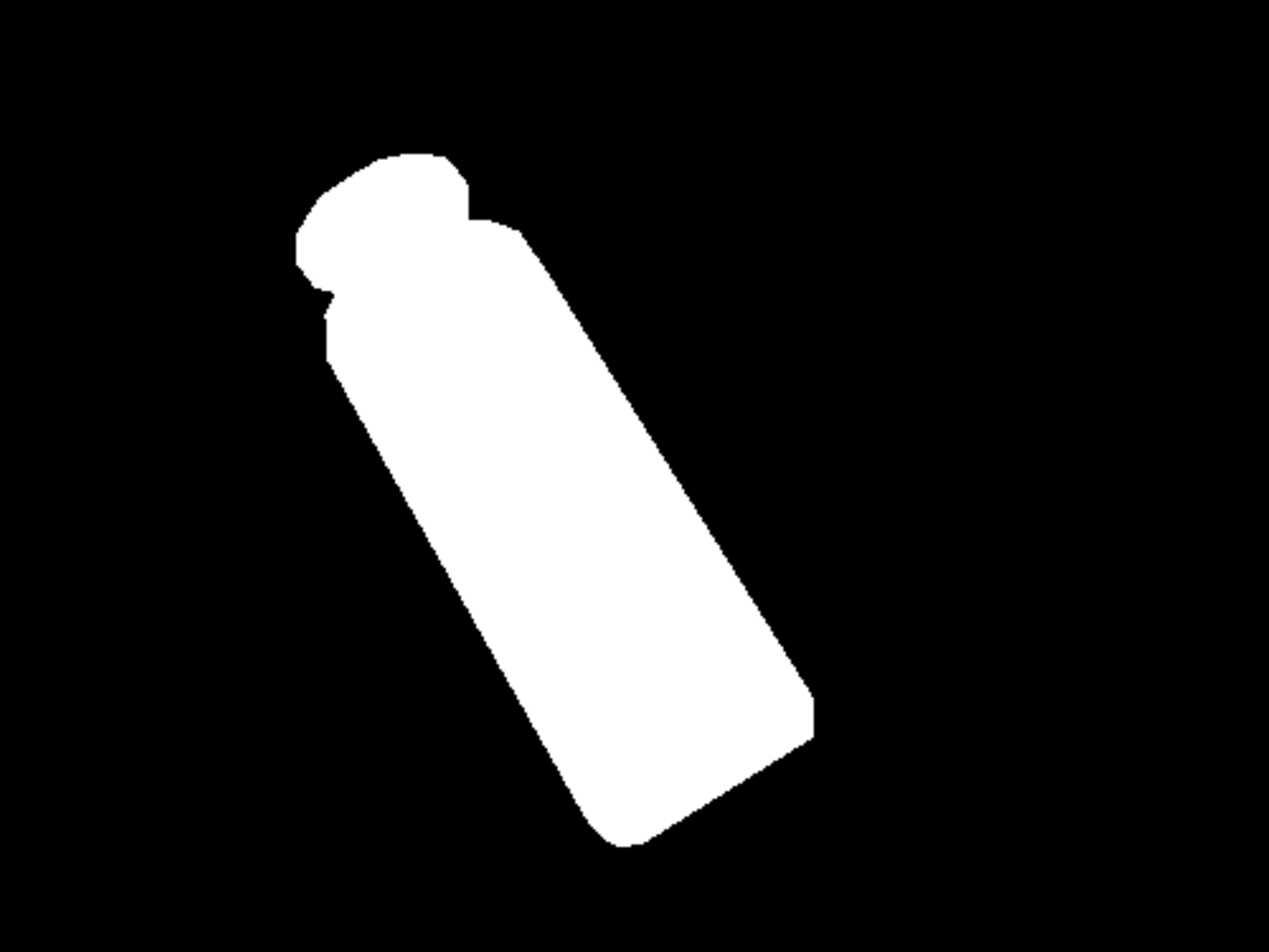}
	 \includegraphics[width=\figobjectmatchingw, height=\figobjectmatchingsw]{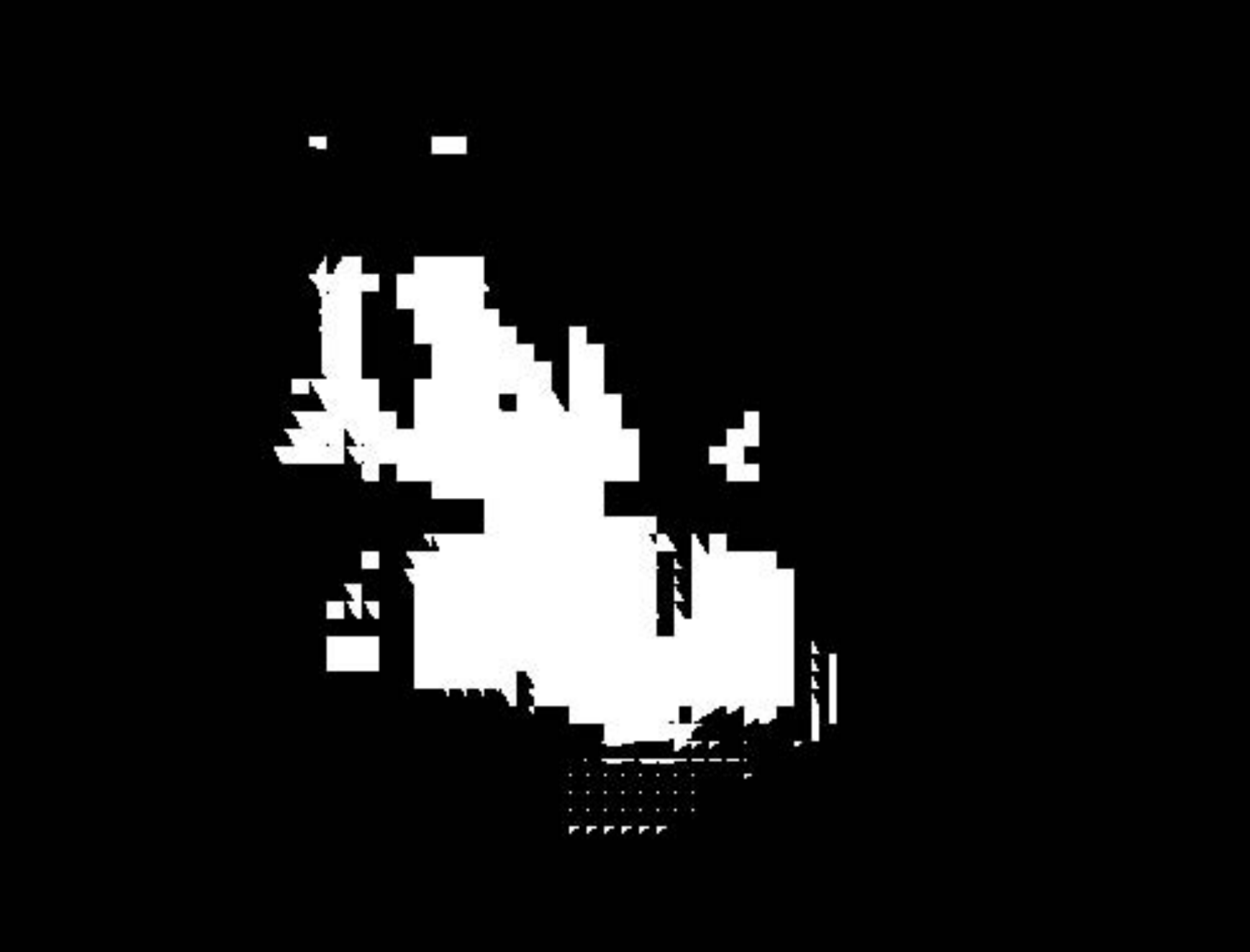}
	 \includegraphics[width=\figobjectmatchingw, height=\figobjectmatchingsw]{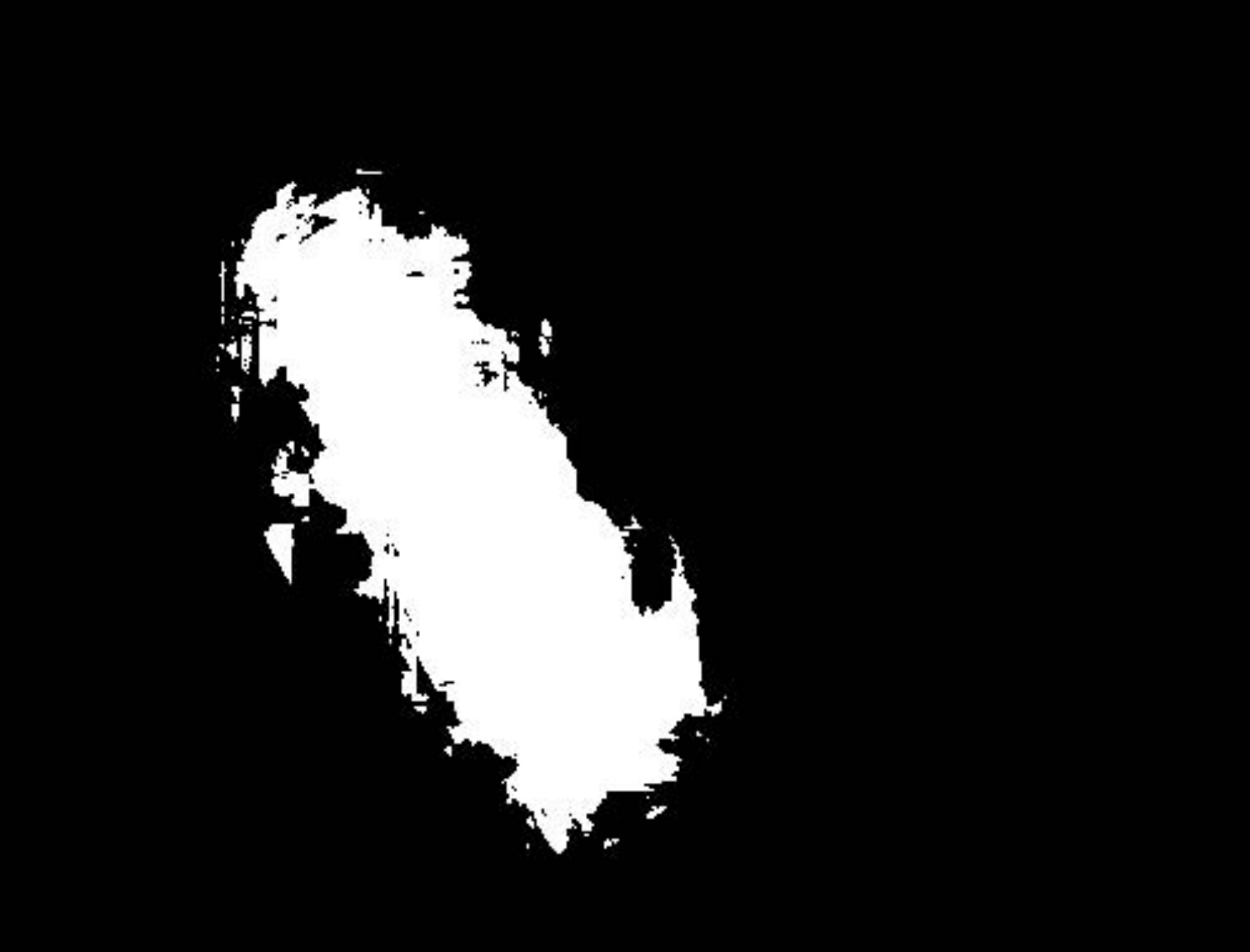}
	 \includegraphics[width=\figobjectmatchingw, height=\figobjectmatchingsw]{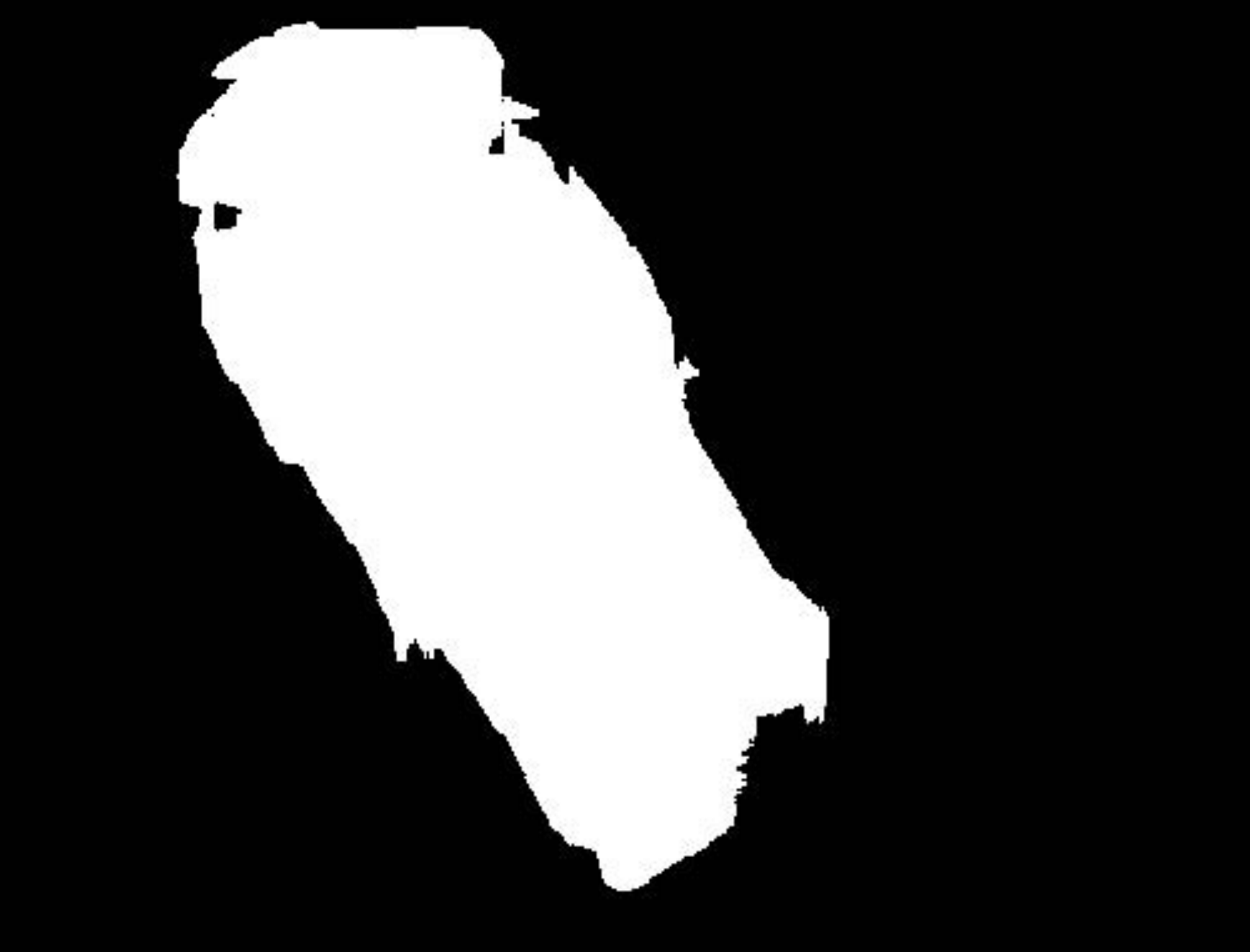}
	 \includegraphics[width=\figobjectmatchingw, height=\figobjectmatchingsw]{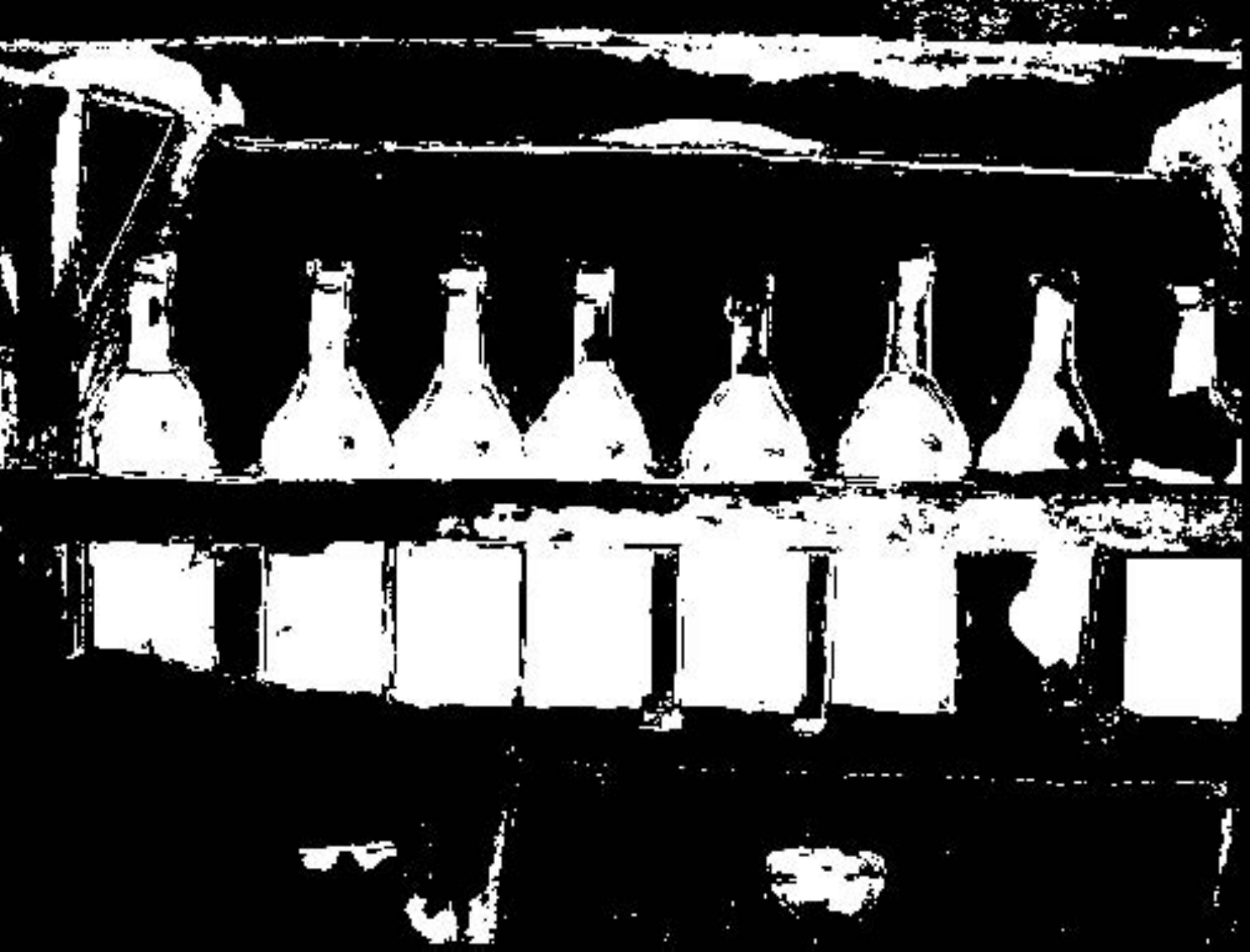} \\

	 \includegraphics[width=\figobjectmatchingw, height=\figobjectmatchingsw]{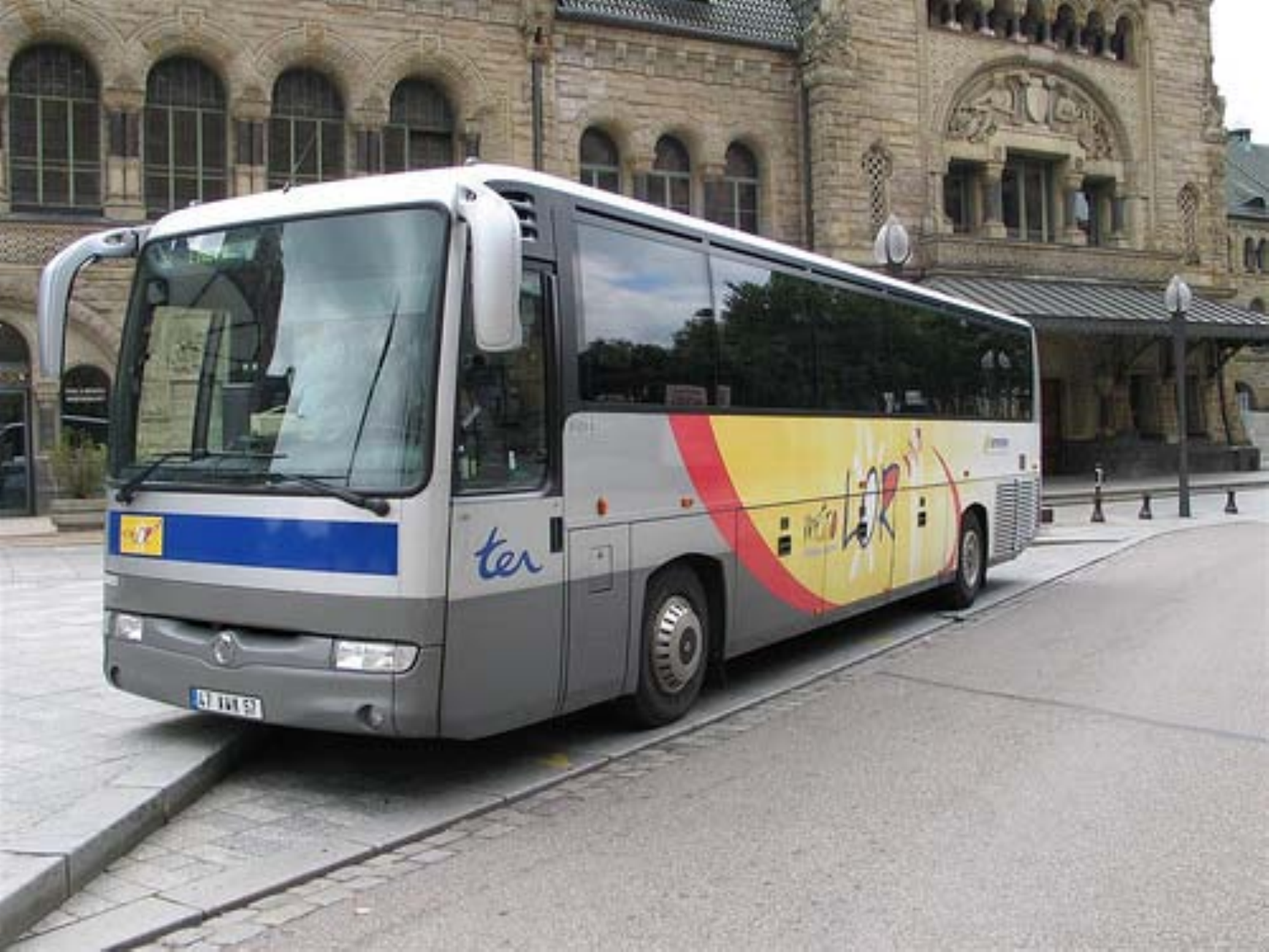}
	 \includegraphics[width=\figobjectmatchingw, height=\figobjectmatchingsw]{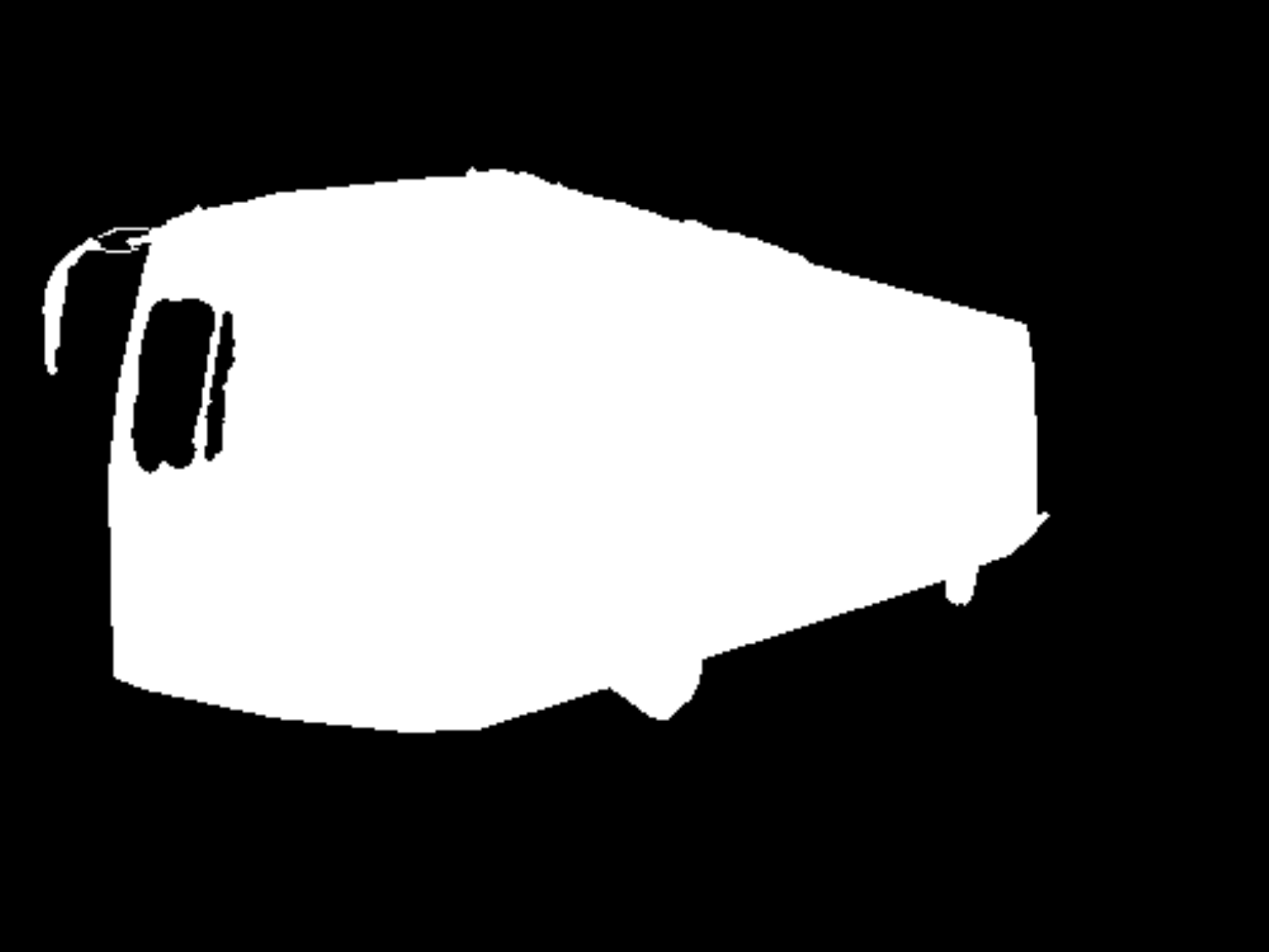}
	 \includegraphics[width=\figobjectmatchingw, height=\figobjectmatchingsw]{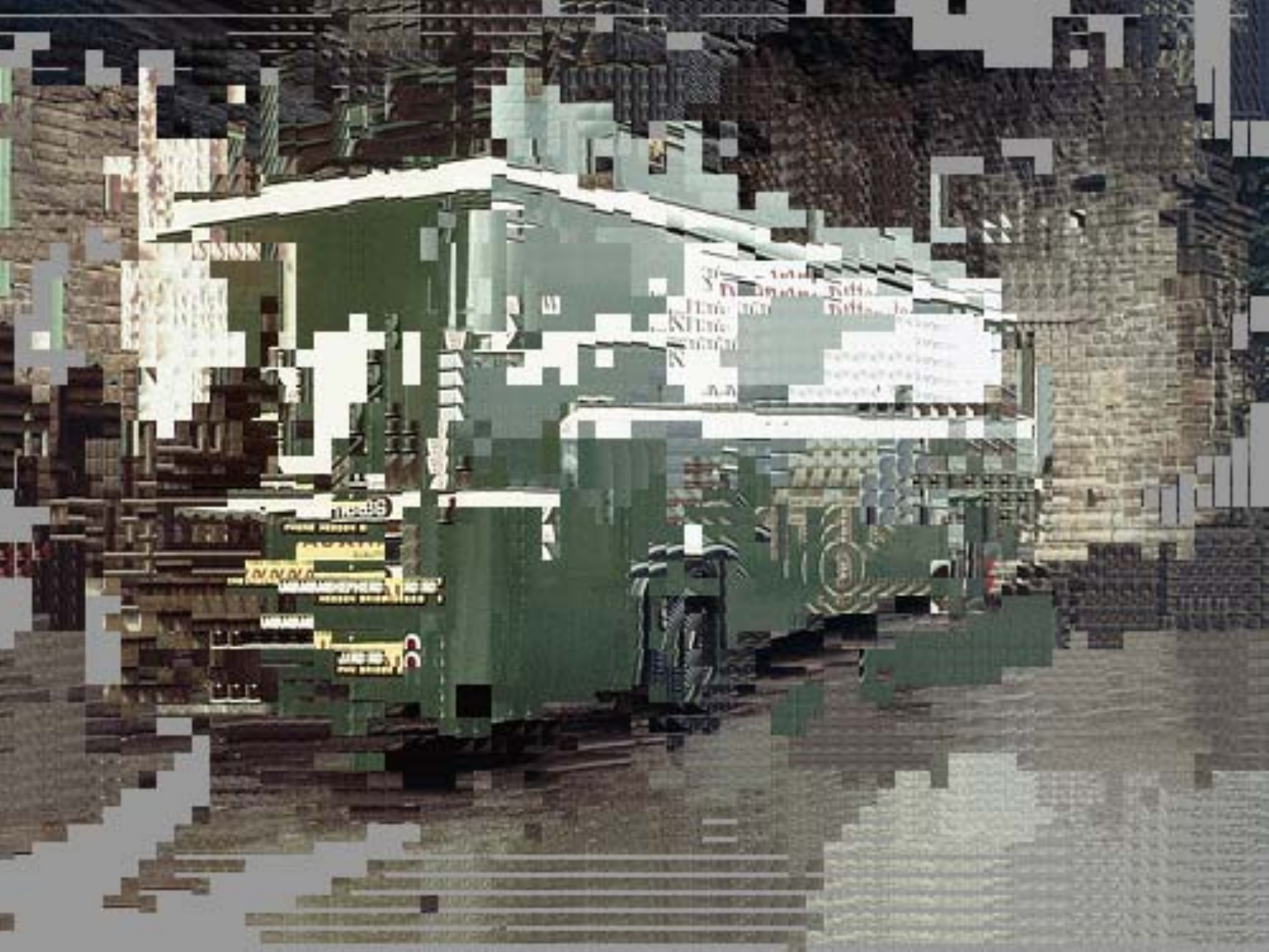}
     \includegraphics[width=\figobjectmatchingw, height=\figobjectmatchingsw]{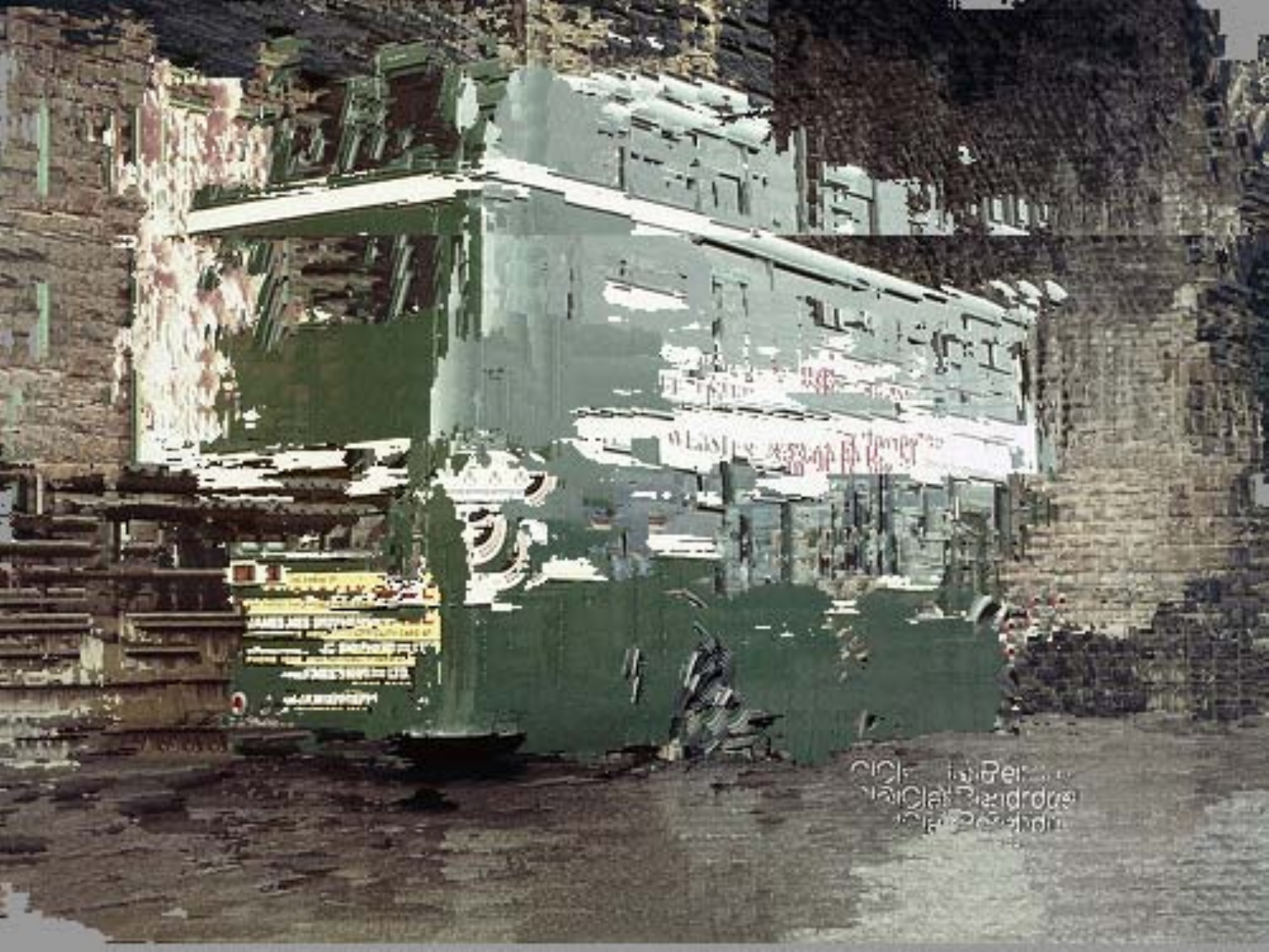}
    \includegraphics[width=\figobjectmatchingw, height=\figobjectmatchingsw]{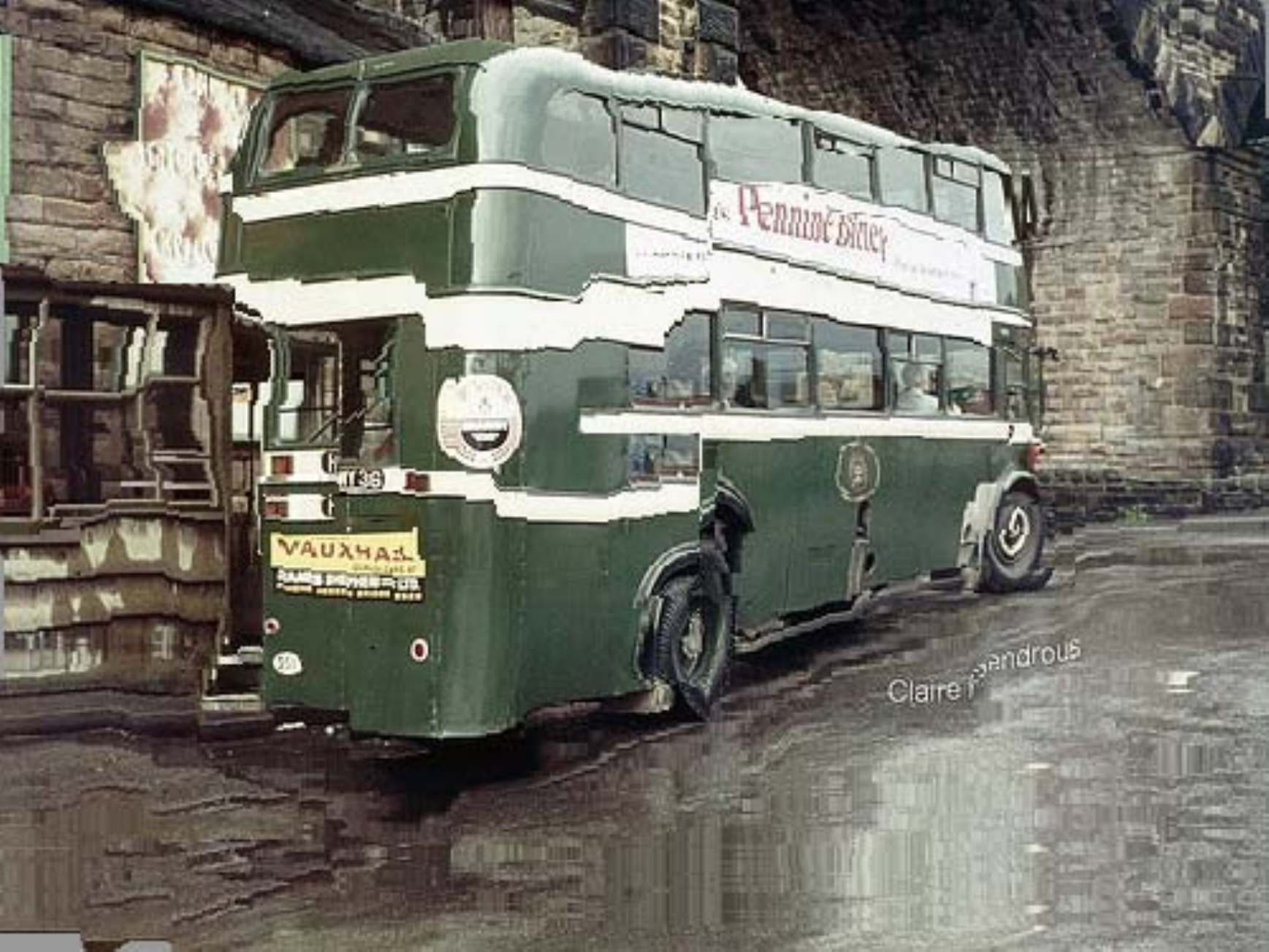}
     \includegraphics[width=\figobjectmatchingw, height=\figobjectmatchingsw]{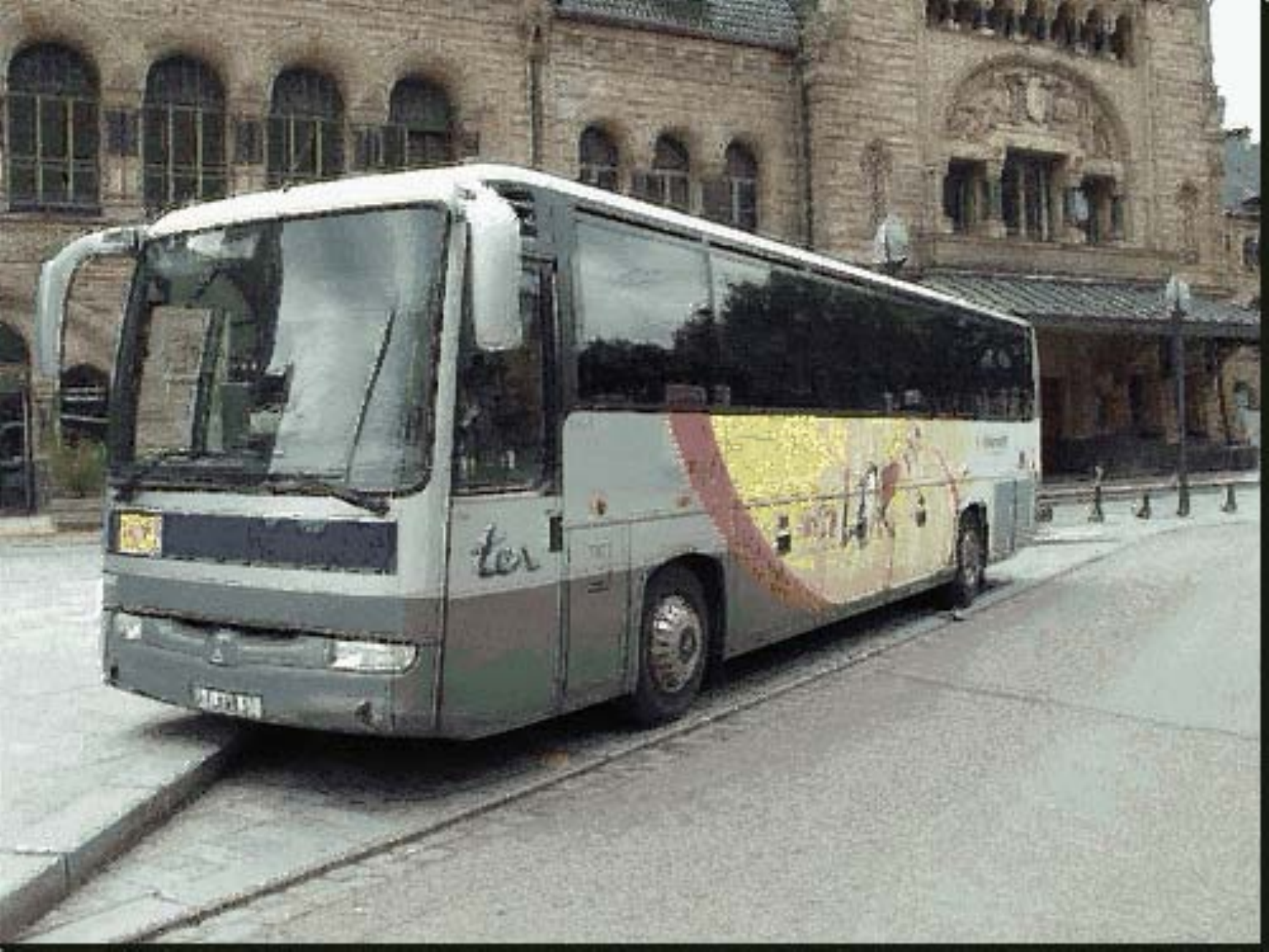} \\

 	 \includegraphics[width=\figobjectmatchingw, height=\figobjectmatchingsw]{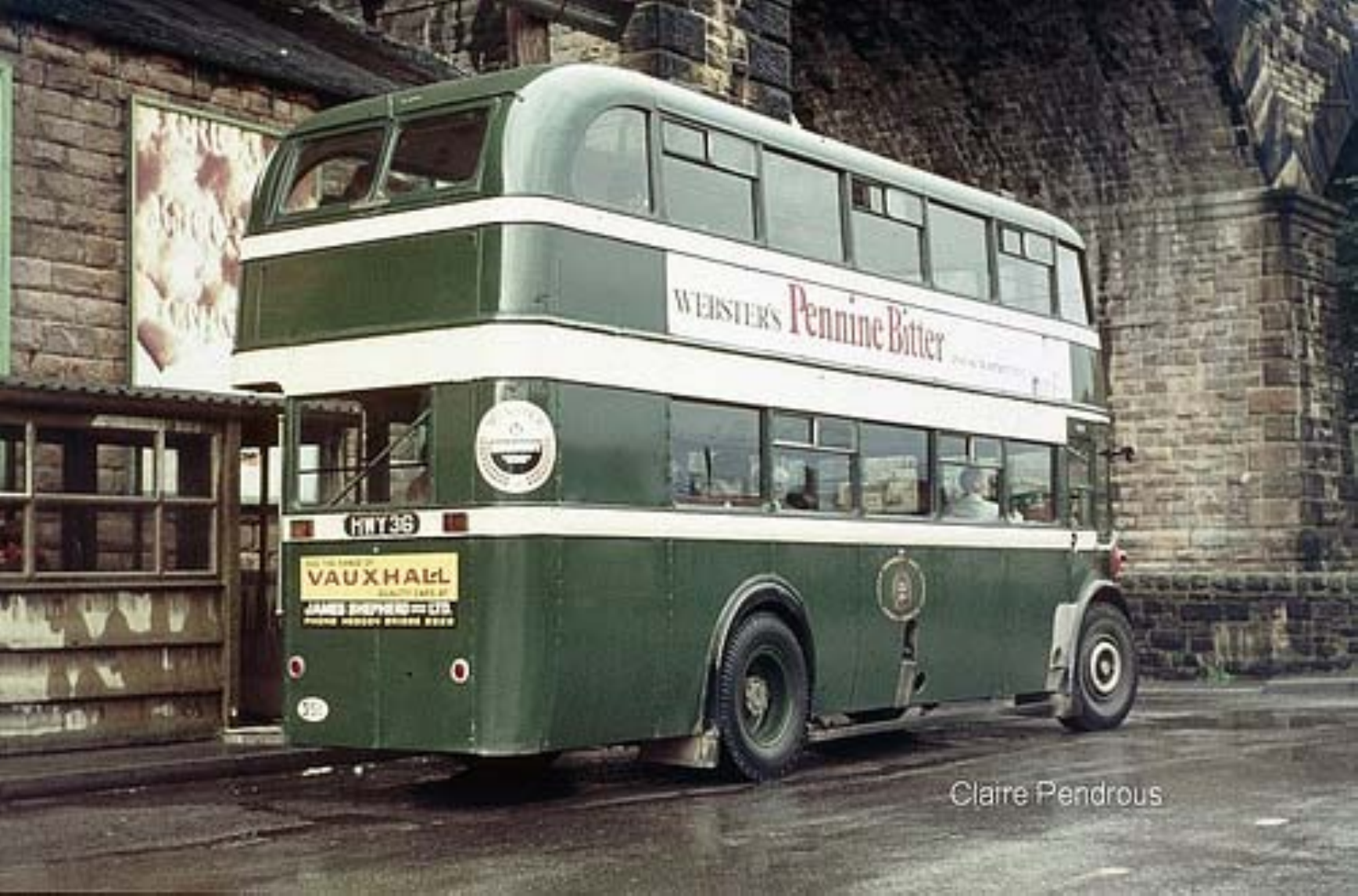}
	 \includegraphics[width=\figobjectmatchingw, height=\figobjectmatchingsw]{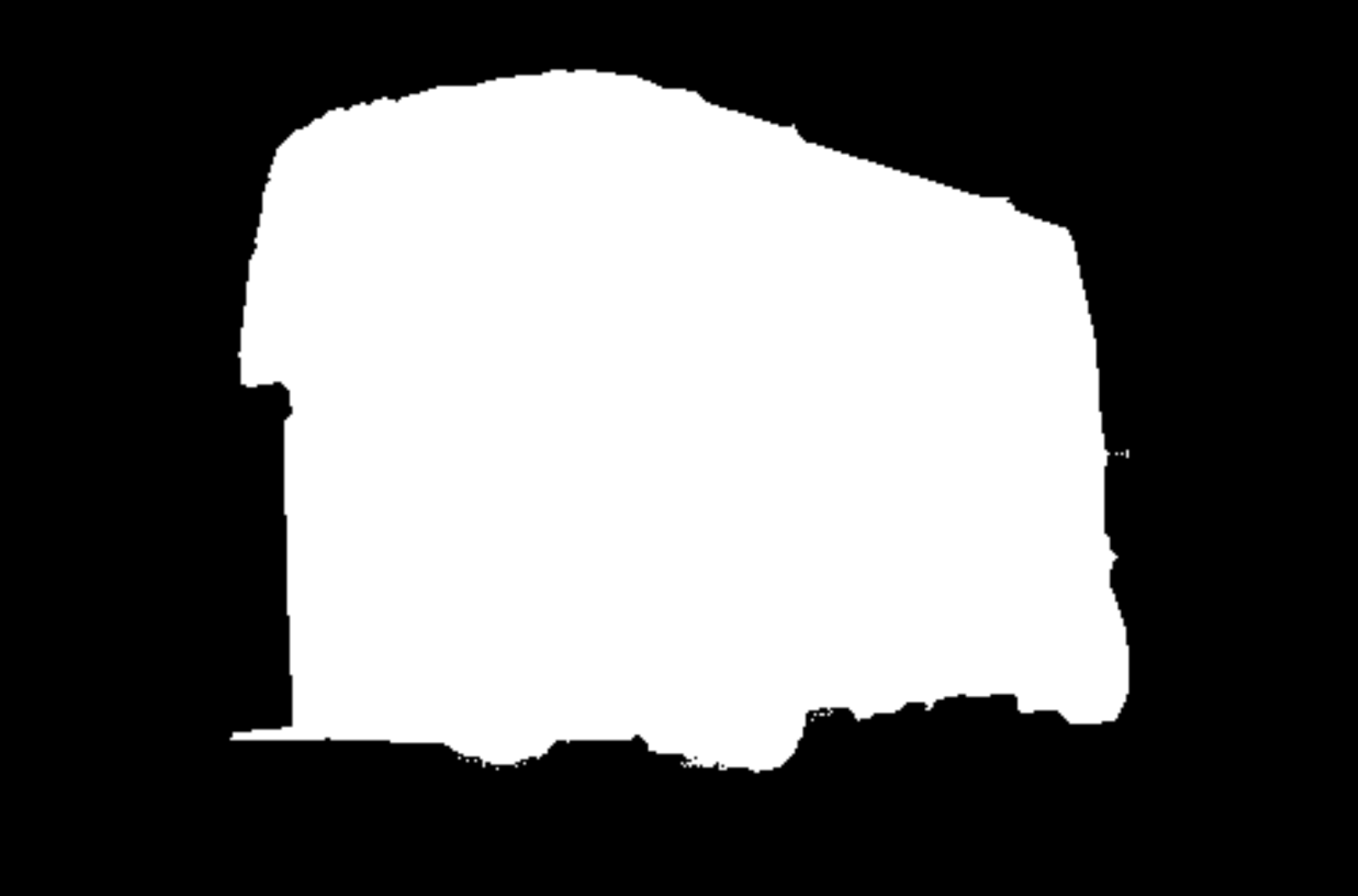}
	 \includegraphics[width=\figobjectmatchingw, height=\figobjectmatchingsw]{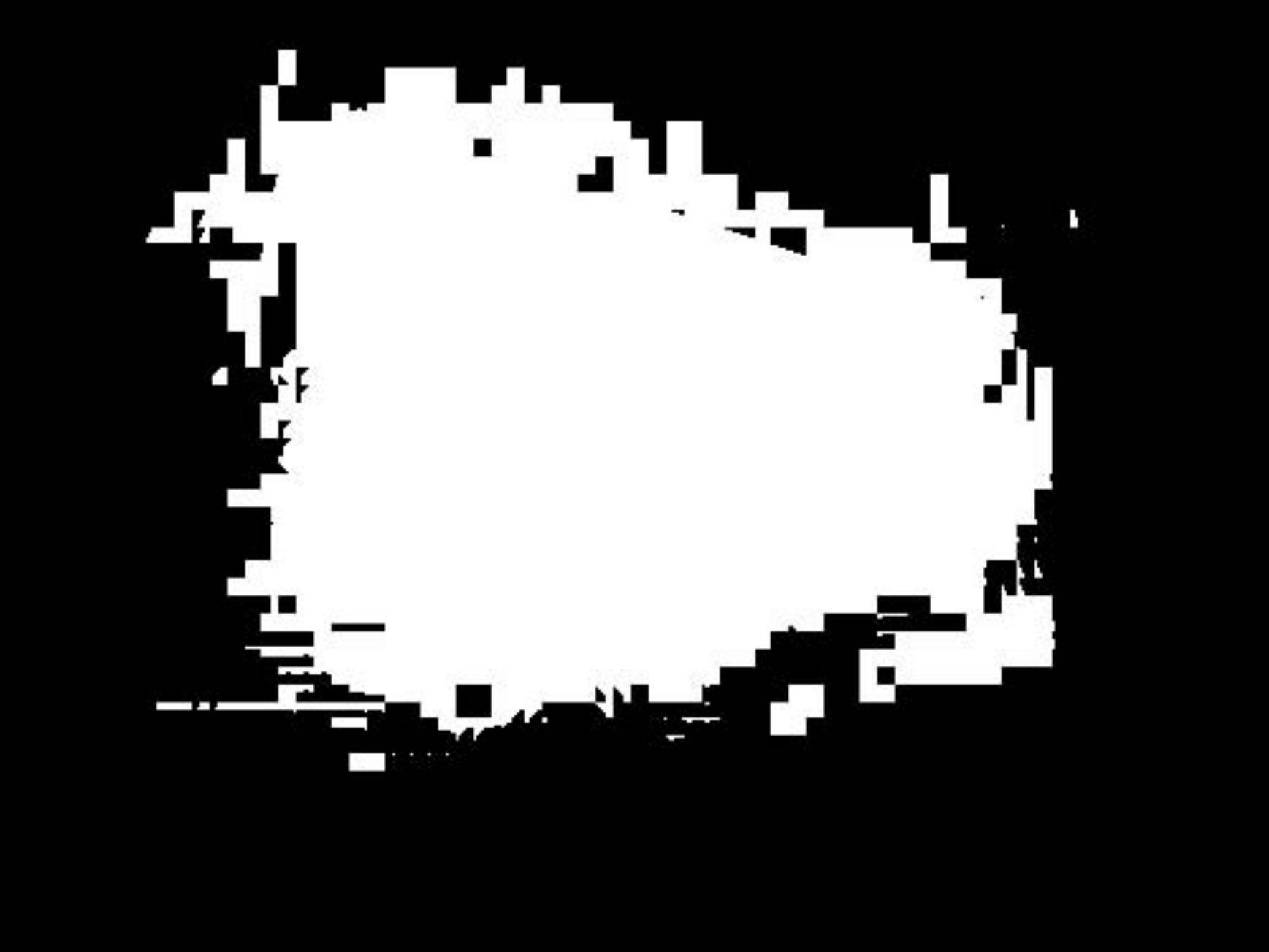}
	 \includegraphics[width=\figobjectmatchingw, height=\figobjectmatchingsw]{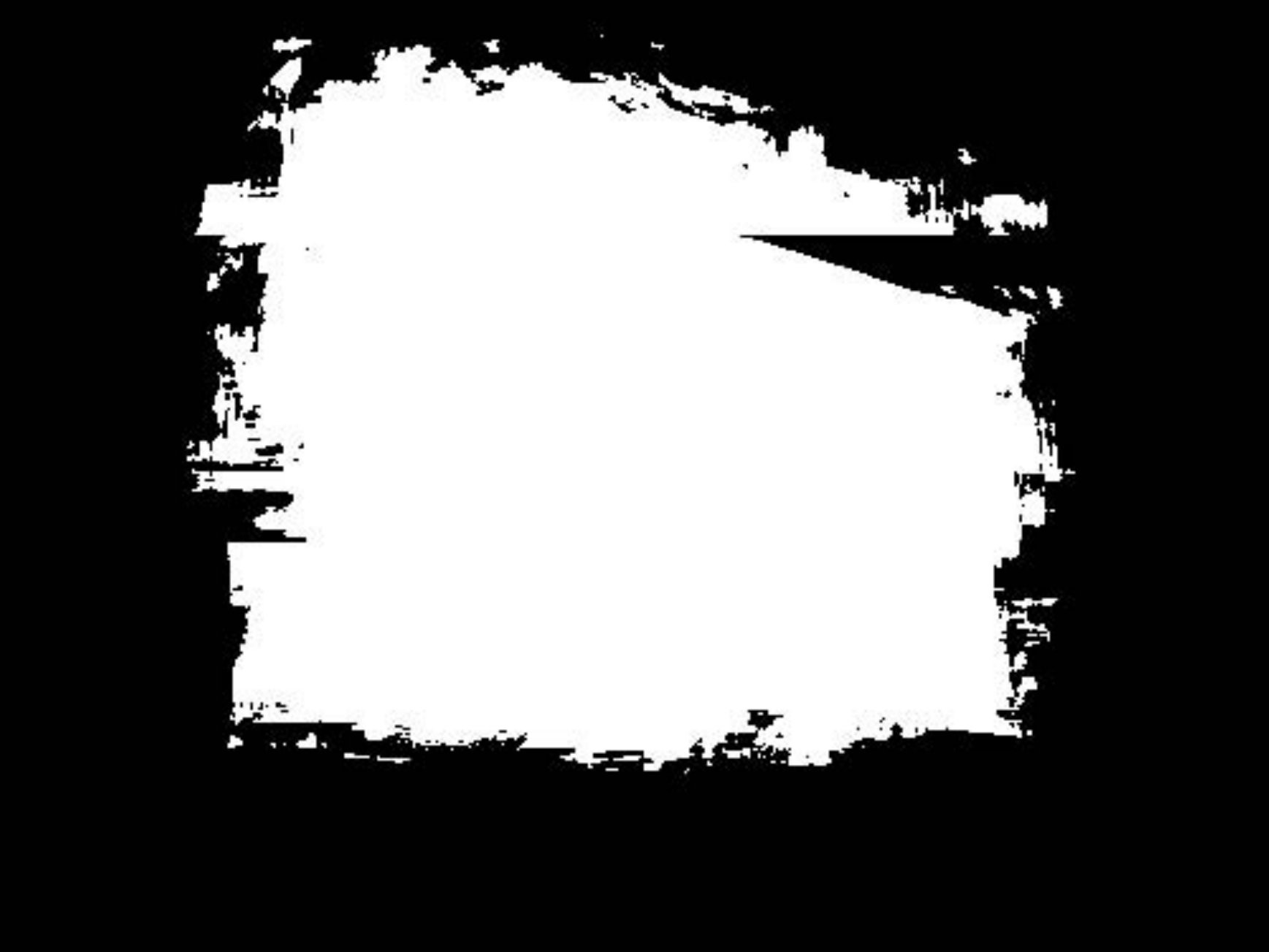}
	 \includegraphics[width=\figobjectmatchingw, height=\figobjectmatchingsw]{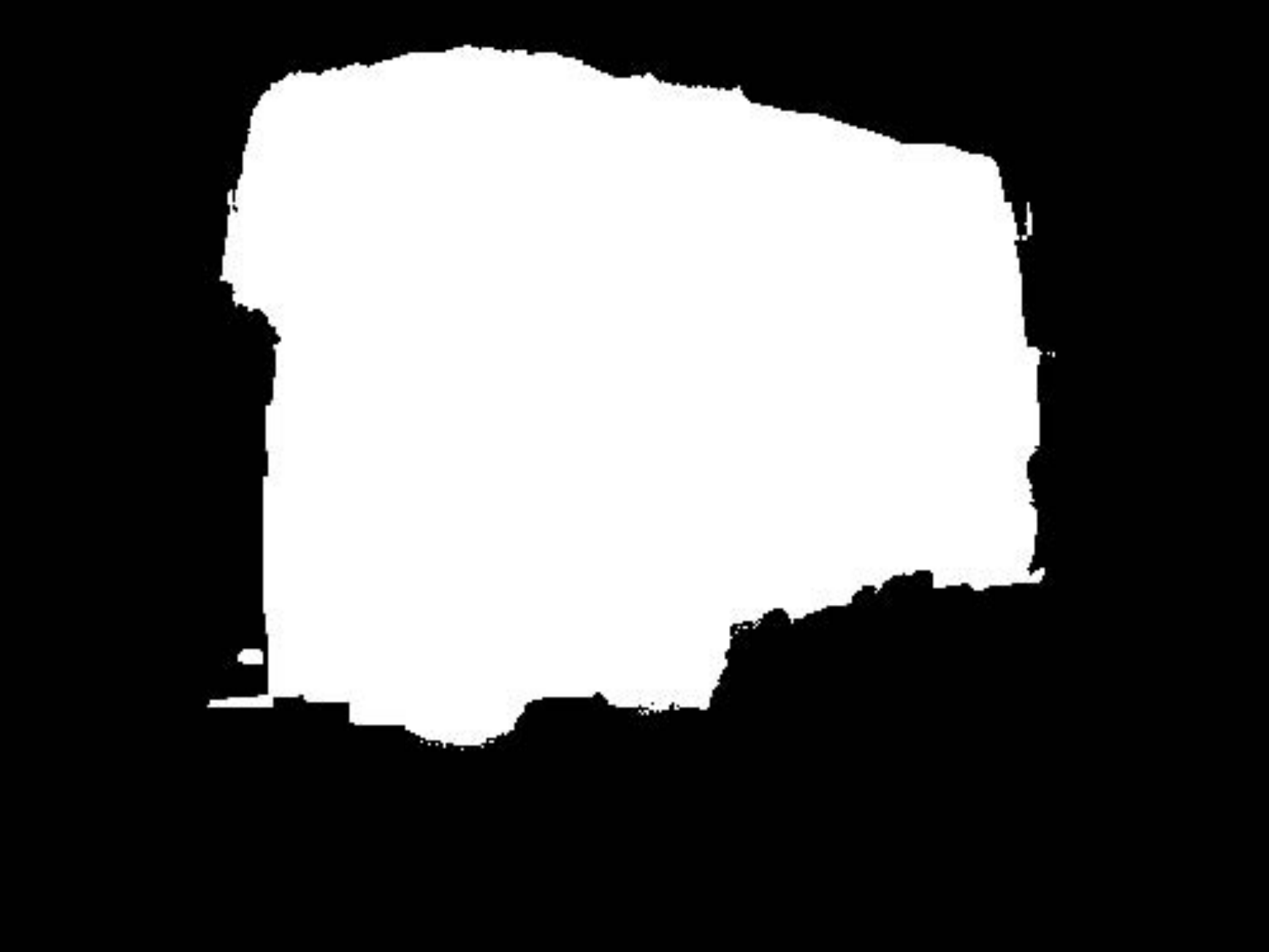}
	 \includegraphics[width=\figobjectmatchingw, height=\figobjectmatchingsw]{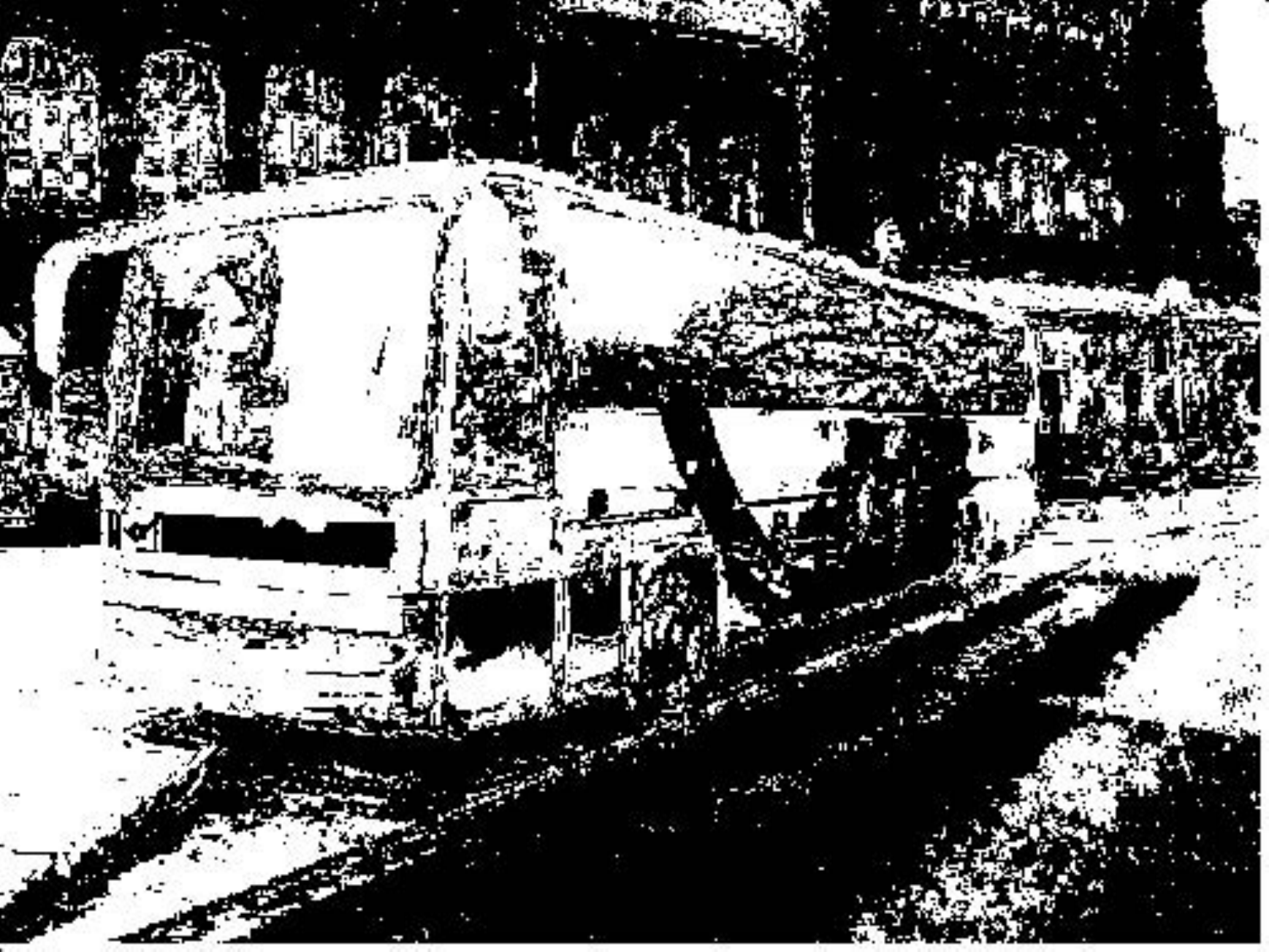} \\

\vspace{\vspacedistmid}
\caption{Qualitative comparison.
		 We show some example results of compared methods on the Pascal dataset.
         The first column stacks matching image pairs.
         The second column are ground truth label of input images.
         Columns 3--6 show the warping results
		from the exemplar image  to
		the test image  via pixel correspondences
		by our method, DSP, SIFT flow and CSH, respectively.
}  \label{fig:pascal_matching_seg}
\vspace{\vspacedist}
\end{figure*}

Figure \ref{fig:pascal_matching_seg} shows some example results of compared methods.
The results show that our method is more robust than other methods under large object
appearance variations (e.g., first example)
and cluttered backgrounds (e.g., third and fourth examples).

\begin{table} \center
{
\begin{tabular}{ r | c     c    c  c  }
\hline
\multirow{1}{*}{{{method}}} &\multicolumn{1}{c}{{Ours (patch layer)}} &\multicolumn{1}{c}{{DSP}}
&\multicolumn{1}{c}{{SIFT Flow}} &\multicolumn{1}{c}{{CSH}} \\
\hline
	{LT-ACC}     &\textbf{0.729} & 0.727 &0.711 &0.6173\\
	{Time (s)}     &\textbf{0.18} & 1.64 &9.38  &0.42\\
	\hline

\end{tabular}
}
\caption{Intra-class image matching performance on the Pascal dataset.
		The best results are in bold.}
		\label{tab:pascal_matching_acc}
\vspace{\vspacedist}
\end{table}

\subsection{Analysis of feature learning}\label{feature_learning}
In this section, we examine several factors
that may affect the performance of our proposed matching algorithms.
We randomly pick 20 pairs of images for each object class
on the Caltech-101 dataset (thus in total 2020 pairs of images).
The parameters are set as following unless otherwise specified.
We use K-means dictionary learning and K-means
triangle encoding for our method.
The dictionary is learned from $10^{6}$ image patches extracted from
200 Background\_Google class images in Caltech-101.
The dictionary size is set to $100$.
The patch size for extracting pixel features is
$11\times 11$-pixels region centered at that pixel.
Patch features are then calculated by max-pooling pixel
features within each non-overlapping
patch of $7\times 7$ pixels.

\paragraph{Evaluation of different dictionary learning methods and encoding schemes}
Here we examine the importance of dictionary learning and
feature encoding methods with respect to the final dense matching accuracy.
First, we compare the K-means dictionary learning method,
which has been used in our experiments in the previous section,
with two other dictionary learning algorithms,
namely K-SVD \cite{aharon2006ksvd} and
randomly sampled patches (RA) \cite{coates2011importance}.

\begin{table} \center
\vspace{\vspacedistfront}
{
\begin{tabular}{ r | c   c|  c   | c  c  }
\hline
{{{Encoder}}} &\multicolumn{2}{c|}{{KT}} &\multicolumn{1}{c|}{{KT}}
&\multicolumn{1}{c}{{OMP-K}} &\multicolumn{1}{c}{{SA}} \\
 {Dictionary}&{K-SVD} &{RA} &{K-means}  &  \multicolumn{2}{c}{{K-means}}     \\
\hline
	{LT-ACC}     &0.789	& 0.792 &0.803 &0.755 & 0.667 \\
 	{IOU}     	  &0.467 	& 0.481 &0.505 &0.386 & 0.014	\\
 	{LOC-ERR}    &0.354	& 0.336 &0.324 &0.497 & 1.621	\\
	\hline

\end{tabular}
}
\caption{Object matching performance using different dictionary
learning and encoding methods on Caltech-101 in matching accuracy.
Definition of the acronyms:
KT (K-means triangle),
OMP-K (orthogonal matching pursuit),
SA (soft assignment),
RA (random sampling).
}
\label{tab:caltech_dic_enc}
\vspace{\vspacedist}
\end{table}

As we can see from Table \ref{tab:caltech_dic_enc},
{ different dictionary learning methods do not have a significant
impact on the final the matching results.}
Even using randomly sampled patches as the dictionary
can achieve encouraging matching performance.
Different learning methods lead to similar matching accuracies.
As concluded in \cite{coates2011importance},
the main value of the dictionary is to provide basis,
and how to construct the dictionary is less critical than the choice of encoding.
For the application of pixel matching, we show that this conclusion holds too.

Then, we compare three encoding schemes:
K-means triangle (KT) \cite{coates2011importance},
OMP-k \cite{coates2011importance} and soft assignment (SA) \cite{liu2011defense}
to evaluate the impact of different encoding schemes.
We apply the OMP encoding with a sparsity level $k= 10 $.
According to the Table \ref{tab:caltech_dic_enc},
KT encoding achieves the best matching result,
which marginally outperforms other encoding methods.
It shows that the encoding scheme has a larger impact on the feature performance than dictionary learning.

In our model, there are two properties that pixel features should obtain.
The first is sparsity.
Since we use max-pooling to form our patch features,
some degree of sparsity contributes to the improvement.
Lack of sparsity in pixel features may decrease the power of patch features.
Based on the KT method, roughly one half of the features will be set to 0.
SA results in dense features and it performs very poorly in our framework.
This is very different from image classification applications \cite{liu2011defense}.

The other property is smoothness.
As mentioned in \cite{huang2012learning}, they use the congealing method to align  face images,
which reduces entropy by performing local hill-climbing in the transformation parameters.
The smoothness of that optimization landscape is a key factor to their successful alignment.
In our method, we might have faced similar situations.
To find the dense correspondence,
 we optimize the object function via belief propagation (BP).
Without the smoothness, the algorithm can easily get stuck at a local minimum.
OMP-k features are more sparse than the KT features, but OMP-k features
are not sufficiently smooth to perform well in our framework.

The results in Table \ref{tab:caltech_dic_enc} show that KT encoding
performs better than other two methods in our framework.
This observation deviates from the case of generic image classification, in which
many encoding methods (KT, SA, sparse coding, soft thresholding, etc.) have
performed similarly \cite{liu2011defense,coates2011importance}.

 \paragraph{Evaluation of the dictionary size}
In this subsection, we evaluate the impact of the dictionary size
 on the performance of dense matching.
 As dictionary size equals to the feature dimension,
 larger dictionary implies more patch information for each feature,
 which may lead to better matching performance. At the same time,
 the trade-off is that a lager feature dimension requires more
 computation and slows down the matching procedure.

\begin{table*} \center
{
\begin{tabular}{ r | c  c  c  c c  c | c  c c }
\hline
\multirow{1}{*}{{{Method}}} &\multicolumn{6}{c|}{{Ours}} &\multicolumn{1}{c}{{DSP}}
&\multicolumn{1}{c}{{SIFT Flow}} &\multicolumn{1}{c}{{CSH}} \\
\multirow{1}{*}{{{Dictionary Size}}} & 64 & 100 & 144 & 196 & 289 & 400 & & & \\

\hline
	{LT-ACC }  &0.764 & 0.765 &0.767 &\textbf{0.769} & 0.767 &0.765 & 0.757 &0.741 &0.594 \\
{CPU time (s)}  &\textbf{0.041} & 0.052 &0.065 &0.118 & 0.523 & 1.17 & 0.557 & 2.83  &0.165 \\
	\hline

\end{tabular}
}
\caption{Evaluation of different dictionary sizes w.r.t. the final matching accuracy.
  We can see that with a dictionary size
  of 100, our method has already outperformed DSP in accuracy and CPU time. } \label{tab:dic_time_matching}
\vspace{\vspacedist}
\end{table*}

 We evaluate six dictionary sizes (64, 100, 144, 196, 289, 400)
 on object matching performance, while keeping other parameters fixed.
 As shown in Table \ref{tab:dic_time_matching}, using a dictionary of size 196
 considerably outperforms the matching performance of size 64.
 Beyond this point, we observe slightly decreased accuracies.
 However, the  CPU time grows
 tremendously from 0.04 seconds per image matching with a dictionary size of  64 to 0.12
 seconds using a dictionary of size 196, as shown in Table \ref{tab:dic_time_matching}.
 It can be seen that even using a dictionary size of 64 we can achieve high accuracy
 performance.
 This experiment result shows that using a bigger dictionary size (longer feature length) leads to
 slightly better matching accuracy in a wide range.
Based on these observations, we have chosen the dictionary size to be 100
in our object matching and scene segmentation experiments
as a balance of the accuracy and CPU time.

Also note that with this choice, our feature dimension is
actually smaller than the SIFT's dimension of 128.
This fact has contributed to the faster speed of our method,
compared to methods like SIFT flow.

\paragraph{Effect of the patch size for extracting pixel features}
\begin{figure}
\vspace{\vspacedistfront}
\centering
\subfigure[]{
	\label{fig:pooling_patch_size:1}
	\includegraphics[width=0.431\textwidth]{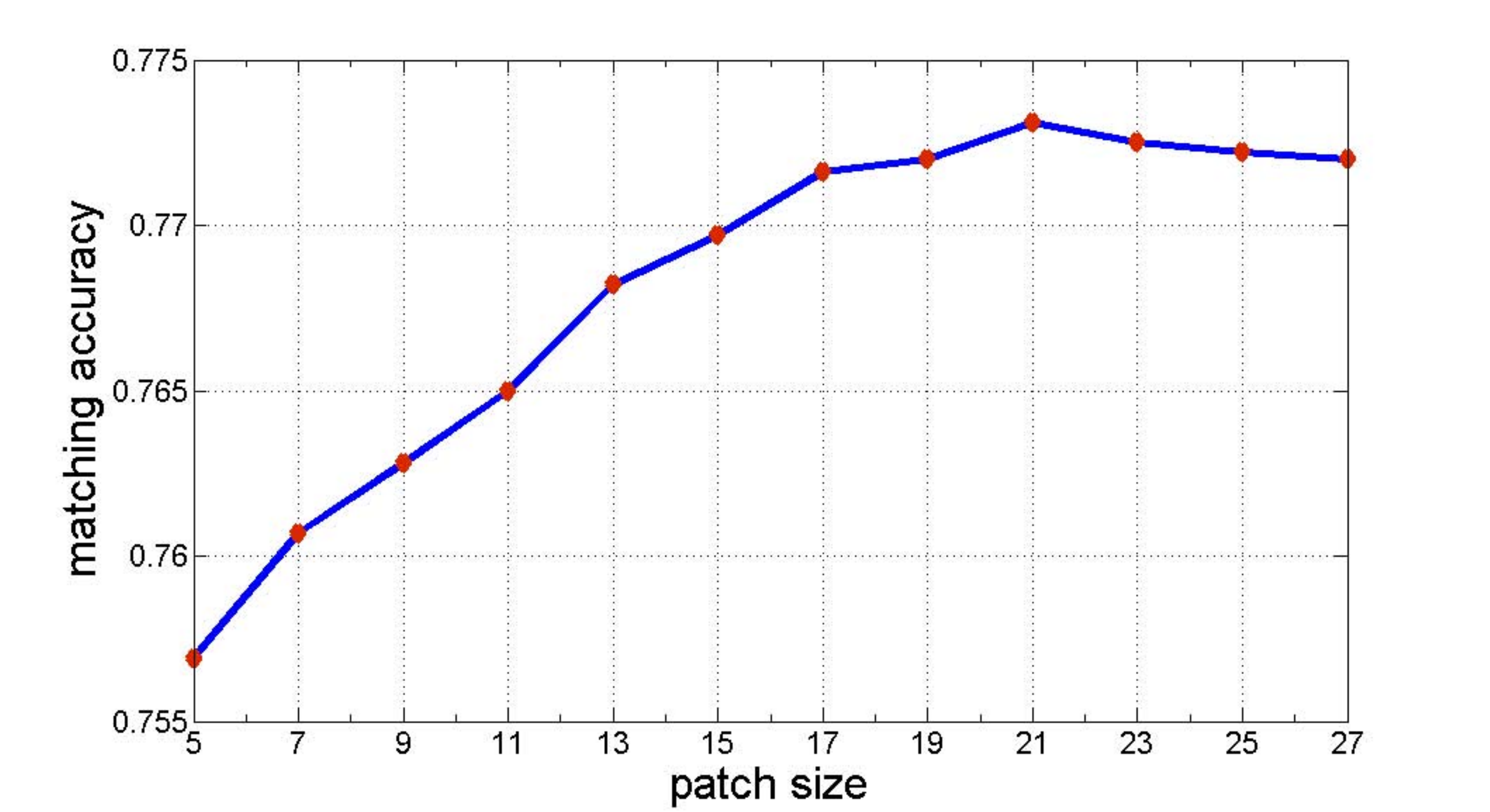}
	}
\subfigure[]{
    \label{fig:pooling_patch_size:2}
	\includegraphics[width=0.431\textwidth]{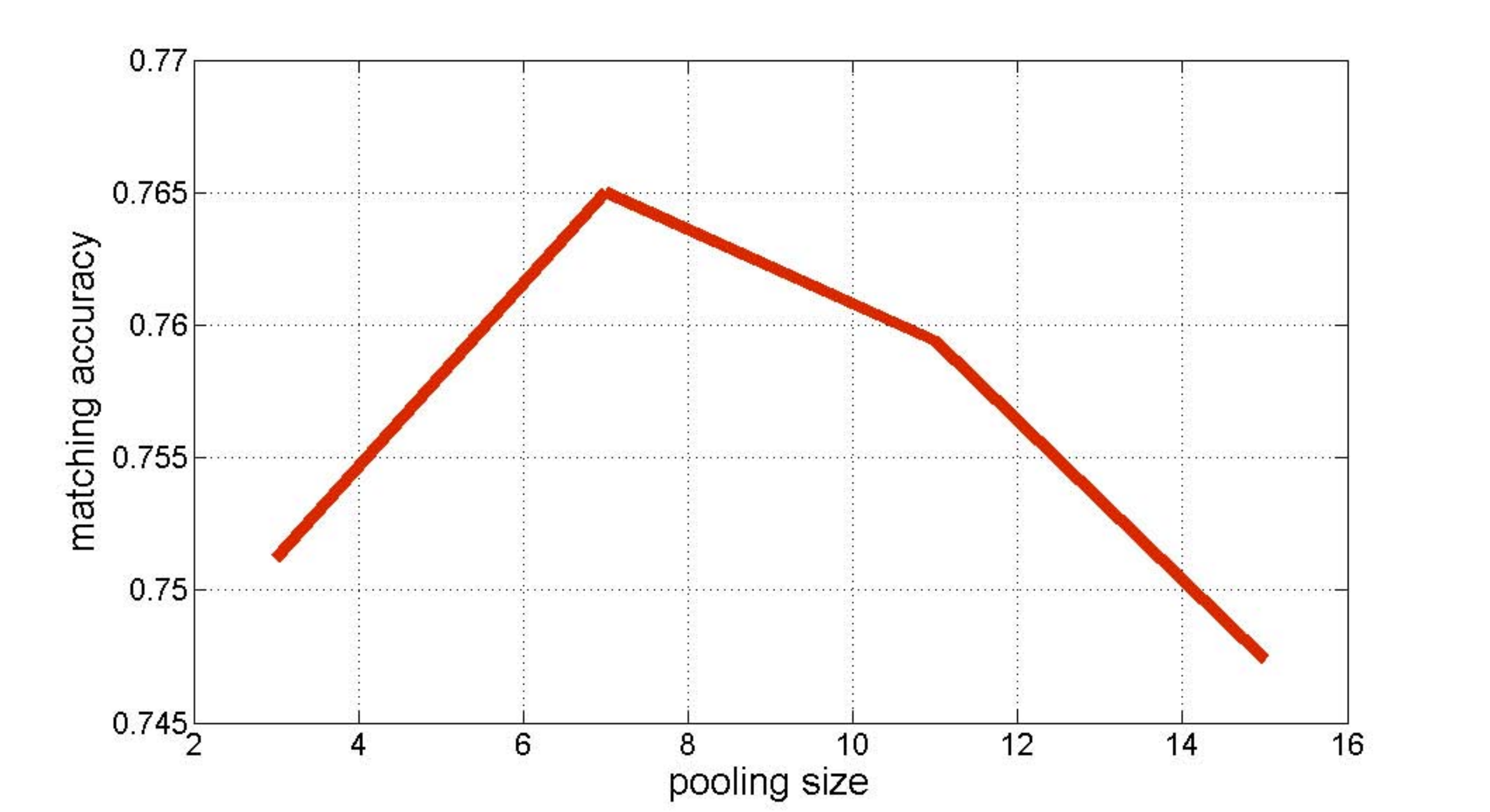} } %
\subfigure[]{
    \label{fig:pooling_patch_size:3}
	\includegraphics[width=0.431\textwidth]{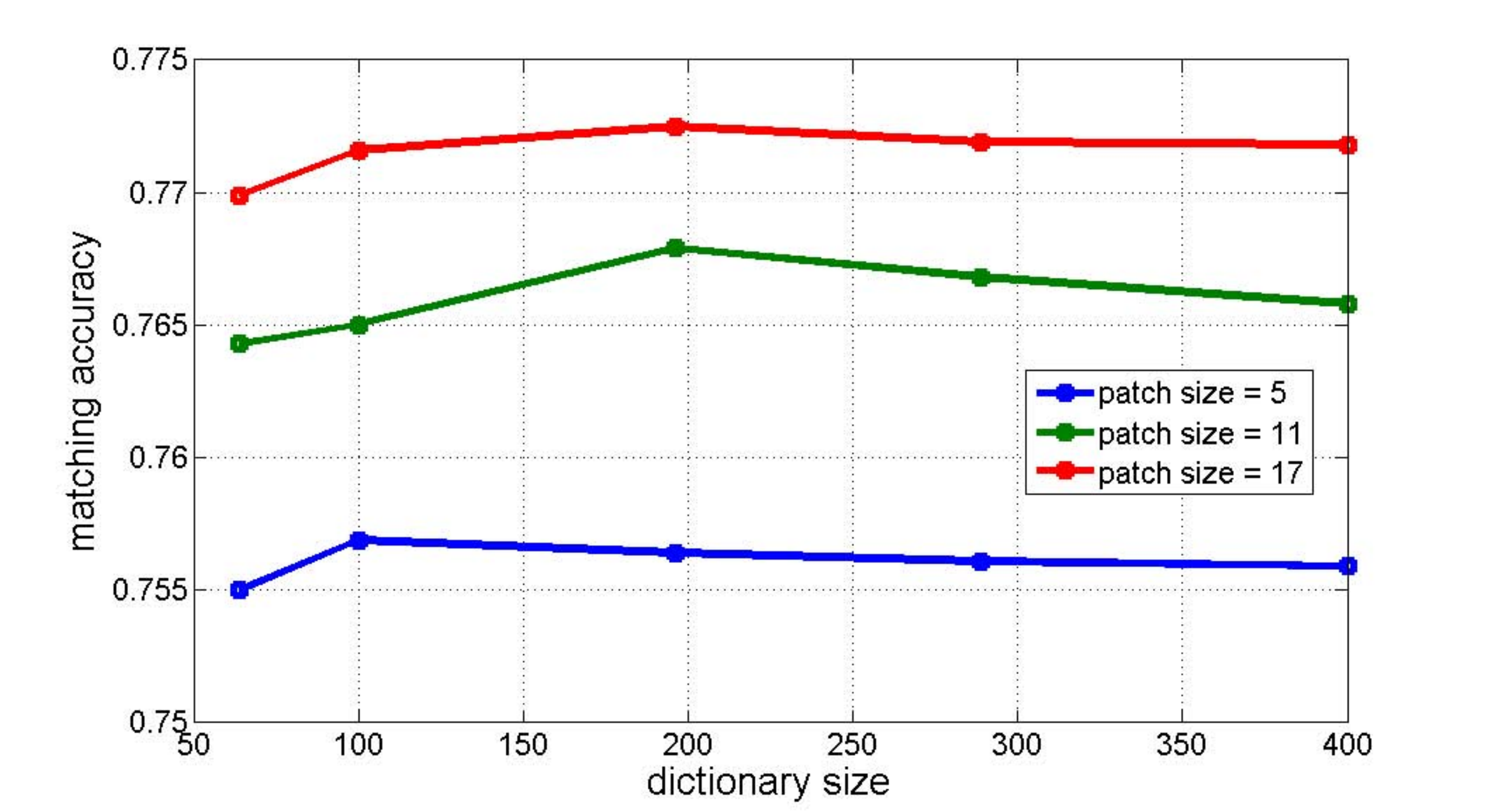} }%

\vspace{\vspacedistmid}
     \caption{
      The impact of changes in model setup of feature learning process.
      \subref{fig:pooling_patch_size:1} shows the matching accuracy using different patch sizes
      for extracting pixel layer features.
      \subref{fig:pooling_patch_size:2} shows the impact of the max-pooling size for obtaining the
      patch layer features.
      \subref{fig:pooling_patch_size:3} shows the performance of different patch sizes and dictionary sizes.
	The impact of the patch size is greater than the changes of the dictionary size. Beyond the
	point of patch size $17\times17$, the gain of increasing the patch size of  the patch layer
	becomes less noticeable.
}
\label{fig:pooling_patch_size}
\vspace{\vspacedist}
\end{figure}

This experiment considers the effect of patch size for
extracting pixel features (from $5\times5$ to $27\times27$ pixels, we evaluate 12 patch sizes).
For each image pixel, we extract a certain size patch around that pixel and
obtain pixel feature by using KT encoding.
Larger patch regions allow us to extract more complex features
as they may contain more information.
On the other hand, it increases the dimensionality of the space that the
algorithm must cover.
The results are shown in Figure \ref{fig:pooling_patch_size:1} and Figure
\ref{fig:pooling_patch_size:3}.
Overall, larger patches for extracting pixel features lead to  better
matching accuracy.
From $5\times5$ to $17\times17$ pixels, the matching performance increases significantly,
while beyond $17 \times 17 $  pixels,  the accuracy  improvement becomes negligible.
From Figure \ref{fig:pooling_patch_size:3}, we can see that the impact of the patch size
for extracting pixel features is much greater than the choice of dictionary sizes.
As shown in Figure \ref{fig:pooling_patch_size:3}, the best performance of
$5\times5$ patch size is obtained by using a dictionary size of 100. Meanwhile,
for patch size of $11\times11$, best performance point shifts to the dictionary size
of 196.
This  suggests that one should choose a larger dictionary when larger patches are used.

\paragraph{Impact of the pooling size at the patch layer}
We examine the impact of the pooling size for obtaining the patch layer features.
As described in Section \ref{pixel_patch_feat},
an image is divided into non-overlapping pooling regions.
Each of those pooling region/patches is represented by a patch feature,
which is obtained by
max-pooling all pixel features within that patch.
As a result,  larger-size pooling regions (patches) result in fewer pooled features.

The experiment result is shown in Figure \ref{fig:pooling_patch_size:2}.
We consider four pooling sizes ($3\times3$, $7\times7$, $11\times11$, $15\times15$ pixels).
The best performance is achieved by the pooling size of $7\times7$ pixels,
regardless of the region
size for extracting pixel layer features.

\paragraph{Impact of the size of training data}

In all the previous experiments, the dictionary is learned form $10^{6}$ patches
extracted from 200 Background\_Google class images in the Caltech-101 dataset.
In this experiment, we evaluate the impact of the training data size.
Multiple dictionaries are learned from different numbers of patches in the Background\_Google class
of Caltech-101 dataset.
The matching accuracy increases very slightly with more sampled patches for training, which is expected.

\begin{figure}
\vspace{\vspacedistfront}
\centering
	\includegraphics[width=0.5\textwidth]{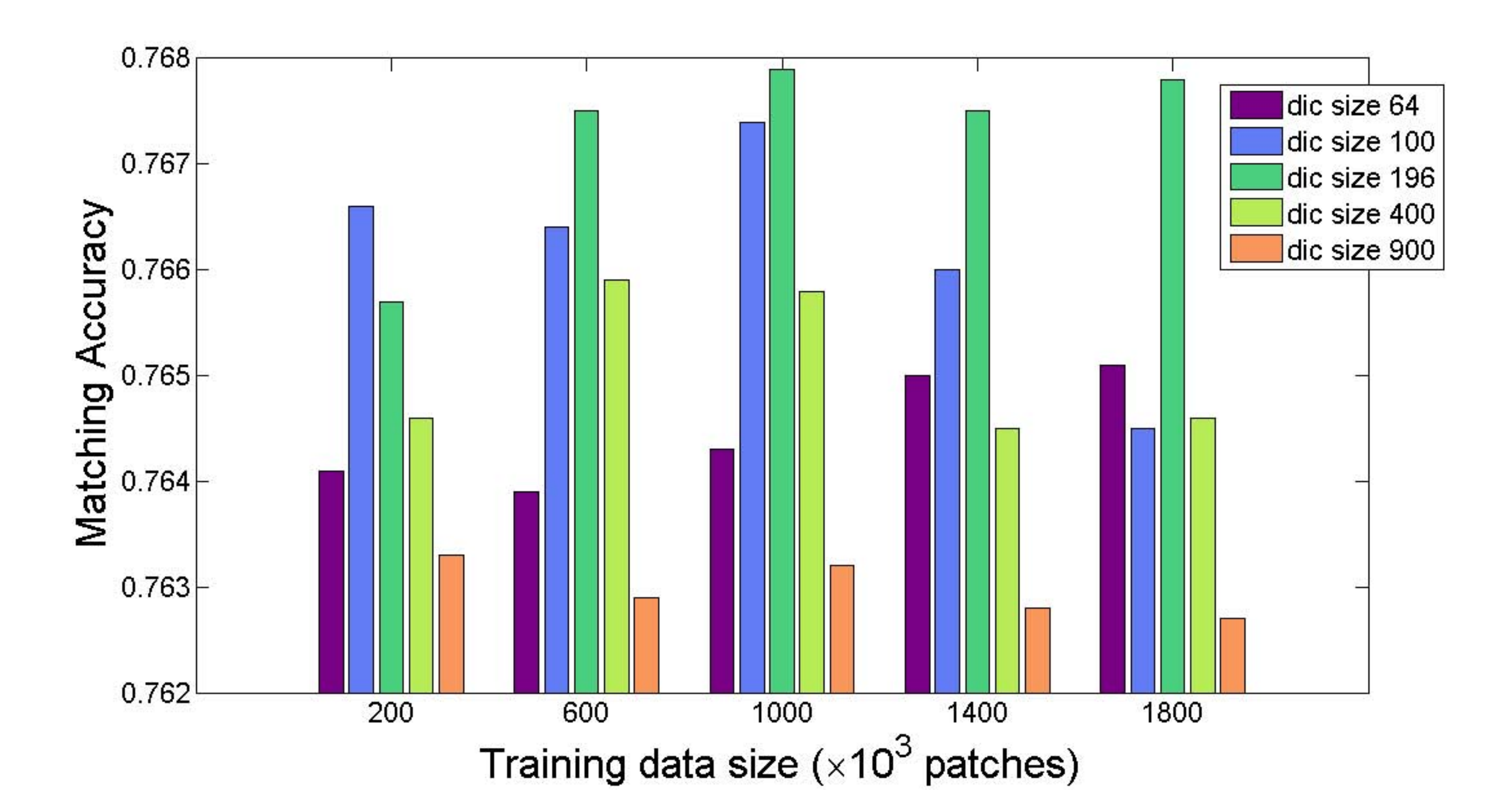}
	\vspace{\vspacedist}
\caption{
Matching accuracies with varying amount of training data and varying dictionary sizes.
}
\label{fig:sample_dic_size}
\vspace{\vspacedist}
\end{figure}

\section{Conclusions}
We have proposed to learn features for pixel correspondence estimation in an unsupervised manner.
A new multi-layer matching algorithm is designed, which naturally
aligns with the unsupervised feature learning pipeline.
For the first time, we show that learned features can work better than those widely-used hand-crafted
features like SIFT on the problem of dense pixel correspondence estimation.

We empirically demonstrate that our proposed algorithm can
robustly match different objects or scenes exhibiting large appearance differences and achieve
state-of-the-art performance in terms of
both matching accuracy and running time. A limitation of the proposed framework is that
currently the system is not very robust to  rotation and scale variations.
We want to pursue this issue in future work.

We have made the code online available at \url{https://bitbucket.org/chhshen/ufl}.

\section*{Acknowledgments}
This work is in part supported by ARC grant FT120100969.
C. Zhang's contribution was made when she was a visiting student at the University of Adelaide.

\bibliographystyle{IEEEtran}
\bibliography{cs}

\begin{thebibliography}{10}
\providecommand{\url}[1]{#1}
\csname url@samestyle\endcsname
\providecommand{\newblock}{\relax}
\providecommand{\bibinfo}[2]{#2}
\providecommand{\BIBentrySTDinterwordspacing}{\spaceskip=0pt\relax}
\providecommand{\BIBentryALTinterwordstretchfactor}{4}
\providecommand{\BIBentryALTinterwordspacing}{\spaceskip=\fontdimen2\font plus
\BIBentryALTinterwordstretchfactor\fontdimen3\font minus
  \fontdimen4\font\relax}
\providecommand{\BIBforeignlanguage}[2]{{%
\expandafter\ifx\csname l@#1\endcsname\relax
\typeout{** WARNING: IEEEtran.bst: No hyphenation pattern has been}%
\typeout{** loaded for the language `#1'. Using the pattern for}%
\typeout{** the default language instead.}%
\else
\language=\csname l@#1\endcsname
\fi
#2}}
\providecommand{\BIBdecl}{\relax}
\BIBdecl

\bibitem{liu2011sift}
C.~Liu, J.~Yuen, and A.~Torralba, ``{SIFT} flow: Dense correspondence across
  scenes and its applications,'' \emph{IEEE Trans. Patt. Analysis \& Machine
  Intell.}, vol.~33, no.~5, pp. 978--994, 2011.

\bibitem{korman2011coherency}
S.~Korman and S.~Avidan, ``Coherency sensitive hashing,'' in \emph{Proc. IEEE
  Int. Conf. Comp. Vis.}, 2011, pp. 1607--1614.

\bibitem{kim2013deformable}
J.~Kim, C.~Liu, F.~Sha, and K.~Grauman, ``Deformable spatial pyramid matching
  for fast dense correspondences,'' in \emph{Proc. IEEE Conf. Comp. Vis. Patt.
  Recogn.}, 2013, pp. 2307--2314.

\bibitem{Barnes2010TGP}
C.~Barnes, E.~Shechtman, D.~Goldman, and A.~Finkelstein, ``The generalized
  {PatchMatch} correspondence algorithm,'' in \emph{Proc. Eur. Conf. Comp.
  Vis.}, 2010.

\bibitem{leordeanu2013sparsetodense}
M.~Leordeanu, A.~Zanfir, and C.~Sminchisescu, ``Locally affine sparse-to-dense
  matching for motion and occlusion estimation,'' in \emph{Proc. IEEE Int.
  Conf. Comp. Vis.}, 2013.

\bibitem{SIFT}
D.~G. Lowe, ``Object recognition from local scale-invariant features,'' in
  \emph{Proc. IEEE Int. Conf. Comp. Vis.}, 1999.

\bibitem{bo2013multipath}
L.~Bo, X.~Ren, and D.~Fox, ``Multipath sparse coding using hierarchical
  matching pursuit,'' in \emph{Proc. IEEE Conf. Comp. Vis. Patt. Recogn.},
  2013, pp. 660--667.

\bibitem{coates2011importance}
A.~Coates and A.~Ng, ``The importance of encoding versus training with sparse
  coding and vector quantization,'' in \emph{Proc. Int. Conf. Mach. Learn.},
  2011, pp. 921--928.

\bibitem{coates2011analysis}
A.~Coates, A.~Y. Ng, and H.~Lee, ``An analysis of single-layer networks in
  unsupervised feature learning,'' in \emph{Int. Conf. Artificial Intell. \&
  Stat.}, 2011, pp. 215--223.

\bibitem{huang2012learning}
G.~B. Huang, M.~Mattar, H.~Lee, and E.~G. Learned-Miller, ``Learning to align
  from scratch,'' in \emph{Proc. Adv. Neural Inf. Process. Syst.}, 2012.

\bibitem{liu2009nonparametric}
C.~Liu, J.~Yuen, and A.~Torralba, ``Nonparametric scene parsing: Label transfer
  via dense scene alignment,'' in \emph{Proc. IEEE Conf. Comp. Vis. Patt.
  Recogn.}, 2009, pp. 1972--1979.

\bibitem{everingham2014pascal}
\BIBentryALTinterwordspacing
M.~Everingham, S.~Eslami, L.~Van~Gool, C.~Williams, J.~Winn, and A.~Zisserman,
  ``The {Pascal} visual object classes challenge: A retrospective,'' \emph{Int.
  J. Comp. Vis.}, 2014. [Online]. Available:
  \url{http://dx.doi.org/10.1007/s11263-014-0733-5}
\BIBentrySTDinterwordspacing

\bibitem{rubinstein2013unsupervised}
M.~Rubinstein, A.~Joulin, J.~Kopf, and C.~Liu, ``Unsupervised joint object
  discovery and segmentation in internet images,'' in \emph{Proc. IEEE Conf.
  Comp. Vis. Patt. Recogn.}, 2013, pp. 1939--1946.

\bibitem{duchenne2011graph}
O.~Duchenne, A.~Joulin, and J.~Ponce, ``A graph-matching kernel for object
  categorization,'' in \emph{Proc. IEEE Int. Conf. Comp. Vis.}, 2011, pp.
  1792--1799.

\bibitem{berg2005shapematch}
A.~C. Berg, T.~Berg, and J.~Malik, ``Shape matching and object recognition
  using low distortion correspondences,'' in \emph{Proc. IEEE Conf. Comp. Vis.
  Patt. Recogn.}, vol.~1, 2005, pp. 26--33.

\bibitem{leordeanu2005spectral}
M.~Leordeanu and M.~Hebert, ``A spectral technique for correspondence problems
  using pairwise constraints,'' in \emph{Proc. IEEE Int. Conf. Comp. Vis.},
  vol.~2, 2005.

\bibitem{tola2008fast}
E.~Tola, V.~Lepetit, and P.~Fua, ``A fast local descriptor for dense
  matching,'' in \emph{Proc. IEEE Conf. Comp. Vis. Patt. Recogn.}, 2008.

\bibitem{zelnik2012sifts}
L.~Zelnik-Manor, ``On {SIFTs} and their scales,'' in \emph{Proc. IEEE Conf.
  Comp. Vis. Patt. Recogn.}, 2012, pp. 1522--1528.

\bibitem{quoc2013buildinghigh}
Q.~V. Le, M.~Ranzato, R.~Monga, M.~Devin, G.~Corrado, K.~Chen, J.~Dean, and
  A.~Y. Ng, ``Building high-level features using large scale unsupervised
  learning,'' in \emph{Proc. Int. Conf. Mach. Learn.}, 2012.

\bibitem{NIPS2011ICA}
Q.~V. Le, A.~Karpenko, J.~Ngiam, and A.~Y. Ng, ``{ICA} with reconstruction cost
  for efficient overcomplete feature learning,'' in \emph{Proc. Adv. Neural
  Inf. Process. Syst.}, 2011, pp. 1017--1025.

\bibitem{AISTATS2012ZhouSL12}
G.~Zhou, K.~Sohn, and H.~Lee, ``Online incremental feature learning with
  denoising autoencoders,'' in \emph{Proc. Int. Conf. Artificial Intell. \&
  Stat.}, vol.~22, 2012, pp. 1453--1461.

\bibitem{Sparse2010}
J.~Wright, Y.~Ma, J.~Mairal, G.~Sapiro, T.~S. Huang, and S.~Yan, ``Sparse
  representation for computer vision and pattern recognition,''
  \emph{Proceedings of the IEEE}, vol.~98, no.~6, pp. 1031--1044, 2010.

\bibitem{Mairal2010JMLR}
J.~Mairal, F.~Bach, J.~Ponce, and G.~Sapiro, ``Online learning for matrix
  factorization and sparse coding,'' \emph{J. Mach. Learn. Res.}, vol.~11, pp.
  19--60, 2010.

\bibitem{Lee1999}
D.~D. Lee and H.~S. Seung, ``Learning the parts of objects by non-negative
  matrix factorization,'' \emph{Nature}, vol. 401, no. 6755, pp. 788--791,
  1999.

\bibitem{NIPS2011ADAM}
A.~Coates and A.~Y. Ng, ``Selecting receptive fields in deep networks,'' in
  \emph{Proc. Adv. Neural Inf. Process. Syst.}, 2011, pp. 2528--2536.

\bibitem{Tuytelaars00widebaseline}
T.~Tuytelaars and L.~{Van Gool}, ``Wide baseline stereo matching based on
  local, affinely invariant regions,'' in \emph{Proc. British Machine Vis.
  Conf.}, 2000, pp. 412--425.

\bibitem{Bay:2008:SRF}
H.~Bay, A.~Ess, T.~Tuytelaars, and L.~Van~Gool, ``Speeded-up robust features
  (surf),'' \emph{Comput. Vis. Image Underst.}, vol. 110, no.~3, pp. 346--359,
  2008.

\bibitem{Tola10}
E.~Tola, V.~Lepetit, and P.~Fua, ``{DAISY}: An efficient dense descriptor
  applied to wide baseline stereo,'' \emph{{IEEE} Trans. Pattern Anal. Mach.
  Intell.}, vol.~32, no.~5, pp. 815--830, May 2010.

\bibitem{Heikkila:2009}
M.~Heikkil\"{a}, M.~Pietik\"{a}inen, and C.~Schmid, ``Description of interest
  regions with local binary patterns,'' \emph{Pattern Recogn.}, vol.~42, no.~3,
  pp. 425--436, 2009.

\bibitem{aharon2006ksvd}
M.~Aharon, M.~Elad, and A.~Bruckstein, ``{K-SVD}: An algorithm for designing
  overcomplete dictionaries for sparse representation,'' \emph{IEEE Trans.
  Signal Process.}, vol.~54, no.~11, pp. 4311--4322, 2006.

\bibitem{boureau2010theoretical}
Y.-L. Boureau, J.~Ponce, and Y.~LeCun, ``A theoretical analysis of feature
  pooling in visual recognition,'' in \emph{Proc. Int. Conf. Mach. Learn.},
  2010, pp. 111--118.

\bibitem{Ihler05loopybelief}
A.~T. Ihler, J.~W.~F. {III}, and A.~S. Willsky, ``Loopy belief propagation:
  Convergence and effects of message errors,'' \emph{J. Mach. Learn. Res.},
  vol.~6, pp. 905--936, 2005.

\bibitem{Felzenszwalb2006Effi}
\BIBentryALTinterwordspacing
P.~Felzenszwalb and D.~Huttenlocher, ``Efficient belief propagation for early
  vision,'' \emph{Int. J. Comp. Vis.}, vol.~70, no.~1, pp. 41--54, 2006.
  [Online]. Available: \url{http://dx.doi.org/10.1007/s11263-006-7899-4}
\BIBentrySTDinterwordspacing

\bibitem{liu2011defense}
L.~Liu, L.~Wang, and X.~Liu, ``In defense of soft-assignment coding,'' in
  \emph{Proc. IEEE Int. Conf. Comp. Vis.}, 2011, pp. 2486--2493.

\end{thebibliography}

\end{document}